\documentclass[10pt,twocolumn,letterpaper]{article}
\pdfoutput=1
\usepackage[table]{xcolor}
\usepackage{cvpr}
\usepackage{times}
\usepackage{epsfig}
\usepackage{graphicx, subfigure}
\usepackage{amsmath}
\usepackage{amssymb}
\usepackage{psfrag}
\usepackage{mathrsfs}
\usepackage{cite}
\usepackage{multirow}
\usepackage{makecell}
\usepackage{placeins}
\usepackage{balance}
\usepackage{capt-of}
\usepackage[page]{appendix}

\makeatletter
\def\old@comma{,}
\catcode`\,=13
\def,{%
	\ifmmode%
	\old@comma\discretionary{}{}{}%
	\else%
	\old@comma%
	\fi%
}
\makeatother

\usepackage[pagebackref=true,breaklinks=true,letterpaper=true,colorlinks,bookmarks=false]{hyperref}

\cvprfinalcopy 

\def\cvprPaperID{2581} 

\ifcvprfinal\fi
\begin{document}

\title{CASENet: Deep Category-Aware Semantic Edge Detection}

\author{
Zhiding Yu\thanks{The authors contributed equally.}\\
Carnegie Mellon University\\
{\tt\small yzhiding@andrew.cmu.edu}
\and
Chen Feng\footnotemark[1]{} \ \ \ Ming-Yu Liu\thanks{This work was done during the affiliation with MERL.} \ \ \ Srikumar Ramalingam\footnotemark[2]{}\\
Mitsubishi Electric Research Laboratories (MERL)\\
{\tt\small cfeng@merl.com, mingyul@nvidia.com, srikumar@cs.utah.edu}
}

\maketitle
\thispagestyle{empty}

\begin{abstract}
   Boundary and edge cues are highly beneficial in improving a wide variety of vision tasks such as semantic segmentation, object recognition, stereo, and object proposal generation. Recently, the problem of edge detection has been revisited and significant progress has been made with deep learning. While classical edge detection is a challenging binary problem in itself, the category-aware semantic edge detection by nature is an even more challenging multi-label problem. We model the problem such that each edge pixel can be associated with more than one class as they appear in contours or junctions belonging to two or more semantic classes. To this end, we propose a novel end-to-end deep semantic edge learning architecture based on ResNet and a new skip-layer architecture where category-wise edge activations at the top convolution layer share and are fused with the same set of bottom layer features. We then propose a multi-label loss function to supervise the fused activations. We show that our proposed architecture benefits this problem with better performance, and we outperform the current state-of-the-art semantic edge detection methods by a large margin on standard data sets such as SBD and Cityscapes.
\end{abstract}

\section{Introduction}
\label{sec.intro}

\definecolor{blk_color_0}{rgb}{1.000,0.400,0.000}
\definecolor{blk_color_1}{rgb}{1.000,0.000,0.365}
\definecolor{blk_color_2}{rgb}{1.000,0.000,0.031}
\definecolor{blk_color_3}{rgb}{0.000,1.000,1.000}
\definecolor{blk_color_4}{rgb}{1.000,0.569,0.000}
\definecolor{blk_color_5}{rgb}{0.667,1.000,0.000}
\definecolor{blk_color_6}{rgb}{0.000,0.031,1.000}
\definecolor{blk_color_7}{rgb}{1.000,0.667,0.000}
\definecolor{blk_color_8}{rgb}{0.498,1.000,0.000}
\definecolor{blk_color_9}{rgb}{0.000,0.933,1.000}
\definecolor{blk_color_10}{rgb}{0.000,1.000,0.667}
\definecolor{blk_color_11}{rgb}{0.933,1.000,0.000}
\definecolor{blk_color_12}{rgb}{0.000,0.933,1.000}
\definecolor{blk_color_13}{rgb}{0.769,1.000,0.000}
\definecolor{blk_color_14}{rgb}{0.000,0.498,1.000}
\definecolor{blk_color_15}{rgb}{1.000,0.169,0.000}
\definecolor{blk_color_16}{rgb}{0.000,1.000,0.733}
\definecolor{blk_color_17}{rgb}{0.769,1.000,0.000}
\definecolor{blk_color_18}{rgb}{0.000,1.000,0.400}
\definecolor{blk_color_19}{rgb}{1.000,0.000,0.667}

\begin{figure}
\centering
\resizebox{0.476\textwidth}{!}{
\begin{tabular}{@{}ccccccc@{}}
\cellcolor{blk_color_0} building+pole &
\cellcolor{blk_color_1} road+sidewalk &
\cellcolor{blk_color_2} road &
\cellcolor{blk_color_3} sidewalk+building &
\cellcolor{blk_color_4} building+traffic sign &
\cellcolor{blk_color_5} building+car &
\cellcolor{blk_color_6} \textcolor{white}{road+car} \\
\cellcolor{blk_color_7} building &
\cellcolor{blk_color_8} building+vegetation &
\cellcolor{blk_color_9} road+pole &
\cellcolor{blk_color_10} building+sky &
\cellcolor{blk_color_11} pole+car &
\cellcolor{blk_color_12} building+person &
\cellcolor{blk_color_13} pole+vegetation
\end{tabular}
}
\centering
\subfigure[Input image]{\label{fig:1a}\includegraphics[width=0.2363\textwidth]{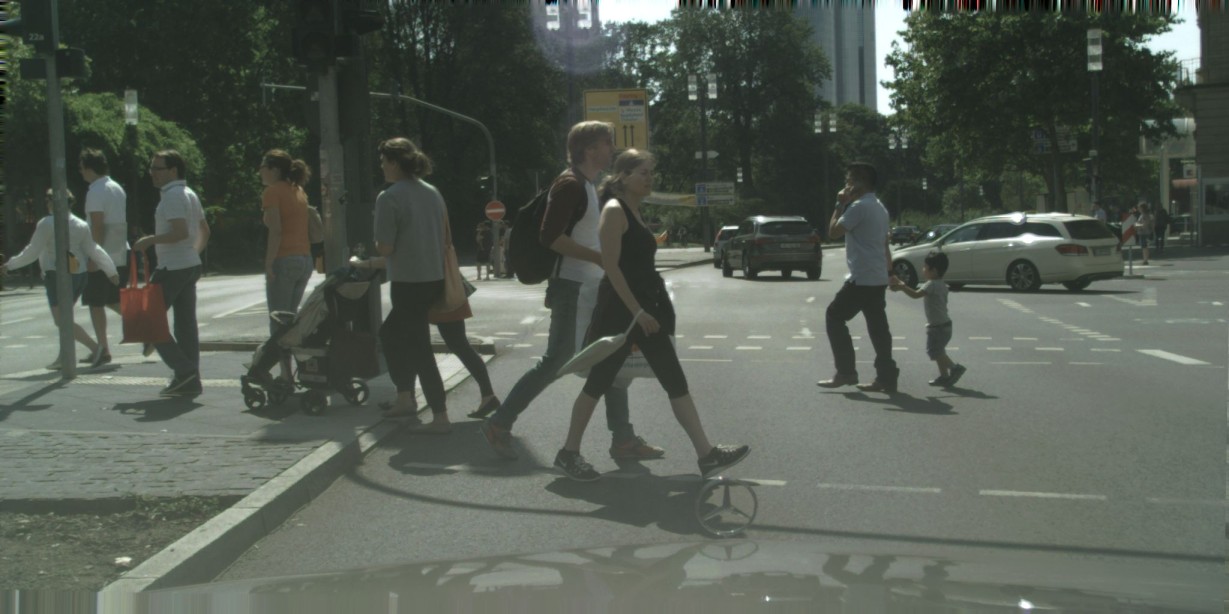}}
\subfigure[Ground truth]{\label{fig:1b}\includegraphics[width=0.2363\textwidth]{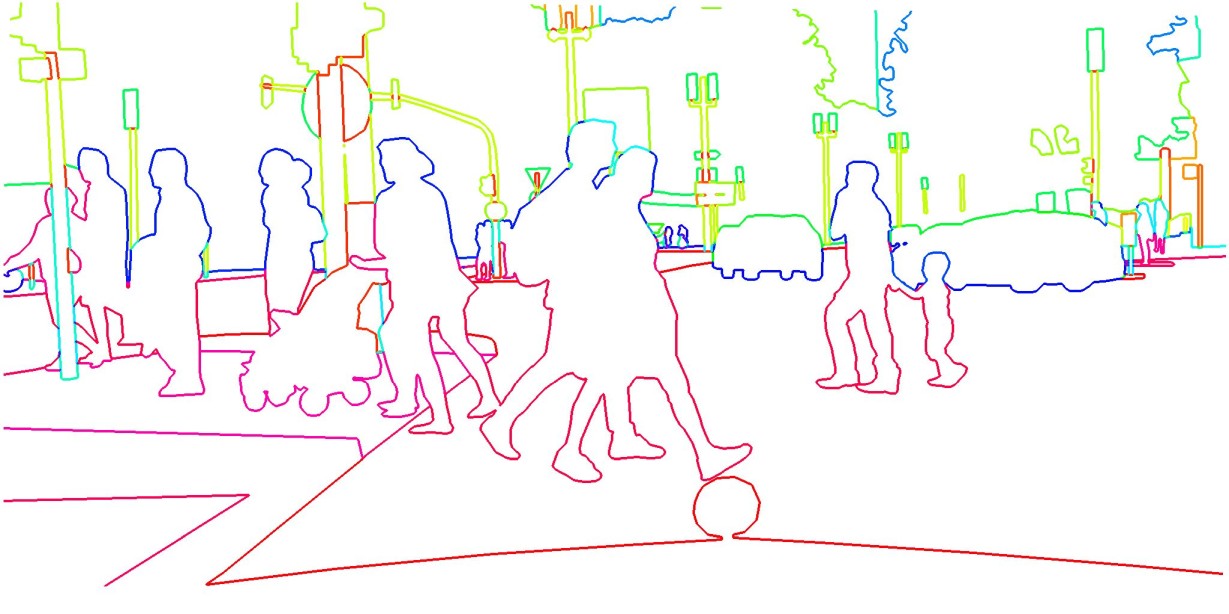}}
\subfigure[CASENet output]{\label{fig:1c}\includegraphics[width=0.478\textwidth]{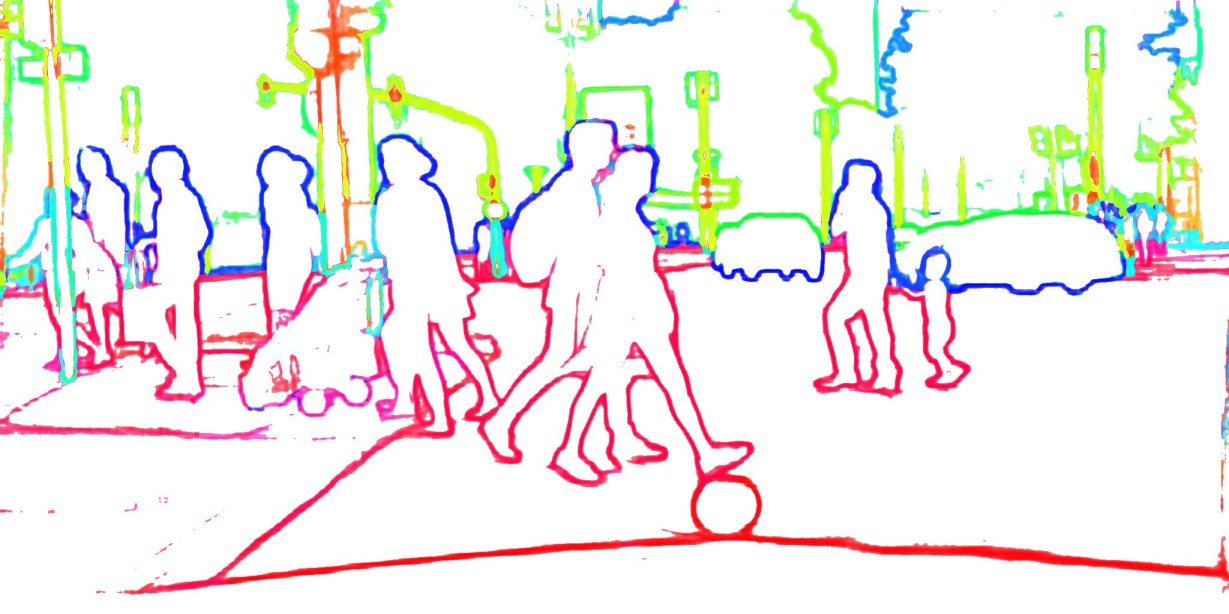}}
\caption{Edge detection and categorization with our approach. Given a street view image, our goal is to simultaneously detect the boundaries and assign each edge pixel with one or more semantic categories. (b) and (c) are color coded by HSV where hue and saturation together represent the composition and associated strengths of categories. Best viewed in color.}
\label{fig:1}
\end{figure}

\begin{figure*}
\centering
\includegraphics[width=1\textwidth]{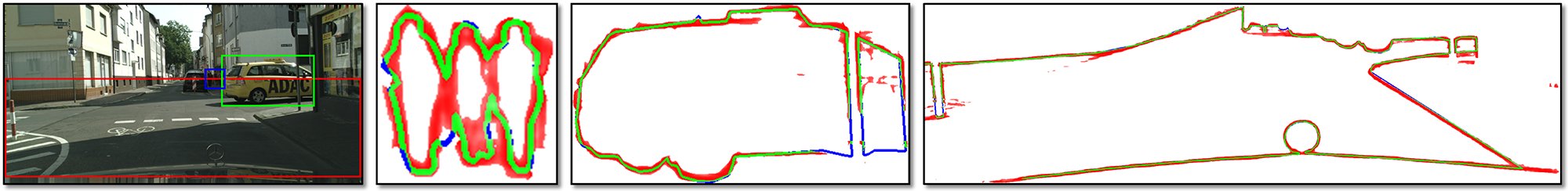}
\caption{An example of a test image and zoomed edge maps corresponding to bounding box regions. The visualized edge maps belong to the categories of person, car and road, respectively. Green and blue denote correctly detected and missed edge pixels.}
\label{fig:2}
\end{figure*}

Figure~\ref{fig:1} shows an image of a road scene from Cityscapes dataset~\cite{Cordts2016Cityscapes} with several object categories such as building, ground, sky, and car. In particular, we study the problem of simultaneously detecting edge pixels and classifying them based on association to one or more of the object categories~\cite{prasad2006learning,Hariharan2011}. For example, an edge pixel lying on the contour separating building and pole can be associated with both of these object categories. In Figure~\ref{fig:1}, we visualize the boundaries and list the colors of typical category combinations such as ``building+pole'' and ``road+sidewalk''. In our problem, every edge pixel is denoted by a vector whose individual elements denote the strength of pixel's association with different semantic classes. While most edge pixels will be associated with only two object categories, in the case of junctions~\cite{Maire2014} one may expect the edge pixel to be associated with three or even more. We therefore do not restrict the number of object categories a pixel can be associated with, and formulate our task as a multi-label learning problem. In this paper, we propose CASENet, a deep network able to detect category-aware semantic edges.  Given $K$ defined semantic categories, the network essentially produces $K$ separate edge maps where each map indicates the edge probability of a certain category. An example of separately visualized edge maps on a test image is given in Figure~\ref{fig:2}.

The problem of edge detection has been shown to be useful for a number of computer vision tasks such as segmentation~\cite{Arbelaez2011,yu2015generalized,Chen2016,Bertasius2015_hfl,bertasius2016semantic}, object proposal~\cite{Bertasius2015_hfl}, 3d shape recovery~\cite{Karsch2013}, and 3d reconstruction~\cite{Shan2014}. By getting a better understanding of the edge classes and using them as prior or constraints, it is reasonable to expect some improvement in these tasks. With a little extrapolation, it is not difficult to see that a near-perfect semantic edge, without any additional information, can solve semantic segmentation, depth estimation~\cite{lineCurvedObjects,hoiem2005geometric}, image-based localization~\cite{Ramalingam2010}, and object detection~\cite{Ferrari2010}. We believe that it is important to improve the accuracy of semantic edge detection to a certain level for moving towards a holistic scene interpretation.

Early work tends to treat edge information as low-level cues to enhance other applications. However, the availability of large training data and the progress in deep learning methods have allowed one to make significant progress for the edge detection problem in the last few years. In particular, there have been newer data sets~\cite{Hariharan2011}. The availability of large-scale semantic segmentation data sets~\cite{Cordts2016Cityscapes} can also be easily processed to obtain semantic edge data set as these two problems can be seen as dual problems.

\subsection{Related works}

The definition of boundary or edge detection has evolved over time from low-level to high-level features:
simple edge filters~\cite{Canny1986}, depth edges~\cite{gupta2013perceptual}, object boundaries~\cite{Martin2004}, and semantic contours~\cite{Hariharan2011}. In some sense, the evolution of edge detection algorithms captures the progress in computer vision from simple convolutional filters such as Sobel~\cite{Kittler1983} or Canny~\cite{Canny1986} to fully developed deep neural networks.

\paragraph{Low-level edges}
Early edge detection methods used simple convolutional filters such as Sobel~\cite{Kittler1983} or Canny~\cite{Canny1986}.

\paragraph{Depth edges}
Some previous work focuses on labeling contours into convex, concave, and occluding ones from synthetic line drawings~\cite{lineCurvedObjects} and real world images under restricted settings~\cite{gupta2013perceptual,Fouhey14c}. Indoor layout estimation can also be seen as the identification of concave boundaries (lines folding walls, ceilings, and ground)~\cite{Hedau2009}. By recovering occluding boundaries~\cite{hoiem-ROB}, it was shown that the depth ordering of different layers in the scene can be obtained.

\paragraph{Perceptual edges} A wide variety of methods are driven towards the extraction of perceptual boundaries~\cite{Martin2004}. Dollar \textit{et al.}~\cite{Dollar2006} used boosted decision trees on different patches to extract edge maps. Lim \textit{et al.}~\cite{Lim2013} computed sketch tokens which are object boundary patches using random forests. Several other edge detection methods include statistical edges~\cite{Konishi2003}, multi-scale boundary detection~\cite{Ren2008}, and point-wise mutual information (PMI) detector~\cite{isola14crisp}. More recently, Dollar and Zitnick~\cite{Dollar2013} proposed a realtime fast edge detection method using structured random forests. Latest methods~\cite{Xie2015,Bertasius2015_hfl,yang2016object,kokkinos2016pushing} using deep neural networks have pushed the detection performance to state-of-the-art.

\paragraph{Semantic edges}
The origin of semantic edge detection can be possibly pinpointed to ~\cite{prasad2006learning}. As a high level task, it has also been used implicitly or explicitly in many problems related to segmentation~\cite{Wang2007} and reconstruction~\cite{hoiem2005geometric}. In some sense, all semantic segmentation methods~\cite{farabet2013learning,girshick2014rich,pinheiro2014,sharma2014recursive,liu2015layered,Long2015,Chen2015,vemulapalli2016gaussian,Cordts2016Cityscapes} can be loosely seen as semantic edge detection since one can easily obtain edges, although not necessarily an accurate one, from the segmentation results. There are papers that specifically formulate the problem statement as binary or category-aware semantic edge detection~\cite{prasad2006learning,Ferrari2010,Hariharan2011,Bertasius2015_hfl,bertasius2016semantic,yang2016object,maninis2016convolutional,khoreva2016weakly}. Hariharan \textit{et al.}~\cite{Hariharan2011} introduced the Semantic Boundaries Dataset (SBD) and proposed inverse detector which combines both bottom-up edge and top-down detector information to detect category-aware semantic edges. HFL~\cite{Bertasius2015_hfl} first uses VGG~\cite{Simonyan2015} to locate binary semantic edges and then uses deep semantic segmentation networks such as FCN~\cite{Long2015} and DeepLab~\cite{Chen2015} to obtain category labels. The framework, however, is not end-to-end trainable due to the separated prediction process.

\paragraph{DNNs for edge detection}
Deep neural networks recently became popular for edge detection. Related work includes SCT based on sparse coding~\cite{Maire2014}, $N^4$ fields~\cite{ganin2014n}, deep contour~\cite{Shen2015}, deep edge~\cite{Bertasius2015_de}, and CSCNN~\cite{Hwang2015}. One notable method is the holistically-nested edge detection (HED)~\cite{Xie2015} which trains and predicts edges in an image-to-image fashion and performs end-to-end training.

\subsection{Contributions}
Our work is related to HED in adopting a nested architecture but we extend the work to the more difficult category-aware semantic edge detection problem. Our main contributions in this paper are summarized below:
\begin{itemize}
\item To address edge categorization, we propose a multi-label learning framework which allows improved edge learning than traditional multi-class framework.
\item We propose a novel nested architecture without deep supervision on ResNet~\cite{He2016}, where bottom features are only used to augment top classifications. We show that deep supervision may not be beneficial in our problem.
\item We outperform previous state-of-the-art methods by significant margins on SBD and Cityscapes datasets.
\end{itemize}

\section{Problem Formulation}
Given an input image, our goal is to compute the semantic edge maps corresponding to pre-defined categories. More formally, for an input image $\mathbf{I}$ and $K$ defined semantic categories, we are interested in obtaining $K$ edge maps $\{\mathbf{Y}_1,\cdots,\mathbf{Y}_K\}$, each having the same size as $\mathbf{I}$. With a network having the parameters $\mathbf{W}$, we denote $\mathbf{Y}_k(\mathbf{p}|\mathbf{I},\mathbf{W}) \in [0,1]$ as the network output indicating the computed edge probability on the $k$-th semantic category at pixel $\mathbf{p}$.

\subsection{Multi-label loss function}
\label{subsec:loss}
Possibly driven by the multi-class nature of semantic segmentation, several related works on category-aware semantic edge detection have more or less looked into the problem from the multi-class learning perspective. Our intuition is that this problem by nature should allow one pixel belonging to multiple categories simultaneously, and should be addressed by a multi-label learning framework.

We therefore propose a multi-label loss. Suppose each image $\mathbf{I}$ has a set of label images $\{\mathbf{\bar{Y}}_1,\cdots,\mathbf{\bar{Y}}_K\}$, where $\mathbf{\bar{Y}}_k$ is a binary image indicating the ground truth of the $k$-th class semantic edge. The multi-label loss is formulated as:
\begin{align}\label{eq:loss}
\mathcal{L}(\mathbf{W})= & \sum_{k} \mathcal{L}_k(\mathbf{W}) \\ \nonumber
	= & \sum_{k} \sum_{\mathbf{p}} \{ - \beta \mathbf{\bar{Y}}_k(\mathbf{p})\log \mathbf{Y}_k(\mathbf{p}|\mathbf{I};\mathbf{W}) \\ \nonumber
	  & - (1-\beta) (1-\mathbf{\bar{Y}}_k(\mathbf{p}))\log(1-\mathbf{Y}_k(\mathbf{p}|\mathbf{I};\mathbf{W}))\},
\end{align}
\noindent where $\beta$ is the percentage of non-edge pixels in the image to account for skewness of sample numbers, similar to~\cite{Xie2015}.

\section{Network Architecture}
\begin{figure*}[t]
\begin{minipage}{\textwidth}
\centering
\subfigure[Basic Network]{\label{fig:naive}\includegraphics[height=.19\textheight]{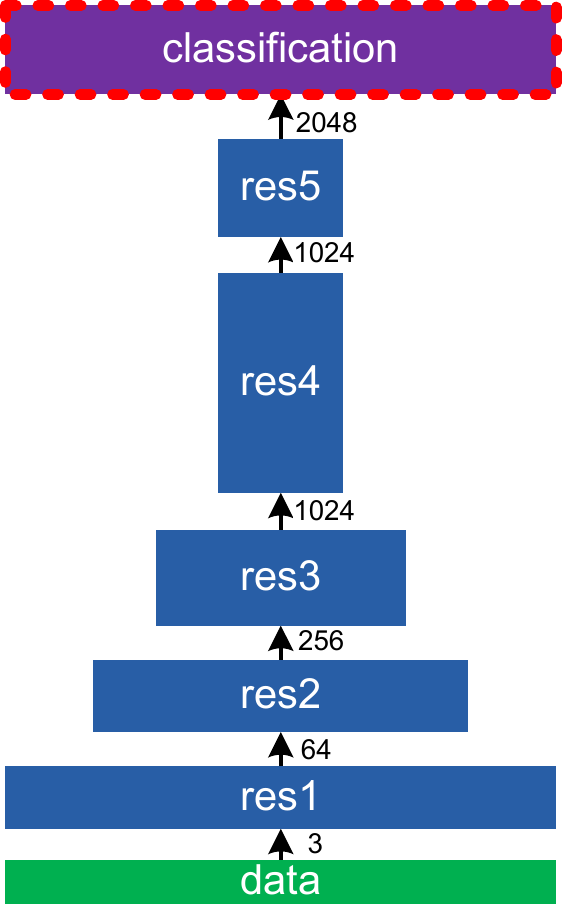}}\qquad
\subfigure[Deeply Supervised Network (DSN)]{\label{fig:dsn}\includegraphics[height=.19\textheight]{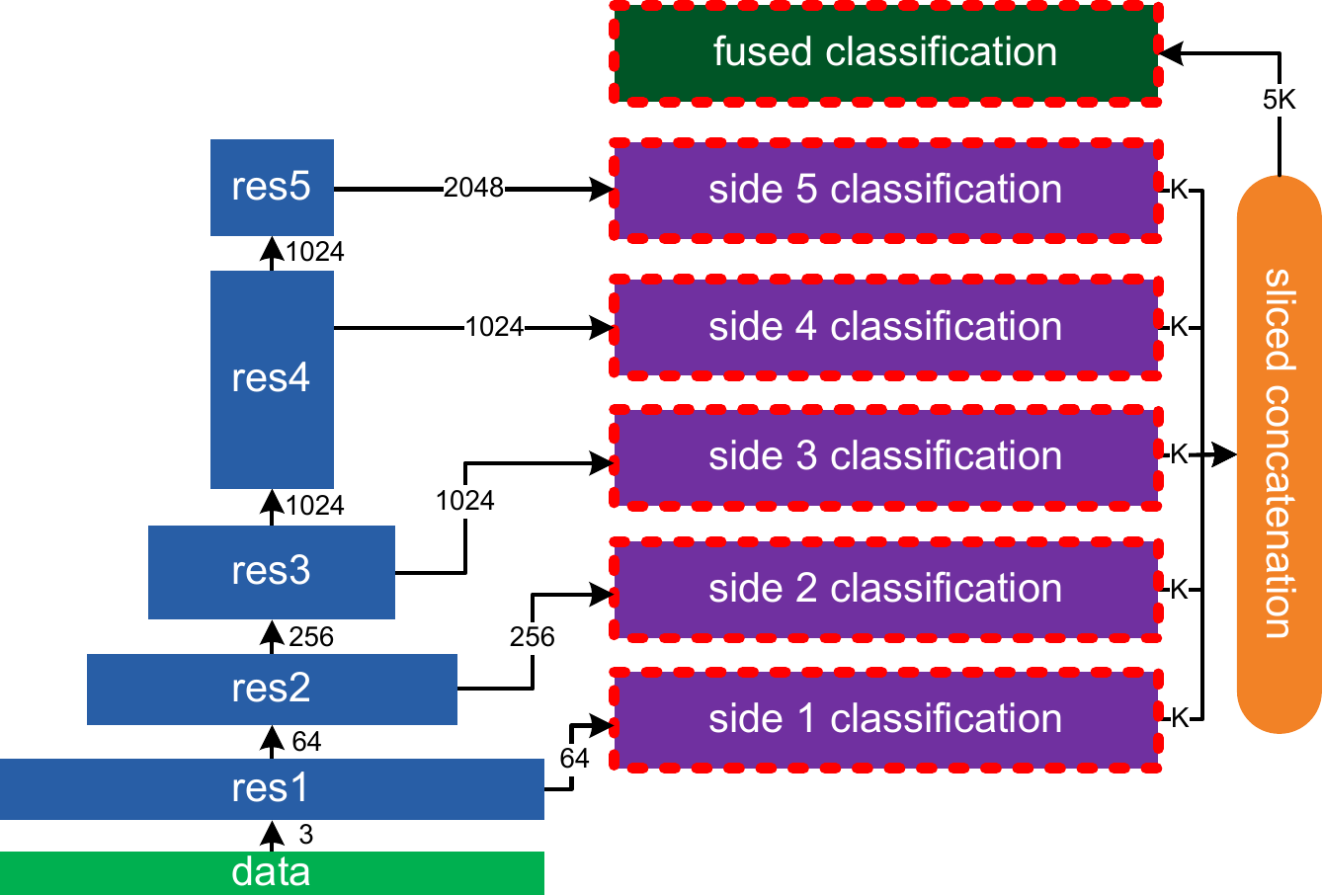}}\qquad
\subfigure[CASENet]{\label{fig:dsbd}\includegraphics[height=.19\textheight]{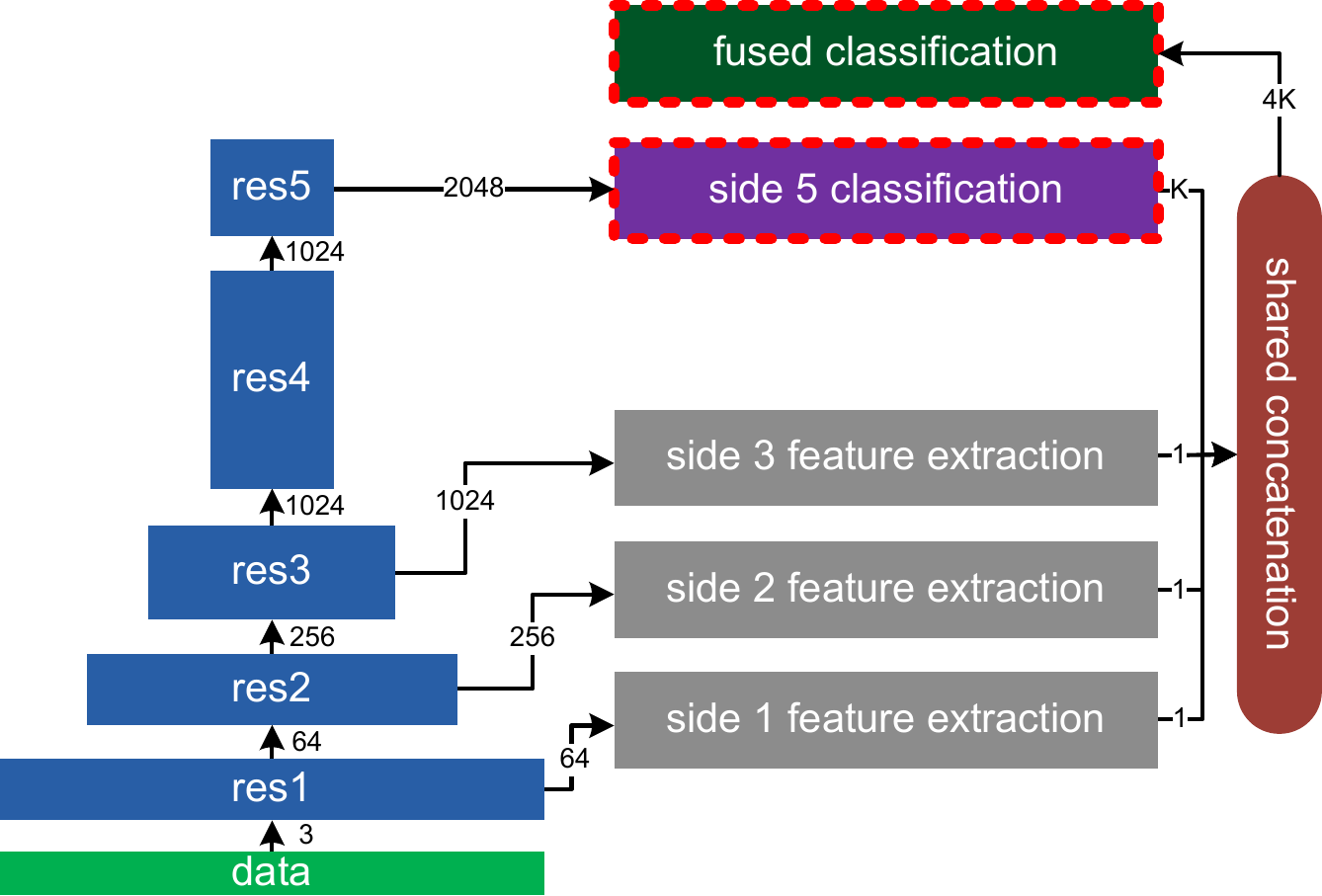}}
\end{minipage}
\begin{minipage}{0.20\textwidth}
\centering
\subfigure[Classification]{\label{fig:module_class}\includegraphics[height=.06\textheight]{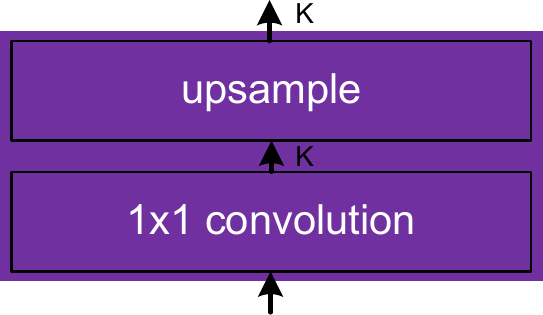}} \\
\subfigure[Side Feature]{\label{fig:modele_detect}\includegraphics[height=.06\textheight]{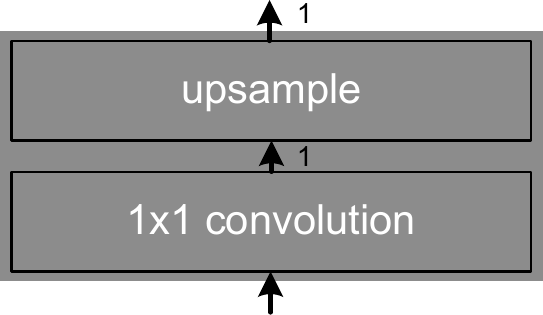}}
\end{minipage}
\begin{minipage}{0.81\textwidth}
\centering
\subfigure[Fused Classification]{\label{fig:fused_class}\includegraphics[height=.13\textheight]{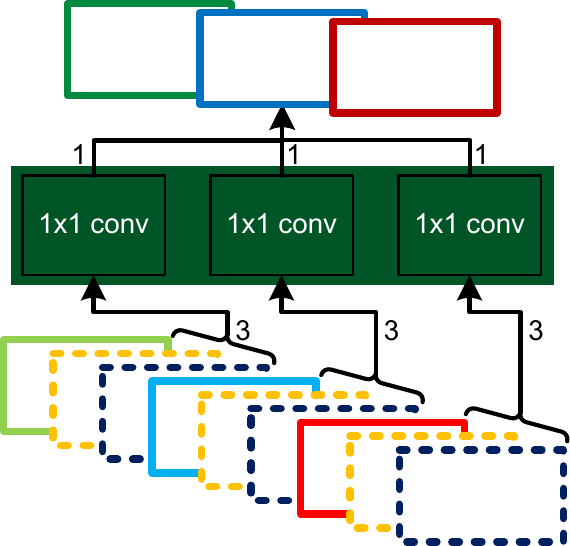}}\quad
\subfigure[Sliced Concatenation]{\label{fig:slice_concat}\includegraphics[height=.13\textheight]{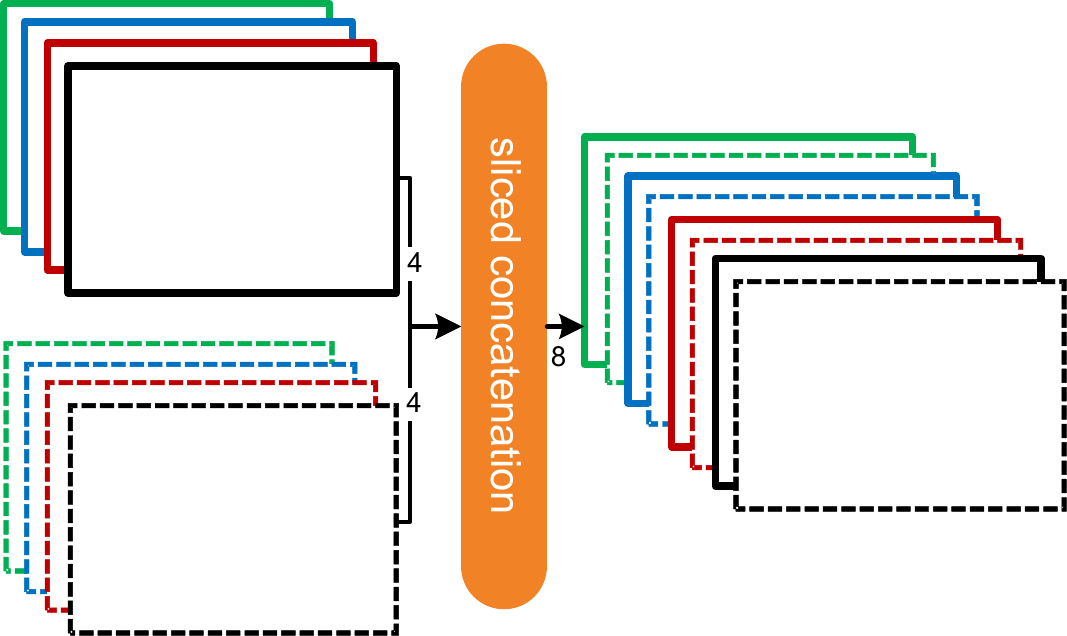}}\quad
\subfigure[Shared Concatenation]{\label{fig:shared_concat}\includegraphics[height=.13\textheight]{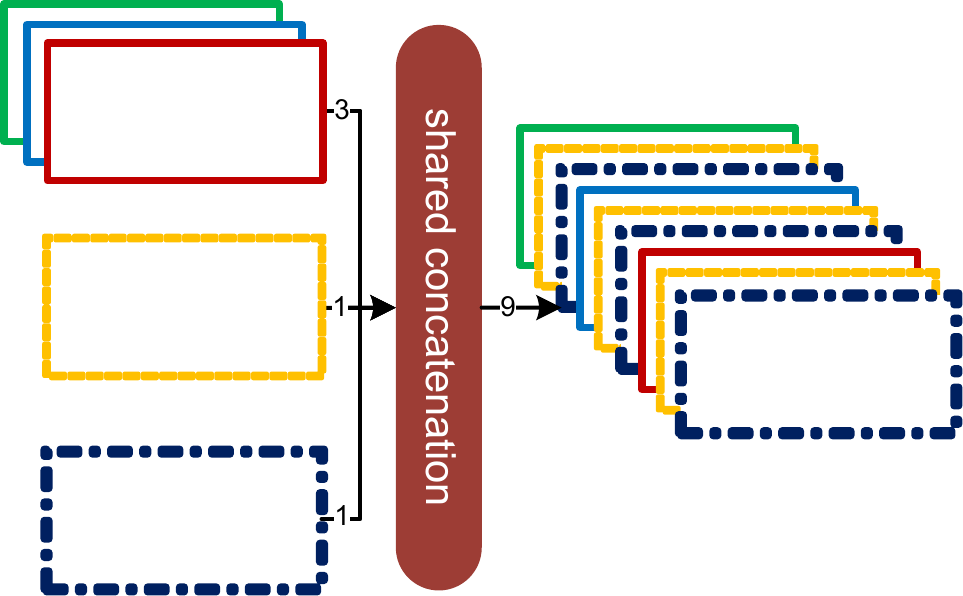}}
\end{minipage}
\caption{Three CNN architectures designed in this paper are shown in (a)-(c). A solid rectangle represents a composite block of CNN layers. Any decrease of its width indicates a drop of spatial resolution of this block's output feature map by a factor of 2. A number besides an arrow indicates the number of channels of the block's output features. A blue solid rectangle is a stack of ResNet blocks. A purple solid rectangle is our classification module. A dotted red outline indicates that block's output is supervised by our loss function in equation \ref{eq:loss}. A gray solid rectangle is our side feature extraction module. A dark green solid rectangle is our fused classification module performing $K$-grouped $1 \times 1$ convolution. (d)-(h) depicts more details of various modules used in (a)-(c), where outlined rectangles illustrate input and output feature maps. Best viewed in color.}
\label{fig:main}
\end{figure*}

We propose CASENet, an end-to-end trainable convolutional neural network (CNN) architecture (shown in Fig.~\ref{fig:dsbd}) to address category-aware semantic edge detection. Before describing CASENet, we first propose two alternative network architectures which one may come up with straightforwardly given the abundant previous literature on edge detection and semantic segmentation. Although both architectures can also address our task, we will analyze  issues associated with them, and address these issues by proposing the CASENet architecture.

\subsection{Base network}
We address the edge detection problem under the fully convolutional network framework. We adopt ResNet-101 by removing the original average pooling and fully connected layer, and keep the bottom convolution blocks. We further modify the base network in order to better preserve low-level edge information. We change the stride of the first and fifth convolution blocks (``res1'' and ``res5'' in Fig.~\ref{fig:main}) in ResNet-101 from 2 to 1.  We also introduce dilation factors to subsequent convolution layers to maintain the same receptive field sizes as the original ResNet, similar to~\cite{He2016}.

\subsection{Basic architecture}
\label{subsec:basic_network}
A very natural architecture one may come up with is the Basic architecture shown in Fig.~\ref{fig:naive}. On top of the base network, we add a classification module (Fig.~\ref{fig:module_class}) as a $1 \times 1$ convolution layer, followed by bilinear up-sampling (implemented by a $K$-grouped deconvolution layer) to produce a set of $K$ activation maps $\{\mathbf{A}_1, \cdots, \mathbf{A}_K\}$, each having the same size as the image. We then model the probability of a pixel belonging to the $k$-th class edge using the sigmoid unit given by $\mathbf{Y}_k(\mathbf{p})= \sigma(\mathbf{A}_k(\mathbf{p}))$, which is presented in the Eq. (\ref{eq:loss}). Note that $\mathbf{Y}_k(\mathbf{p})$ is not mutually exclusive.

\subsection{Deeply supervised architecture}
\label{subsec:deeply_supervised_network}
One of the distinguishing features of the holistically-nested edge detection (HED) network~\cite{Xie2015} is the nested architecture with deep supervision~\cite{Lee2015}. The basic idea is to impose losses to bottom convolution sides besides the top network loss. In addition, a fused edge map is obtained by supervising the linear combination of side activations.

Note that HED only performs binary edge detection. We extended this architecture to handle $K$ channels for side outputs and $K$ channels for the final output. We refer to this as deeply supervised network (DSN), as depicted in Fig.~\ref{fig:dsn}. In this network, we connect an above-mentioned classification module to the output of each stack of residual blocks, producing 5 side classification activation maps $\{\mathbf{A}^{(1)},\dots,\mathbf{A}^{(5)}\}$, where each of them has $K$-channels. We then fuse these 5 activation maps through a sliced concatenation layer (the color denotes the channel index in Fig.~\ref{fig:slice_concat}) to produce a $5K$-channel activation map:
\begin{equation}
 \mathbf{A}^{f} = \{\mathbf{A}^{(1)}_1,\dots,\mathbf{A}^{(5)}_1,\mathbf{A}^{(1)}_2,\dots,\mathbf{A}^{(5)}_2,\dots,\mathbf{A}^{(5)}_K\}
\end{equation}
$\mathbf{A}^{f}$ is fed into our fused classification layer which performs $K$-grouped $1 \times 1$ convolution (Fig.~\ref{fig:fused_class}) to produce a $K$-channel activation map $\mathbf{A}^{(6)}$. Finally, 6 loss functions are computed on $\{\mathbf{A}^{(1)},\dots,\mathbf{A}^{(6)}\}$ using the Equation~\ref{eq:loss} to provide deep supervision to this network.

Note that the reason we perform sliced concatenation in conjunction with grouped convolution instead of the corresponding conventional operations is as follows. Since the 5 side activations are supervised, we implicitly constrain each channel of those side activations to carry information that is most relevant to the corresponding class.

With sliced concatenation and grouped convolution, the fused activation for a pixel $\mathbf{p}$ is given by:
\begin{equation}
\mathbf{A}^{(6)}_k(\mathbf{p}) = W_k^T[\mathbf{A}^{(1)}_k(\mathbf{p})^T, \cdots, \mathbf{A}^{(5)}_k(\mathbf{p})^T]
\end{equation}
This essentially integrates corresponding class-specific activations from different scales as the finally fused activations. Our experiments empirically support this design choice.

\subsection{CASENet architecture}
\label{subsec:casenet}
Upon reviewing the Basic and DSN architectures, we notice several potential associated issues in the category-aware semantic edge detection task:

First, the receptive field of the bottom side is limited. As a result it may be unreasonable to require the network to perform semantic classification at an early stage, given that context information plays an important role in semantic classification. We believe that semantic classification should rather happen on top where features are encoded with high-level information.

Second, bottom side features are helpful in augmenting top classifications, suppressing non-edge pixels and providing detailed edge localization and structure information. Hence, they should be taken into account in edge detection.

Our proposed CASENet architecture (Fig.~\ref{fig:dsbd}) is motivated by addressing the above issues. The network adopts a nested architecture which to some extent shares similarity to DSN but also contains several key improvements. We summarize these improvements below:

\begin{enumerate}
\item Replace the classification modules at bottom sides to the feature extraction modules.
\item Put the classification module and impose supervision only at the top of the network.
\item Perform shared concatenation (Fig.~\ref{fig:shared_concat}) instead of sliced concatenation.
\end{enumerate}

The difference between side feature extraction and side classification is that the former only outputs a single channel feature map $\mathbf{F}^{(j)}$ rather than $K$ class activations. The shared concatenation replicates the bottom features $\mathbf{F} = \{\mathbf{F}^{(1)},\mathbf{F}^{(2)},\mathbf{F}^{(3)}\}$ from Side-1-3 to separately concatenate with each of the $K$ top activations:
\begin{equation}
\mathbf{A}^{f}=
\{\mathbf{F}, \mathbf{A}^{(5)}_1, \mathbf{F}, \mathbf{A}^{(5)}_2, \mathbf{F}, \mathbf{A}^{(5)}_3,\dots, \mathbf{F}, \mathbf{A}^{(5)}_K\}.
\end{equation}
The resulting concatenated activation map is again fed into the fused classification layer with $K$-grouped convolution to produce a $K$-channel activation map $\mathbf{A}^{(6)}$.

In general, CASENet can be thought of as a joint edge detection and classification network by letting lower level features participating and augmenting higher level semantic classification through a skip-layer architecture.

\section{Experiments}
In this paper, we compare CASENet\footnote{Source code available at: \url{http://www.merl.com/research/license\#CASENet}.} with previous state-of-the-art methods, including InvDet~\cite{Hariharan2011}, HFL~\cite{Bertasius2015_hfl}, weakly supervised object boundaries~\cite{khoreva2016weakly}, as well as several baseline network architectures.

\subsection{Datasets}
We evaluate the methods on SBD~\cite{Hariharan2011}, a standard dataset for benchmarking semantic edge detection. Besides SBD, we also extend our evaluation to Cityscapes~\cite{Cordts2016Cityscapes}, a popular semantic segmentation dataset with pixel-level high quality annotations and challenging street view scenarios. To the best of our knowledge, our paper is the first work to formally report semantic edge detection results on this dataset.

\paragraph{SBD} The dataset consists of 11355 images from the PASCAL VOC2011~\cite{pascal-voc-2011} trainval set, divided into 8498 training and 2857 test images\footnote{There has been a clean up of the dataset with a slightly changed image number. We also report the accordingly updated InvDet results.}. This dataset has semantic boundaries labeled with one of 20 Pascal VOC classes.

\paragraph{Cityscapes} The dataset contains 5000 images divided into 2975 training, 500 validation and 1525 test images. Since the labels of test images are currently not available, we treat the validation images as test set in our experiment.

\subsection{Evaluation protocol}
On both SBD and Cityscapes, the edge detection accuracy for each class is evaluated using the official benchmark code and ground truth from~\cite{Hariharan2011}. We keep all settings and parameters as default, and report the maximum F-measure (MF) at optimal dataset scale (ODS), and average precision (AP) for each class. Note that for Citiscapes, we follow~\cite{Hariharan2011} exactly to generate ground truth boundaries with single pixel width for evaluation, and reduce the sizes of both ground truth and predicted edge maps to half along each dimension considering the speed of evaluation.

\subsection{Implementation details}
We trained and tested CASENet, HED~\cite{Xie2015}, and the proposed baseline architectures using the \textit{Caffe} library~\cite{jia2014caffe}.

\paragraph{Training labels}
Considering the misalignment between human annotations and true edges, and the label ambiguity of pixels near boundaries, we generate slightly thicker ground truth edges for network training. This can be done by looking into neighbors of a pixel and seeking any difference in segmentation labels. The pixel is regarded as an edge pixel if such difference exists. In our paper, we set the maximum range of neighborhood to be $2$. Under the multi-label framework, edges from different classes may overlap.

\paragraph{Baselines}
Since several main comparing methods such as HFL and HED use VGG or VGG based architectures for edge detection and categorization, we also adopt the CASENet and other baseline architectures on VGG (denoted as CASENet-VGG). In particular, we remove the max pooling layers after conv4, and keep the resolutions of conv5, fc6 and fc7 the same as conv4 (1/8 of input). Similar to~\cite{Chen2015}, both fc6 and fc7 are treated as convolution layers with $3\times3$ and $1\times1$ convolution and dimensions set to 1024. Dilation factors of 2 and 4 are applied to conv5 and fc6.

To compare our multi-label framework with multi-class, we generate ground truth with non-overlapping edges of each class, reweight the softmax loss similar to our paper, and replace the top with a 21-class reweighted softmax loss.

\paragraph{Initialization}
In our experiment, we initialize the convolution blocks of ResNet/VGG in CASENet and all comparing baselines with models pre-trained on MS COCO~\cite{MS_COCO}.

\paragraph{Hyper-parameters}
We unify the hyper-parameters for all comparing methods with the same base network, and set most of them following HED. In particular, we perform SGD with iteration size of 10, and fix loss weight to be 1, momentum 0.9, and weight decay 0.0005. For methods with ResNet, we set the learning rate, step size, gamma and crop size to $1e-7$ / $5e-8$, $10000$ / $20000$, $0.1$ / $0.2$ and $352\times 352$ / $472\times 472$ respectively for SBD and Cityscapes. For VGG, the learning rate is set to $1e-8$ while others remain the same as ResNet on SBD. For baselines with softmax loss, the learning rate is set to 0.01 while other parameters remain the same. The iteration numbers on SBD and Cityscapes are empirically set to $22000$ and $40000$.

\paragraph{Data augmentation}
During training, we enable random mirroring and cropping on both SBD and Cityscapes. We additionally augment the SBD data by resizing each image with scaling factors \{0.5, 0.75, 1.0, 1.25, 1.5\}, while no such augmentation is performed on Cityscapes.

\subsection{Results on SBD}
Table~\ref{tb:main} shows the MF scores of different methods performing category-wise edge detection on SBD, where CASENet outperforms previous methods. Upon using the benchmark code from~\cite{Hariharan2011}, one thing we notice is that the recall scores of the curves are not monotonically increasing, mainly due to the fact that post-processing is taken after thresholding in measuring the precision and recall rates. This is reasonable since we have not taken any postprocessing operations on the obtained raw edge maps. We only report the MF on SBD since AP is not well defined under such situation. The readers may kindly refer to supplementary materials for class-wise precision recall curves.

\paragraph{Multi-label or multi-class?}
We compare the proposed multi-label loss with the reweighted softmax loss under the Basic architecture. One could see that using softmax leads to significant performance degradation on both VGG and ResNet, supporting our motivation in formulating the task as a multi-label learning problem, in contrast to the well accepted concept which addresses it in a multi-class way.

\paragraph{Is deep supervision necessary?}
We compare CASENet with baselines network architectures including Basic and DSN depicted in Fig.~\ref{fig:main}. The result empirically supports our intuition that deep supervision on bottom sides may not be necessary. In particular, CASENet wins frequently on per-class MF as well as the final mean MF score. Our observation is that the annotation quality to some extent influenced the network learning behavior and evaluation, leading to less performance distinctions across different methods. Such distinction becomes more obvious on Cityscapes.

\paragraph{Is top supervision necessary?}
One might further question the necessity of imposing supervision on Side-5 activation in CASENet. We use CASENet$^-$ to denote the same CASENet architecture without Side-5 supervision during training. The improvement upon adding Side-5 supervision indicates that a supervision on higher level side activation is helpful. Our intuition is that Side-5 supervision helps Side-5 focusing more on the classification of semantic classes with less influence from interacting with bottom layers.

\begin{figure}
\centering
\definecolor{blk_color_p}{rgb}{0.169,0.000,1.000}
\definecolor{blk_color_b}{rgb}{1.000,0.902,0.000}
\definecolor{blk_color_pb}{rgb}{0.000,1.000,0.533}
\setlength{\tabcolsep}{0mm}
\begin{tabular}{ccc}
\centering
\makecell{\includegraphics[width=0.32\columnwidth]{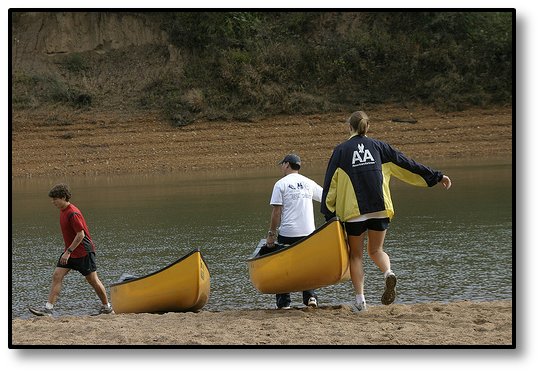}} &
\makecell{\includegraphics[width=0.32\columnwidth]{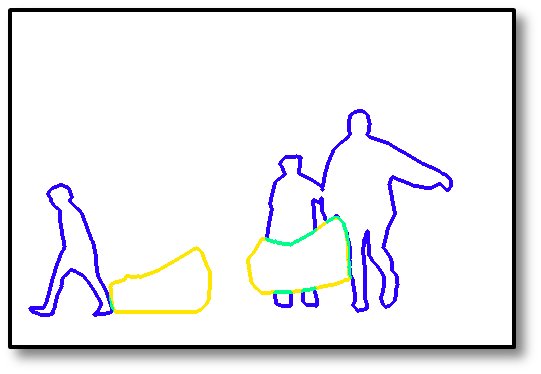}} &
\makecell{
\begin{tabular}{c}
\cellcolor{blk_color_p} \textcolor{white}{Person} \\
\cellcolor{blk_color_b} Boat \\
\cellcolor{blk_color_pb} Boat+Person \\
\end{tabular}
}
\\[-3pt]
\hline
\small DSN & \small CASENet$^-$ & \small CASENet \\ [-2pt]
\makecell{\includegraphics[width=0.32\columnwidth]{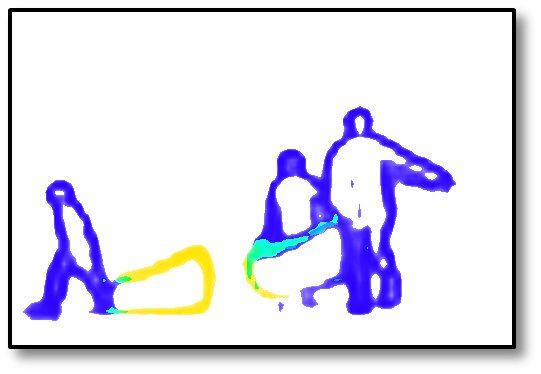}} &
\makecell{\includegraphics[width=0.32\columnwidth]{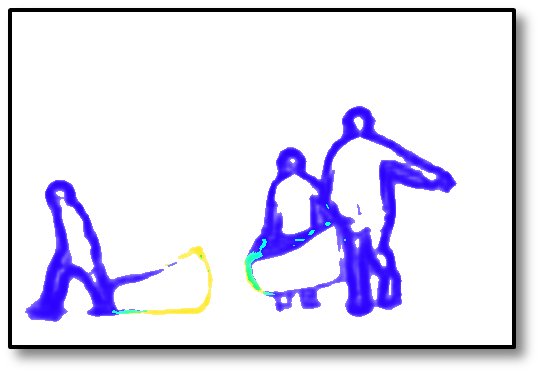}} &
\makecell{\includegraphics[width=0.32\columnwidth]{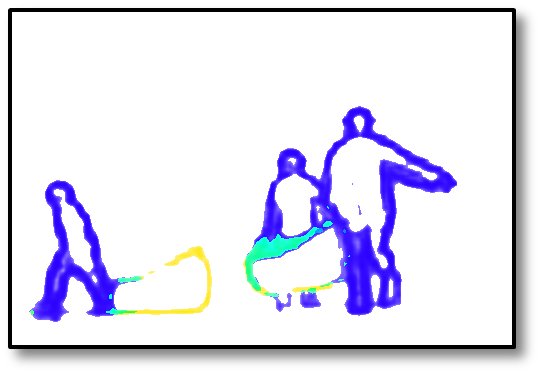}} \\ [-4pt]
\makecell{\includegraphics[width=0.32\columnwidth]{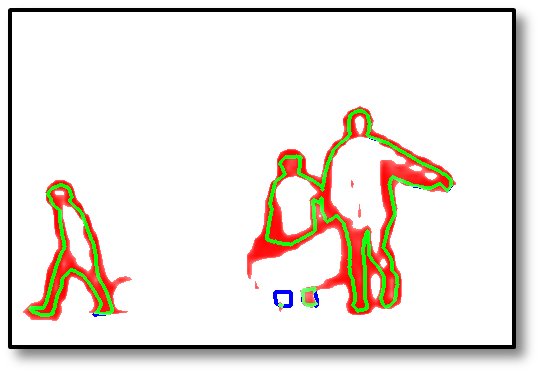}} &
\makecell{\includegraphics[width=0.32\columnwidth]{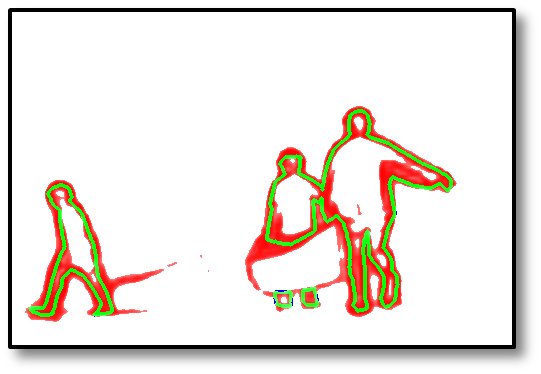}} &
\makecell{\includegraphics[width=0.32\columnwidth]{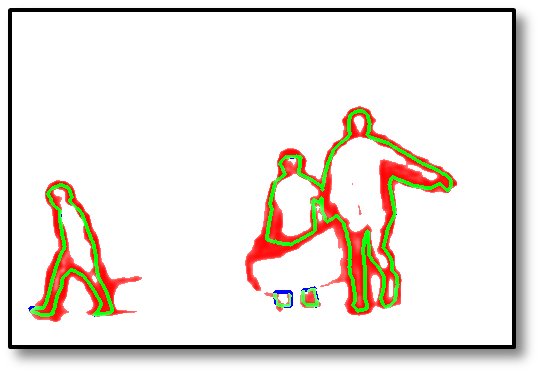}} \\ [-4pt]
\makecell{\includegraphics[width=0.32\columnwidth]{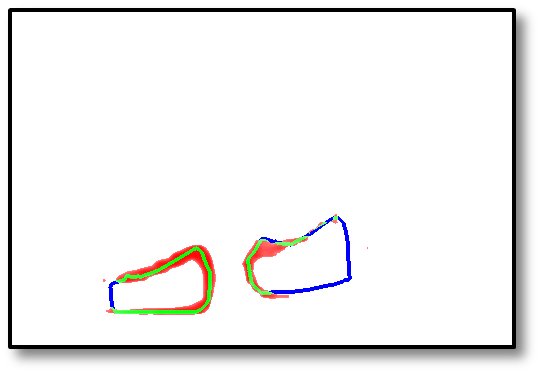}} &
\makecell{\includegraphics[width=0.32\columnwidth]{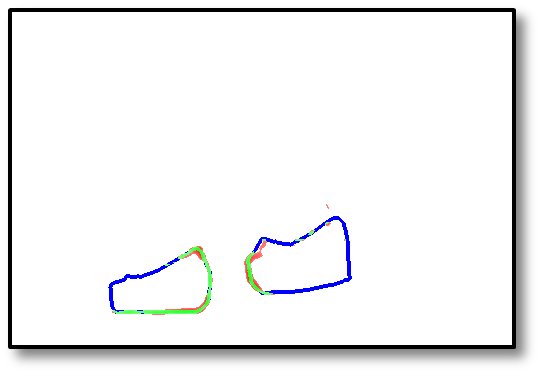}} &
\makecell{\includegraphics[width=0.32\columnwidth]{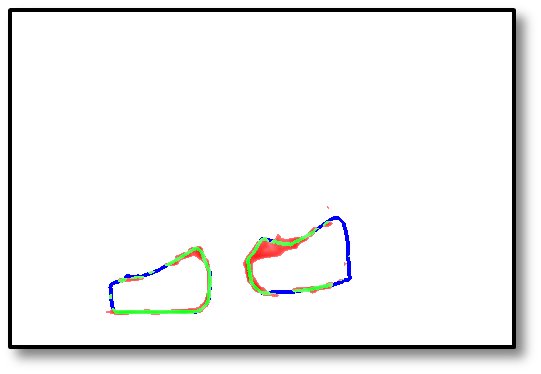}} \\
\end{tabular}
\hfill
\caption{\label{fig:compare}Example results on the SBD dataset. First row: Input and ground truth image and color codes of categories. Second row: Results of different edge classes, where the same color code is used as in Fig.~\ref{fig:1}. Third row: Results of person edge only. Last row: Results of boat edge only. Green, blue, red and white respectively denote true positive, false negative, false positive and true negative pixels, at the threshold of 0.5. Best viewed in color.}
\end{figure}

\begin{figure}
\centering
\setlength{\tabcolsep}{0mm}
\begin{tabular}{cccc}
\small DSN-Boat & \small DSN-Person & \small CASENet$^-$ & \small CASENet \\
\includegraphics[width=0.25\columnwidth]{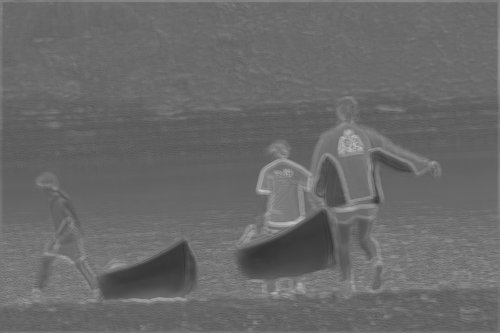}&		\includegraphics[width=0.25\columnwidth]{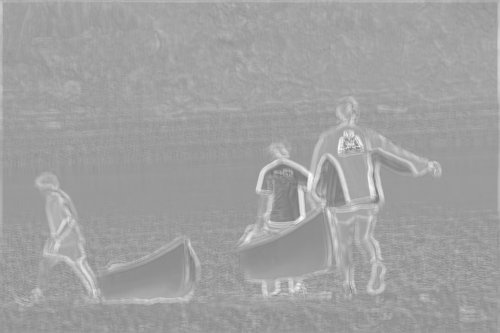}&		\includegraphics[width=0.25\columnwidth]{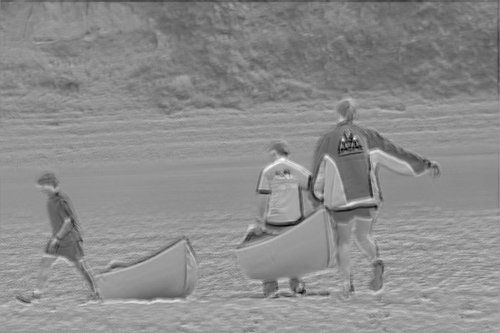}&		\includegraphics[width=0.25\columnwidth]{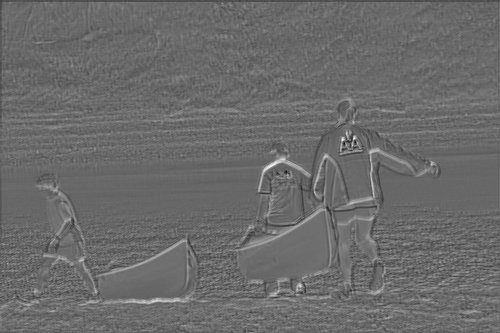}\\ [-4pt]
\multicolumn{4}{c}{\small Side1} \\ [-2pt]
\includegraphics[width=0.25\columnwidth]{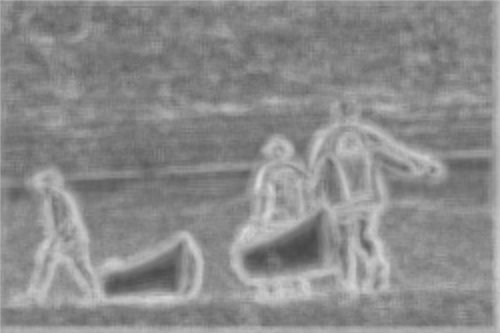}&	\includegraphics[width=0.25\columnwidth]{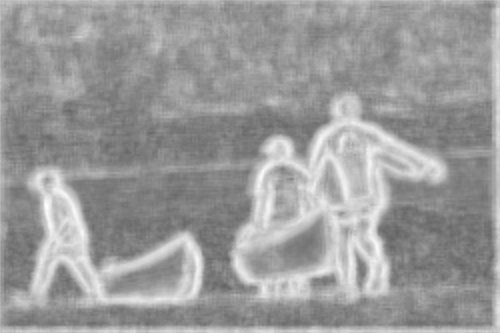}&	\includegraphics[width=0.25\columnwidth]{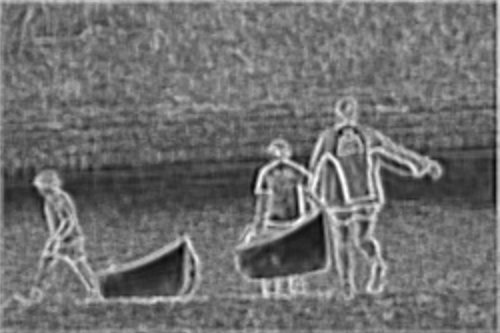}&	\includegraphics[width=0.25\columnwidth]{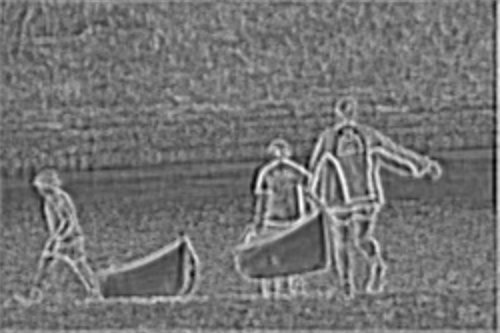}\\ [-4pt]
\multicolumn{4}{c}{\small Side2} \\ [-2pt]
\includegraphics[width=0.25\columnwidth]{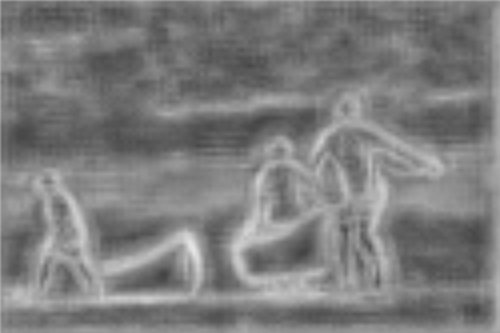}&	\includegraphics[width=0.25\columnwidth]{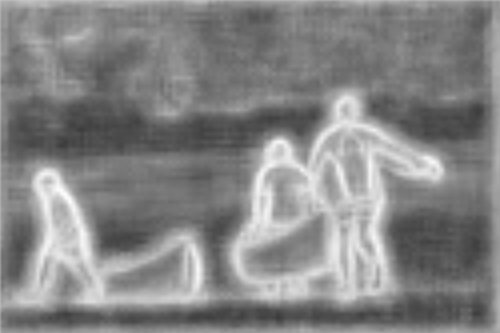}&	\includegraphics[width=0.25\columnwidth]{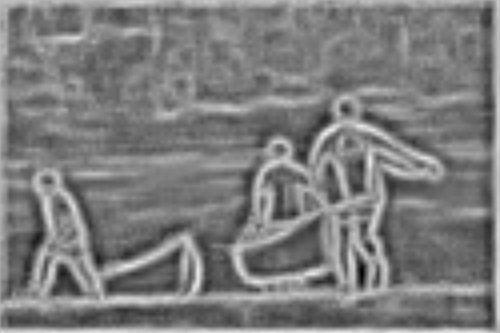}&	\includegraphics[width=0.25\columnwidth]{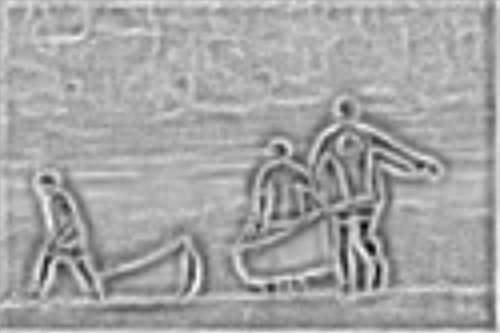}\\ [-4pt]
\multicolumn{4}{c}{\small Side3} \\ [-2pt]
\includegraphics[width=0.25\columnwidth]{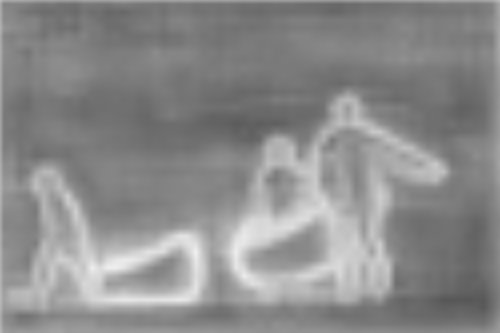}&	\includegraphics[width=0.25\columnwidth]{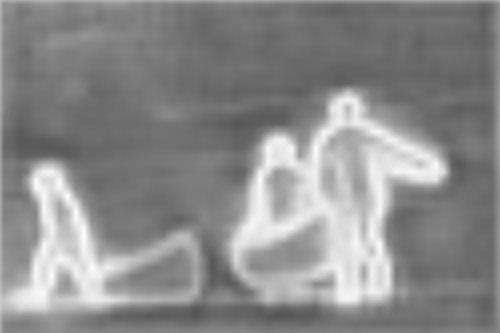}&	\includegraphics[width=0.25\columnwidth]{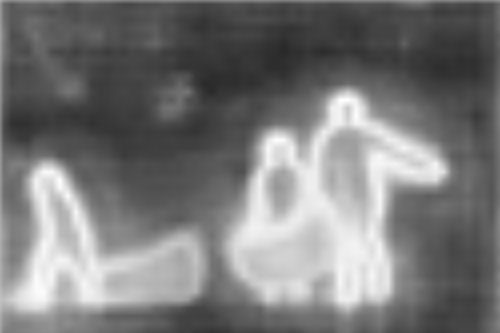}&	\includegraphics[width=0.25\columnwidth]{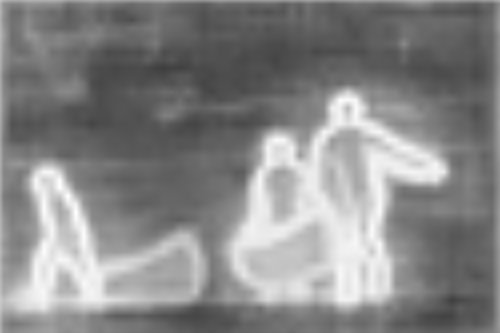}\\ [-4pt]
\multicolumn{2}{c}{\small Side4} & \multicolumn{2}{c}{\small Side5-Person} \\ [-2pt]
\includegraphics[width=0.25\columnwidth]{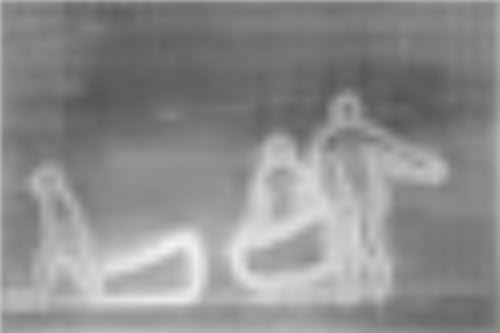}&	\includegraphics[width=0.25\columnwidth]{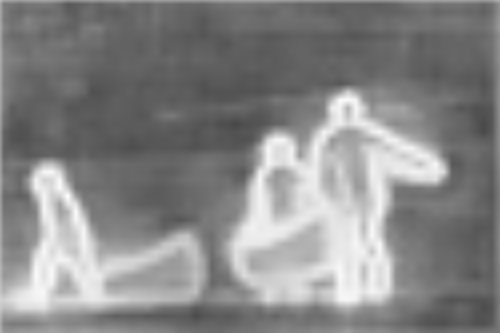}&	\includegraphics[width=0.25\columnwidth]{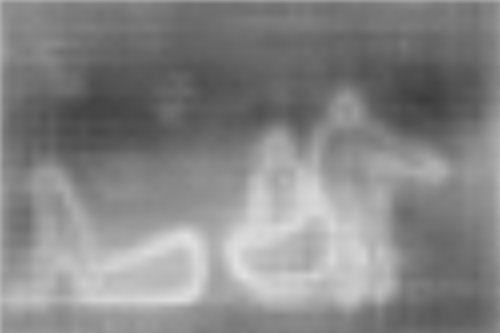}&	\includegraphics[width=0.25\columnwidth]{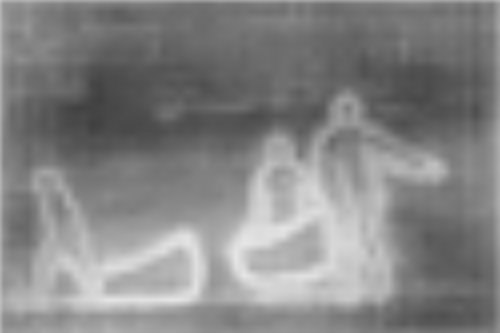}\\ [-4pt]
\multicolumn{2}{c}{\small Side5} & \multicolumn{2}{c}{\small Side5-Boat} \\
\end{tabular}
\hfill
\caption{Side activations on the input image of Fig.~\ref{fig:compare}. The first two columns show the DSN's side classification activations corresponding to the class of Boat and Person, respectively. The last two columns show the side features and classification activations for CASENet$^-$ and CASENet, respectively. Note that the pixel value range of each image is normalized to [0,255] individually inside its corresponding side activation outputs for visualization.\label{fig:sides}}
\end{figure}

\paragraph{Visualizing side activations}
We visualize the results of CASENet, CASENet$^-$ and DSN on a test image in Fig.~\ref{fig:compare}. Overall, CASENet achieves better detection compared to the other two. We further show the side activations of this testing example in Fig.~\ref{fig:sides}, from which one can see that the activations of DSN on Side-1, Side-2 and Side-3 are more blurry than CASENet features. This may be caused by imposing classification requirements on those layers, which seems a bit aggressive given limited receptive field and information and may caused performance degradation. Also one may notice the differences in ``Side5-Person'' and ``Side5-Boat'' between CASENet$^-$ and CASENet, where CASENet's activations overall contain sharper edges, again showing the benefit of Side-5 supervision.

\paragraph{From ResNet to VGG}
CASENet-VGG in Table~\ref{tb:main} shows comparable performance to HFL-FC8. HFL-CRF performs slightly better with the help of CRF postprocessing. The results to some extent shows the effectiveness our learning framework, given HFL uses two VGG networks separately for edge localization and classification. Our method also significantly outperforms the HED baselines from~\cite{khoreva2016weakly}, which gives $44$ / $41$ on MF / AP, and $49$ / $45$ with detection.

\begin{figure}
\centering
\includegraphics[width=0.9\columnwidth]{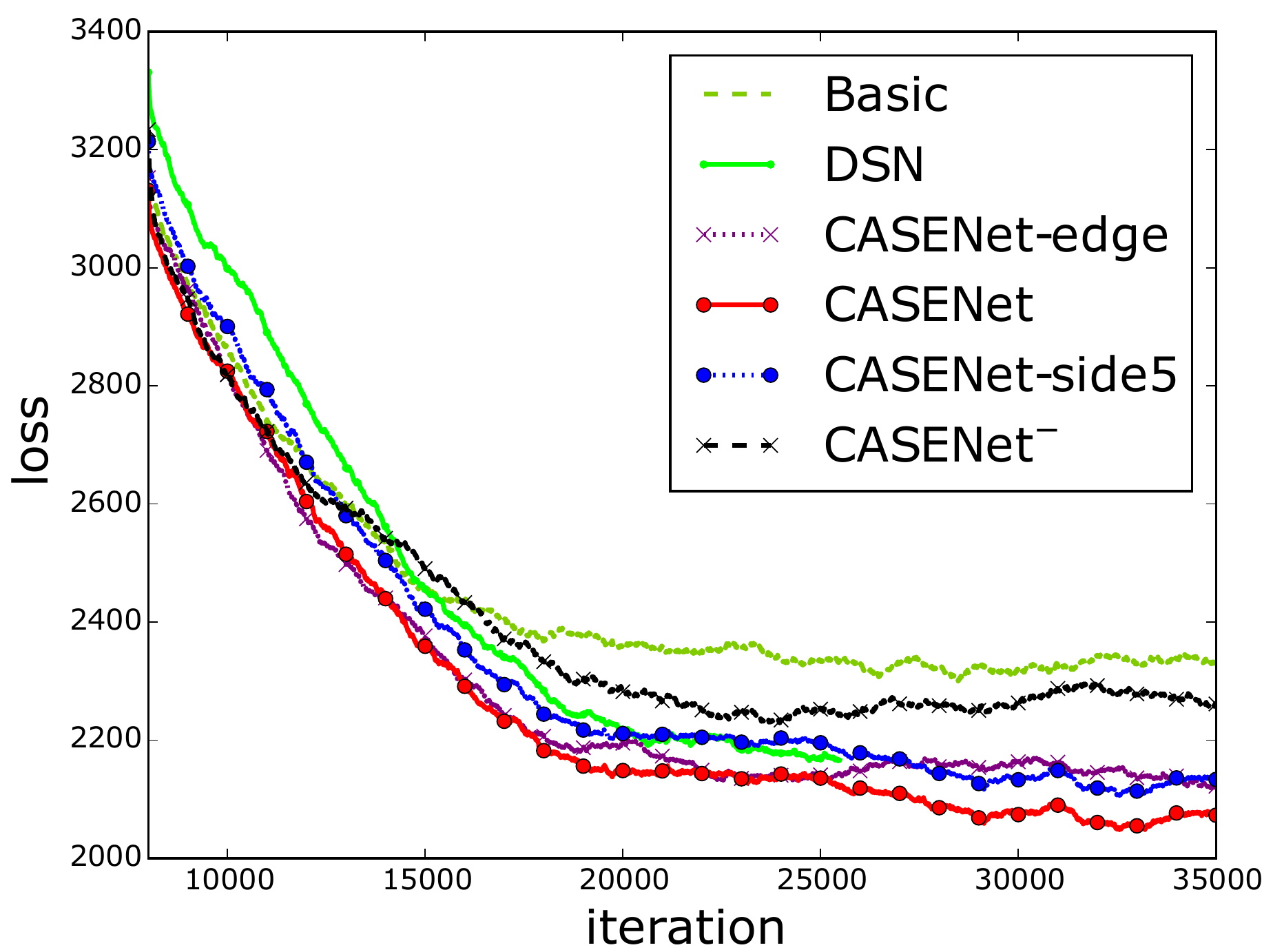}
\caption{Training losses of different variants of CASENet on the SBD dataset. The losses are respectively moving averaged by a kernel length of 8000. All curves means the final fused losses, except for CASENet-side5 which indicates the loss of Side-5's output. Note that CASENet loss is consistently the smallest.
\label{fig:loss}}
\end{figure}

\paragraph{Other variants}
We also investigated several other architectures. For example, we kept the stride of 2 in ``res1''. This downgrades the performance for lower input resolution. Another variant is to use the same CASENet architecture but impose binary edge losses (where a pixel is considered lying on an edge as long as it belongs to the edge of at least one class) on Side-1-3 (denoted as CASENet-edge in Fig.~\ref{fig:loss}). However we found that such supervision seems to be a divergence to the semantic classification at Side-5.

\begin{table*}[!t]
\resizebox{\linewidth}{!}{\begin{tabular}{c|c|l|c|c|c|c|c|c|c|c|c|c|c|c|c|c|c|c|c|c|c|c|c}
Metric & Category & Method	& aero & bike & bird & boat & bottle & bus & car & cat & chair & cow & table & dog & horse & mbike & person & plant & sheep & sofa & train & tv & mean \\
\hline \hline
\multirow{11}{0.07\linewidth}{\centering{MF\\(ODS)}}
& & InvDet & 41.5 & 46.7 & 15.6 & 17.1 & 36.5 & 42.6 & 40.3 & 22.7 & 18.9 & 26.9 & 12.5 & 18.2 & 35.4 & 29.4 & 48.2 & 13.9 & 26.9 & 11.1 & 21.9 & 31.4 & 27.9\\
& Baseline & HFL-FC8 & 71.6 & 59.6 & 68.0 & 54.1 & 57.2 & 68.0 & 58.8 & 69.3 & 43.3 & 65.8 & 33.3 & 67.9 & 67.5 & 62.2 & 69.0 & 43.8 & 68.5 & 33.9 & 57.7 & 54.8 & 58.7\\
& & HFL-CRF & 73.9 & 61.4 & 74.6 & 57.2 & 58.8 & 70.4 & 61.6 & 71.9 & 46.5 & 72.3 & 36.2 & 71.1 & 73.0 & 68.1 & 70.3 & 44.4 & 73.2 & 42.6 & 62.4 & 60.1 & 62.5\\ \cline{2-24}
& & Basic-Softmax & 67.6 & 55.3 & 50.4 & 44.9 & 42.3 & 64.6 & 61.0 & 63.9 & 37.4 & 43.1 & 25.3 & 57.9 & 57.1 & 60.0 & 72.0 & 33.0 & 53.5 & 30.9 & 54.4 & 47.7 & 51.1\\
& VGG & Basic & 70.0 & 58.6 & 62.5 & 50.2 & 51.2 & 65.4 & 60.6 & 66.9 & 39.7 & 47.3 & 31.0 & 60.1 & 59.4 & 60.2 & 74.4 & 38.0 & 56.0 & 35.9 & 60.0 & 53.8 & 55.1\\
& & CASENet & 72.5 & 61.5 & 63.8 & 54.5 & 52.3 & 65.4 & 62.6 & 67.2 & 42.6 & 51.8 & 31.4 & 62.0 & 61.9 & 62.8 & 75.4 & 41.7 & 59.8 & 35.8 & 59.7 & 50.7 & 56.8\\ \cline{2-24}
& & Basic-Softmax & 74.0 & 64.1 & 64.8 & 52.5 & 52.1 & 73.2 & 68.1 & 73.2 & 43.1 & 56.2 & 37.3 & 67.4 & 68.4 & 67.6 & 76.7 & 42.7 & 64.3 & 37.5 & 64.6 & 56.3 & 60.2\\
& & Basic & 82.5 & 74.2 & 80.2 & 62.3 & 68.0 & 80.8 & 74.3 & 82.9 & 52.9 & 73.1 & 46.1 & 79.6 & 78.9 & 76.0 & 80.4 & 52.4 & 75.4 & 48.6 & 75.8 & 68.0 & 70.6\\
& ResNet & DSN & 81.6 & 75.6 & 78.4 & 61.3 & 67.6 & \textbf{82.3} & 74.6 & 82.6 & 52.4 & 71.9 & 45.9 & 79.2 & 78.3 & 76.2 & 80.1 & 51.9 & 74.9 & 48.0 & 76.5 & 66.8 & 70.3\\
& & CASENet$^{-}$ & 83.0 & 74.7 & 79.6 & 61.5 & 67.7 & 80.7 & 74.1 & 82.8 & 53.3 & \textbf{75.0} & 44.5 & 79.8 & \textbf{80.4} & 76.2 & 80.2 & 53.2 & \textbf{77.3} & 47.7 & 75.6 & 66.3 & 70.7\\
& & CASENet & \textbf{83.3} & \textbf{76.0} & \textbf{80.7} & \textbf{63.4} & \textbf{69.2} & 81.3 & \textbf{74.9} & \textbf{83.2} & \textbf{54.3} & 74.8 & \textbf{46.4} & \textbf{80.3} & 80.2 & \textbf{76.6} & \textbf{80.8} & \textbf{53.3} & 77.2 & \textbf{50.1} & \textbf{75.9} & \textbf{66.8} & \textbf{71.4}\\
\end{tabular}}
\hfill
\caption{Results on the SBD benchmark. All MF scores are measured by $\%$.\label{tb:main}}
\end{table*}

\subsection{Results on Cityscapes}

\begin{table*}[!htbp]
\centering
\resizebox{\textwidth}{!}{\begin{tabular}{c|l|c|c|c|c|c|c|c|c|c|c|c|c|c|c|c|c|c|c|c|c}
Metric & Method	& road & sidewalk & building & wall & fence & pole & traffic lgt & traffic sign & vegetation & terrain & sky & person & rider & car & truck & bus & train & motorcycle & bike & mean \\
\hline \hline
\multirow{2}{0.05\linewidth}{\centering{MF\\(ODS)}}
& DSN & 85.4 & 76.4 & 82.6 & \textbf{51.8} & 56.5 & 66.5 & 62.6 & 72.1 & 80.6 & 61.1 & 76.0 & 77.5 & 66.3 & 84.5 & \textbf{52.3} & 67.3 & 49.4 & 56.0 & 76.0 & 68.5\\
& CASENet & \textbf{86.6} & \textbf{78.8} & \textbf{85.1} & 51.5 & \textbf{58.9} & \textbf{70.1} & \textbf{70.8} & \textbf{74.6} & \textbf{83.5} & \textbf{62.9} & \textbf{79.4} & \textbf{81.5} & \textbf{71.3} & \textbf{86.9} & 50.4 & \textbf{69.5} & \textbf{52.0} & \textbf{61.3} & \textbf{80.2} & \textbf{71.3}\\
\hline \hline
\multirow{2}{0.05\linewidth}{\centering{AP}}
& DSN & \textbf{78.0} & 76.0 & 83.9 & 47.9 & 53.1 & 67.9 & 57.9 & 75.9 & 79.9 & 60.2 & 75.0 & 75.4 & 61.0 & 85.8 & \textbf{50.6} & 67.8 & 42.5 & 51.4 & 72.0 & 66.4\\
& CASENet & 77.7 & \textbf{78.6} & \textbf{87.6} & \textbf{49.0} & \textbf{56.9} & \textbf{72.8} & \textbf{70.3} & \textbf{78.9} & \textbf{85.1} & \textbf{63.1} & \textbf{78.4} & \textbf{83.0} & \textbf{70.1} & \textbf{89.5} & 46.9 & \textbf{70.0} & \textbf{48.8} & \textbf{59.6} & \textbf{78.9} & \textbf{70.8}\\
\end{tabular}}
\hfill
\caption{Results on the Cityscapes dataset. All MF and AP scores are measured by $\%$.\label{tb:cityscapes}}
\end{table*}

\begin{figure*}[b]
\definecolor{blk_color_0}{rgb}{0.498,1.000,0.000}
\definecolor{blk_color_1}{rgb}{1.000,0.000,0.031}
\definecolor{blk_color_2}{rgb}{0.000,0.765,1.000}
\definecolor{blk_color_3}{rgb}{1.000,0.667,0.000}
\definecolor{blk_color_4}{rgb}{1.000,0.400,0.000}
\definecolor{blk_color_5}{rgb}{1.000,0.000,0.365}
\definecolor{blk_color_6}{rgb}{1.000,0.569,0.000}
\definecolor{blk_color_7}{rgb}{1.000,0.169,0.000}
\definecolor{blk_color_8}{rgb}{0.000,1.000,0.333}
\definecolor{blk_color_9}{rgb}{0.000,0.933,1.000}
\definecolor{blk_color_10}{rgb}{1.000,0.000,0.667}
\definecolor{blk_color_11}{rgb}{0.000,0.165,1.000}
\definecolor{blk_color_12}{rgb}{0.000,1.000,0.733}
\definecolor{blk_color_13}{rgb}{0.769,1.000,0.000}
\definecolor{blk_color_14}{rgb}{0.000,1.000,0.569}
\definecolor{blk_color_15}{rgb}{1.000,0.502,0.000}
\definecolor{blk_color_16}{rgb}{1.000,0.502,0.000}
\definecolor{blk_color_17}{rgb}{1.000,0.000,0.600}
\definecolor{blk_color_18}{rgb}{0.000,1.000,0.902}
\definecolor{blk_color_19}{rgb}{0.400,0.000,1.000}
\definecolor{blk_color_20}{rgb}{1.000,0.000,0.498}
\definecolor{blk_color_21}{rgb}{0.333,0.000,1.000}
\definecolor{blk_color_22}{rgb}{0.000,1.000,0.835}
\definecolor{blk_color_23}{rgb}{0.600,1.000,0.000}
\definecolor{blk_color_24}{rgb}{0.067,0.000,1.000}
\definecolor{blk_color_25}{rgb}{0.267,1.000,0.000}
\definecolor{blk_color_26}{rgb}{0.000,0.765,1.000}
\definecolor{blk_color_27}{rgb}{1.000,0.333,0.000}
\definecolor{blk_color_28}{rgb}{0.169,0.000,1.000}
\definecolor{blk_color_29}{rgb}{0.067,0.000,1.000}
\centering
\resizebox{1\textwidth}{!}{
\begin{tabular}{@{}cccccccccc@{}}
\cellcolor{blk_color_0} building+vegetation &
\cellcolor{blk_color_1} road &
\cellcolor{blk_color_2} road+traffic sign &
\cellcolor{blk_color_3} building &
\cellcolor{blk_color_4} building+pole &
\cellcolor{blk_color_5} road+sidewalk &
\cellcolor{blk_color_6} building+traffic sign &
\cellcolor{blk_color_7} pole &
\cellcolor{blk_color_8} vegetation &
\cellcolor{blk_color_9} building+person \\
\cellcolor{blk_color_10} sidewalk &
\cellcolor{blk_color_11} \textcolor{white}{sidewalk+vegetation} &
\cellcolor{blk_color_12} sidewalk+pole &
\cellcolor{blk_color_13} pole+vegetation &
\cellcolor{blk_color_14} vegetation+bicycle &
\cellcolor{blk_color_15} building+traffic light &
\cellcolor{blk_color_16} traffic sign &
\cellcolor{blk_color_17} sidewalk+person &
\cellcolor{blk_color_18} sidewalk+traffic sign &
\cellcolor{blk_color_19} \textcolor{white}{road+bicycle} \\
\cellcolor{blk_color_20} person &
\cellcolor{blk_color_21} \textcolor{white}{rider+bicycle} &
\cellcolor{blk_color_22} bicycle &
\cellcolor{blk_color_23} traffic sign+vegetation &
\cellcolor{blk_color_24} \textcolor{white}{vegetation+rider} &
\cellcolor{blk_color_25} building+bicycle &
\cellcolor{blk_color_26} building+rider &
\cellcolor{blk_color_27} pole+traffic sign &
\cellcolor{blk_color_28} \textcolor{white}{person+bicycle} &
\cellcolor{blk_color_29} \textcolor{white}{sidewalk+bicycle}
\end{tabular} }
\centering
\includegraphics[width=0.242\textwidth]{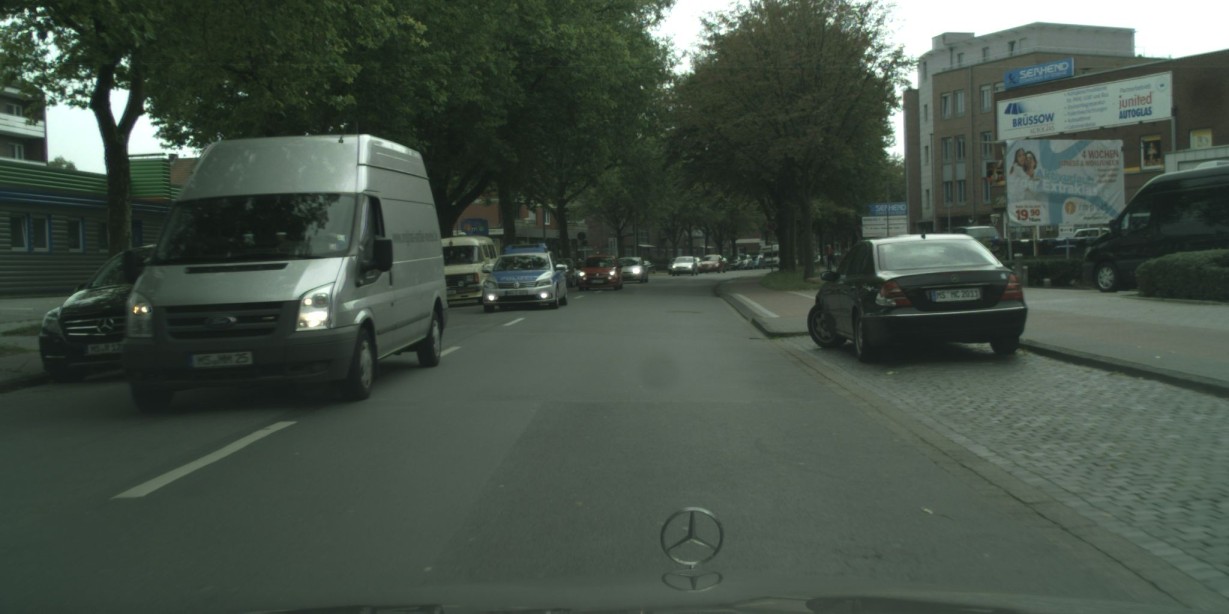}
\includegraphics[width=0.242\textwidth]{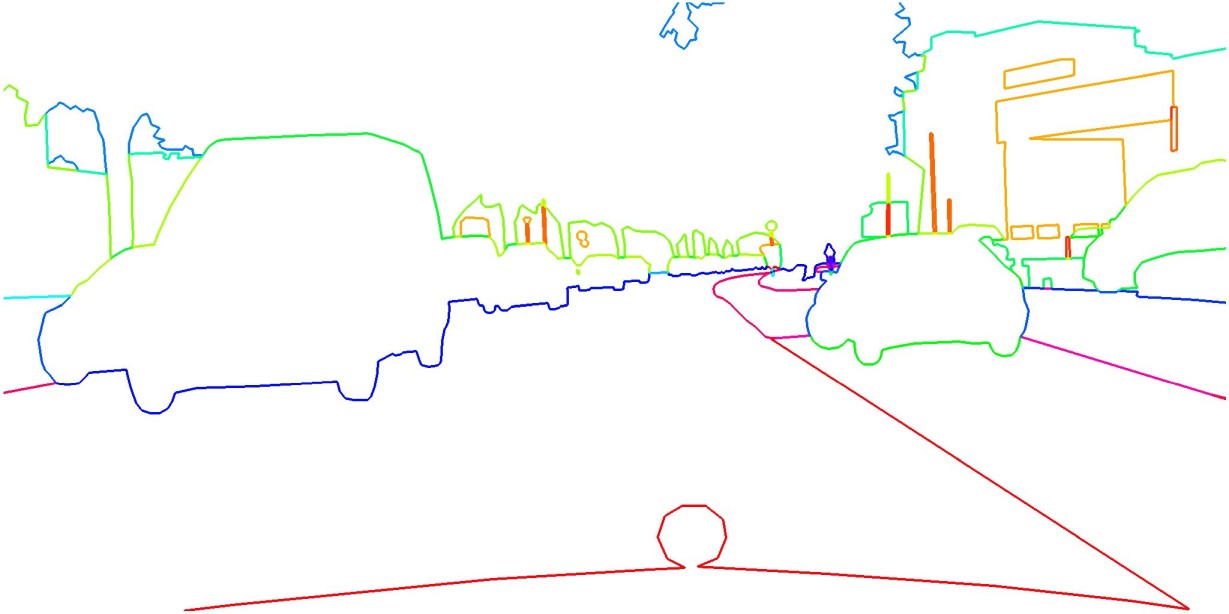}
\includegraphics[width=0.242\textwidth]{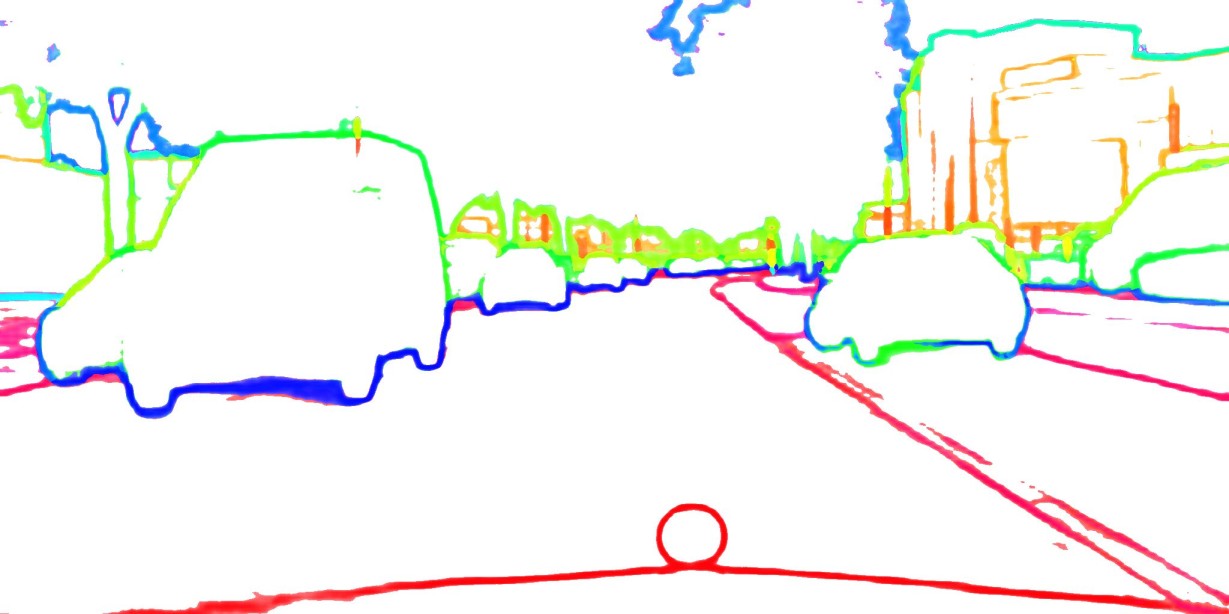}
\includegraphics[width=0.242\textwidth]{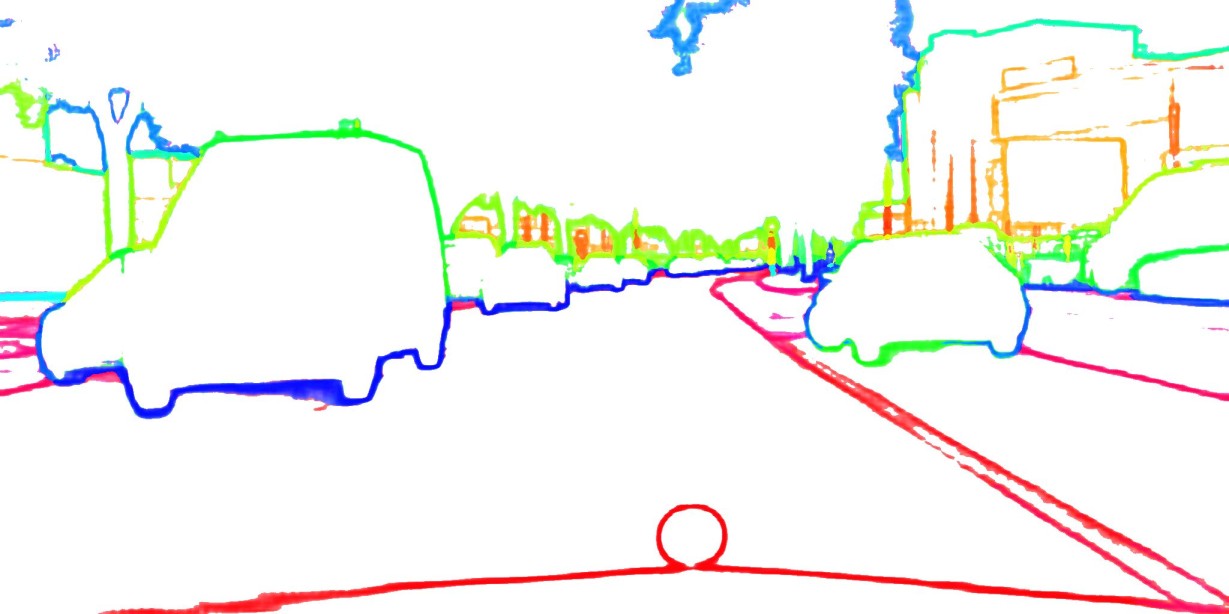}

\includegraphics[width=0.242\textwidth]{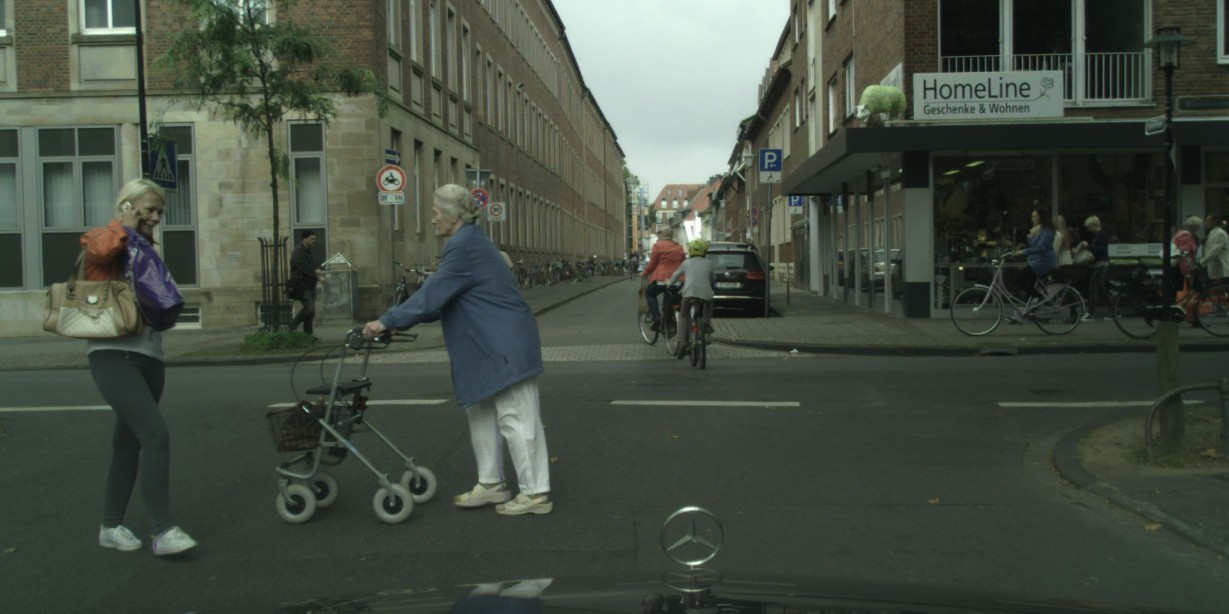}
\includegraphics[width=0.242\textwidth]{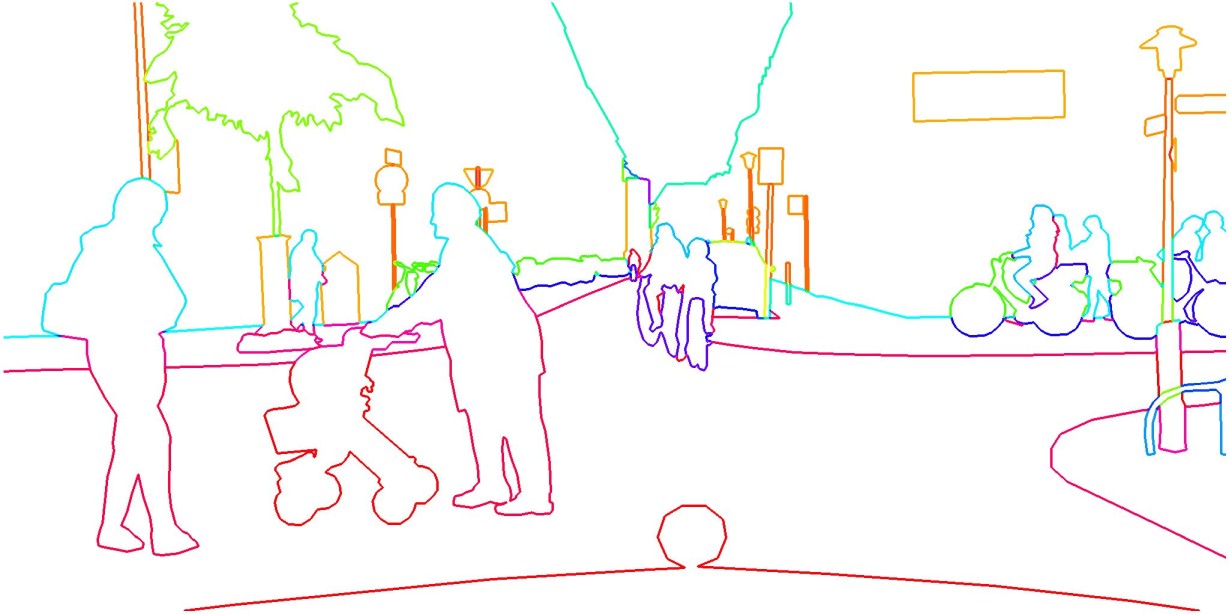}
\includegraphics[width=0.242\textwidth]{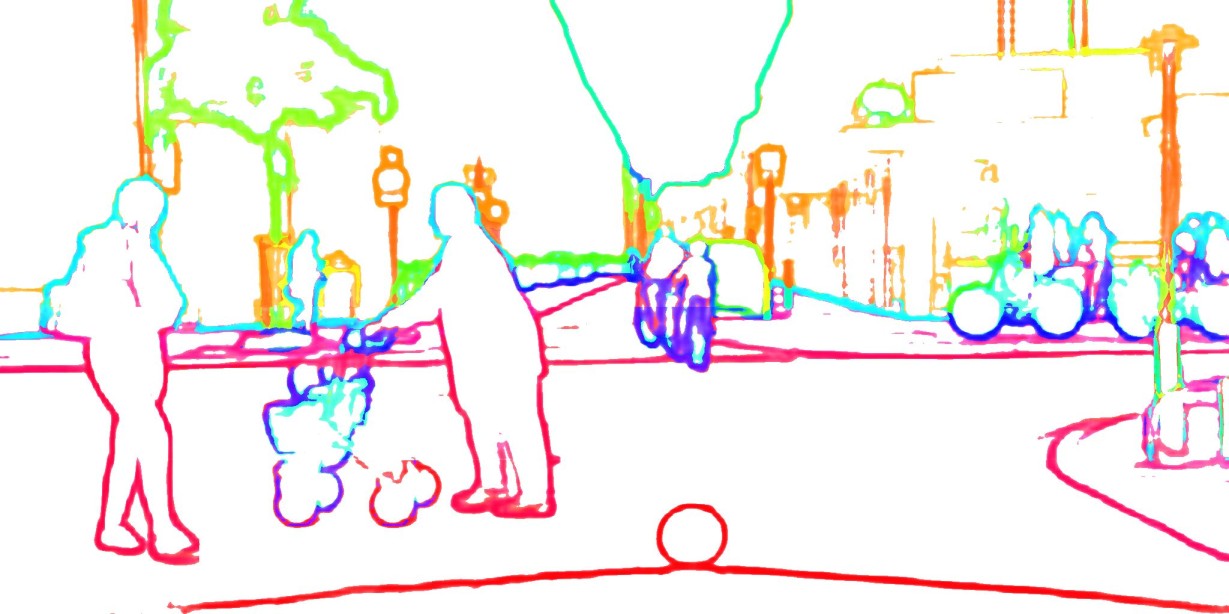}
\includegraphics[width=0.242\textwidth]{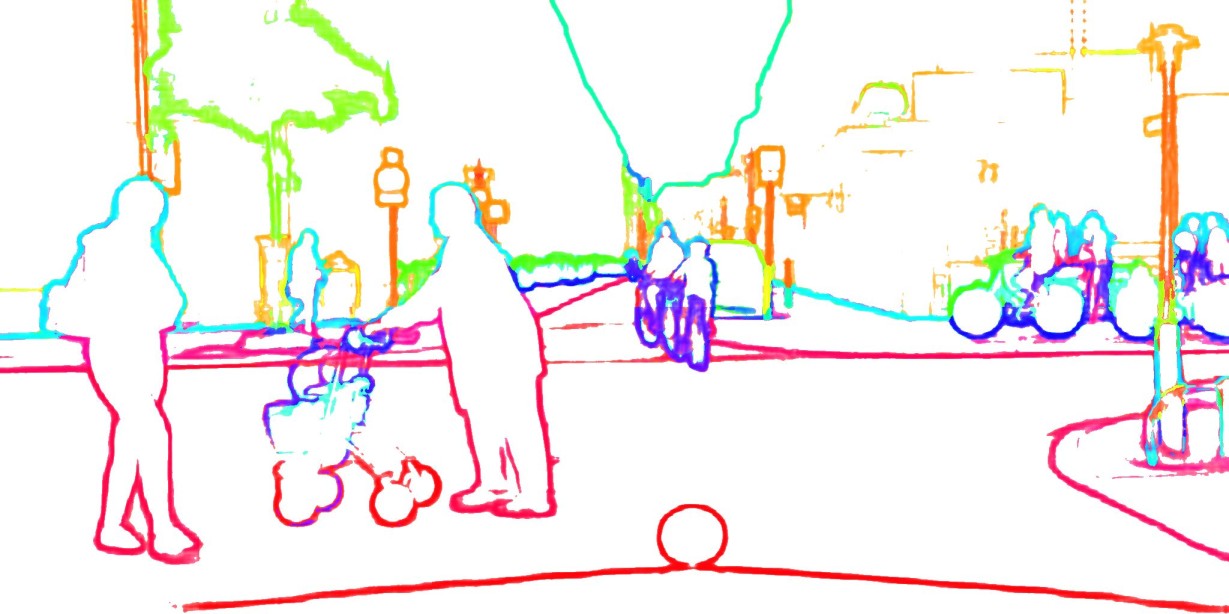}
	
\includegraphics[width=0.242\textwidth]{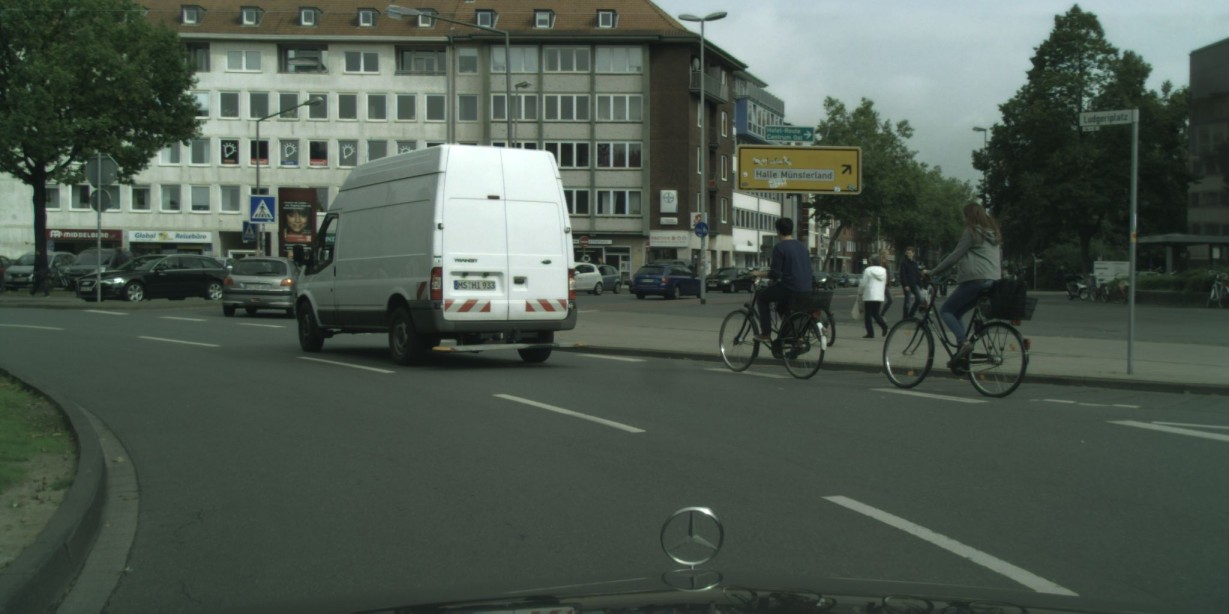}
\includegraphics[width=0.242\textwidth]{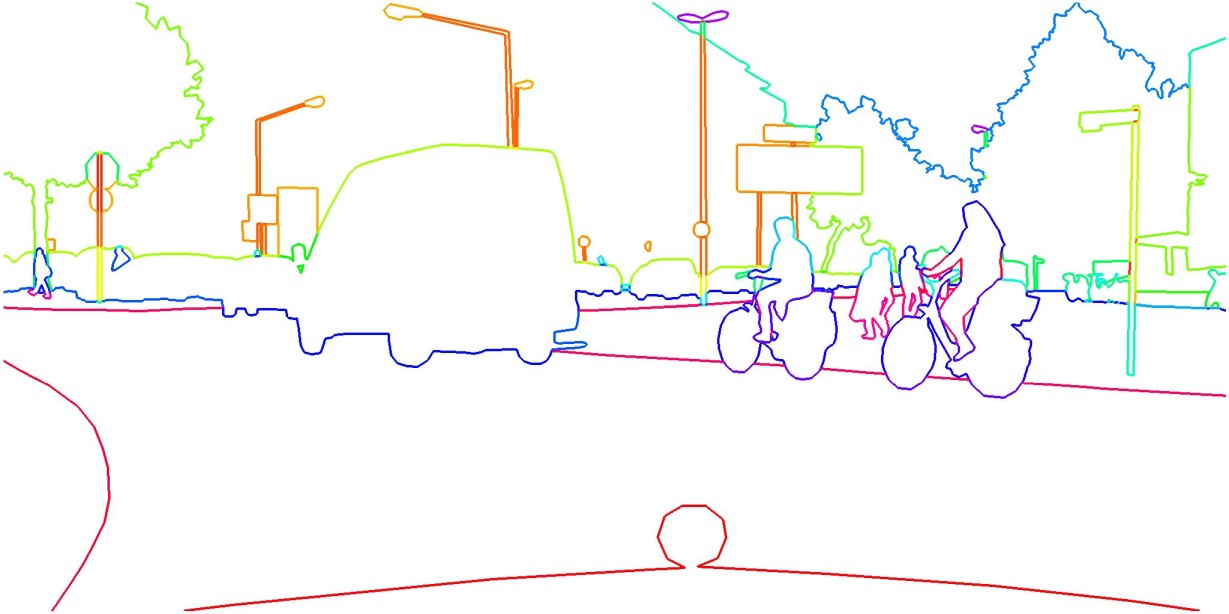}
\includegraphics[width=0.242\textwidth]{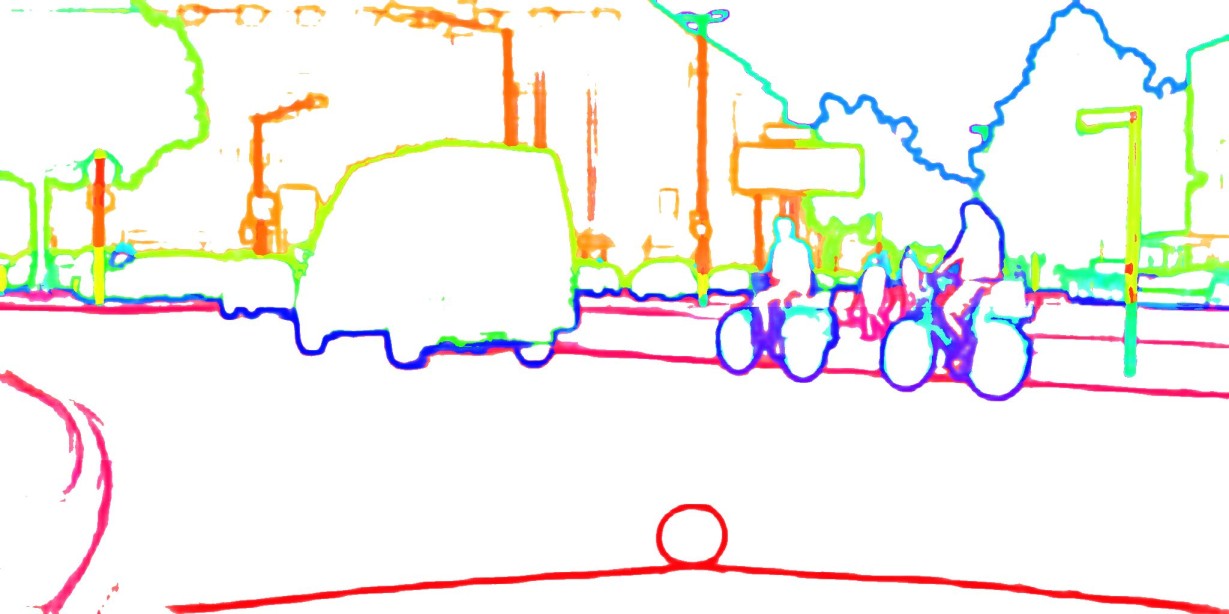}
\includegraphics[width=0.242\textwidth]{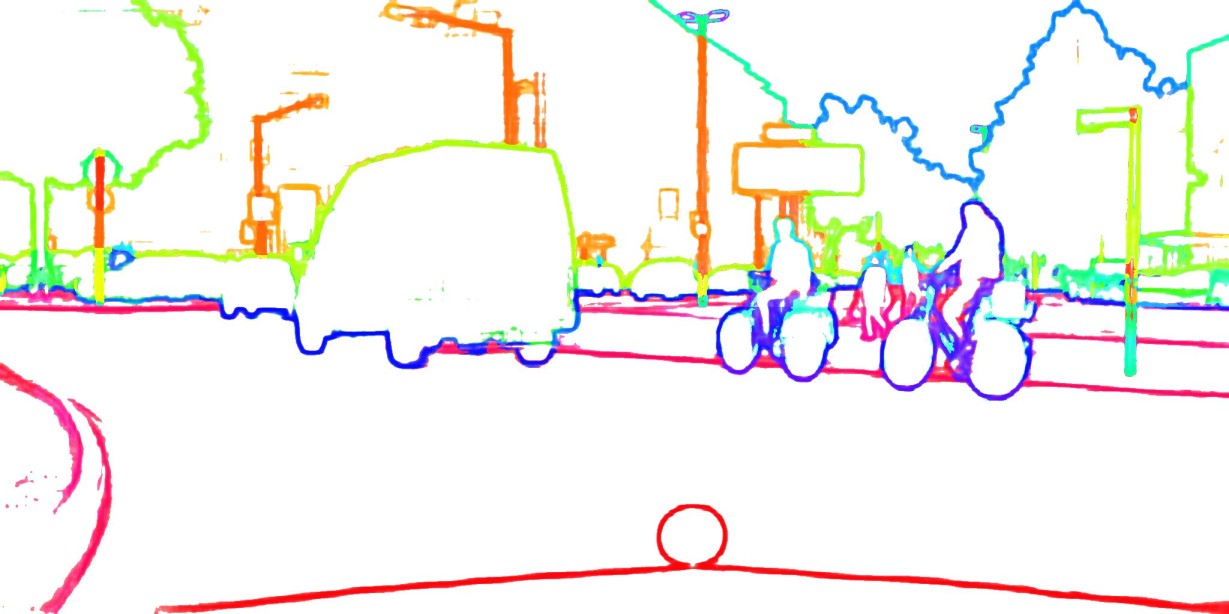}
	
\includegraphics[width=0.242\textwidth]{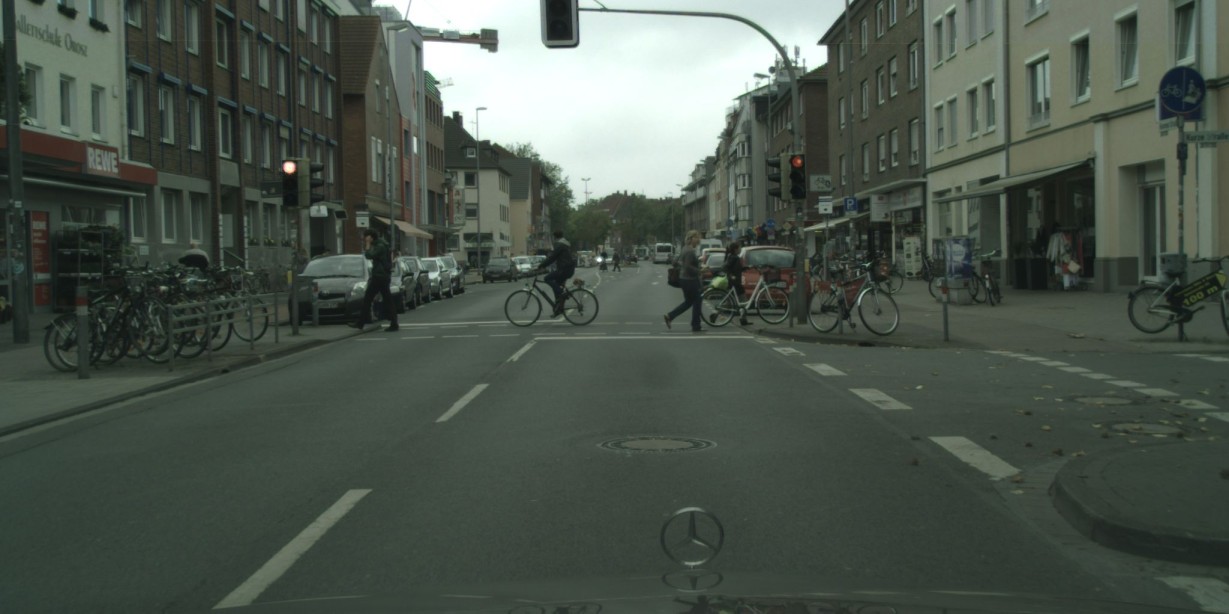}
\includegraphics[width=0.242\textwidth]{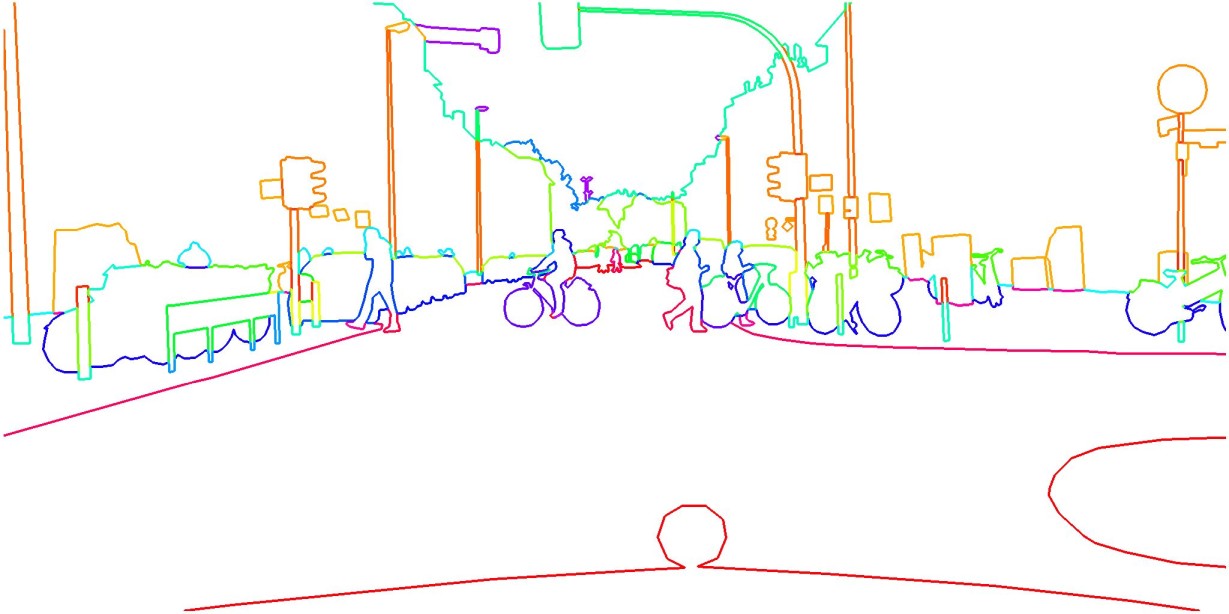}
\includegraphics[width=0.242\textwidth]{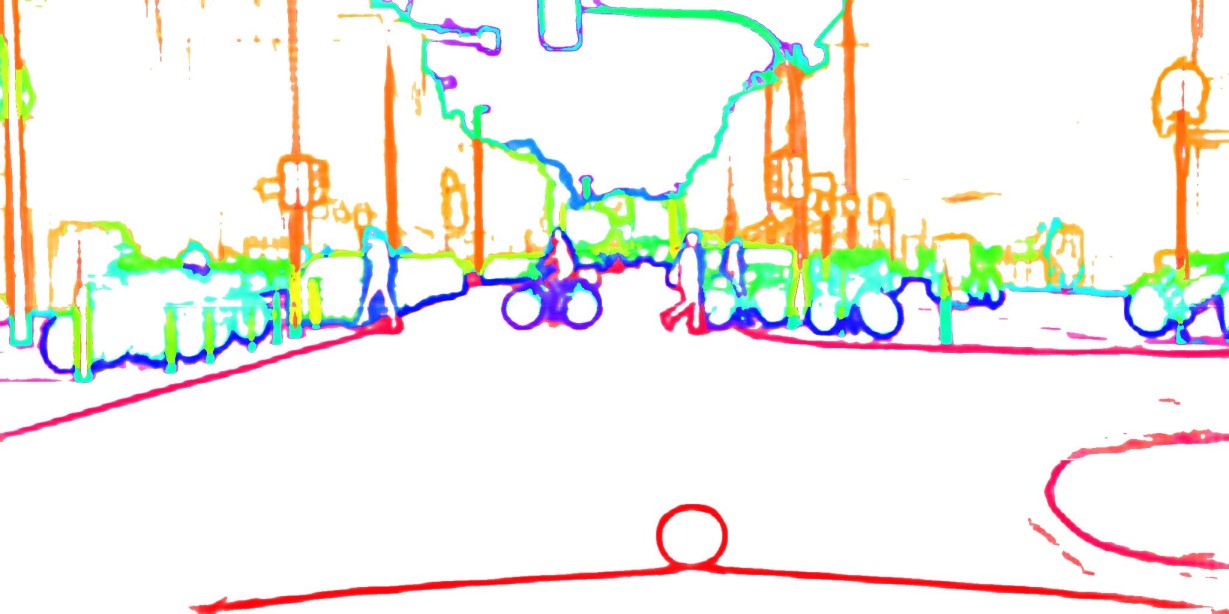}
\includegraphics[width=0.242\textwidth]{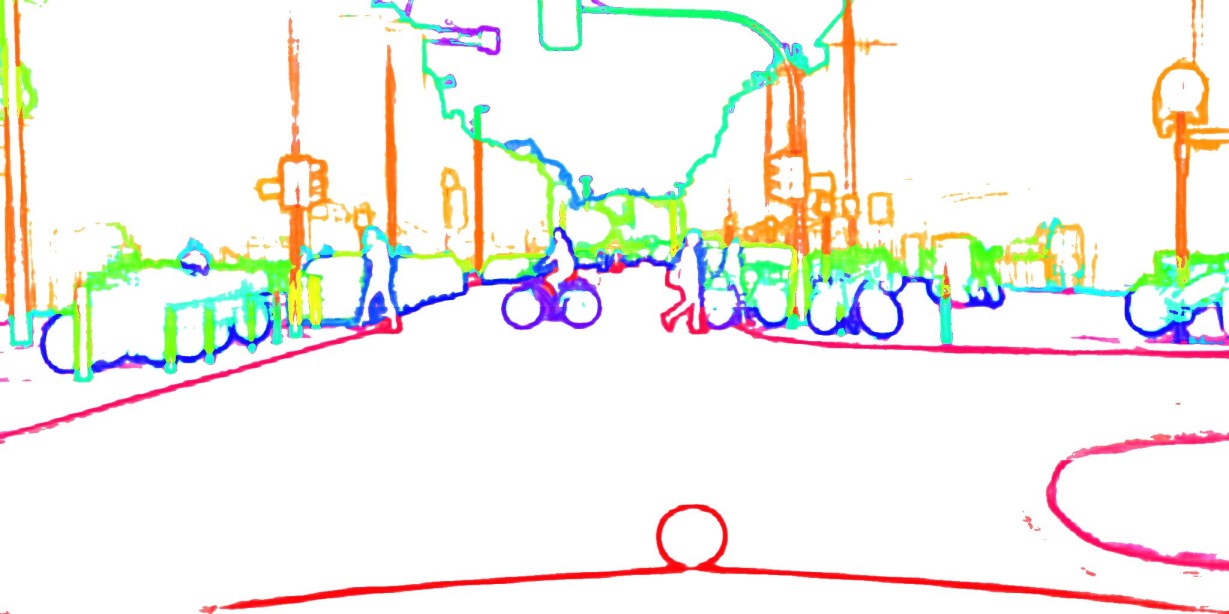}
	
\includegraphics[width=0.242\textwidth]{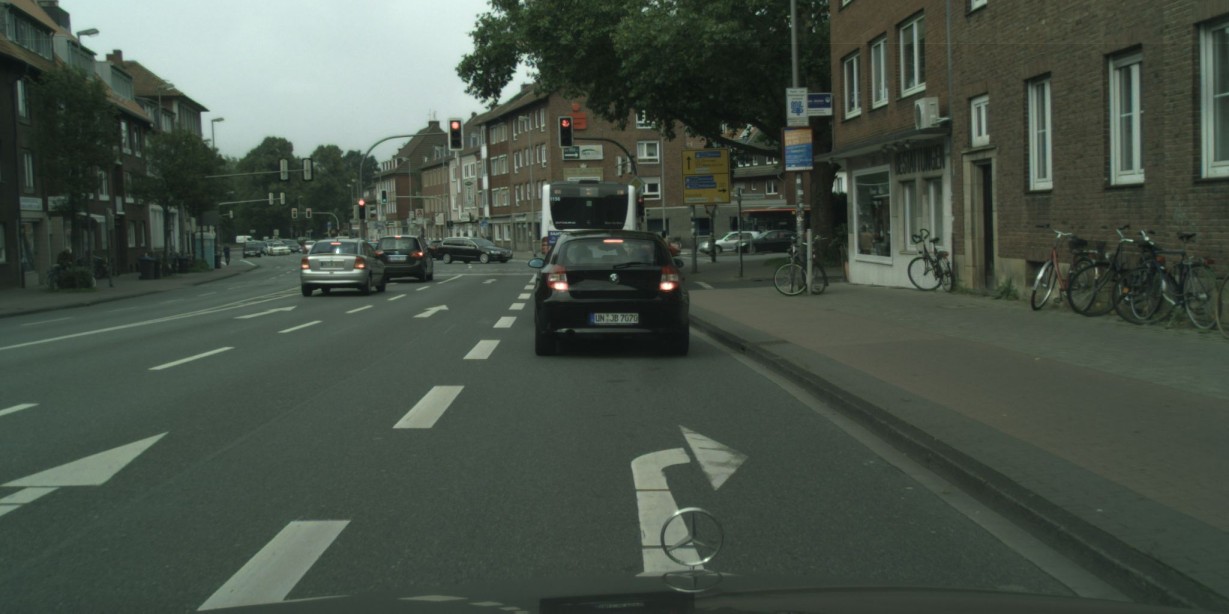}
\includegraphics[width=0.242\textwidth]{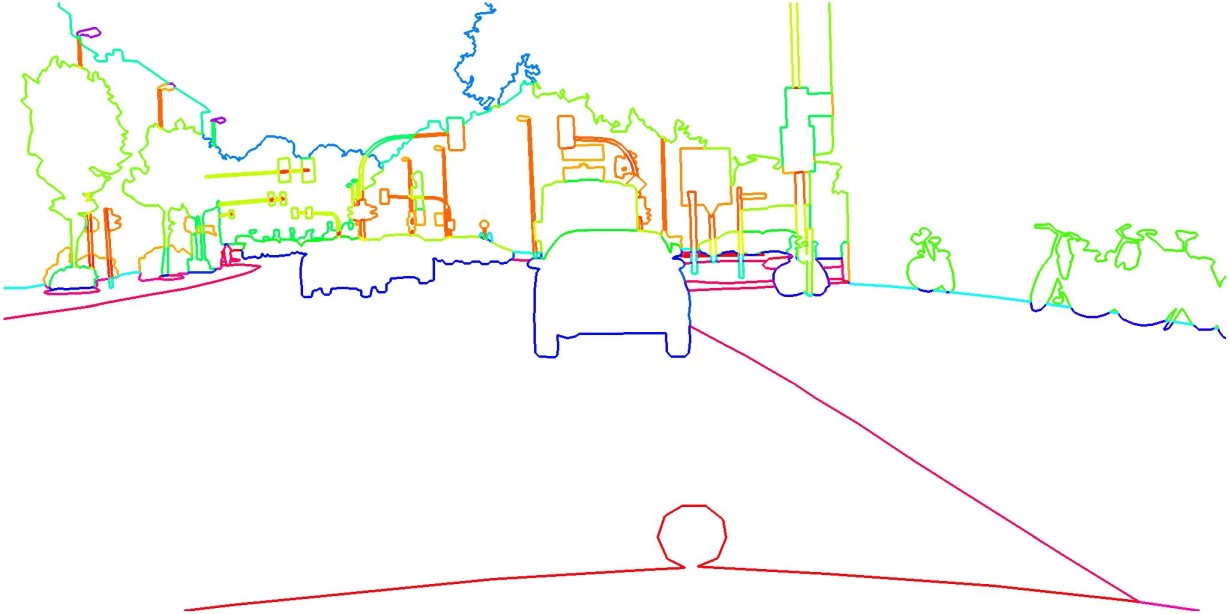}
\includegraphics[width=0.242\textwidth]{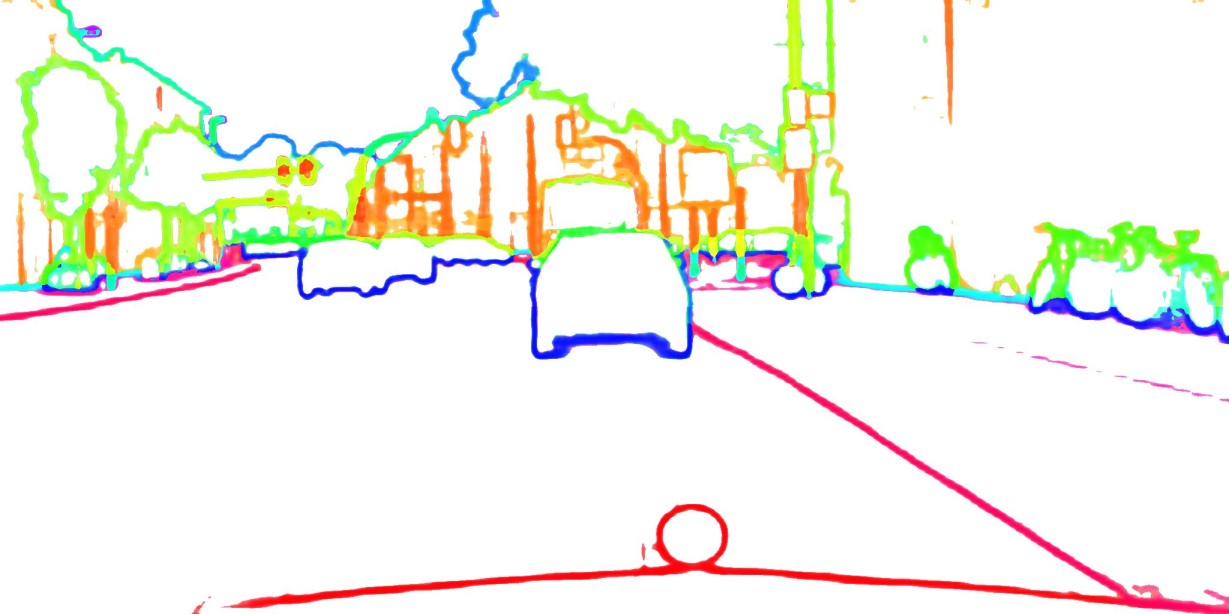}
\includegraphics[width=0.242\textwidth]{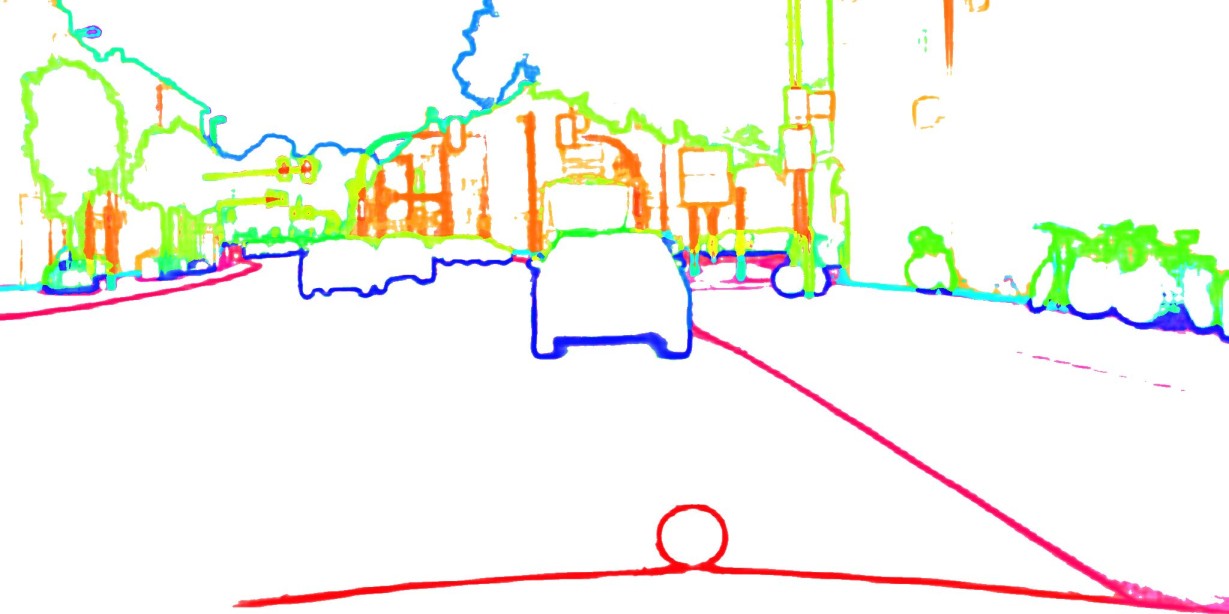}
	
\includegraphics[width=0.242\textwidth]{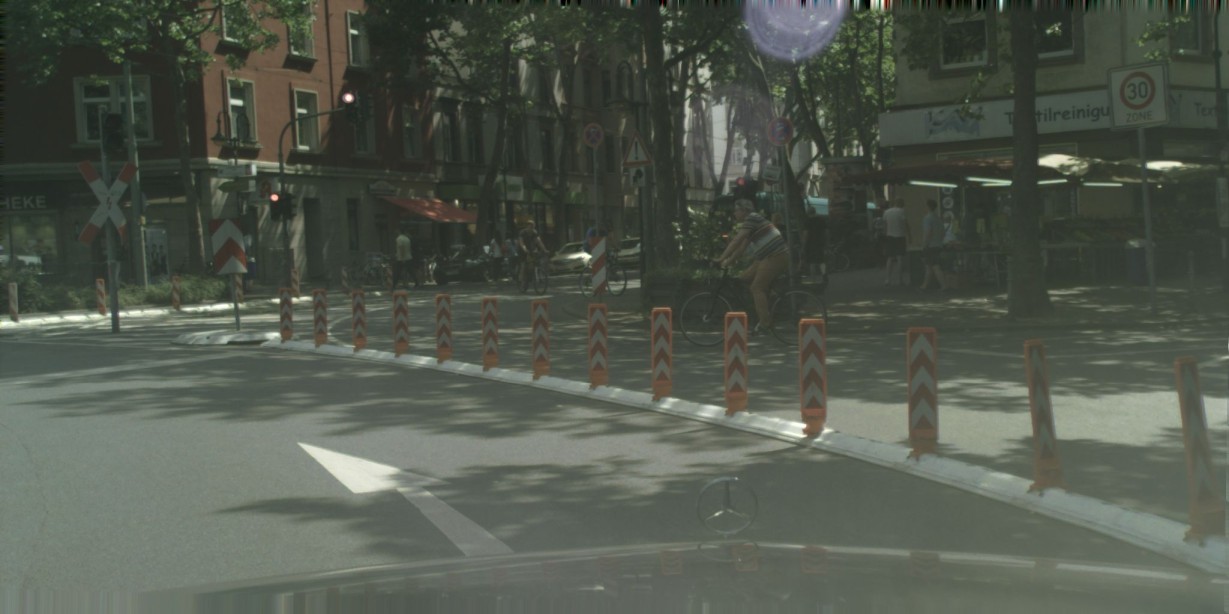}
\includegraphics[width=0.242\textwidth]{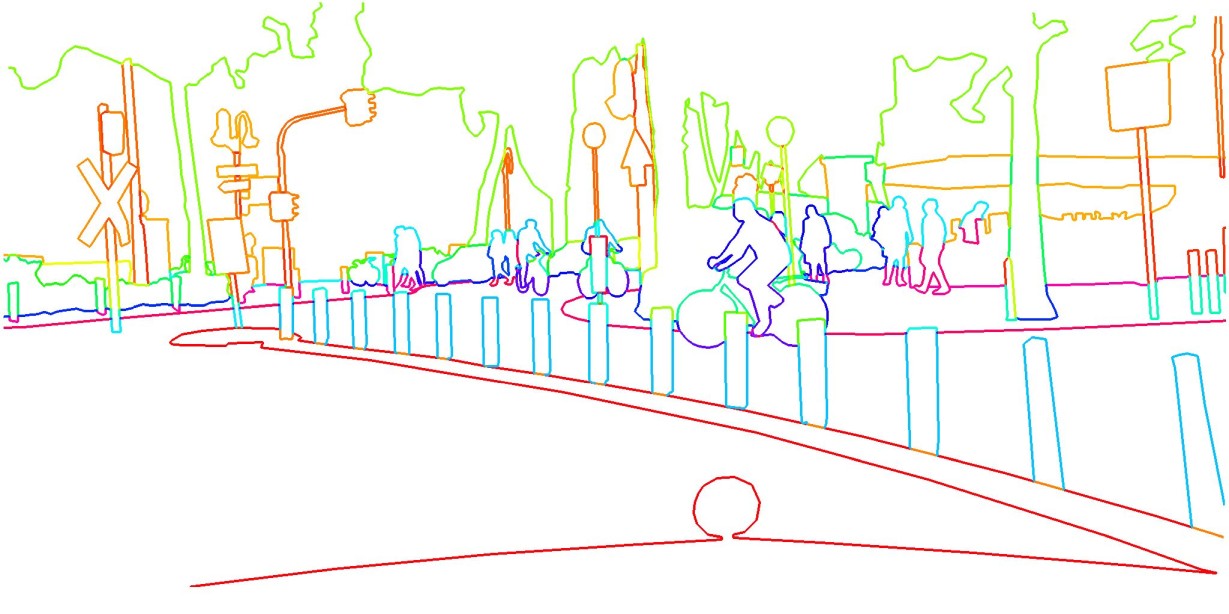}
\includegraphics[width=0.242\textwidth]{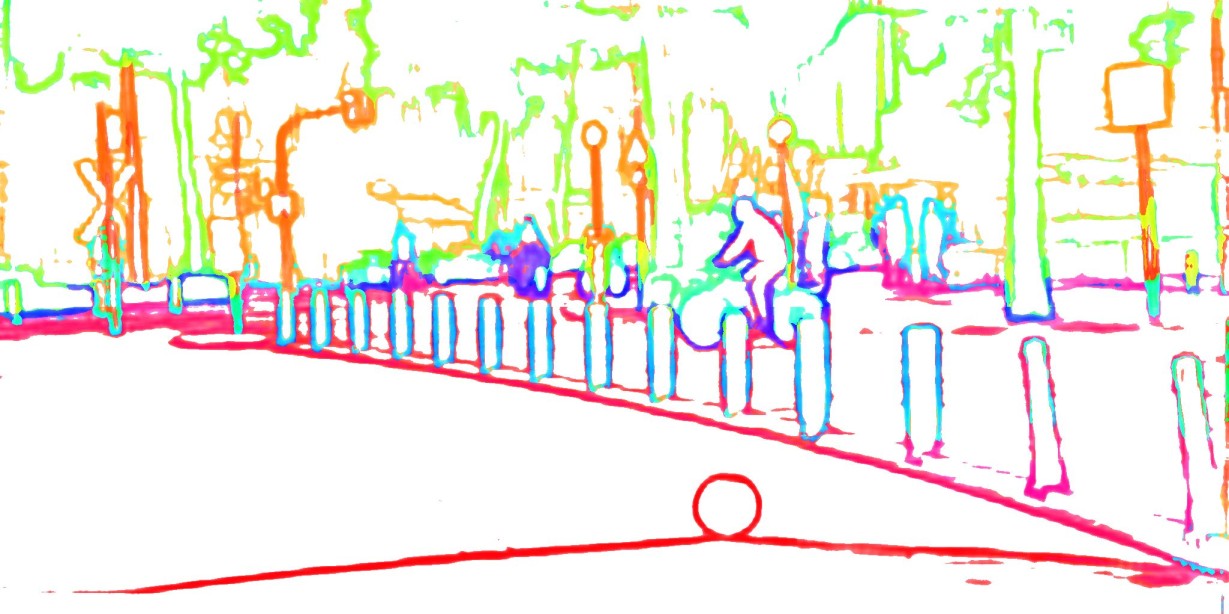}
\includegraphics[width=0.242\textwidth]{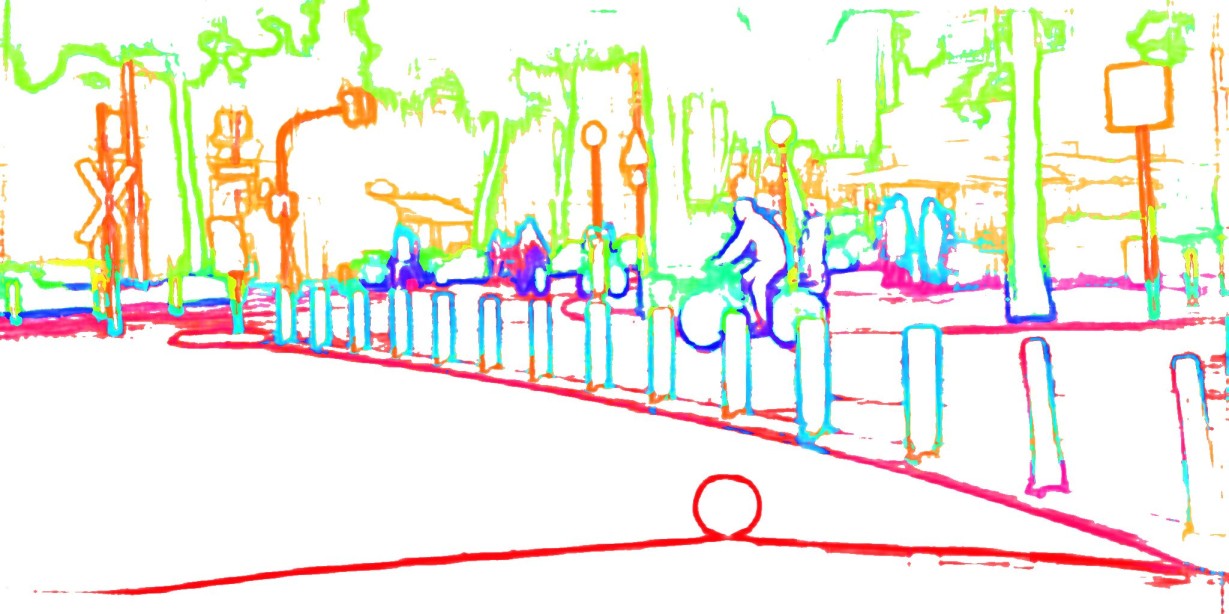}
	
\includegraphics[width=0.242\textwidth]{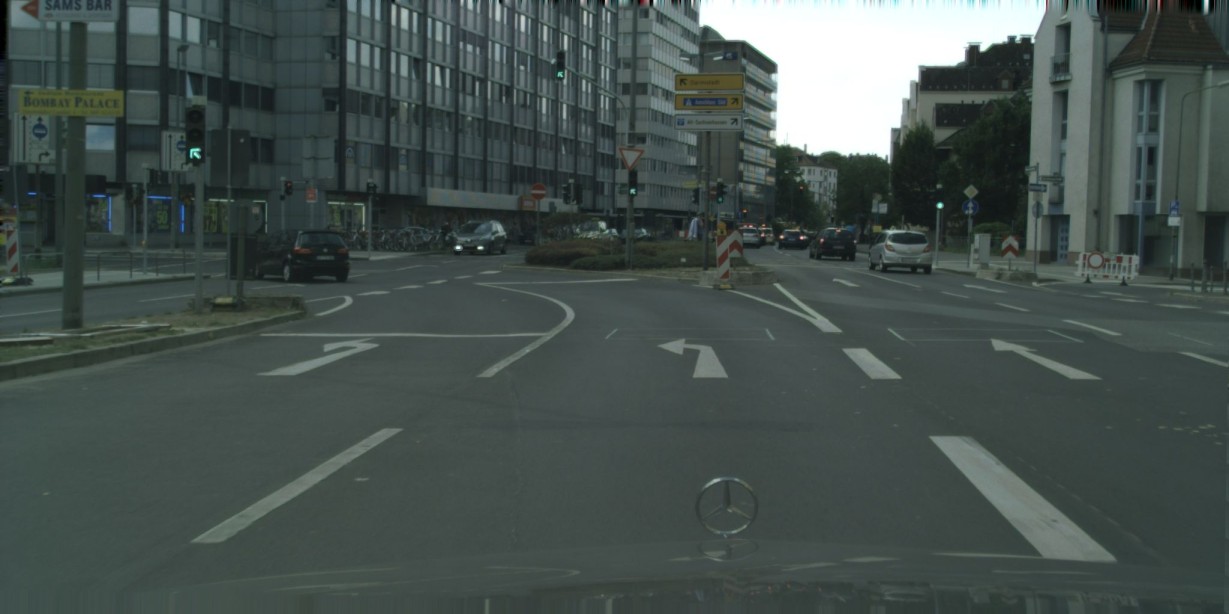}
\includegraphics[width=0.242\textwidth]{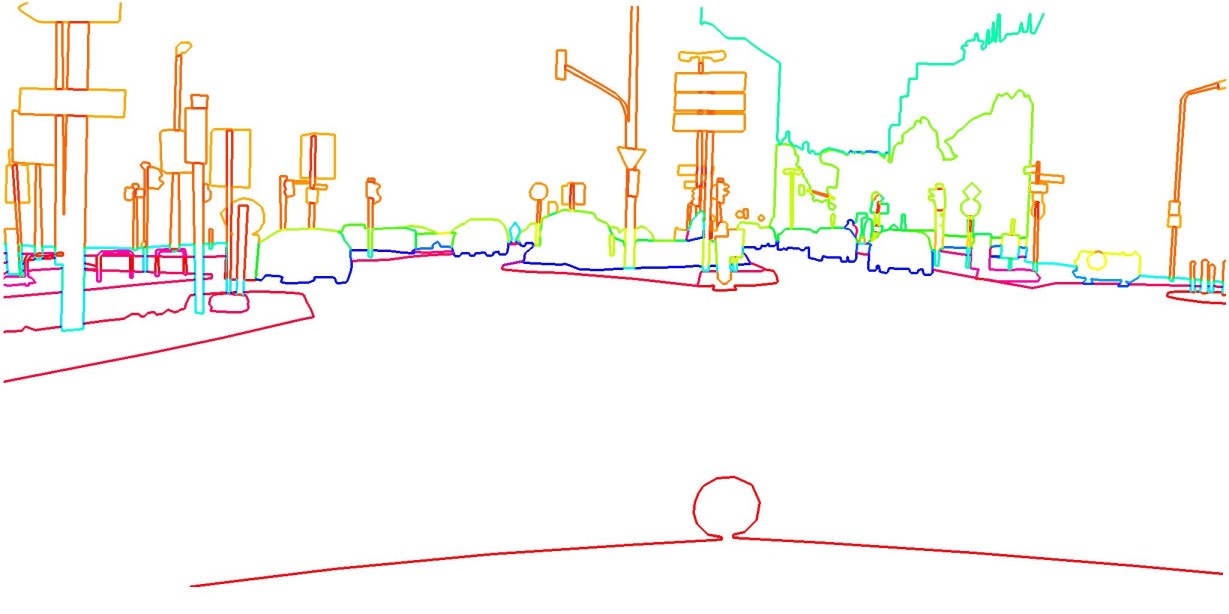}
\includegraphics[width=0.242\textwidth]{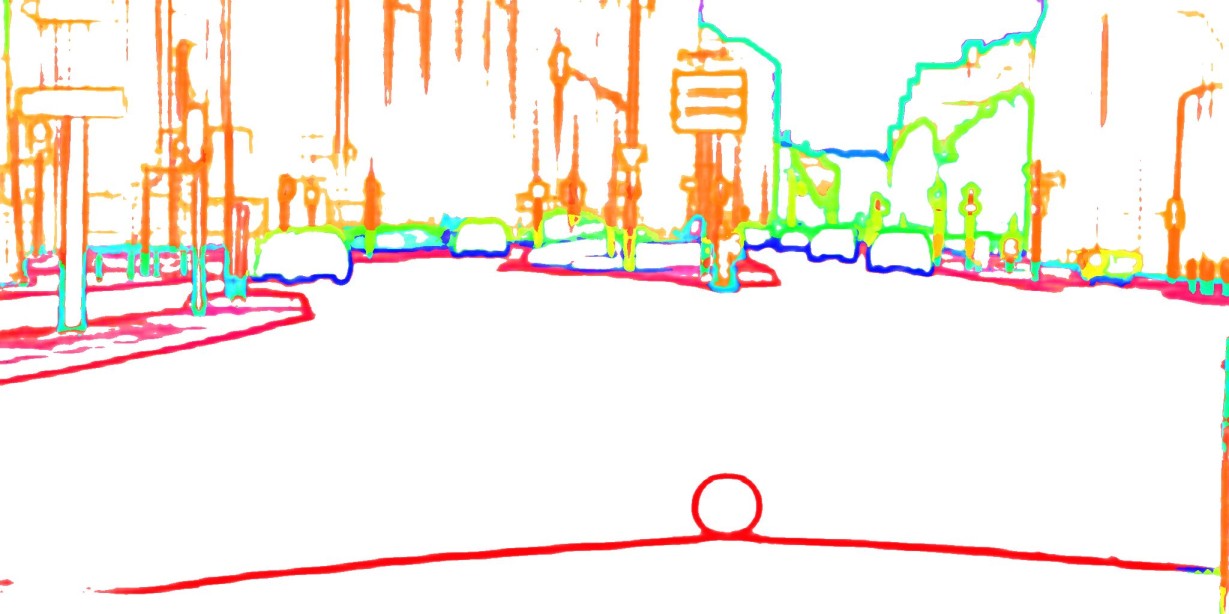}
\includegraphics[width=0.242\textwidth]{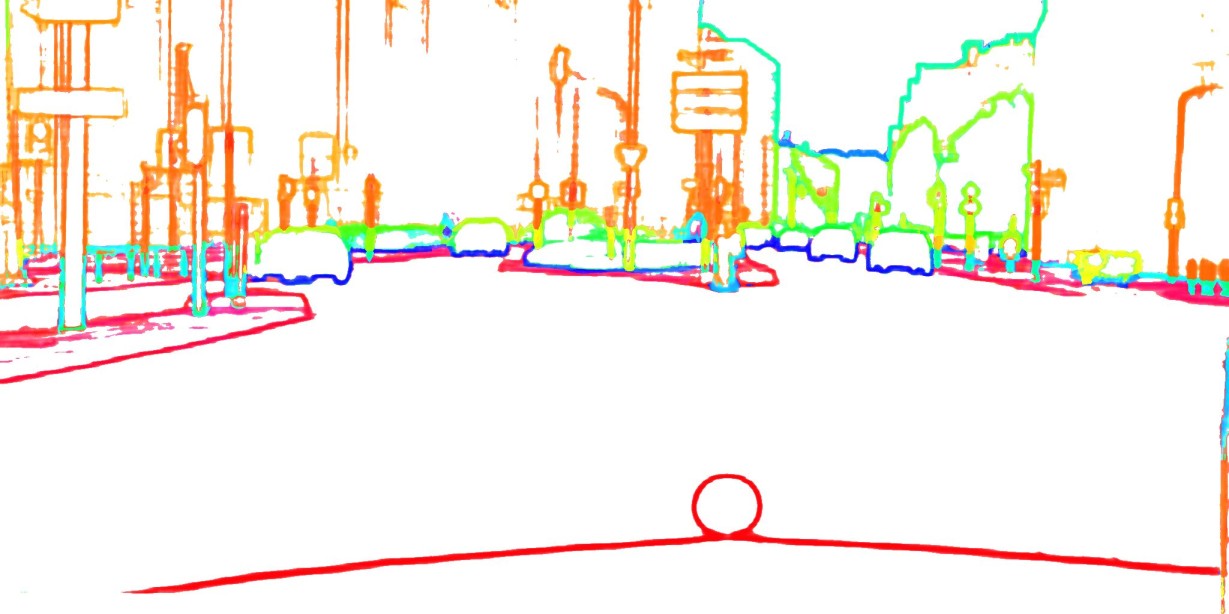}
	
\includegraphics[width=0.242\textwidth]{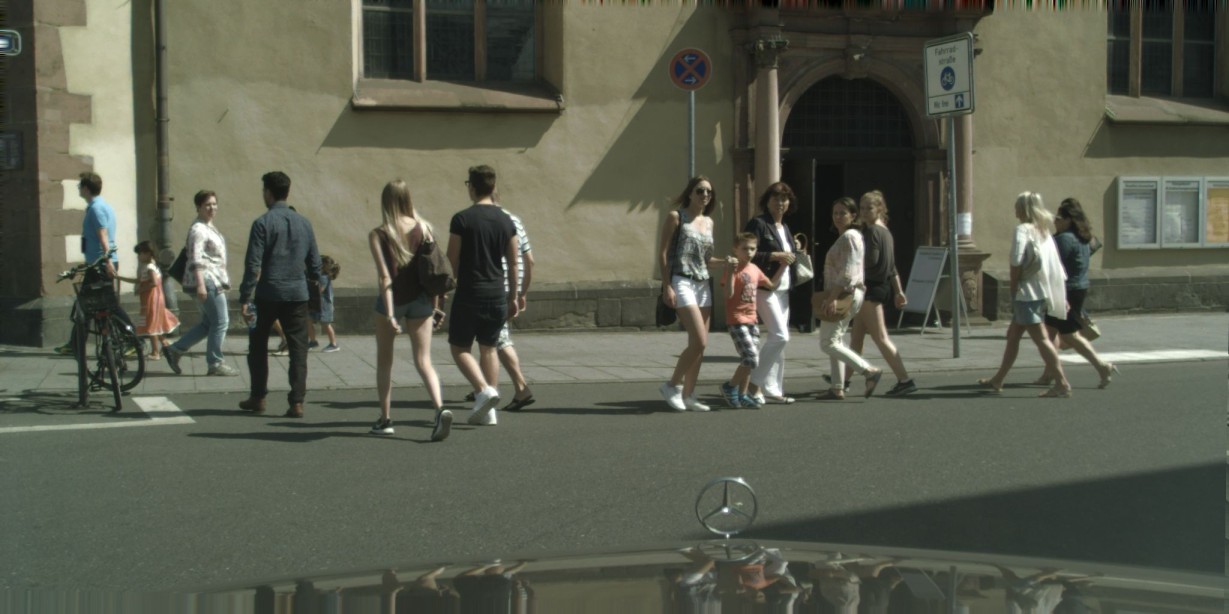}
\includegraphics[width=0.242\textwidth]{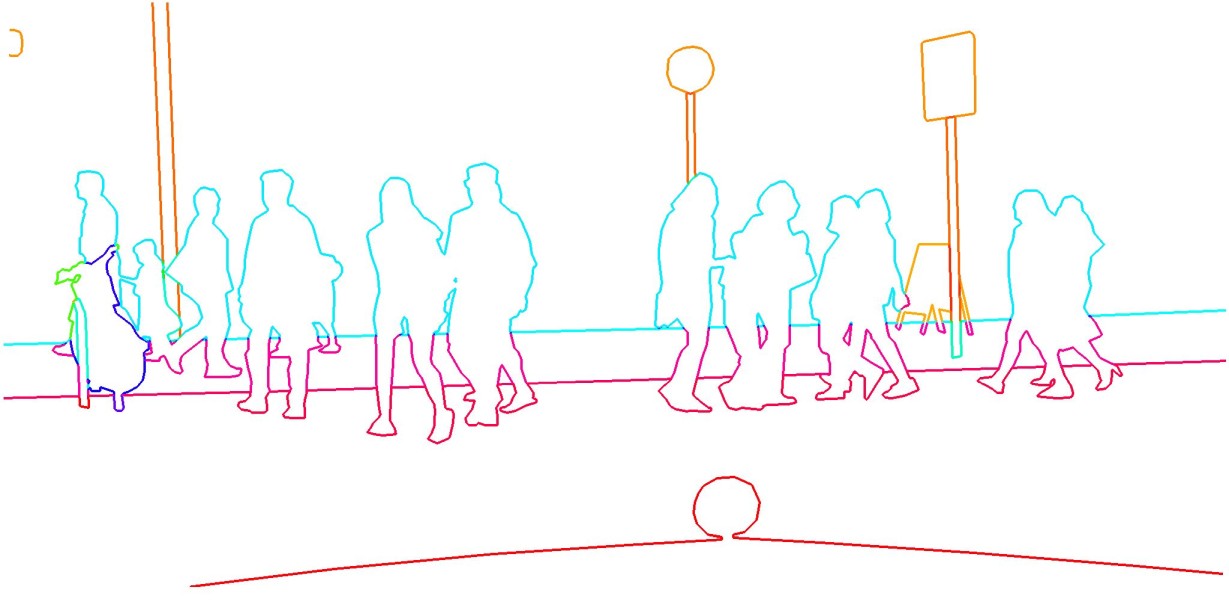}
\includegraphics[width=0.242\textwidth]{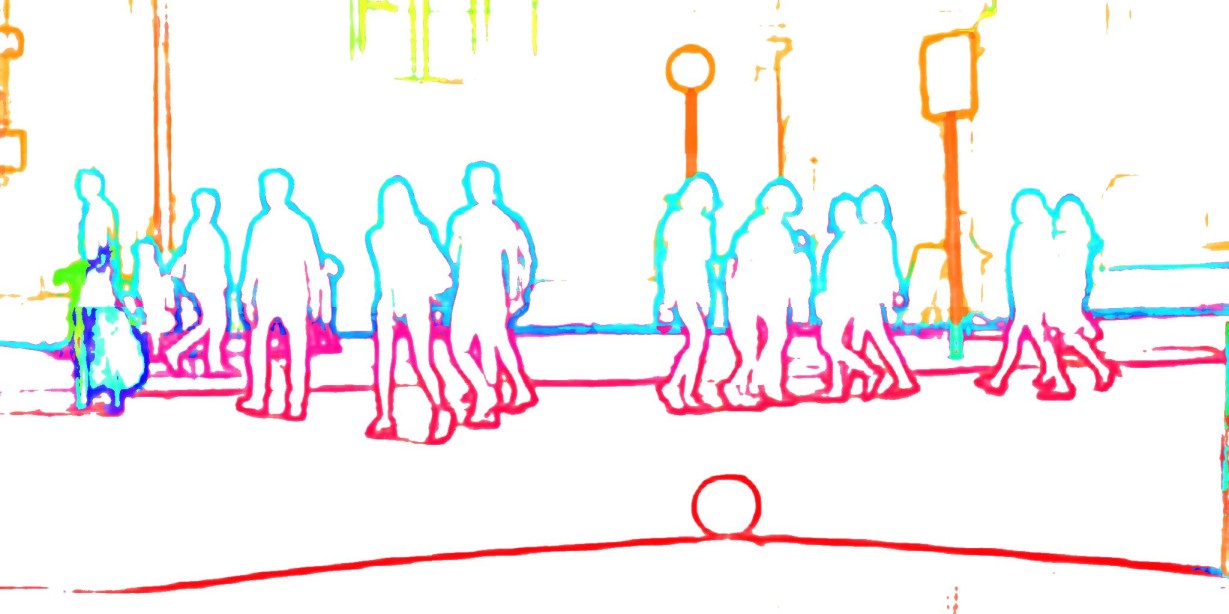}
\includegraphics[width=0.242\textwidth]{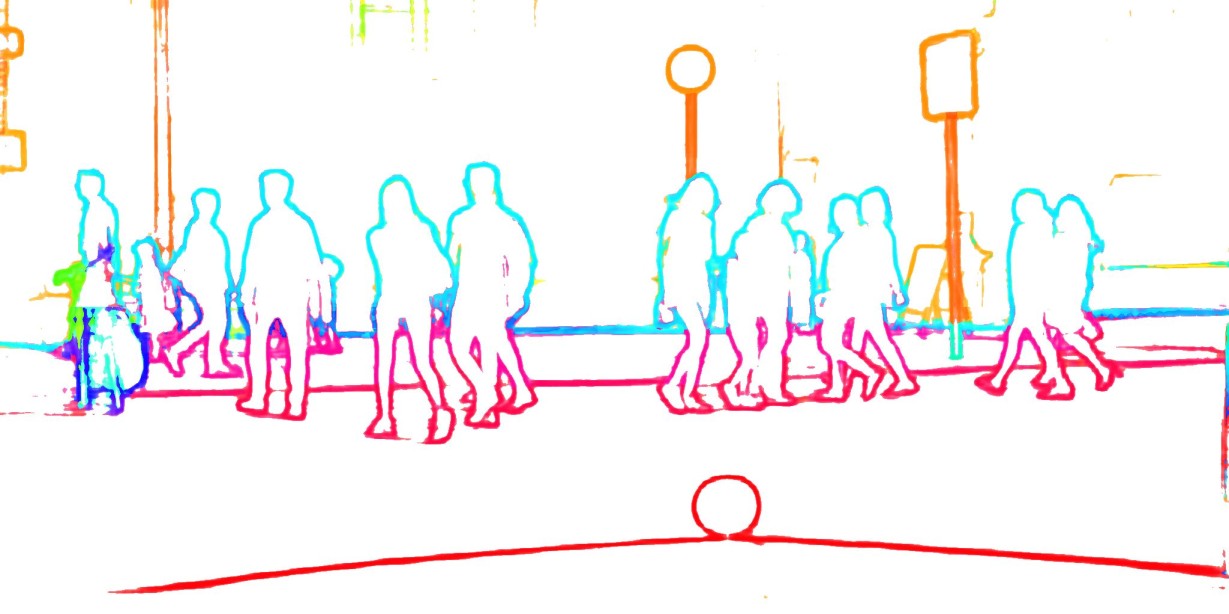}
	
\includegraphics[width=0.242\textwidth]{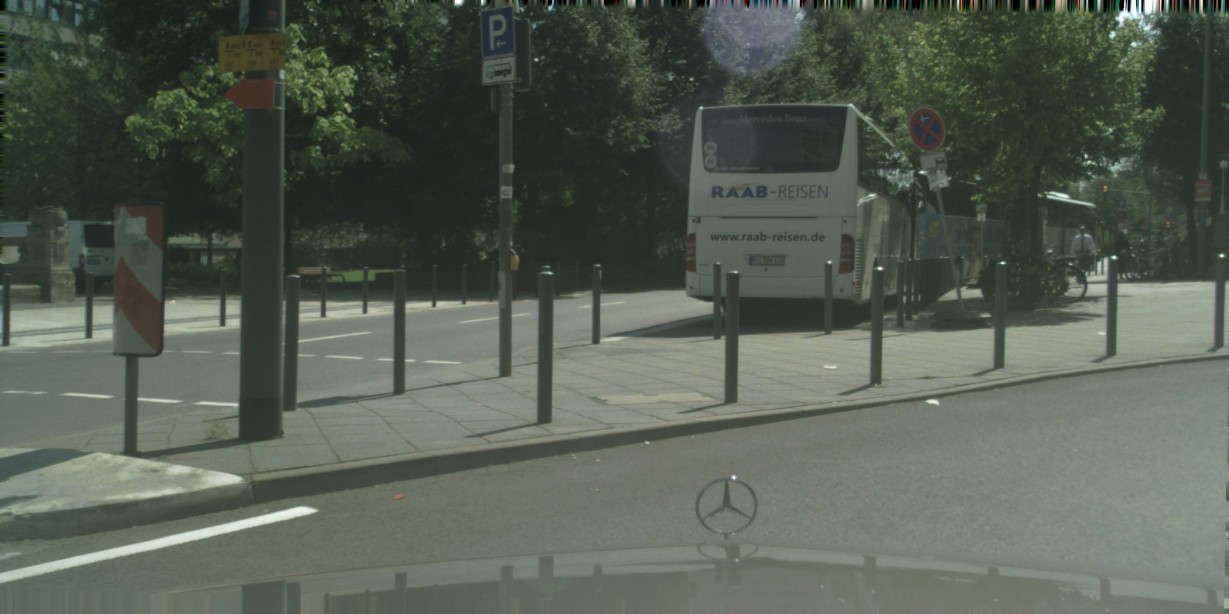}
\includegraphics[width=0.242\textwidth]{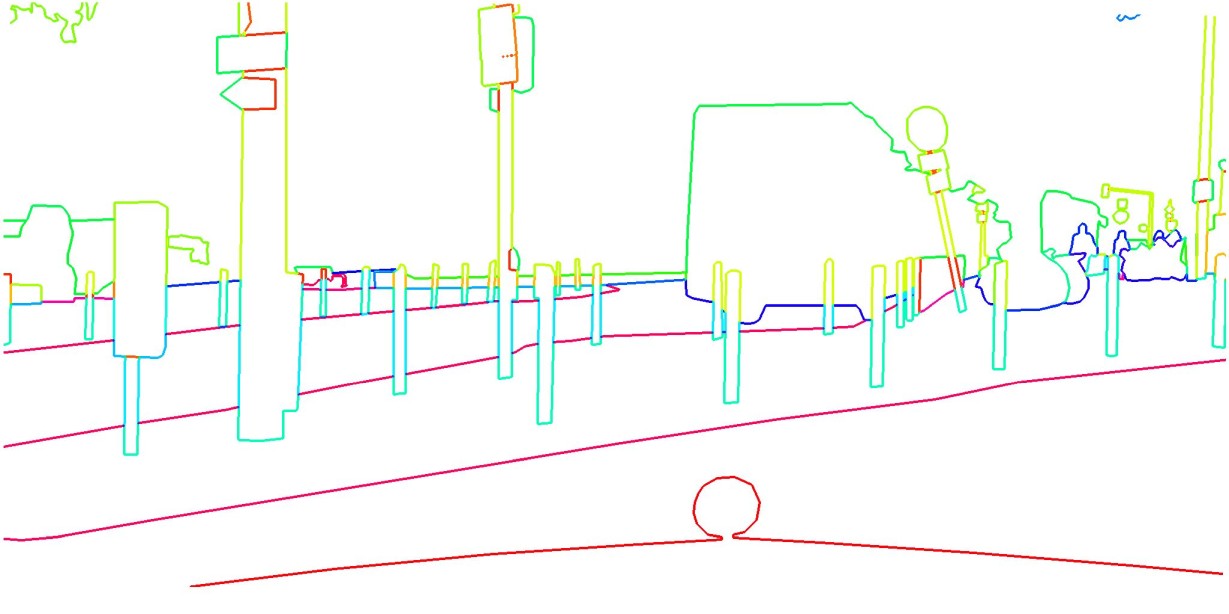}
\includegraphics[width=0.242\textwidth]{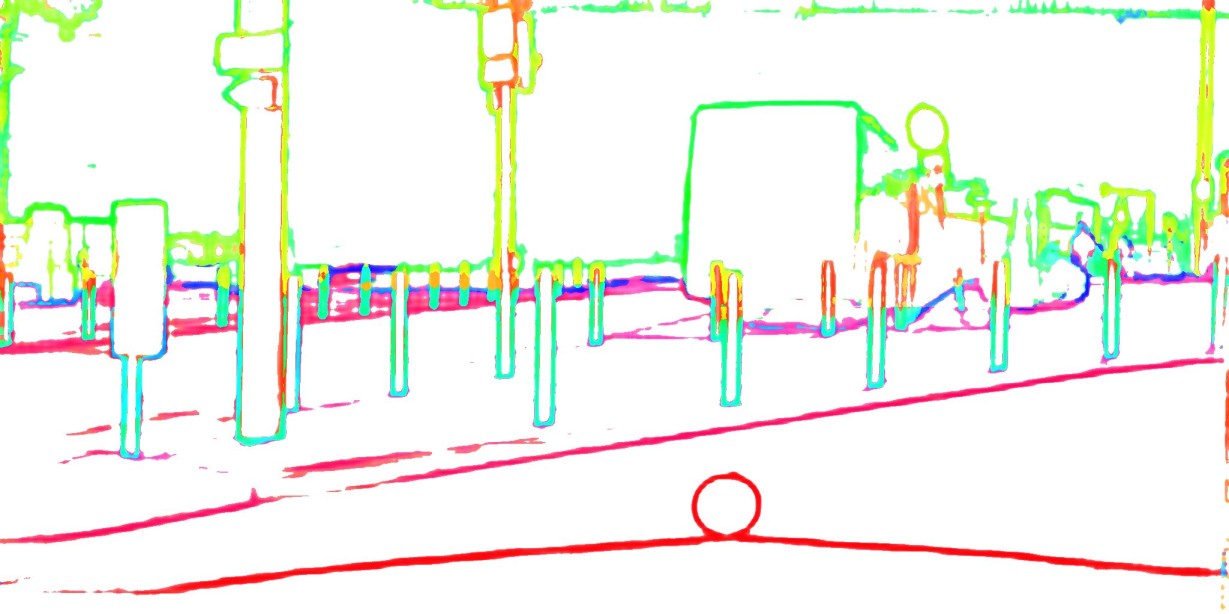}
\includegraphics[width=0.242\textwidth]{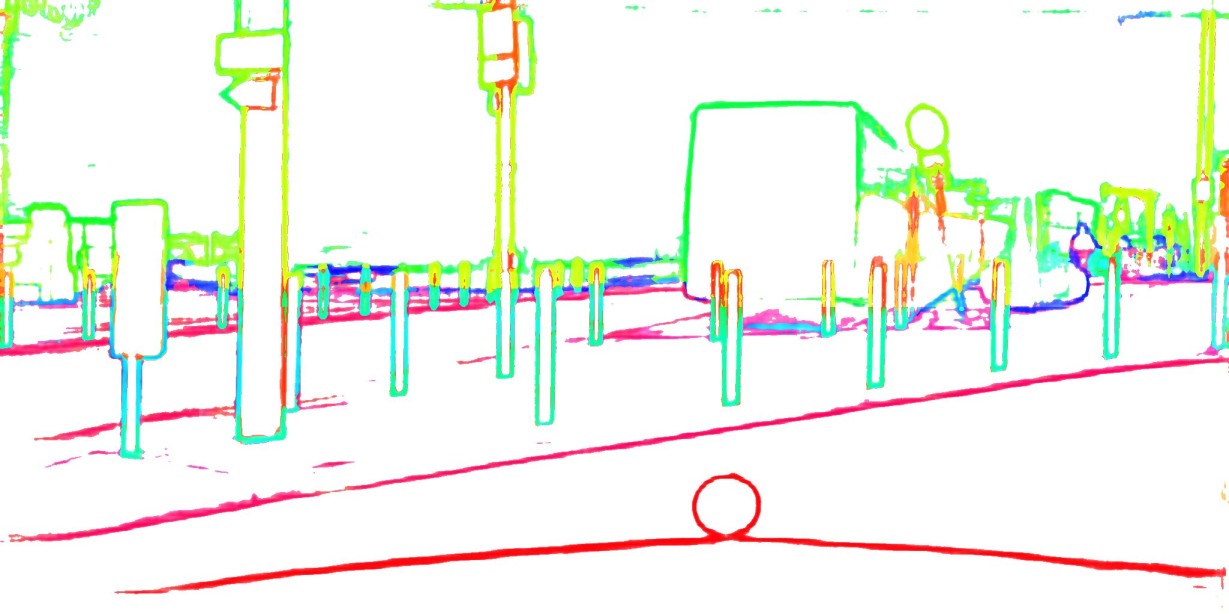}
	
\includegraphics[width=0.242\textwidth]{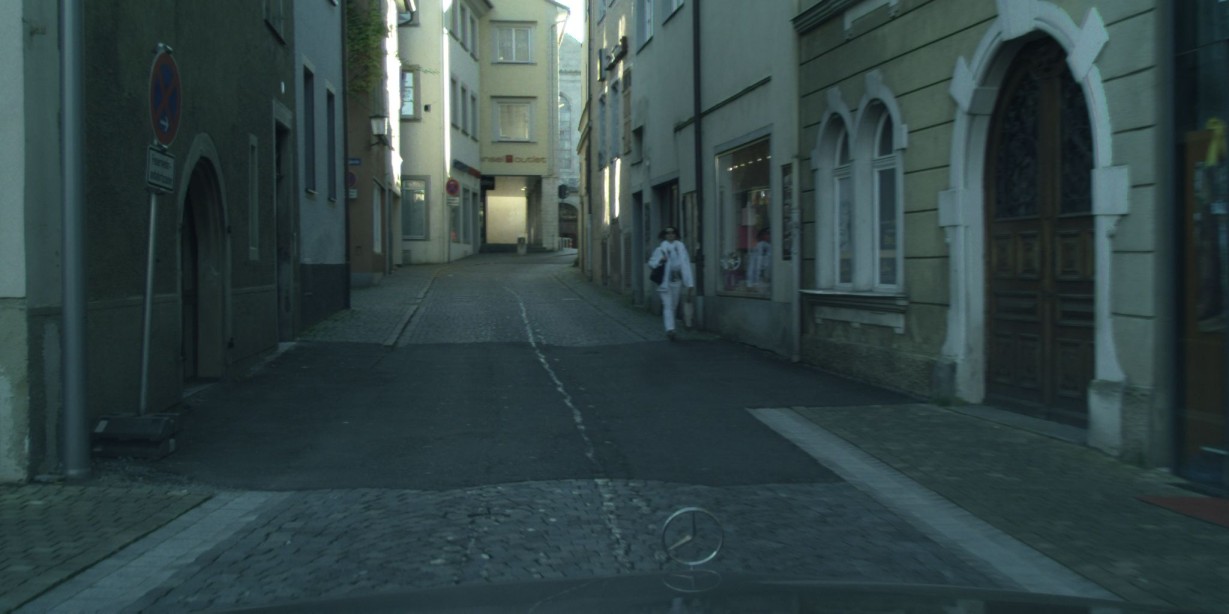}
\includegraphics[width=0.242\textwidth]{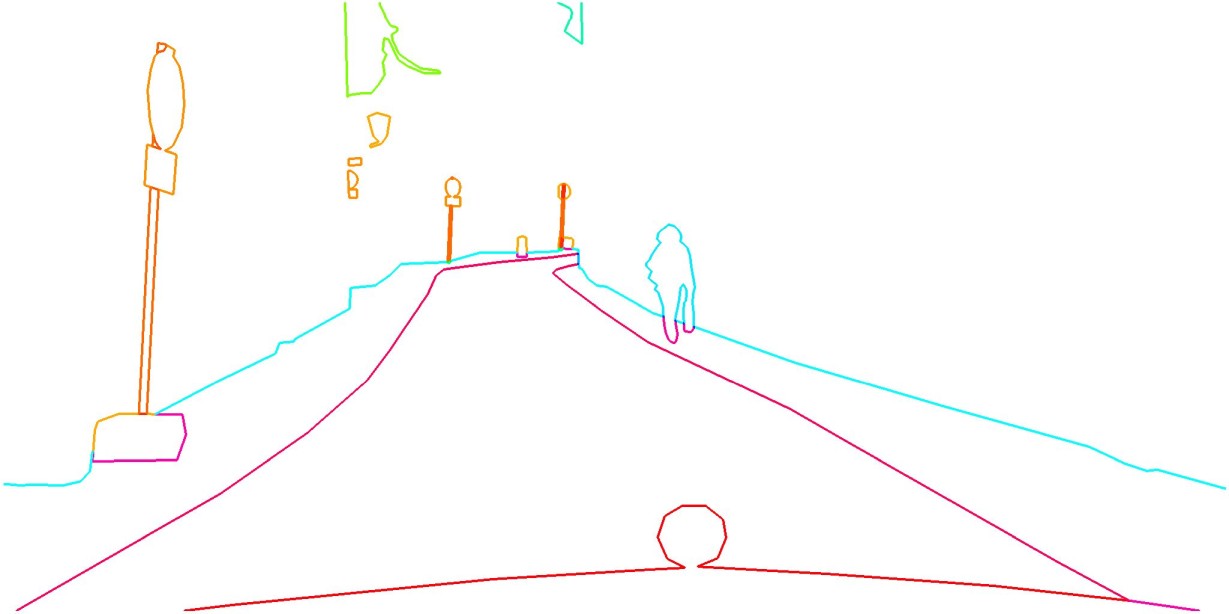}
\includegraphics[width=0.242\textwidth]{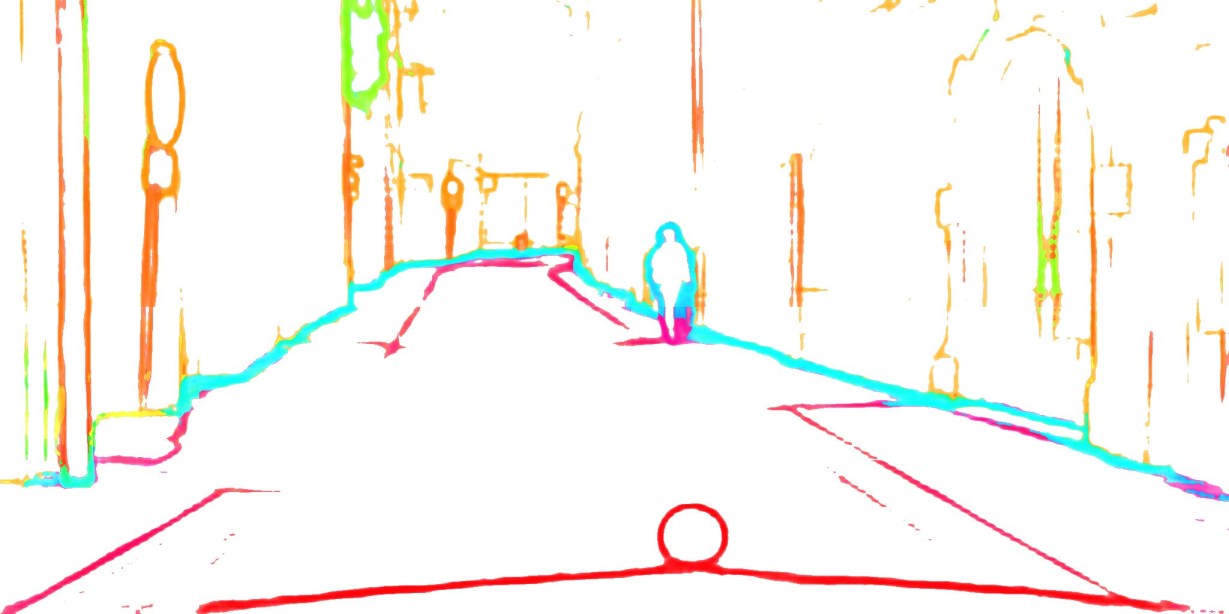}
\includegraphics[width=0.242\textwidth]{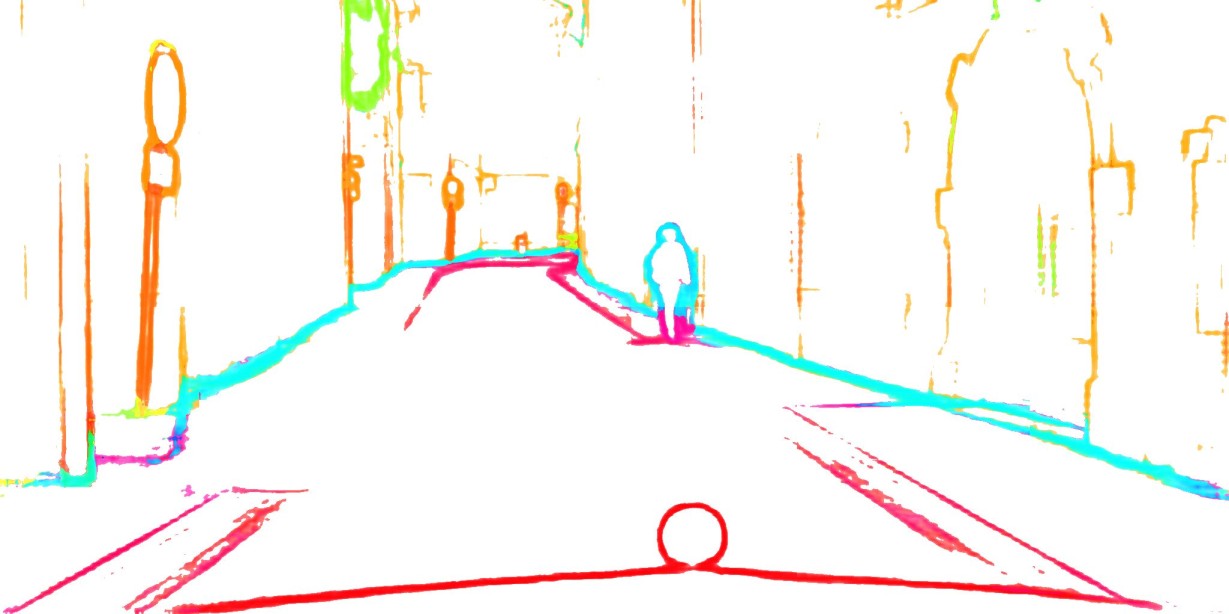}
\caption{Example results on Cityscapes. Columns from left to right: Input, Ground Truth, DSN and CASENet. CASENet shows better detection qualities on challenging objects, while DSN shows slightly more false positives on non-edge pixels. Best viewed in color.
\label{fig:cityscapes}}
\end{figure*}

We also train and test both DSN and CASENet with ResNet as base network on the Cityscapes. Compared to SBD, Cityscapes has relatively higher annotation quality but contains more challenging scenarios. The dataset contains more overlapping objects, which leads to more cases of multi-label semantic boundary pixels and thus may be better to test the proposed method. In Table~\ref{tb:main}, we provide both MF and AP of the comparing methods. To the best of our knowledge, this is the first paper quantitatively reporting the detection performance of category-wise semantic edges on Cityscapes. One could see CASENet consistently outperforms DSN in all classes with a significant margin.
Besides quantitative results, we also visualize some results in Fig.~\ref{fig:cityscapes} for qualitative comparisons.

\section{Concluding Remarks}
In this paper, we proposed an end-to-end deep network for category-aware semantic edge detection. We show that the proposed nested architecture, CASENet, shows improvements over some existing architectures popular in edge detection and segmentation. We also show that the proposed multi-label learning framework leads to better learning behaviors on edge detection. Our proposed method improves over previous state-of-the-art methods with significant margins. In the future, we plan to apply our method to other tasks such as stereo and semantic segmentation.
\FloatBarrier

\balance
{\small
	\bibliographystyle{ieee}
	\bibliography{egbib}
}

\newpage
\begin{appendices}

\section{Multi-label Edge Visualization}
In order to effectively visualize the prediction quality of multi-label semantic edges, the following color coding protocol is used to generate results in Fig.~\ref{fig:1}, Fig.~\ref{fig:loss}, and Fig.~\ref{fig:cityscapes}.
First, we associate each of the $K$ semantic object class a unique value of Hue, denoted as $\mathsf{H}\triangleq[\mathsf{H}_0,\mathsf{H}_1,\cdots,\mathsf{H}_{K-1}]$.
Given a $K$-channel output $\mathbf{Y}$ from our CASENet's fused classification module, where each element $\mathbf{Y}_k(\mathbf{p}) \in [0,1]$ denotes the pixel $\mathbf{p}$'s
predicted confidence of belonging to the $k$-th class, we return an HSV value for that pixel based on the following equations:
\begin{align}
\mathbf{H}(\mathbf{p}) & =  \frac{\sum_k\mathbf{Y}_k(\mathbf{p})\mathsf{H}_k}{\sum_k\mathbf{Y}_k}, \label{eq:viz_hue} \\
\mathbf{S}(\mathbf{p}) & =  255 \max \{\mathbf{Y}_k(\mathbf{p})|k=0,\cdots,K-1\}, \\
\mathbf{V}(\mathbf{p}) & =  255,
\end{align}
which is also how the ground truth color codes are computed (by using $\mathbf{\hat{Y}}$ instead). Note that the edge response maps of testing results are thresholded with 0.5, with the two classes having the strongest responses selected to compute hue based on Eq. (\ref{eq:viz_hue}).

For Cityscapes, we manually choose the following hue values to encode the 19 semantic classes so that the mixed Hue values highlight different multi-label edge types:
\begin{align}\label{eq:hue_for_cityscapes}
\mathsf{H}\triangleq [&359,320,40,80,90,10,20,30,140,340,\nonumber \\
&280,330,350,120,110,130,150,160,170]
\end{align}
The colors and their corresponding class names are illustrated in following Table~\ref{tb:hue}. The way Hue is mixed in equation~\ref{eq:viz_hue} indicates that any strong false positive response or incorrect response strength can lead to hue values shifted from ground truth. This helps to visualize false prediction.

\begin{table}[!h]
	\centering
	\definecolor{blk_color_0}{rgb}{1.000,0.000,0.031}
	\definecolor{blk_color_1}{rgb}{1.000,0.000,0.667}
	\definecolor{blk_color_2}{rgb}{1.000,0.667,0.000}
	\definecolor{blk_color_3}{rgb}{0.667,1.000,0.000}
	\definecolor{blk_color_4}{rgb}{0.498,1.000,0.000}
	\definecolor{blk_color_5}{rgb}{1.000,0.169,0.000}
	\definecolor{blk_color_6}{rgb}{1.000,0.333,0.000}
	\definecolor{blk_color_7}{rgb}{1.000,0.502,0.000}
	\definecolor{blk_color_8}{rgb}{0.000,1.000,0.333}
	\definecolor{blk_color_9}{rgb}{1.000,0.000,0.333}
	\definecolor{blk_color_10}{rgb}{0.667,0.000,1.000}
	\definecolor{blk_color_11}{rgb}{1.000,0.000,0.498}
	\definecolor{blk_color_12}{rgb}{1.000,0.000,0.165}
	\definecolor{blk_color_13}{rgb}{0.000,1.000,0.000}
	\definecolor{blk_color_14}{rgb}{0.165,1.000,0.000}
	\definecolor{blk_color_15}{rgb}{0.000,1.000,0.169}
	\definecolor{blk_color_16}{rgb}{0.000,1.000,0.502}
	\definecolor{blk_color_17}{rgb}{0.000,1.000,0.667}
	\definecolor{blk_color_18}{rgb}{0.000,1.000,0.835}
	\begin{tabular}{cccc}
		\cellcolor{blk_color_0}road &
		\cellcolor{blk_color_1}sidewalk&
		\cellcolor{blk_color_2}building&
		\cellcolor{blk_color_3}wall\\
		\cellcolor{blk_color_4}fence&
		\cellcolor{blk_color_5}pole&
		\cellcolor{blk_color_6}traffic light&
		\cellcolor{blk_color_7}traffic sign\\
		\cellcolor{blk_color_8}vegetation&
		\cellcolor{blk_color_9}terrain&
		\cellcolor{blk_color_10}sky&
		\cellcolor{blk_color_11}person\\
		\cellcolor{blk_color_12}rider&
		\cellcolor{blk_color_13}car&
		\cellcolor{blk_color_14}truck&
		\cellcolor{blk_color_15}bus\\
		\cellcolor{blk_color_16}train&
		\cellcolor{blk_color_17}motorcycle&
		\cellcolor{blk_color_18}bicycle
	\end{tabular}
	\caption{\label{tb:hue}The adopted color codes for Cityscapes semantic classes.}
\end{table}

\section{Additional Results on SBD}

\subsection{Early stage loss analysis}
\begin{figure}[!h]
	\centering
	\includegraphics[width=1\columnwidth]{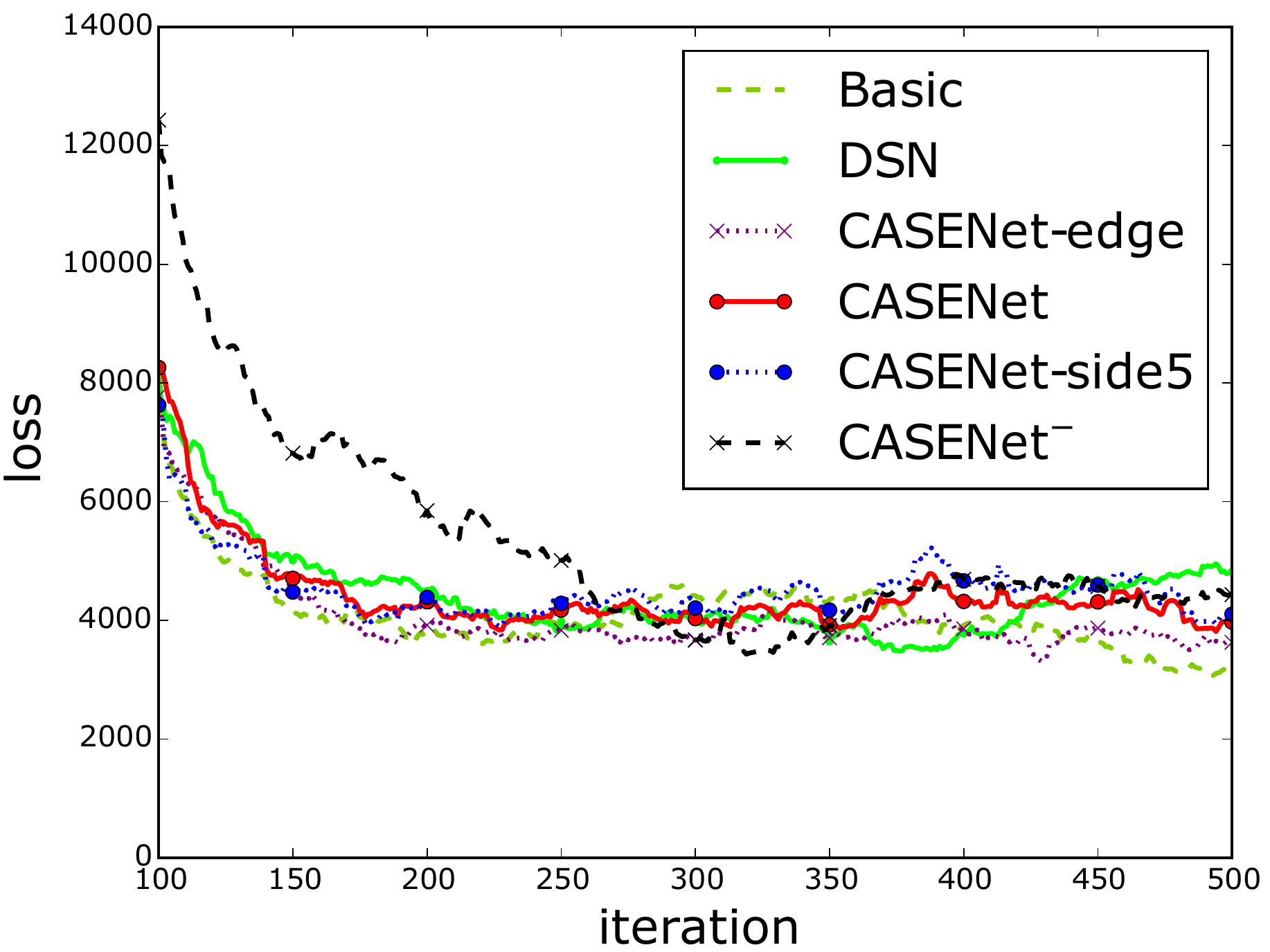}
	\caption{Early stage losses (up to 500 iterations) of different network configurations with a moving average kernel length of 100.}\label{fig:loss_early}
\end{figure}

Fig. \ref{fig:loss_early} shows the losses of different tested network configurations between iteration 100-500. Note that for Fig.~\ref{eq:loss}, loss curves between iteration 0-8000 is not available due to the large averaging kernel size. One can see CASENet's fused loss is initially larger than its side5 loss. It later drops faster and soon become consistently lower than the side5 loss (see Fig.~\ref{eq:loss}).

\subsection{Class-wise prediction examples}
\begin{figure*}[p]
	\centering
	\frame{\includegraphics[width=0.16\textwidth]{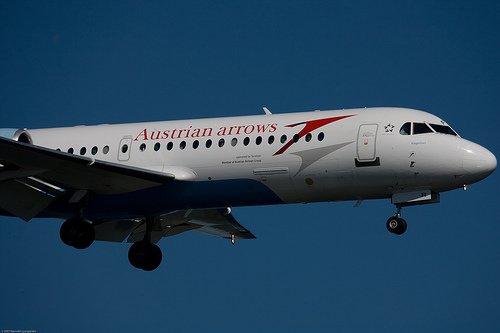}}
	\frame{\includegraphics[width=0.16\textwidth]{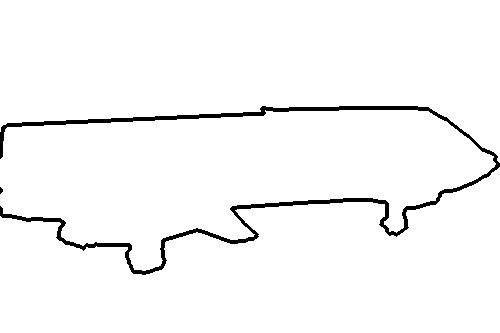}}
	\frame{\includegraphics[width=0.16\textwidth]{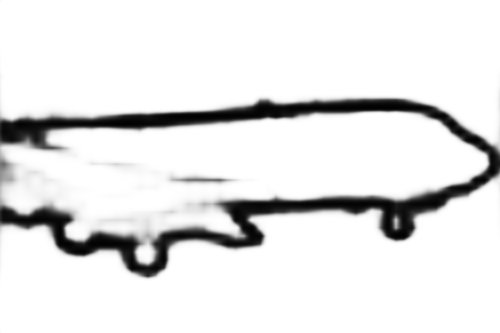}}
	\frame{\includegraphics[width=0.16\textwidth]{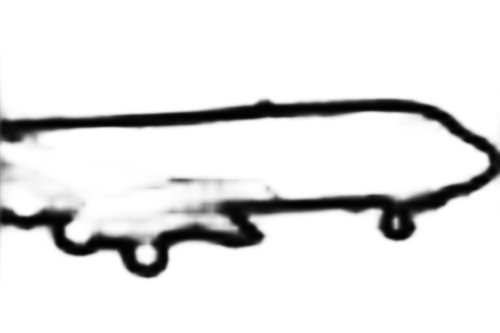}}
	\frame{\includegraphics[width=0.16\textwidth]{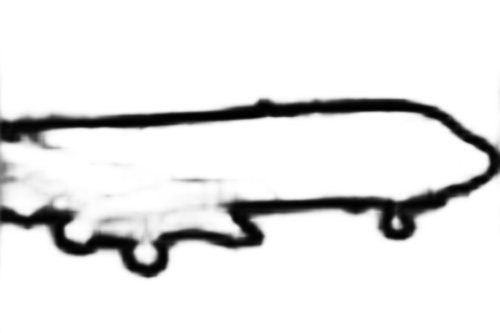}}
	\frame{\includegraphics[width=0.16\textwidth]{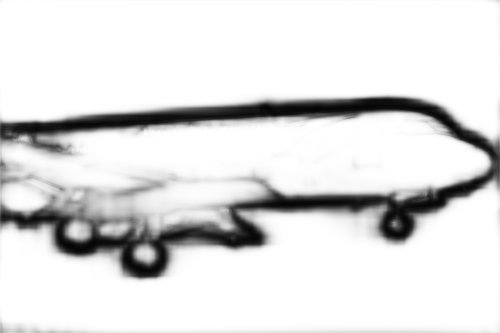}}\\
	\quad\\\vspace{-0.35cm}
	\frame{\includegraphics[width=0.16\textwidth]{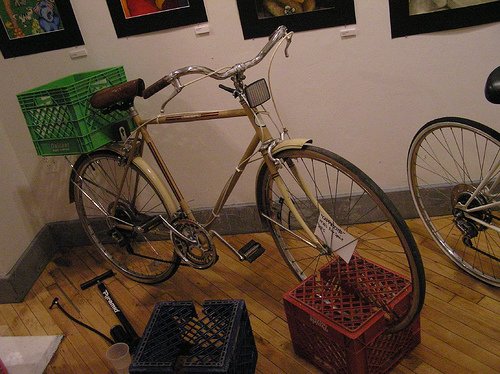}}
	\frame{\includegraphics[width=0.16\textwidth]{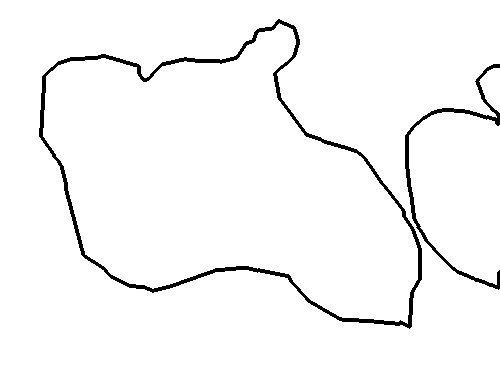}}
	\frame{\includegraphics[width=0.16\textwidth]{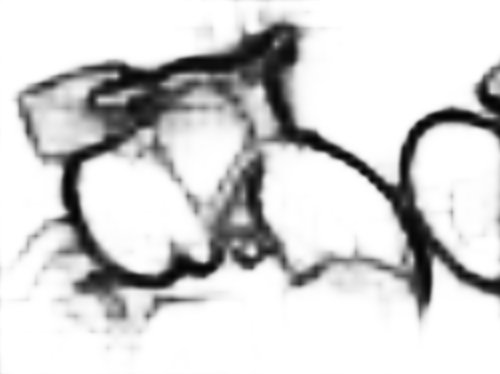}}
	\frame{\includegraphics[width=0.16\textwidth]{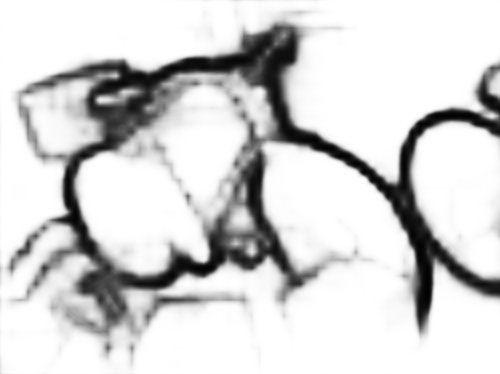}}
	\frame{\includegraphics[width=0.16\textwidth]{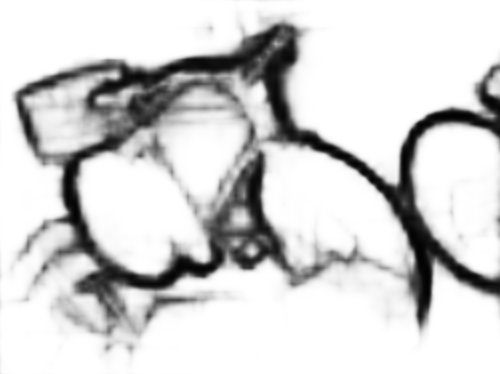}}
	\frame{\includegraphics[width=0.16\textwidth]{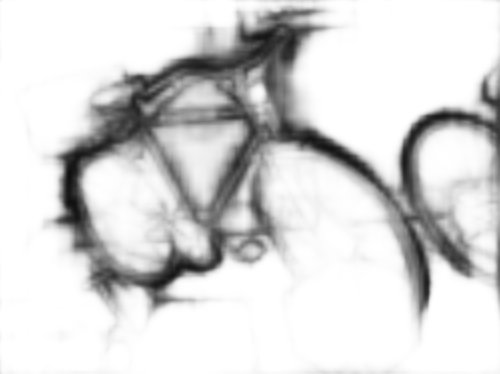}}\\
	\quad\\\vspace{-0.35cm}
	\frame{\includegraphics[width=0.16\textwidth]{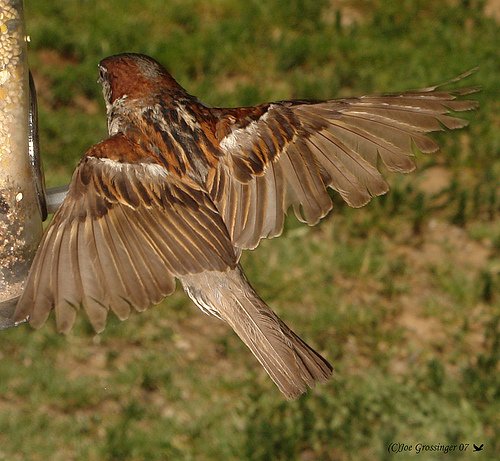}}
	\frame{\includegraphics[width=0.16\textwidth]{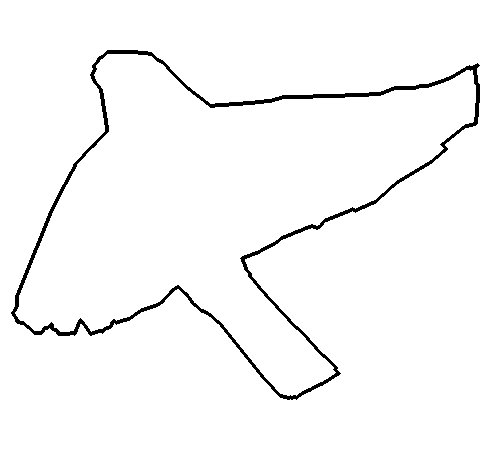}}
	\frame{\includegraphics[width=0.16\textwidth]{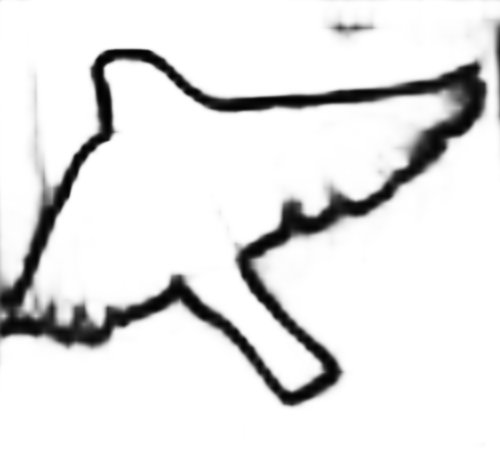}}
	\frame{\includegraphics[width=0.16\textwidth]{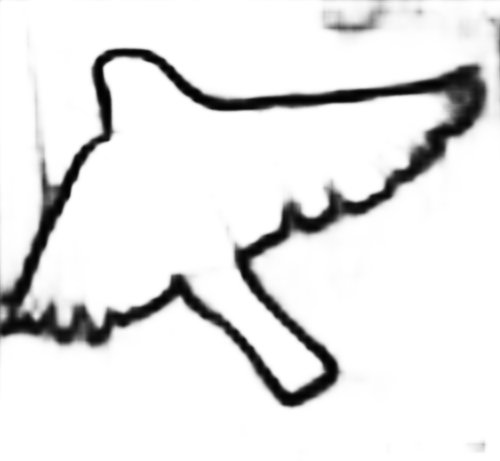}}
	\frame{\includegraphics[width=0.16\textwidth]{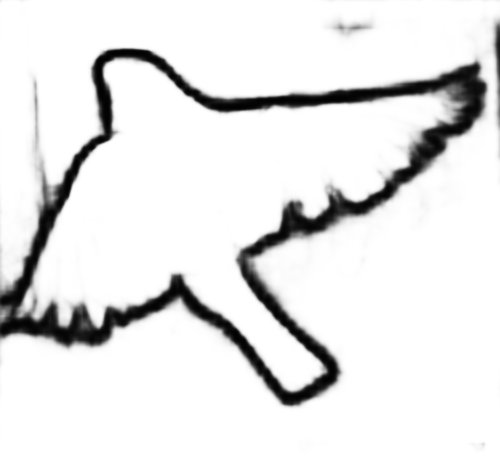}}
	\frame{\includegraphics[width=0.16\textwidth]{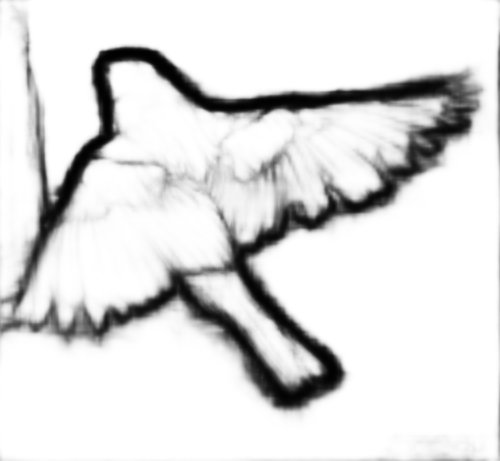}}\\
	\quad\\\vspace{-0.35cm}
	\frame{\includegraphics[width=0.16\textwidth]{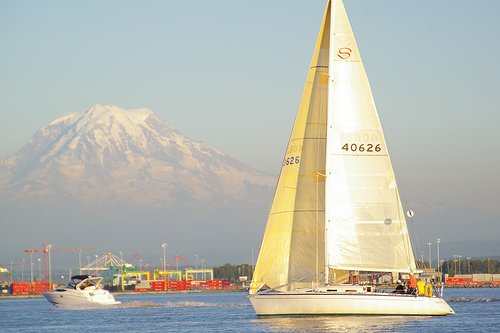}}
	\frame{\includegraphics[width=0.16\textwidth]{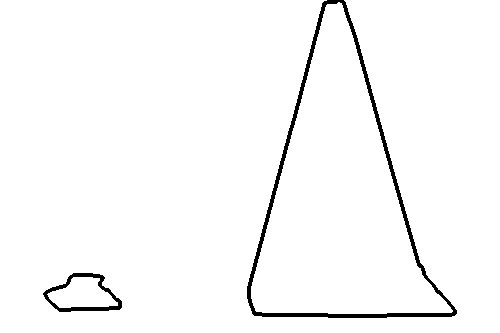}}
	\frame{\includegraphics[width=0.16\textwidth]{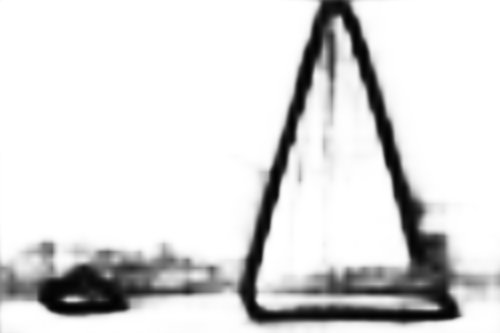}}
	\frame{\includegraphics[width=0.16\textwidth]{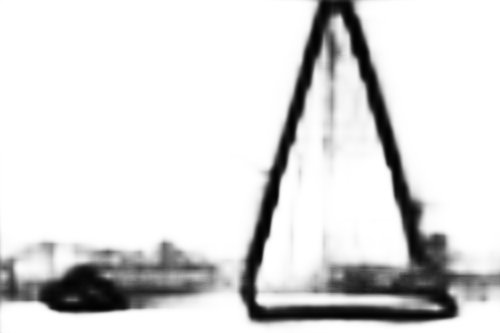}}
	\frame{\includegraphics[width=0.16\textwidth]{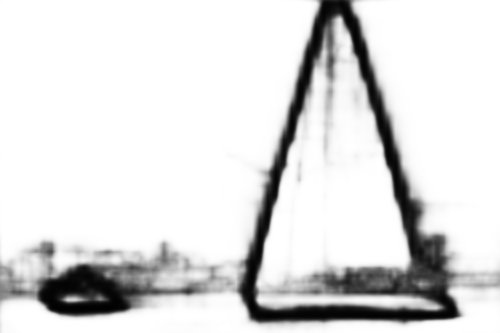}}
	\frame{\includegraphics[width=0.16\textwidth]{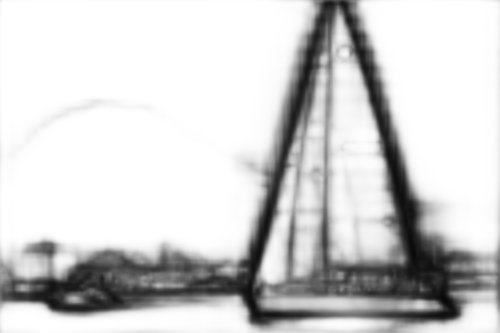}}\\
	\quad\\\vspace{-0.35cm}
	\frame{\includegraphics[width=0.16\textwidth]{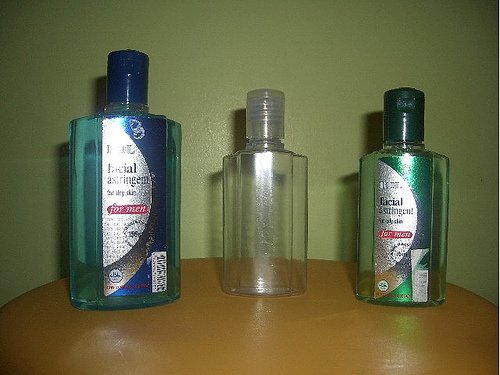}}
	\frame{\includegraphics[width=0.16\textwidth]{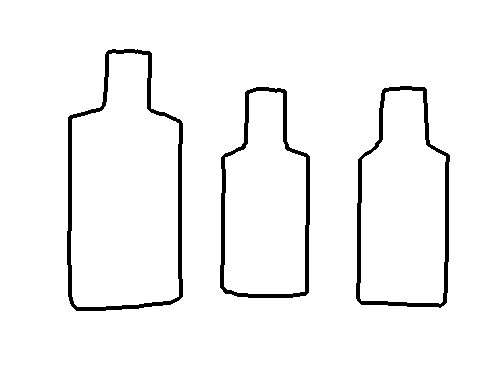}}
	\frame{\includegraphics[width=0.16\textwidth]{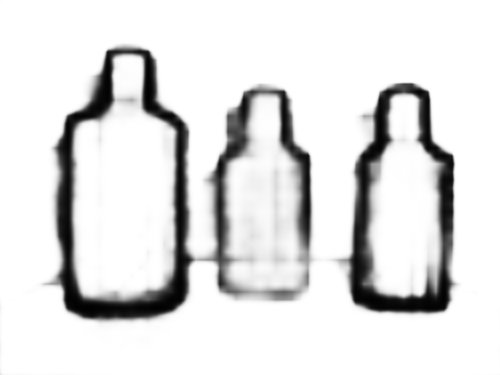}}
	\frame{\includegraphics[width=0.16\textwidth]{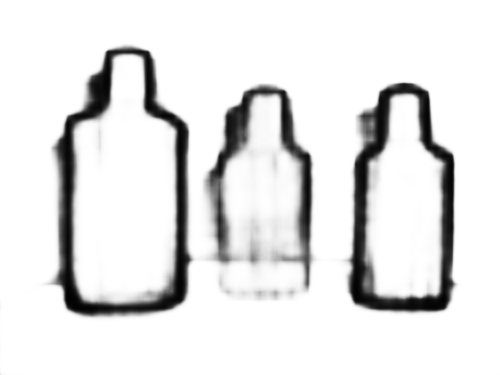}}
	\frame{\includegraphics[width=0.16\textwidth]{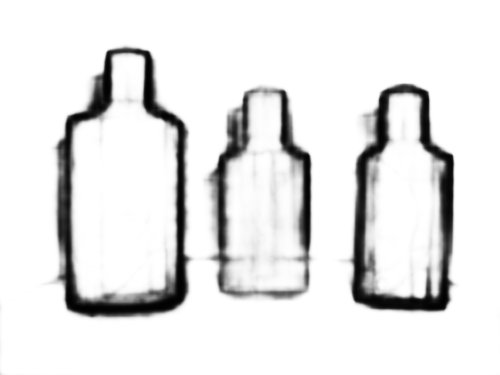}}
	\frame{\includegraphics[width=0.16\textwidth]{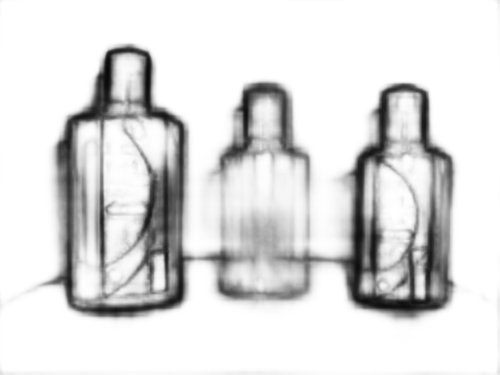}}\\
	\quad\\\vspace{-0.35cm}
	\frame{\includegraphics[width=0.16\textwidth]{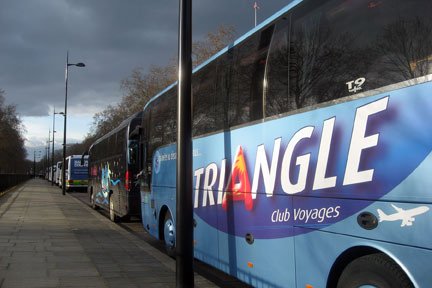}}
	\frame{\includegraphics[width=0.16\textwidth]{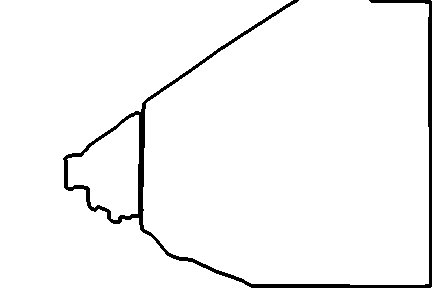}}
	\frame{\includegraphics[width=0.16\textwidth]{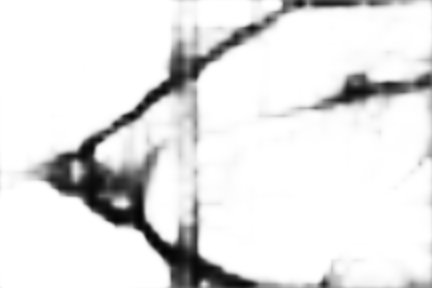}}
	\frame{\includegraphics[width=0.16\textwidth]{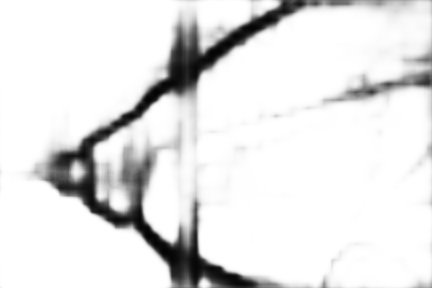}}
	\frame{\includegraphics[width=0.16\textwidth]{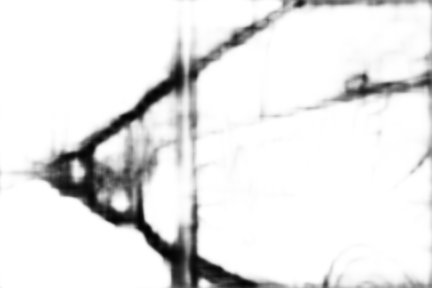}}
	\frame{\includegraphics[width=0.16\textwidth]{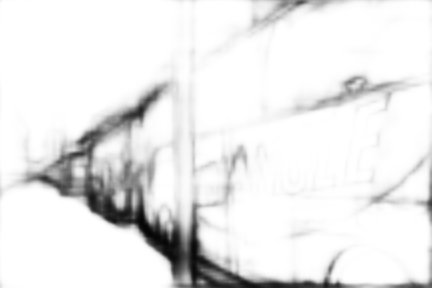}}\\
	\quad\\\vspace{-0.35cm}
	\frame{\includegraphics[width=0.16\textwidth]{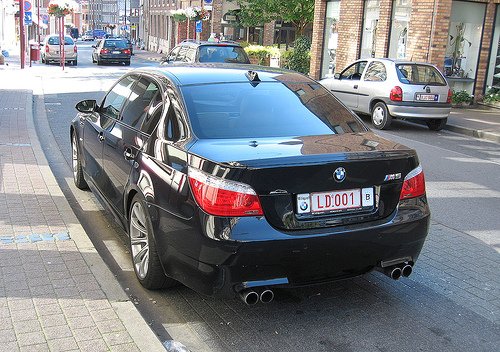}}
	\frame{\includegraphics[width=0.16\textwidth]{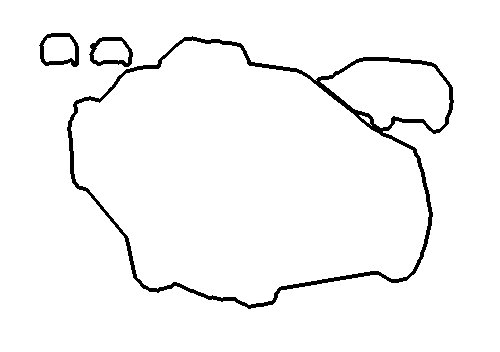}}
	\frame{\includegraphics[width=0.16\textwidth]{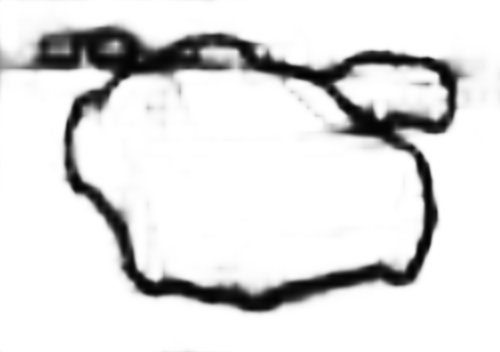}}
	\frame{\includegraphics[width=0.16\textwidth]{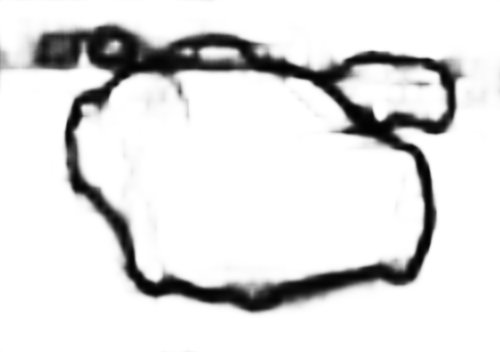}}
	\frame{\includegraphics[width=0.16\textwidth]{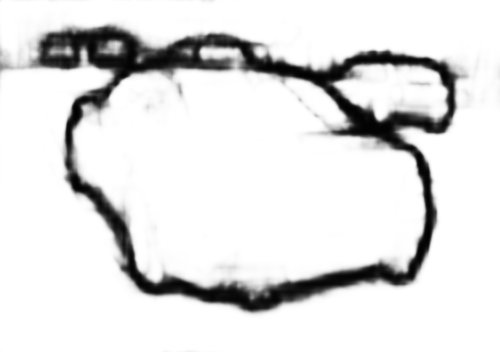}}
	\frame{\includegraphics[width=0.16\textwidth]{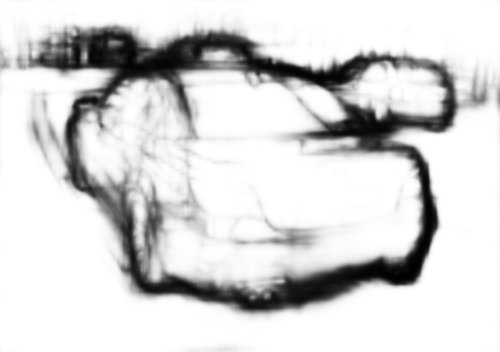}}\\
	\quad\\\vspace{-0.35cm}
	\frame{\includegraphics[width=0.16\textwidth]{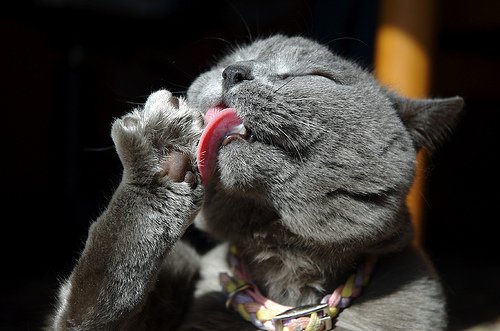}}
	\frame{\includegraphics[width=0.16\textwidth]{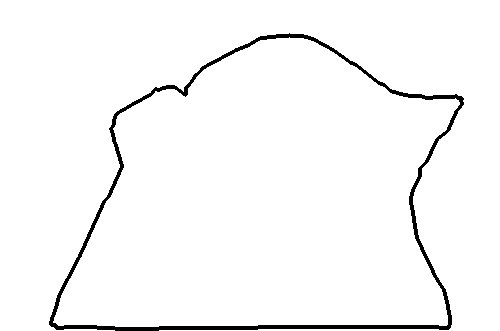}}
	\frame{\includegraphics[width=0.16\textwidth]{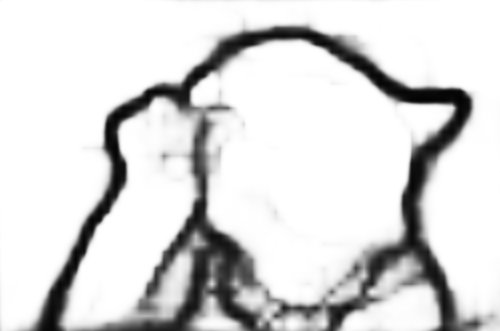}}
	\frame{\includegraphics[width=0.16\textwidth]{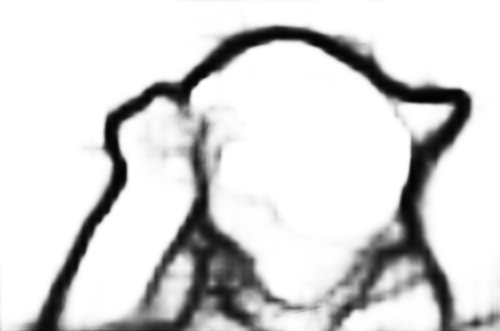}}
	\frame{\includegraphics[width=0.16\textwidth]{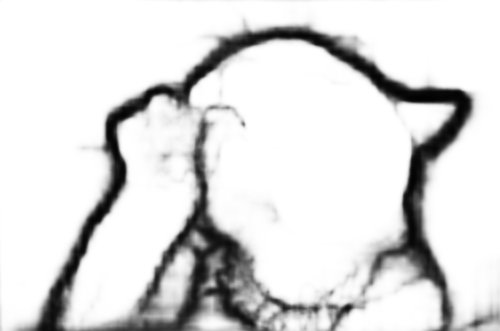}}
	\frame{\includegraphics[width=0.16\textwidth]{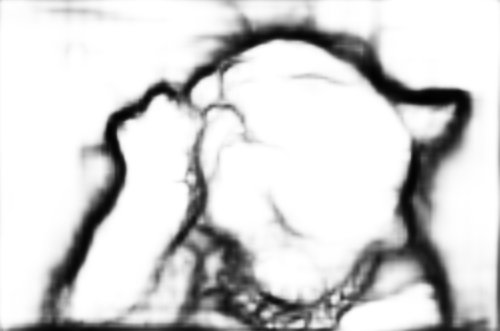}}\\
	\quad\\\vspace{-0.35cm}
	\frame{\includegraphics[width=0.16\textwidth]{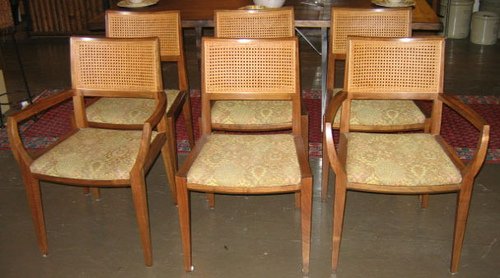}}
	\frame{\includegraphics[width=0.16\textwidth]{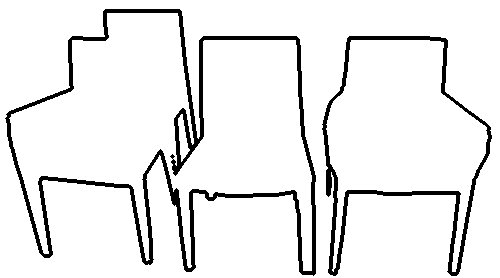}}
	\frame{\includegraphics[width=0.16\textwidth]{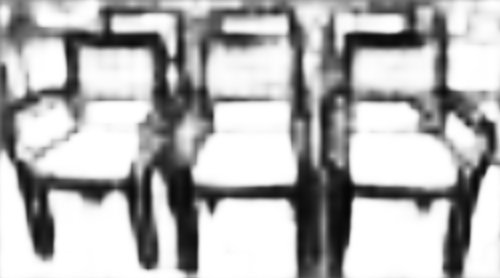}}
	\frame{\includegraphics[width=0.16\textwidth]{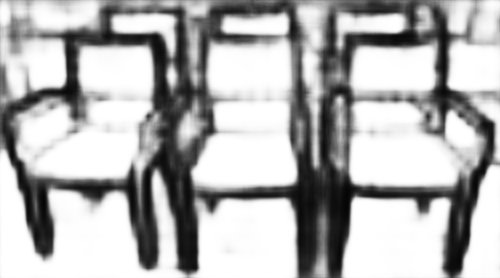}}
	\frame{\includegraphics[width=0.16\textwidth]{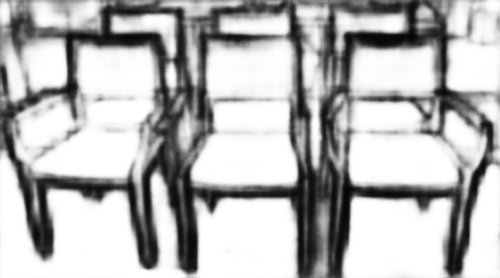}}
	\frame{\includegraphics[width=0.16\textwidth]{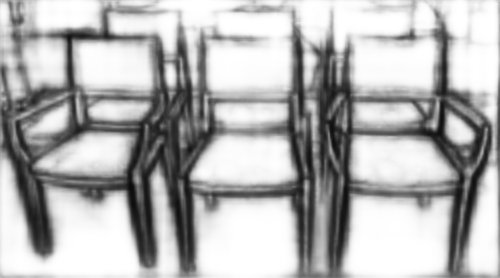}}\\
	\quad\\\vspace{-0.35cm}
	\frame{\includegraphics[width=0.16\textwidth]{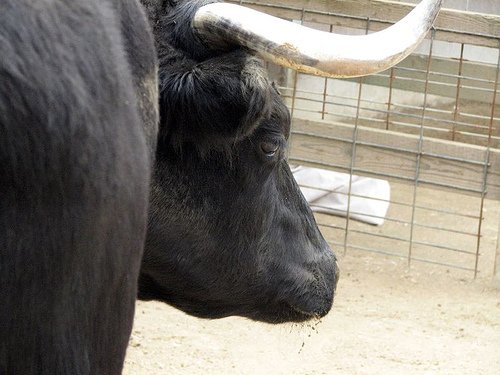}}
	\frame{\includegraphics[width=0.16\textwidth]{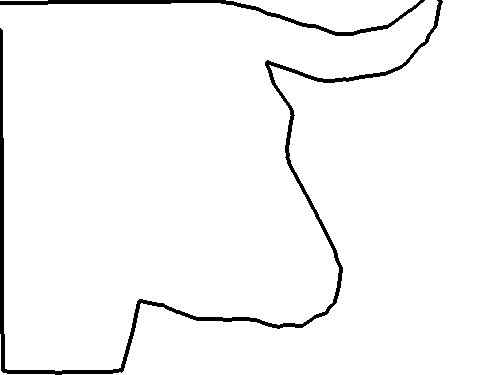}}
	\frame{\includegraphics[width=0.16\textwidth]{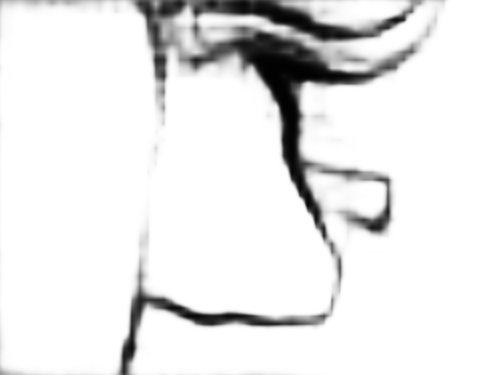}}
	\frame{\includegraphics[width=0.16\textwidth]{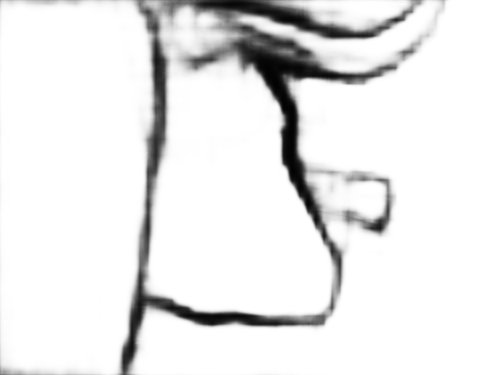}}
	\frame{\includegraphics[width=0.16\textwidth]{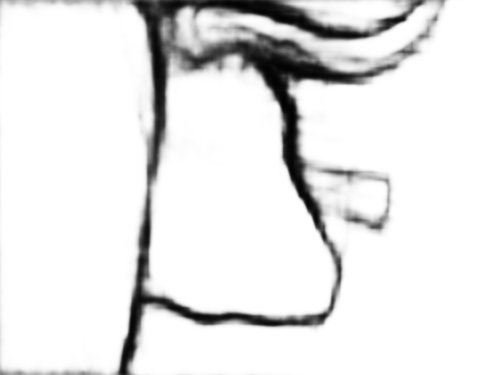}}
	\frame{\includegraphics[width=0.16\textwidth]{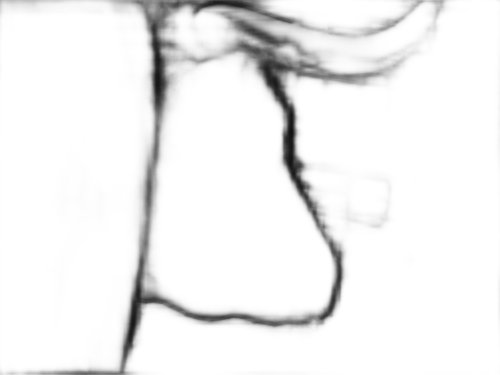}}
	\caption{Class-wise prediction results of comparing methods on the SBD Dataset. Rows correspond to the predicted edges of ``aeroplane'', ``bicycle'', ``bird'', ``boat'', ``bottle'', ``bus'', ``car'', ``cat'', ``chair'' and ``cow''. Columns correspond to original image, ground truth, and results of Basic, DSN, CASENet and CASENet-VGG.}\label{fig:qual_sbd1}
\end{figure*}

\begin{figure*}[p]
	\centering
	\frame{\includegraphics[width=0.16\textwidth]{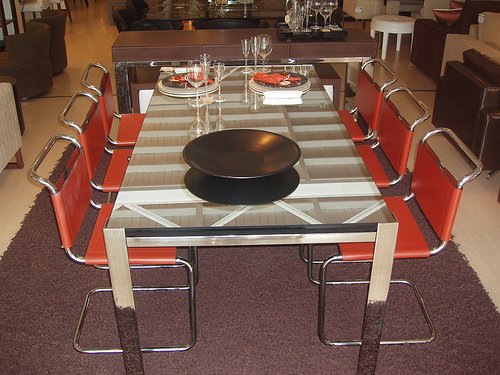}}
	\frame{\includegraphics[width=0.16\textwidth]{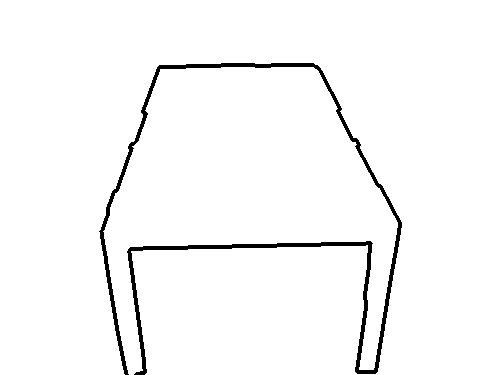}}
	\frame{\includegraphics[width=0.16\textwidth]{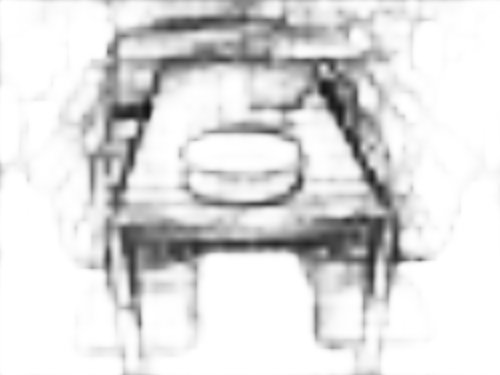}}
	\frame{\includegraphics[width=0.16\textwidth]{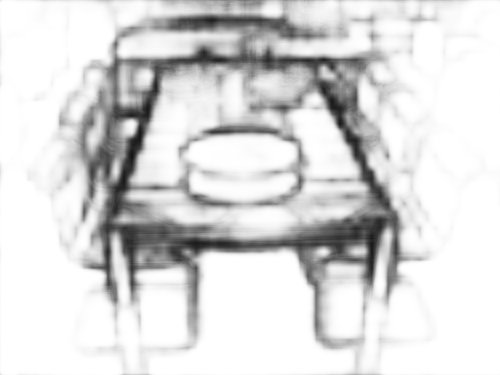}}
	\frame{\includegraphics[width=0.16\textwidth]{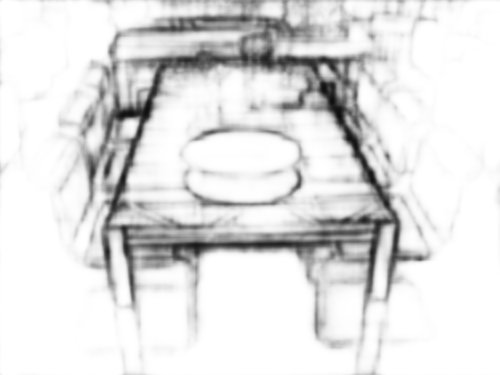}}
	\frame{\includegraphics[width=0.16\textwidth]{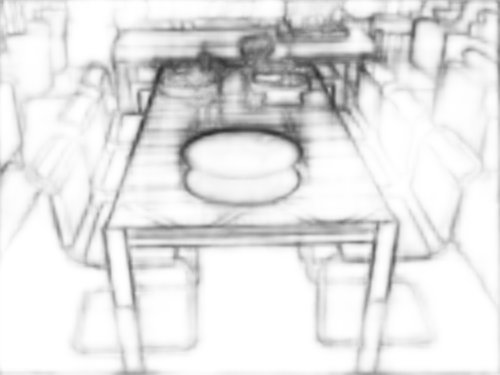}}\\
	\quad\\\vspace{-0.35cm}
	\frame{\includegraphics[width=0.16\textwidth]{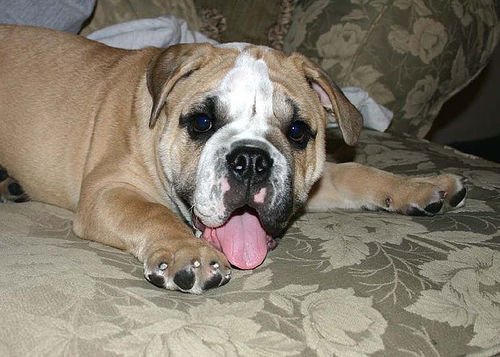}}
	\frame{\includegraphics[width=0.16\textwidth]{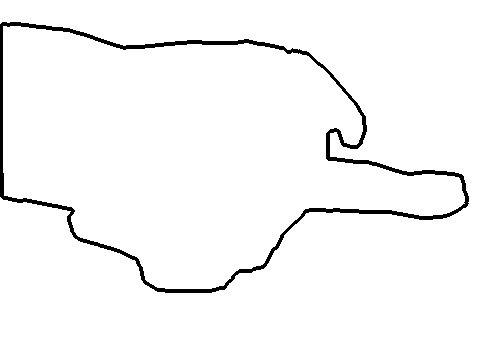}}
	\frame{\includegraphics[width=0.16\textwidth]{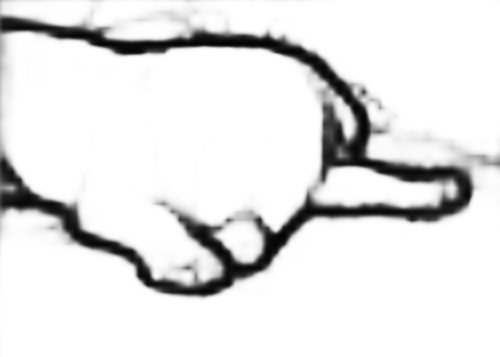}}
	\frame{\includegraphics[width=0.16\textwidth]{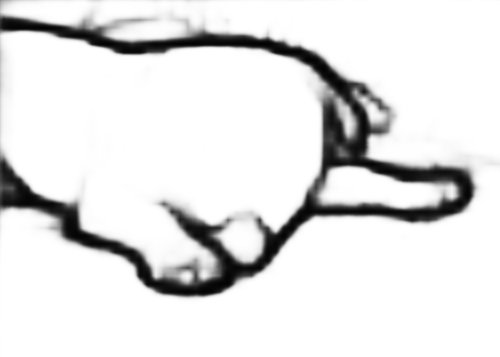}}
	\frame{\includegraphics[width=0.16\textwidth]{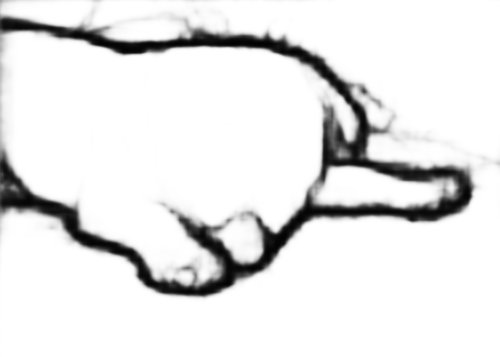}}
	\frame{\includegraphics[width=0.16\textwidth]{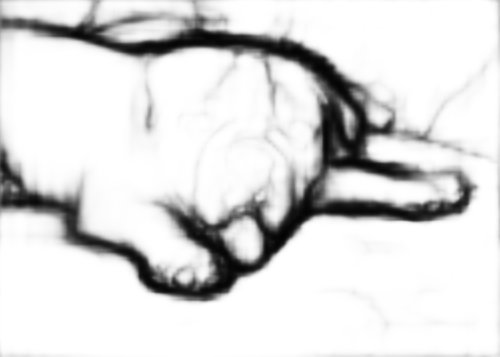}}\\
	\quad\\\vspace{-0.35cm}
	\frame{\includegraphics[width=0.16\textwidth]{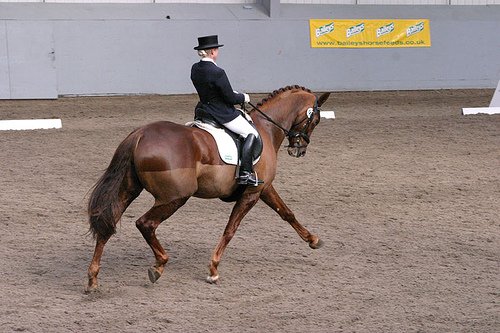}}
	\frame{\includegraphics[width=0.16\textwidth]{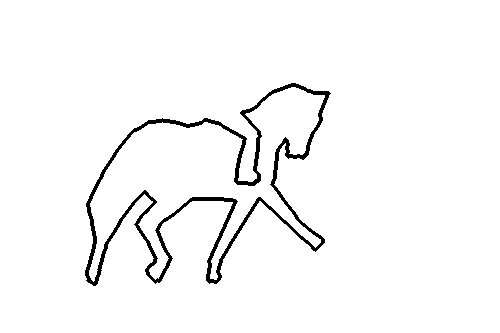}}
	\frame{\includegraphics[width=0.16\textwidth]{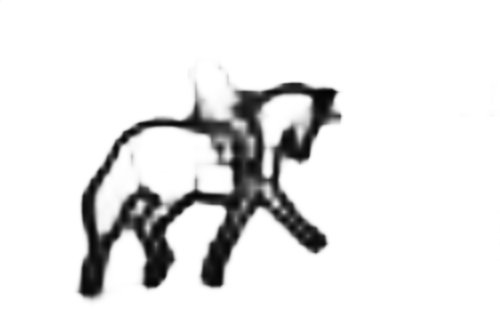}}
	\frame{\includegraphics[width=0.16\textwidth]{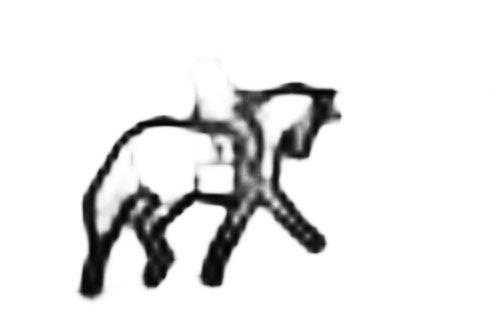}}
	\frame{\includegraphics[width=0.16\textwidth]{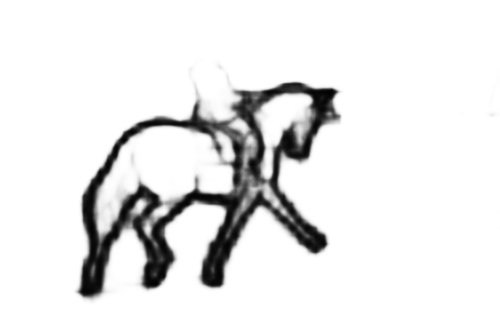}}
	\frame{\includegraphics[width=0.16\textwidth]{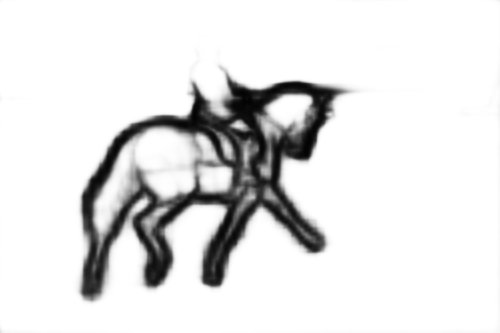}}\\
	\quad\\\vspace{-0.35cm}
	\frame{\includegraphics[width=0.16\textwidth]{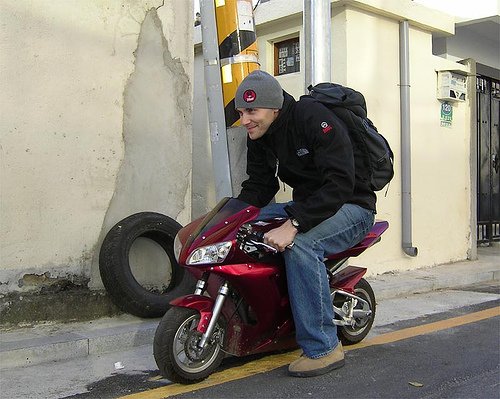}}
	\frame{\includegraphics[width=0.16\textwidth]{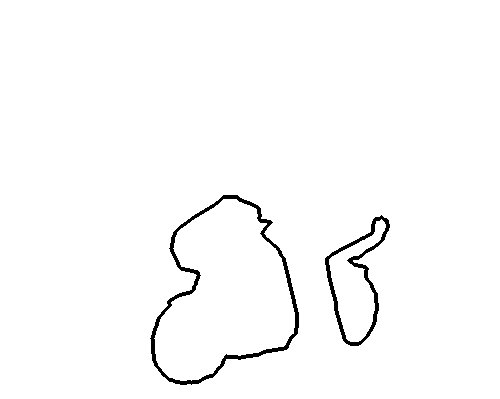}}
	\frame{\includegraphics[width=0.16\textwidth]{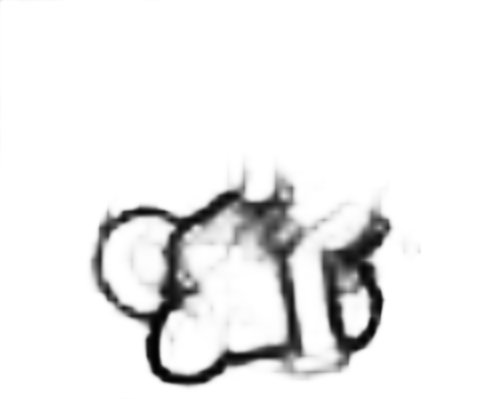}}
	\frame{\includegraphics[width=0.16\textwidth]{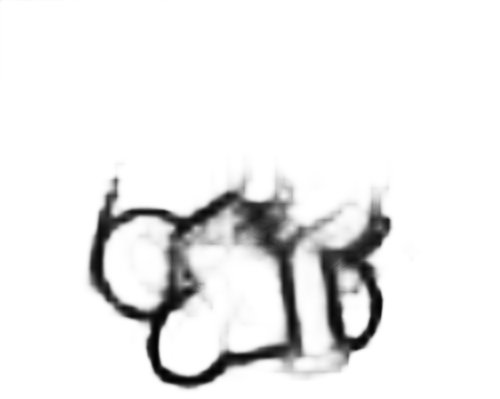}}
	\frame{\includegraphics[width=0.16\textwidth]{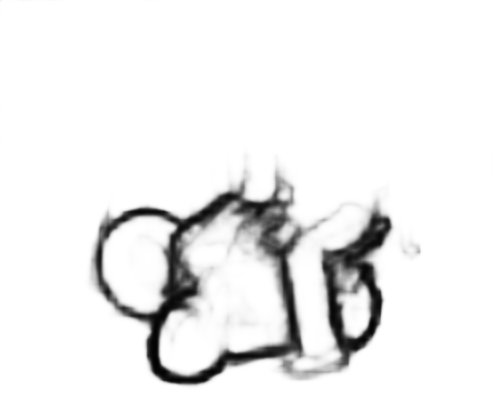}}
	\frame{\includegraphics[width=0.16\textwidth]{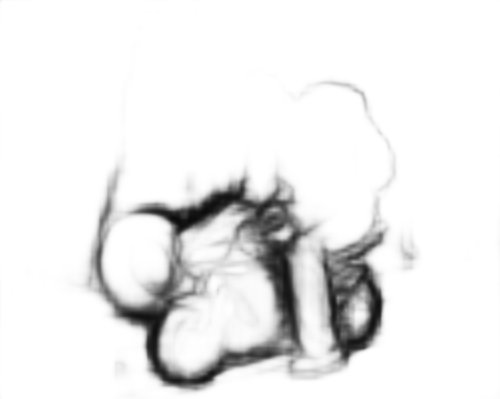}}\\
	\quad\\\vspace{-0.35cm}
	\frame{\includegraphics[width=0.16\textwidth]{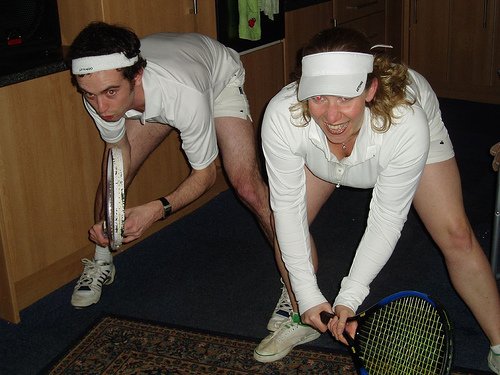}}
	\frame{\includegraphics[width=0.16\textwidth]{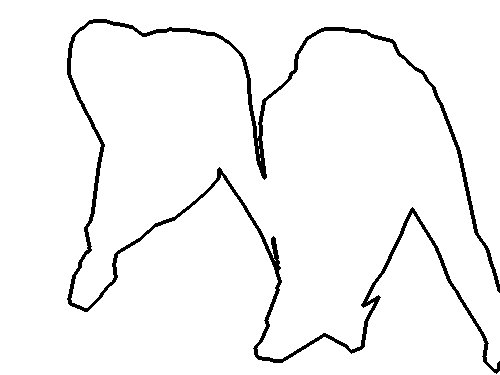}}
	\frame{\includegraphics[width=0.16\textwidth]{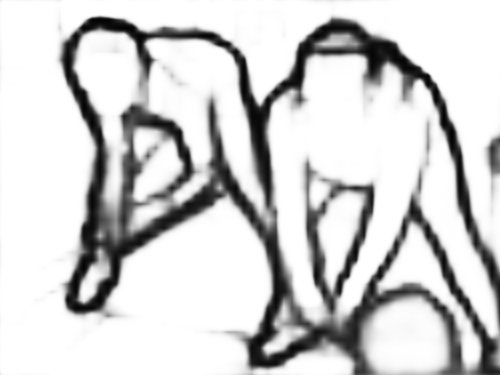}}
	\frame{\includegraphics[width=0.16\textwidth]{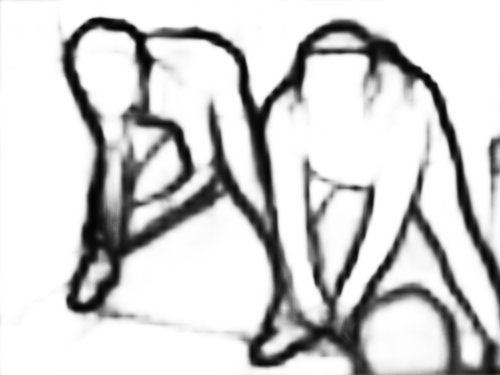}}
	\frame{\includegraphics[width=0.16\textwidth]{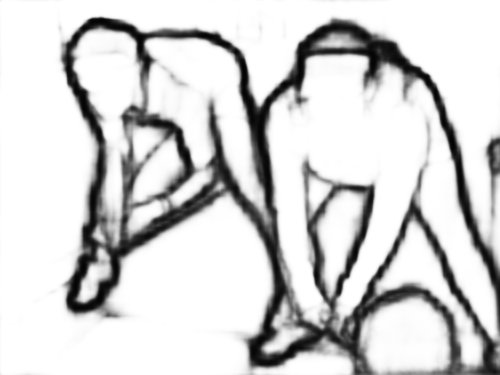}}
	\frame{\includegraphics[width=0.16\textwidth]{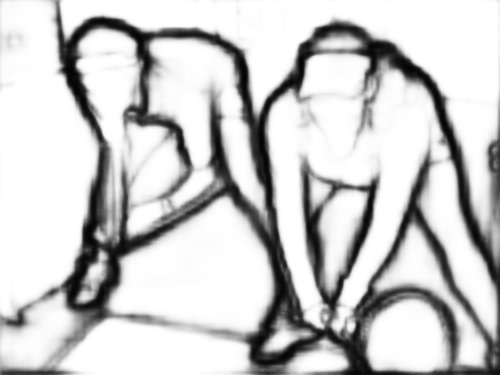}}\\
	\quad\\\vspace{-0.35cm}
	\frame{\includegraphics[width=0.16\textwidth]{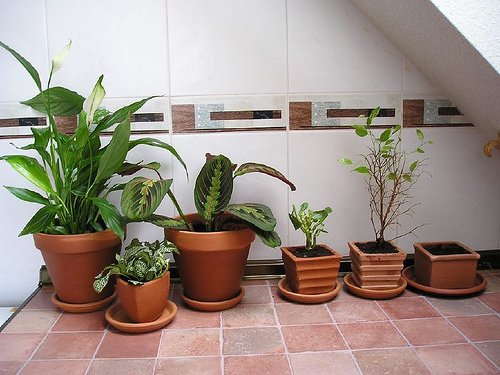}}
	\frame{\includegraphics[width=0.16\textwidth]{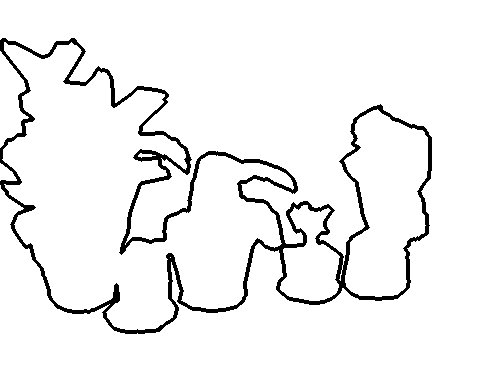}}
	\frame{\includegraphics[width=0.16\textwidth]{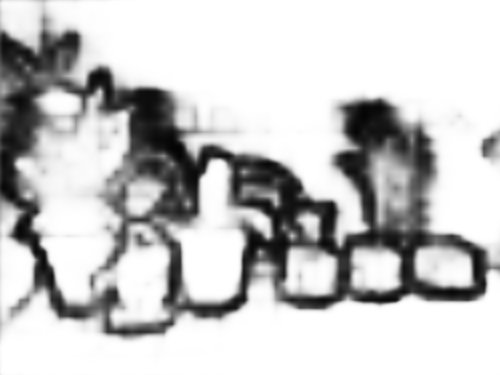}}
	\frame{\includegraphics[width=0.16\textwidth]{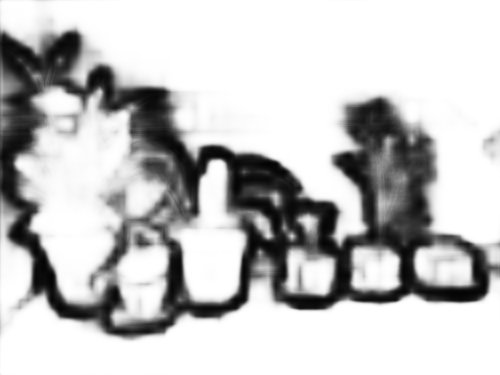}}
	\frame{\includegraphics[width=0.16\textwidth]{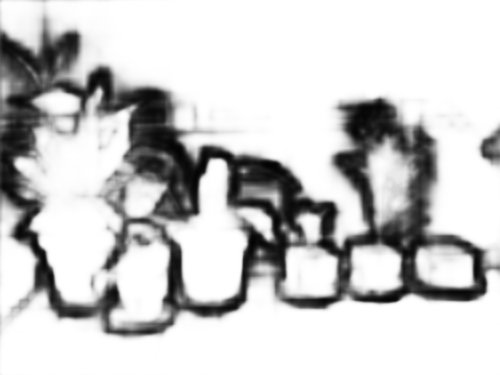}}
	\frame{\includegraphics[width=0.16\textwidth]{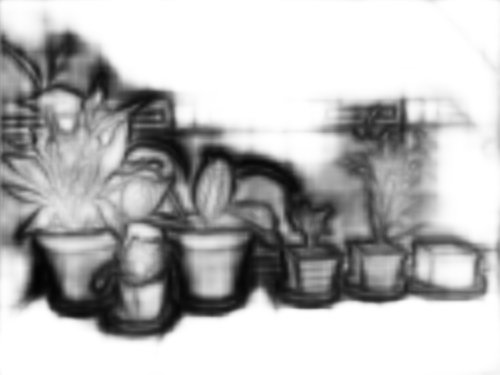}}\\
	\quad\\\vspace{-0.35cm}
	\frame{\includegraphics[width=0.16\textwidth]{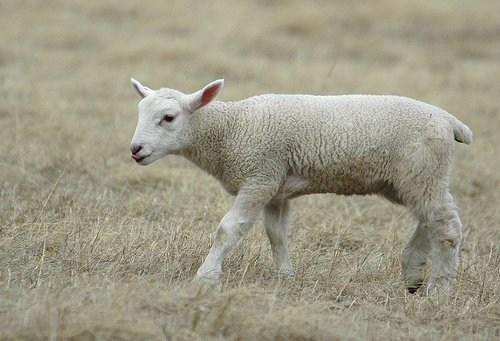}}
	\frame{\includegraphics[width=0.16\textwidth]{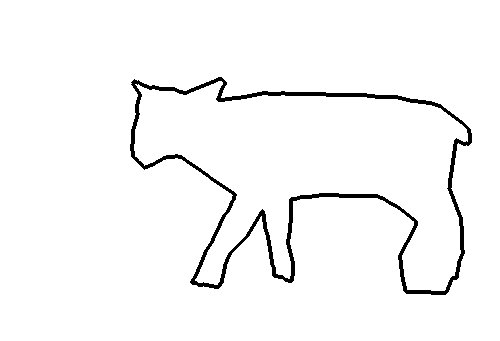}}
	\frame{\includegraphics[width=0.16\textwidth]{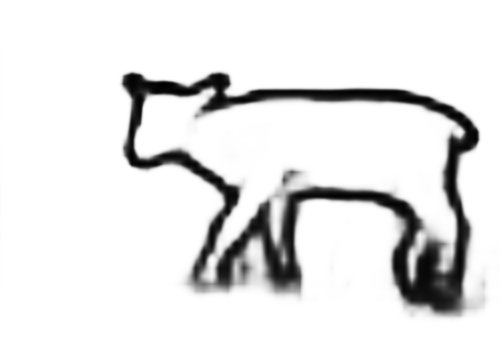}}
	\frame{\includegraphics[width=0.16\textwidth]{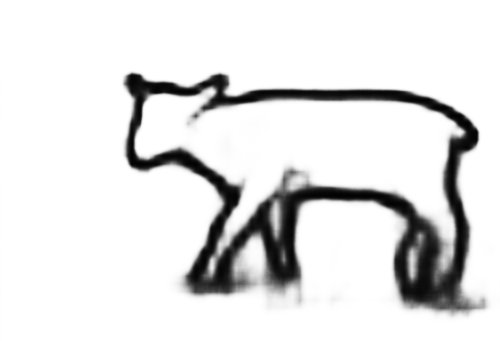}}
	\frame{\includegraphics[width=0.16\textwidth]{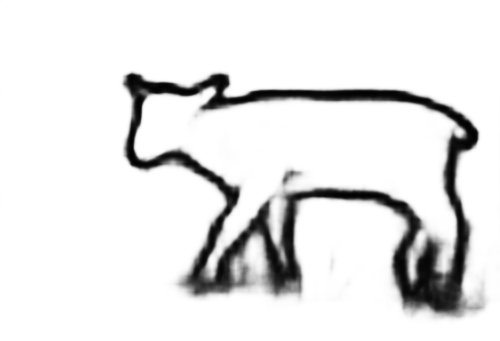}}
	\frame{\includegraphics[width=0.16\textwidth]{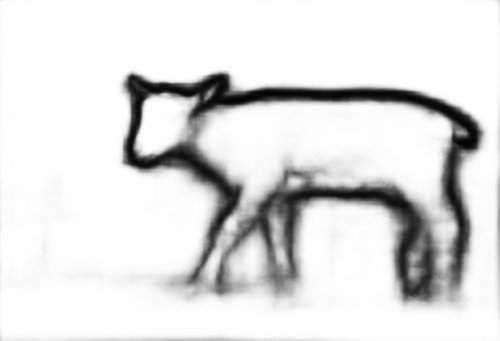}}\\
	\quad\\\vspace{-0.35cm}
	\frame{\includegraphics[width=0.16\textwidth]{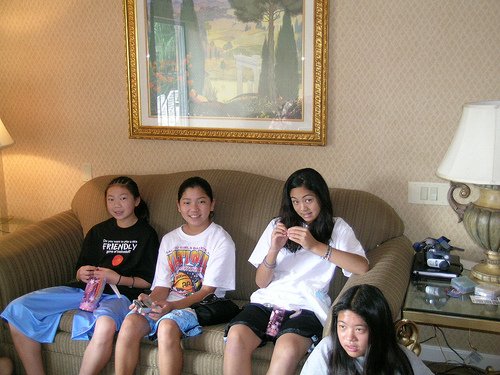}}
	\frame{\includegraphics[width=0.16\textwidth]{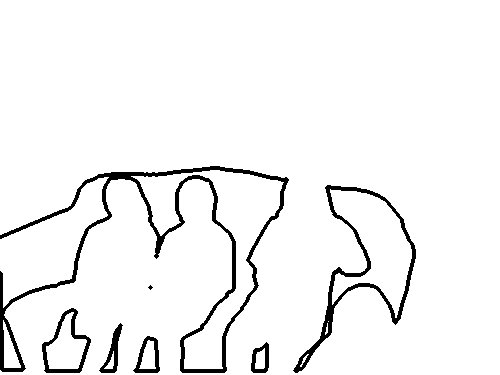}}
	\frame{\includegraphics[width=0.16\textwidth]{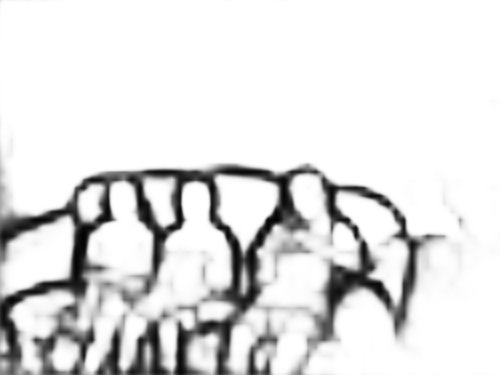}}
	\frame{\includegraphics[width=0.16\textwidth]{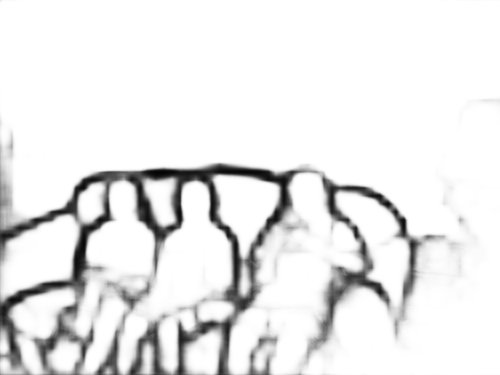}}
	\frame{\includegraphics[width=0.16\textwidth]{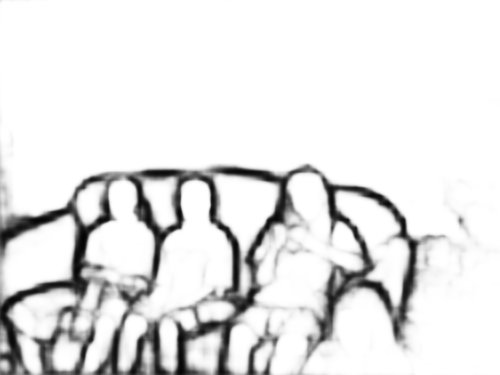}}
	\frame{\includegraphics[width=0.16\textwidth]{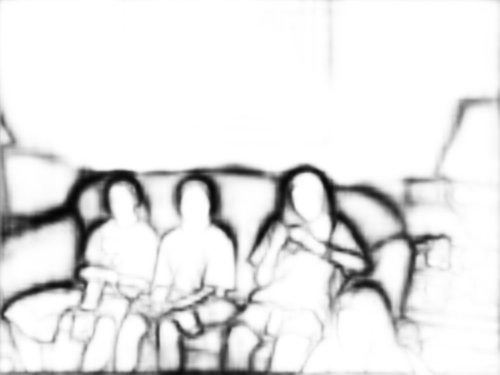}}\\
	\quad\\\vspace{-0.35cm}
	\frame{\includegraphics[width=0.16\textwidth]{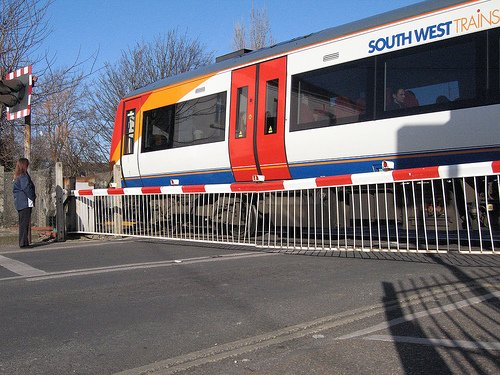}}
	\frame{\includegraphics[width=0.16\textwidth]{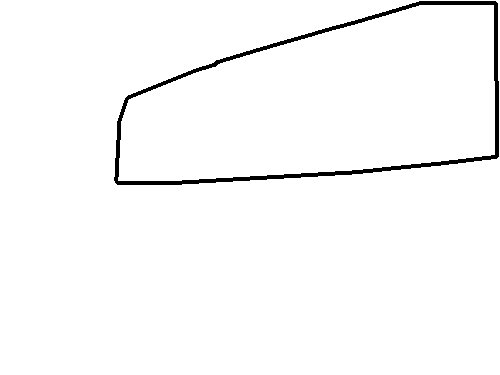}}
	\frame{\includegraphics[width=0.16\textwidth]{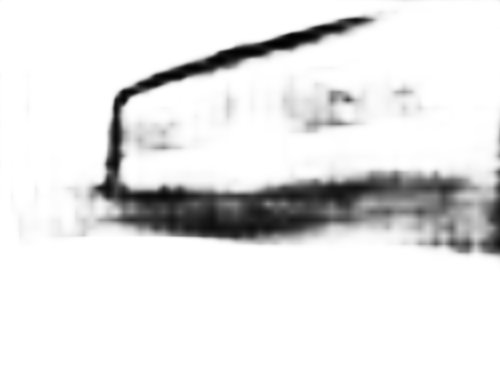}}
	\frame{\includegraphics[width=0.16\textwidth]{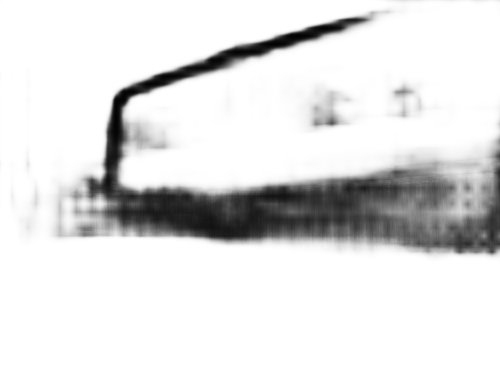}}
	\frame{\includegraphics[width=0.16\textwidth]{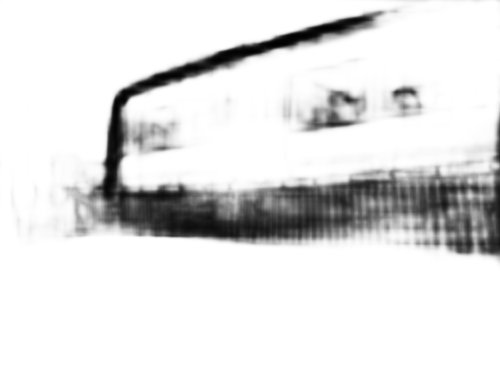}}
	\frame{\includegraphics[width=0.16\textwidth]{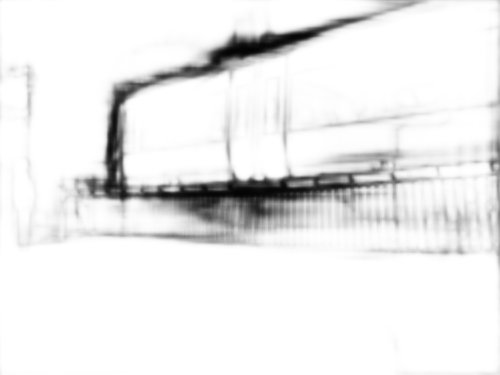}}\\
	\quad\\\vspace{-0.35cm}
	\frame{\includegraphics[width=0.16\textwidth]{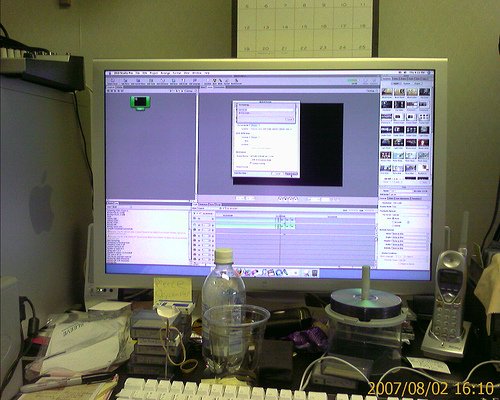}}
	\frame{\includegraphics[width=0.16\textwidth]{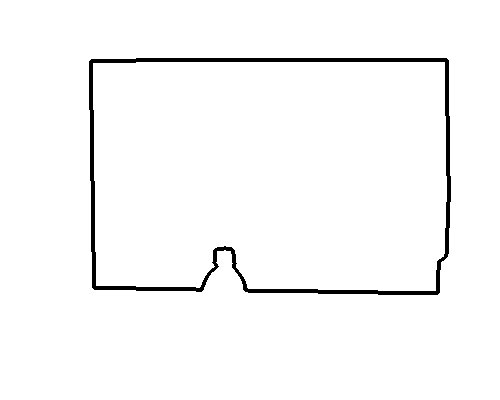}}
	\frame{\includegraphics[width=0.16\textwidth]{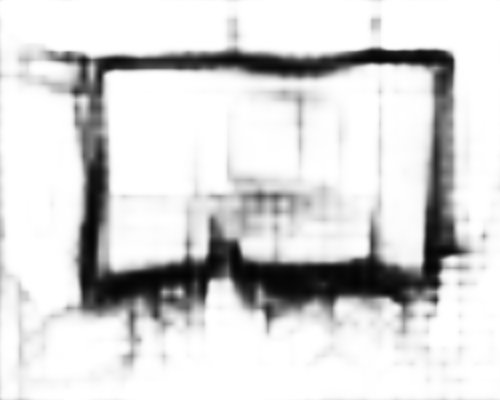}}
	\frame{\includegraphics[width=0.16\textwidth]{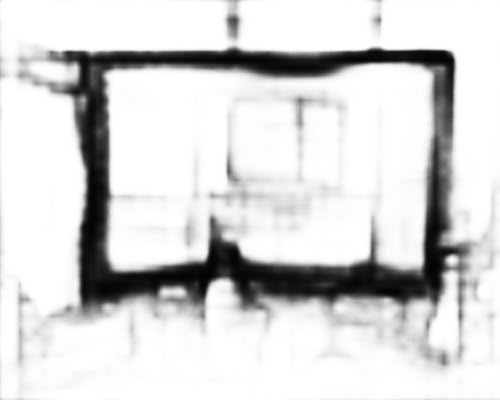}}
	\frame{\includegraphics[width=0.16\textwidth]{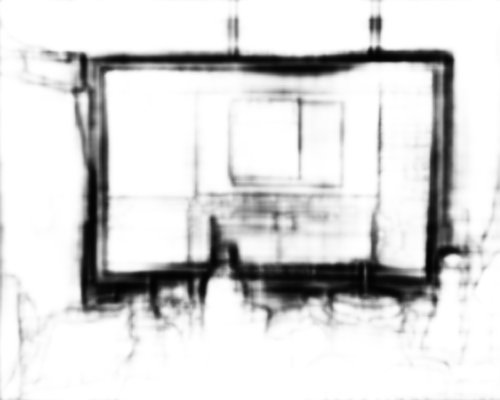}}
	\frame{\includegraphics[width=0.16\textwidth]{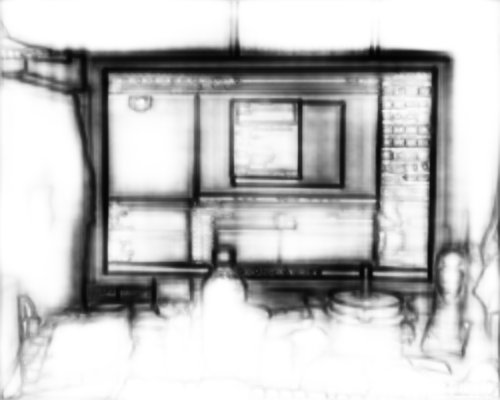}}
	\caption{Class-wise prediction results of comparing methods on the SBD Dataset. Rows correspond to the predicted edges of ``dining table'', ``dog'', ``horse'', ``motorbike'', ``person'', ``potted plant'', ``sheep'', ``sofa'', ``train'' and ``tv monitor''. Columns correspond to original image, ground truth, and results of Basic, DSN, CASENet and CASENet-VGG.}\label{fig:qual_sbd2}
\end{figure*}

We illustrate 20 typical examples of the class-wise edge predictions of different comparing methods in Fig. \ref{fig:qual_sbd1} and \ref{fig:qual_sbd2}, with each example corresponding to one of the SBD semantic category. One can observe that the proposed CASENet slightly but consistently outperforms ResNets with the basic and DSN architectures, by overall showing sharper edges and often having stronger responses on difficult edges.

\begin{figure*}[p]
	\centering
	\frame{\includegraphics[width=0.16\textwidth]{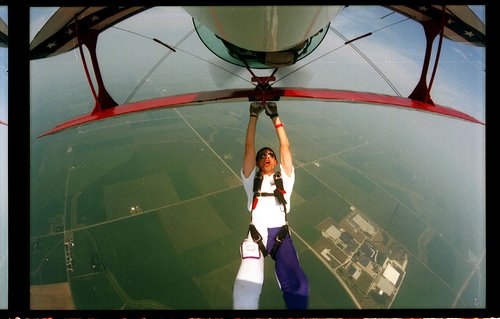}}
	\frame{\includegraphics[width=0.16\textwidth]{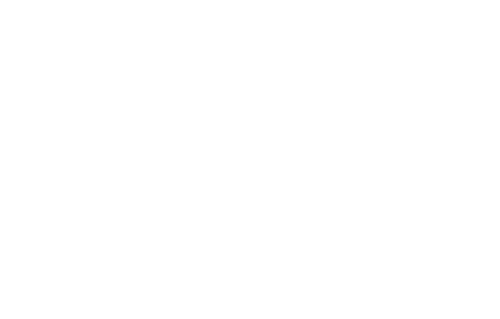}}
	\frame{\includegraphics[width=0.16\textwidth]{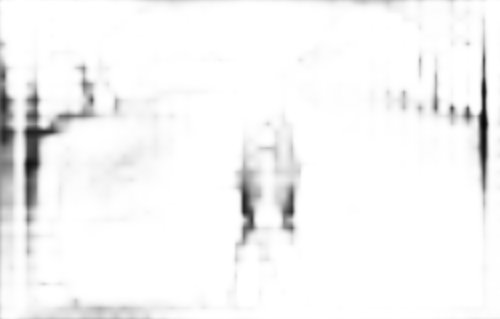}}
	\frame{\includegraphics[width=0.16\textwidth]{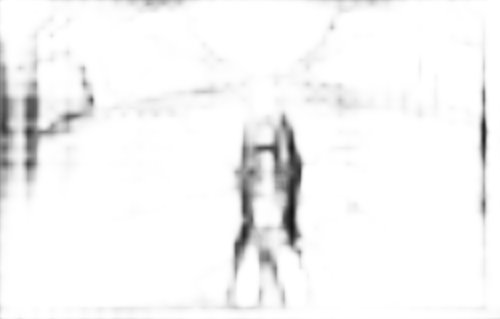}}
	\frame{\includegraphics[width=0.16\textwidth]{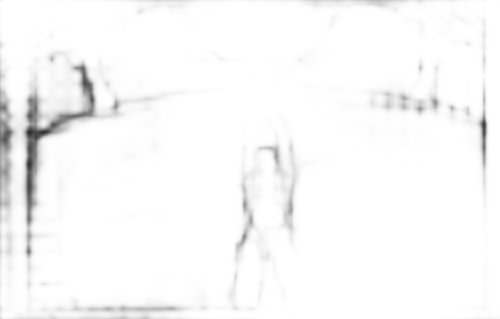}}
	\frame{\includegraphics[width=0.16\textwidth]{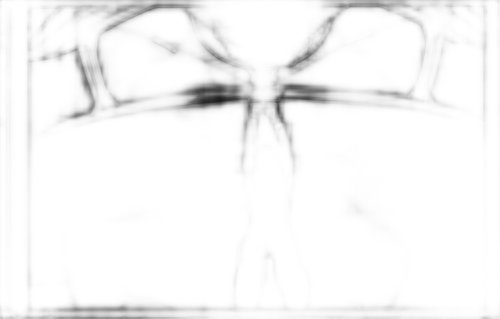}}\\
	\quad\\\vspace{-0.35cm}
	\frame{\includegraphics[width=0.16\textwidth]{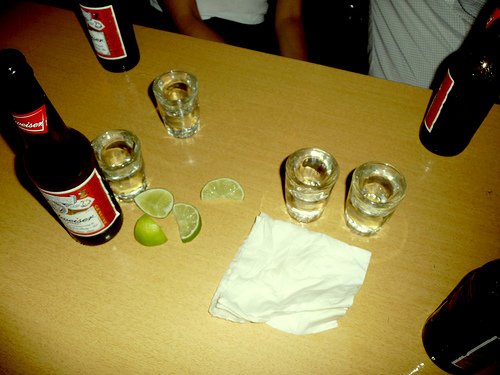}}
	\frame{\includegraphics[width=0.16\textwidth]{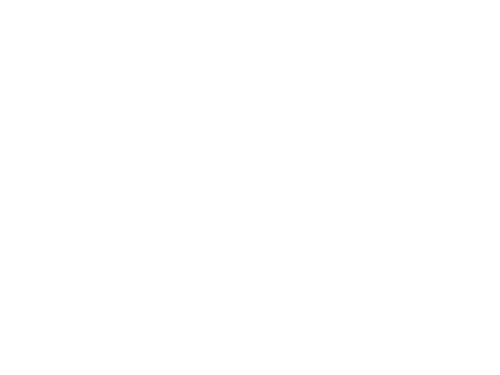}}
	\frame{\includegraphics[width=0.16\textwidth]{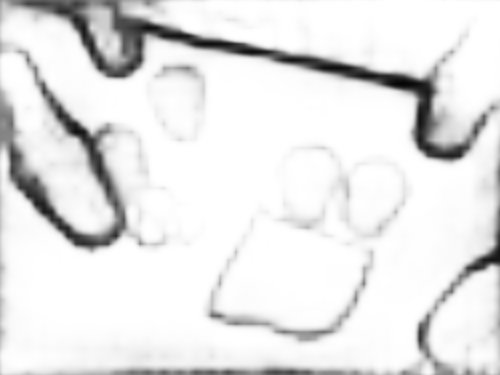}}
	\frame{\includegraphics[width=0.16\textwidth]{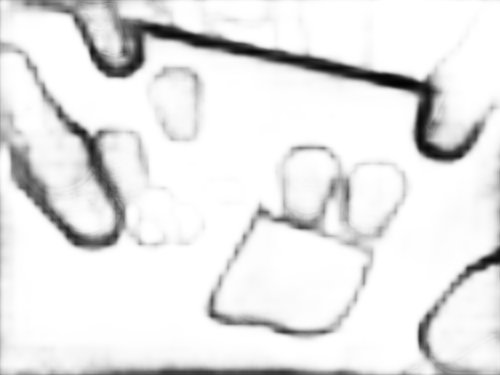}}
	\frame{\includegraphics[width=0.16\textwidth]{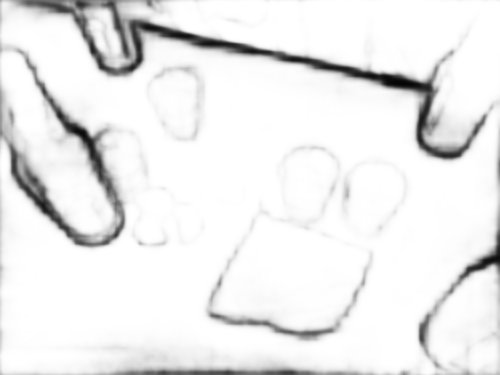}}
	\frame{\includegraphics[width=0.16\textwidth]{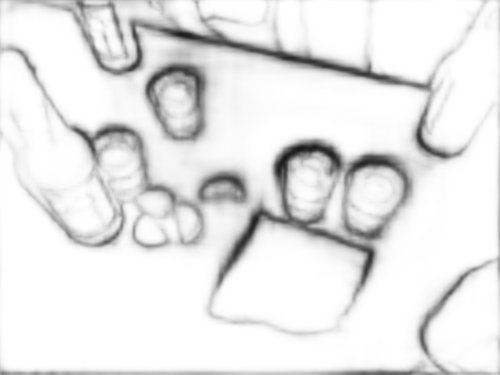}}\\
	\quad\\\vspace{-0.35cm}
	\frame{\includegraphics[width=0.16\textwidth]{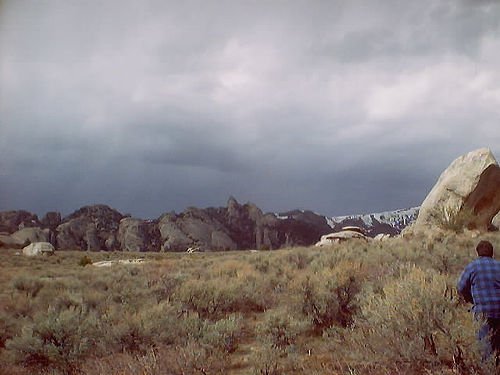}}
	\frame{\includegraphics[width=0.16\textwidth]{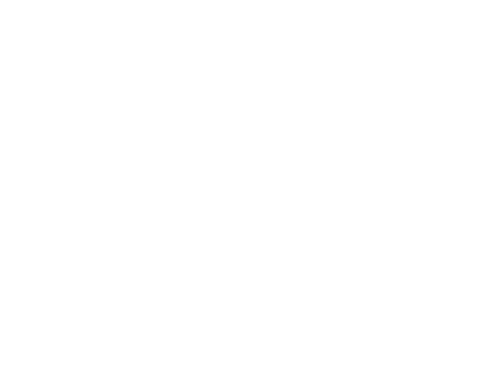}}
	\frame{\includegraphics[width=0.16\textwidth]{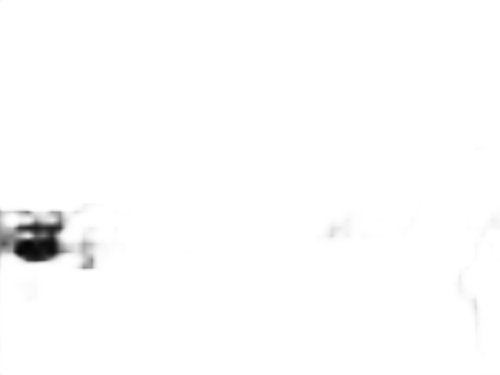}}
	\frame{\includegraphics[width=0.16\textwidth]{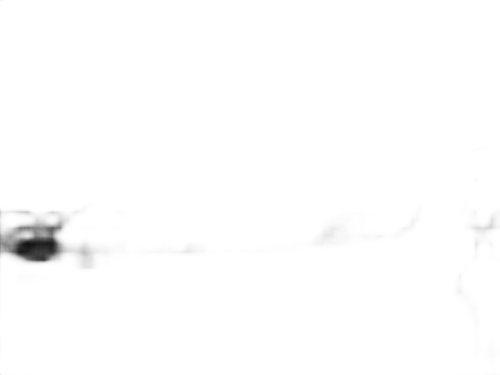}}
	\frame{\includegraphics[width=0.16\textwidth]{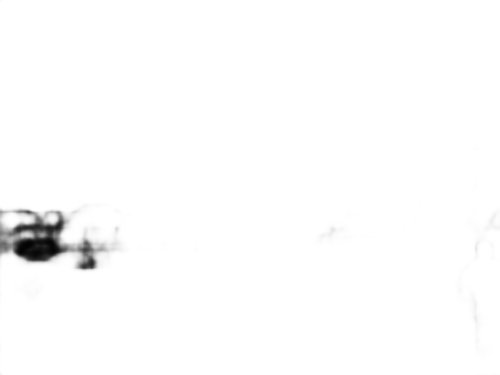}}
	\frame{\includegraphics[width=0.16\textwidth]{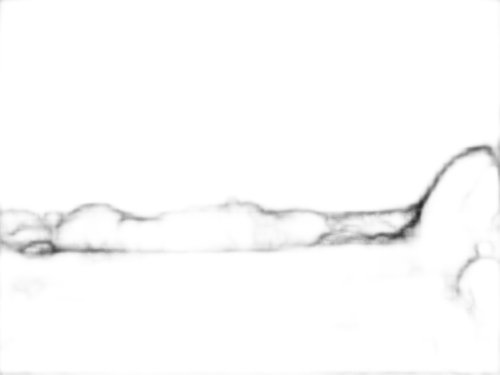}}
	\caption{Difficult or failure cases on the SBD Dataset. Rows correspond to the predicted edges of ``aeroplane'', ``dining table'' and ``sheep''. Columns correspond to original image, ground truth, and results of Basic, DSN, CASENet and CASENet-VGG.}\label{fig:qual_sbd3}
\end{figure*}

Meanwhile, Fig. \ref{fig:qual_sbd3} shows several difficult or failure cases on the SBD Datasets. Interestingly, while the ground truth says there is no ``aeroplane'' in the first row and ``dining table'' in the second, the network is doing decently by giving certain level of edge responses, particularly in the ``dining table'' example. The third row shows an example of the false positive mistakes often made by the networks on small objects. The networks falsely think there is a sheep while it is in fact a rock. When objects become smaller and lose details, such mistakes in general happen more frequently.

\subsection{Class-wise precision-recall curves}
\begin{figure*}[p]
	\centering
	\includegraphics[width=0.194\textwidth]{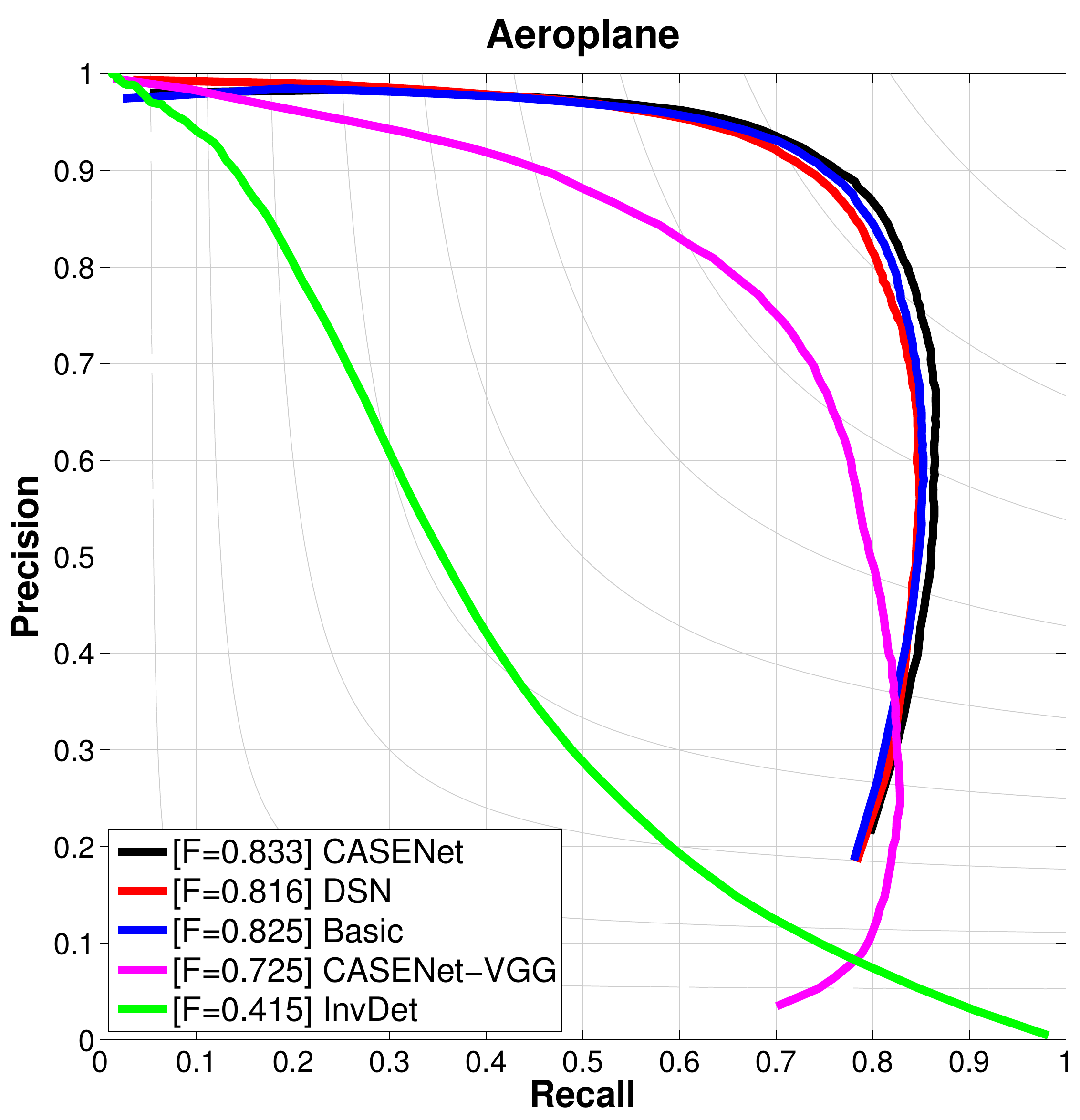}
	\includegraphics[width=0.194\textwidth]{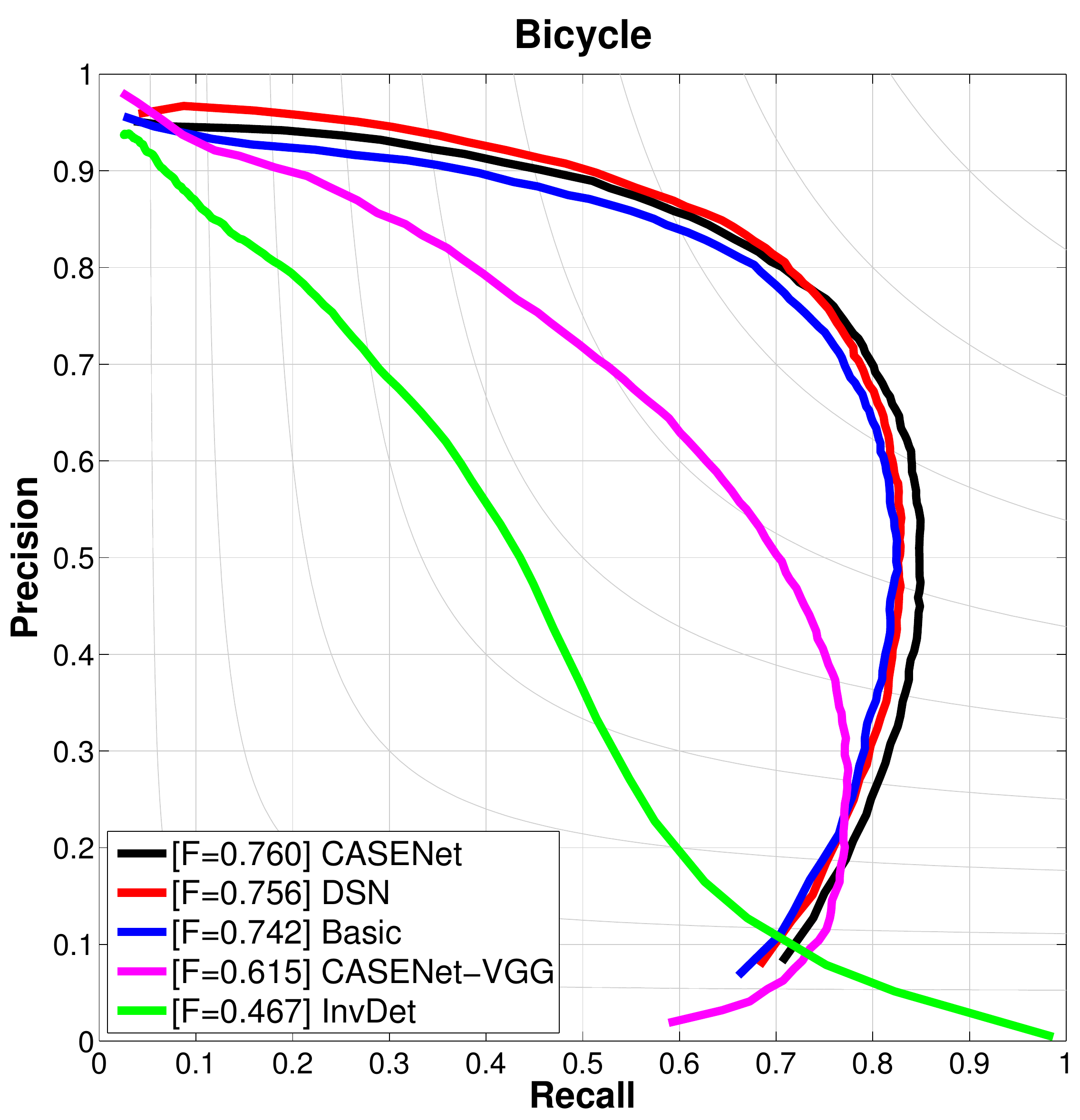}
	\includegraphics[width=0.194\textwidth]{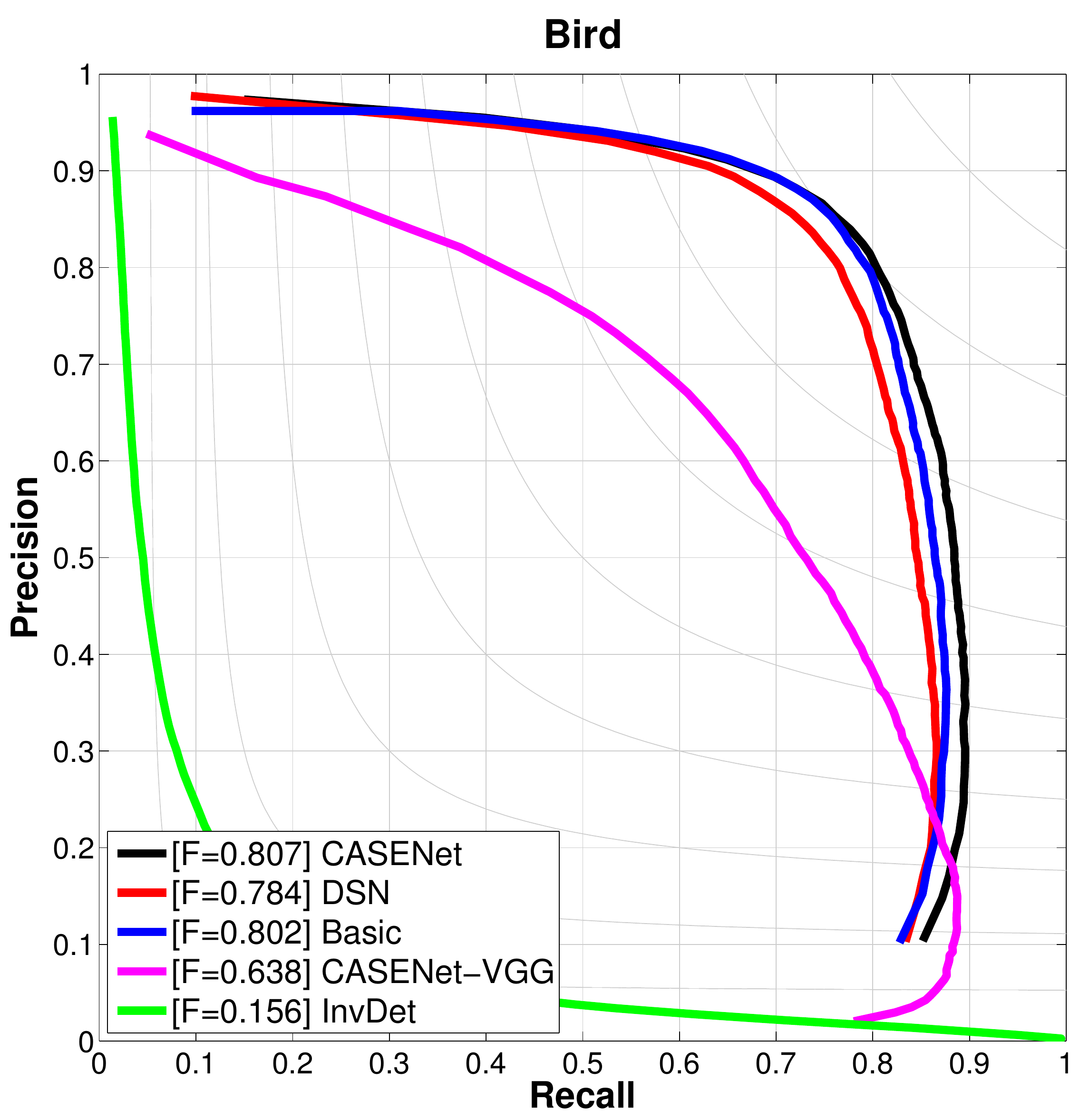}
	\includegraphics[width=0.194\textwidth]{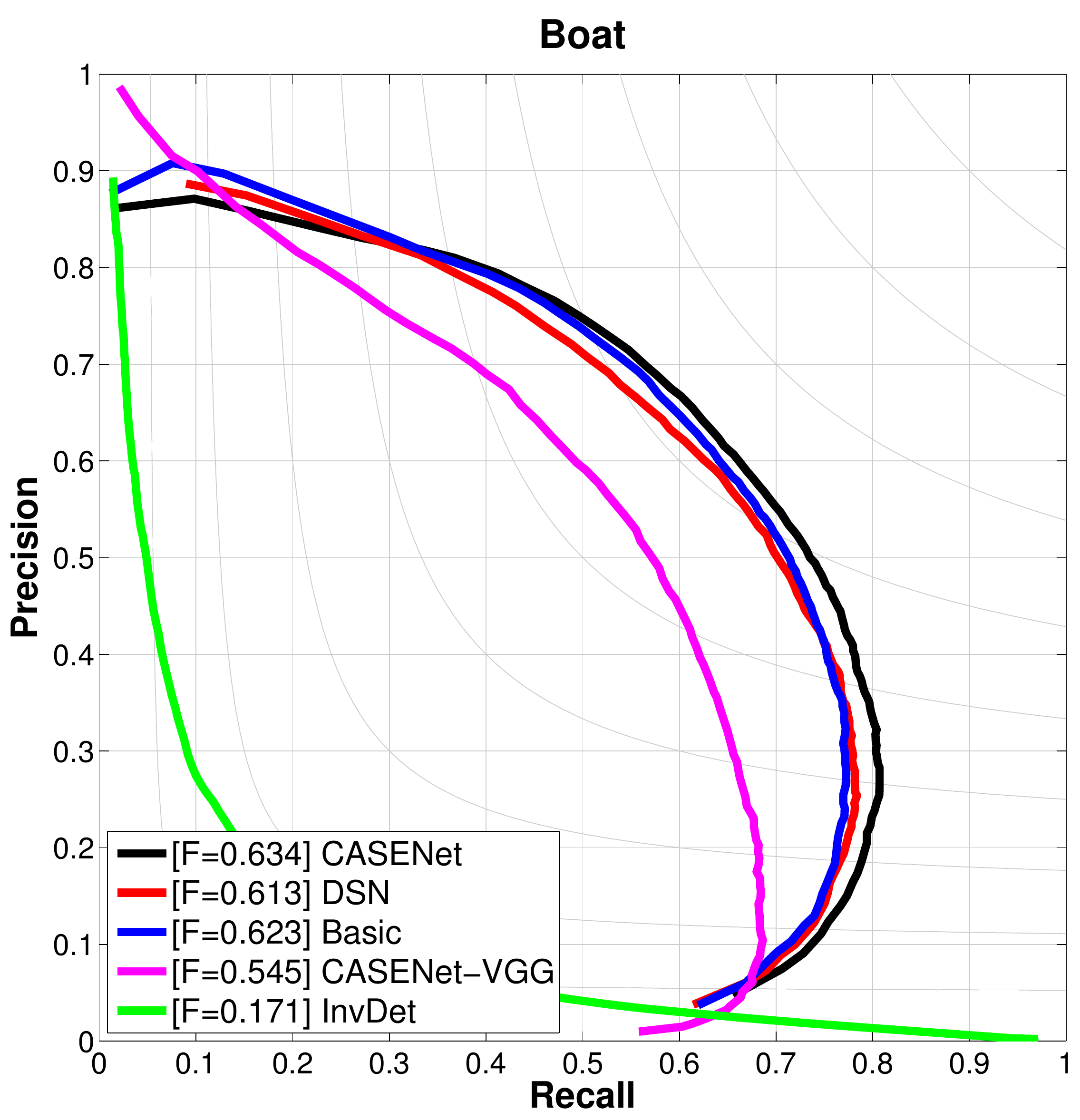}
	\includegraphics[width=0.194\textwidth]{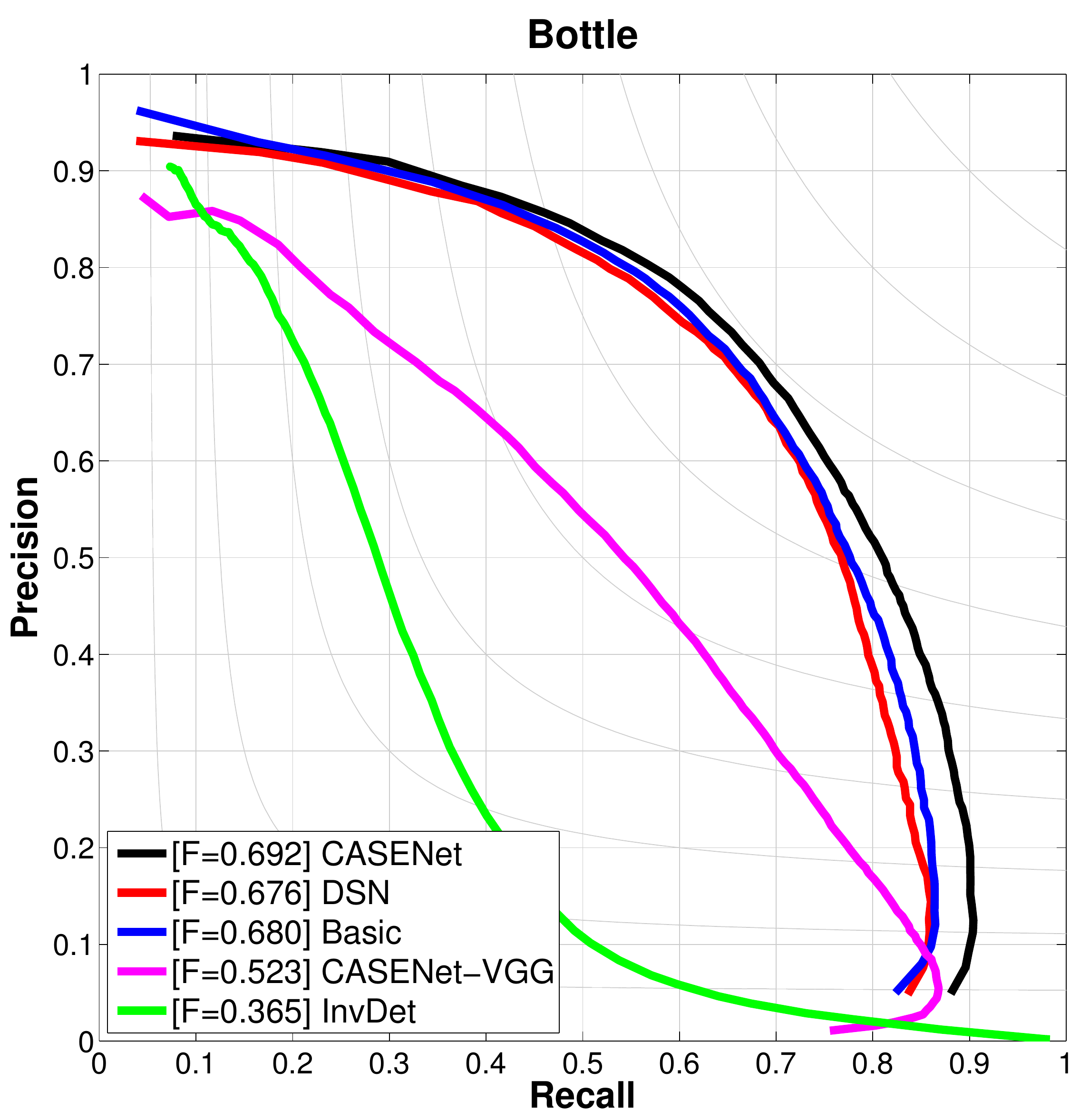}\\
	\quad\\\vspace{-0.35cm}
	\includegraphics[width=0.194\textwidth]{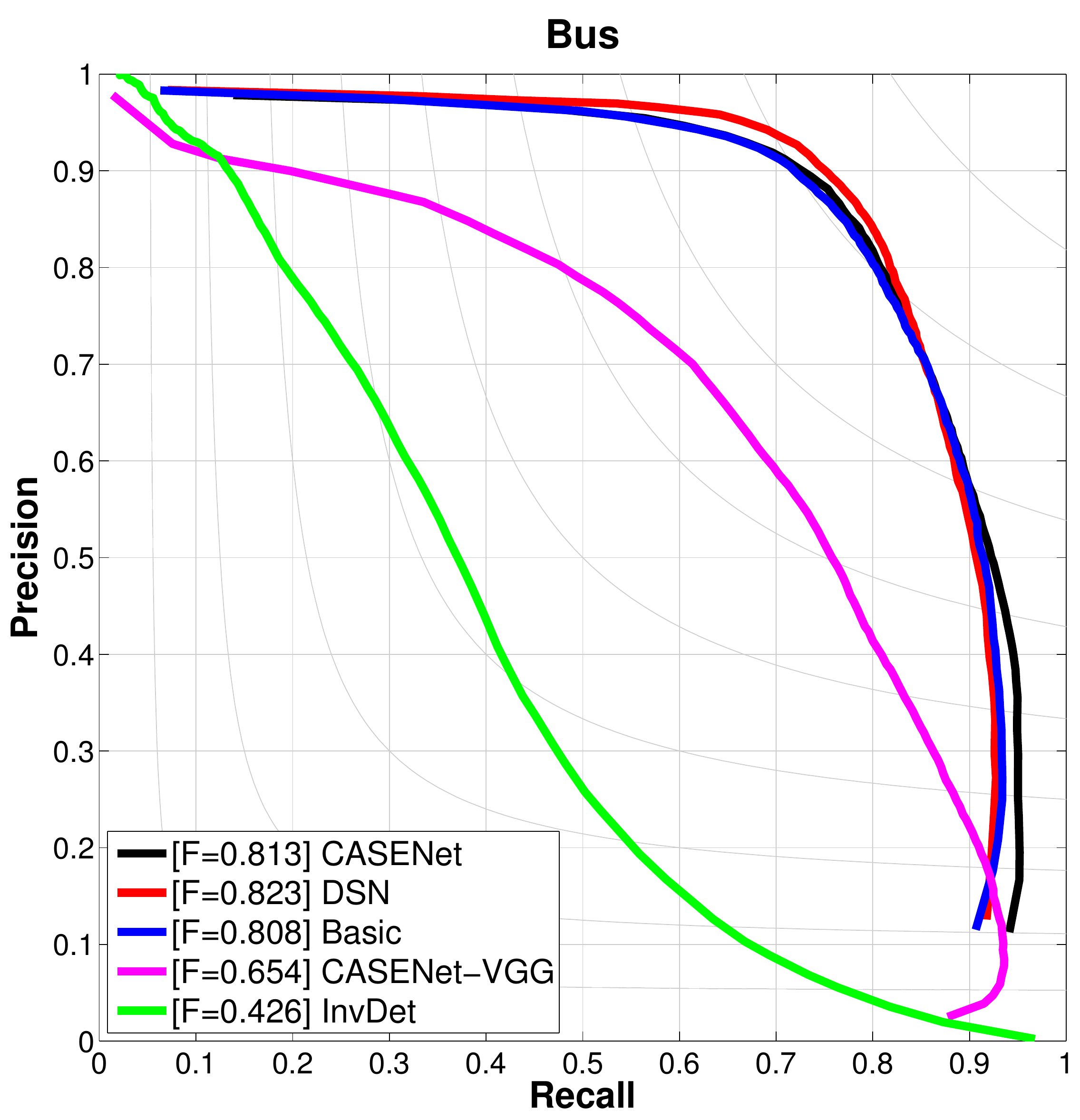}
	\includegraphics[width=0.194\textwidth]{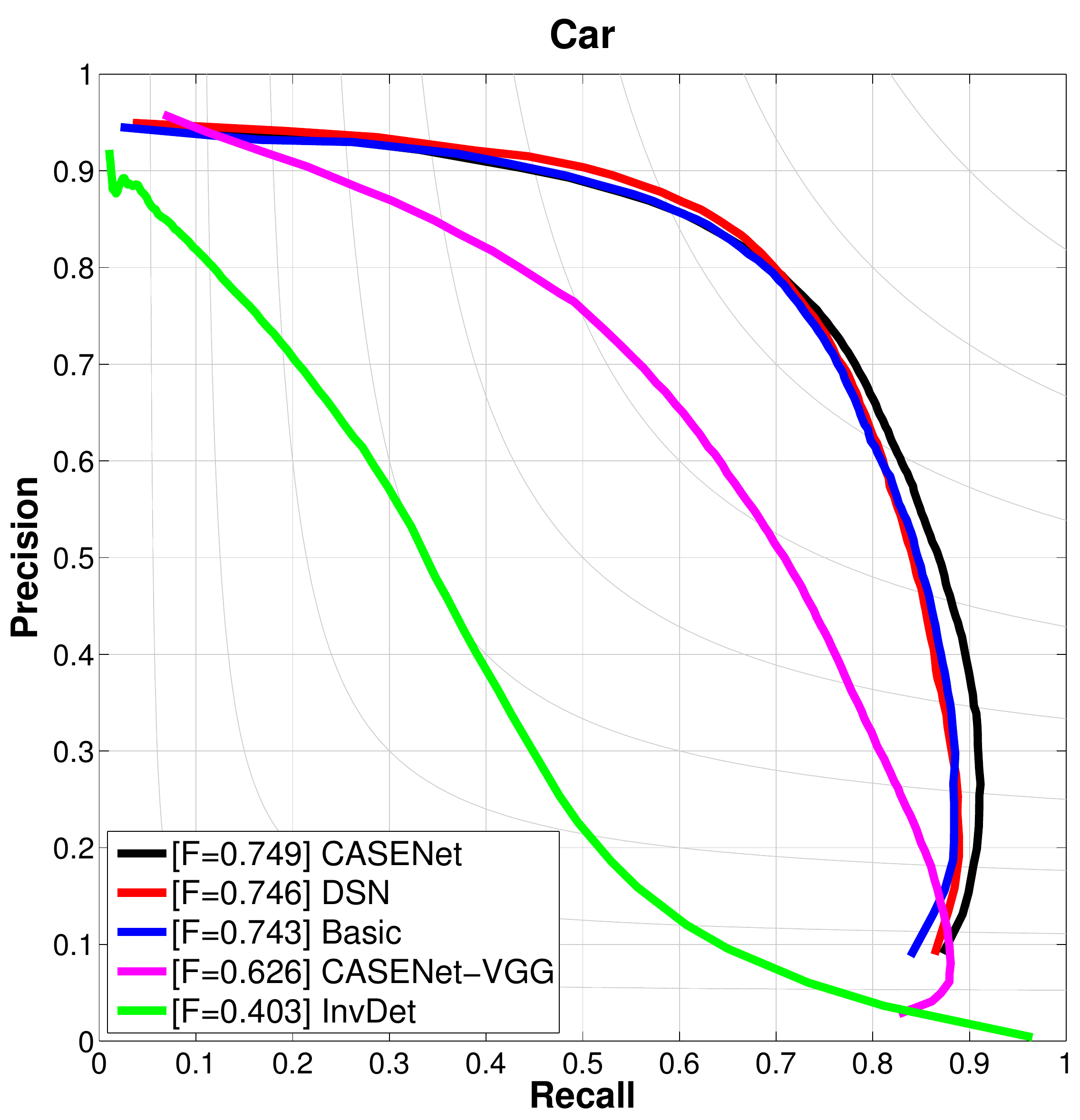}
	\includegraphics[width=0.194\textwidth]{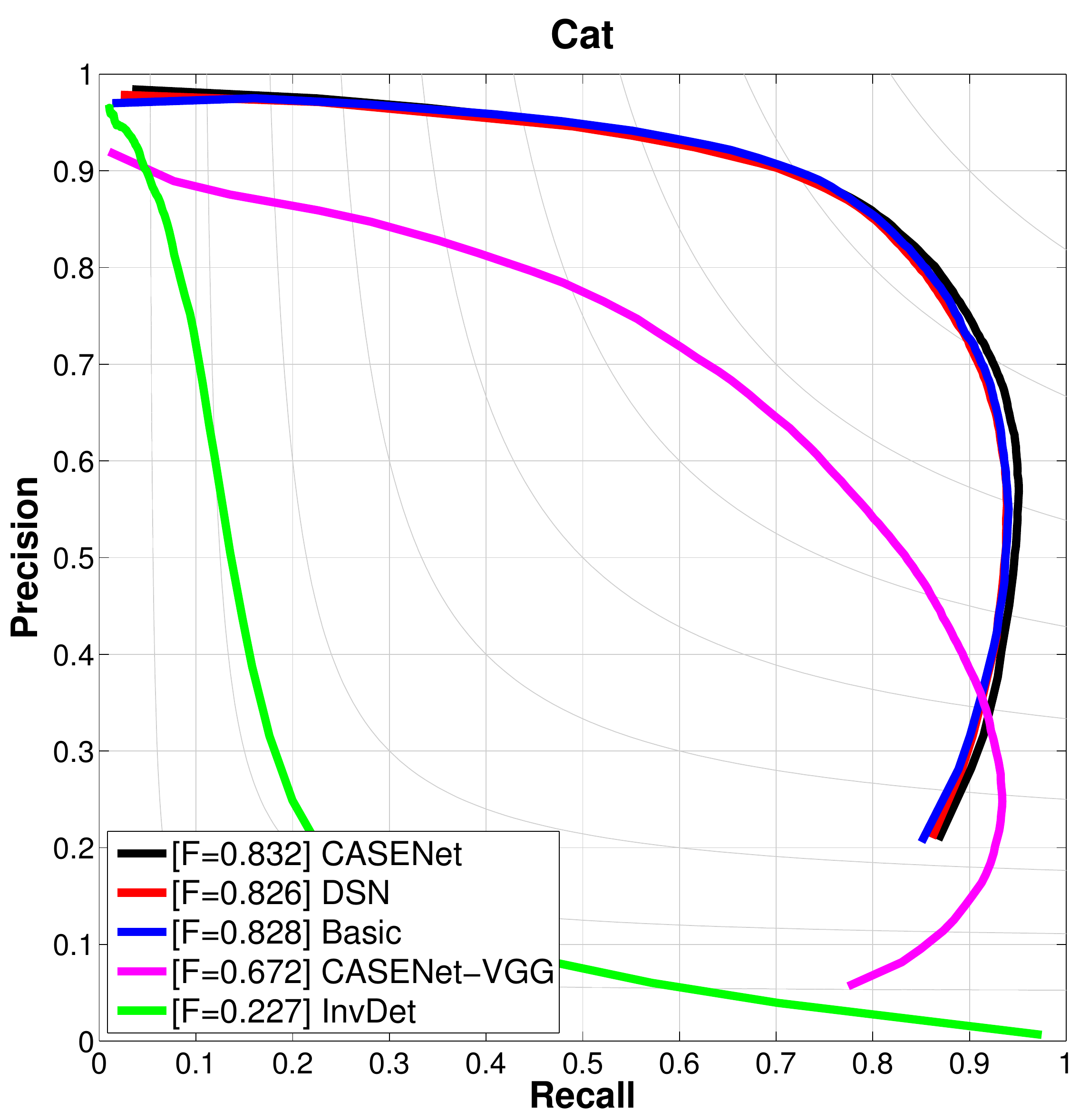}
	\includegraphics[width=0.194\textwidth]{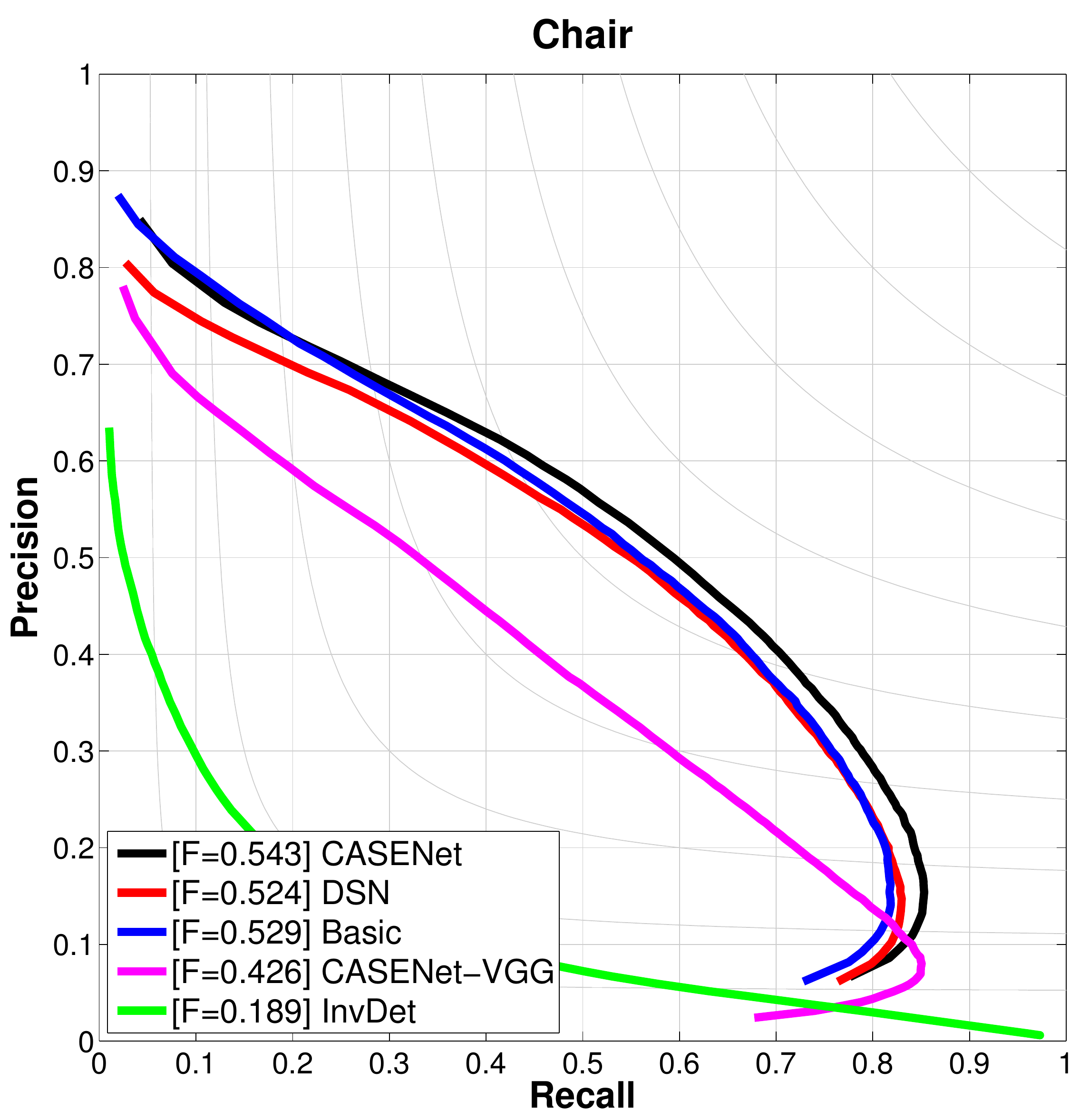}
	\includegraphics[width=0.194\textwidth]{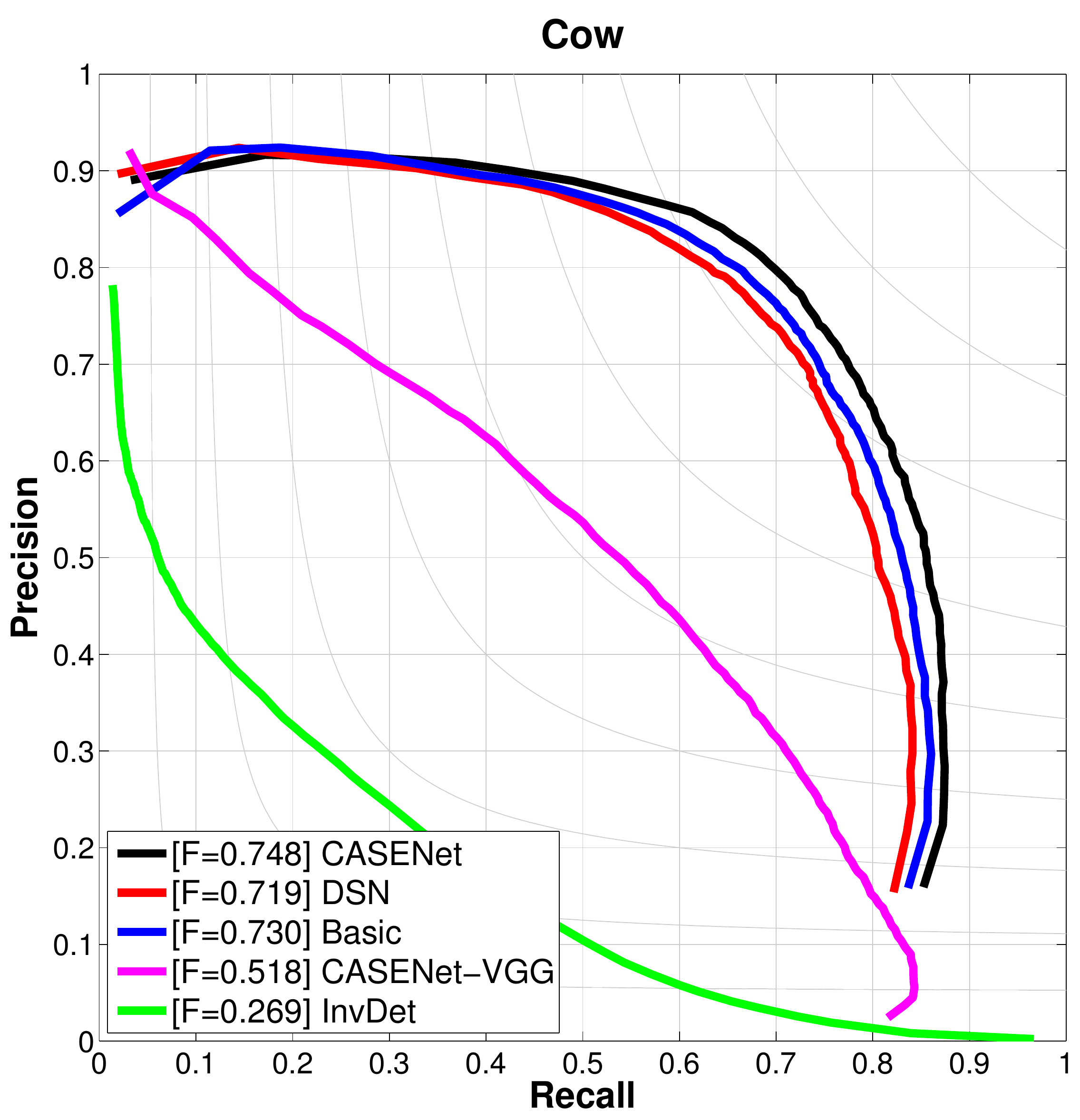}\\
	\quad\\\vspace{-0.35cm}
	\includegraphics[width=0.194\textwidth]{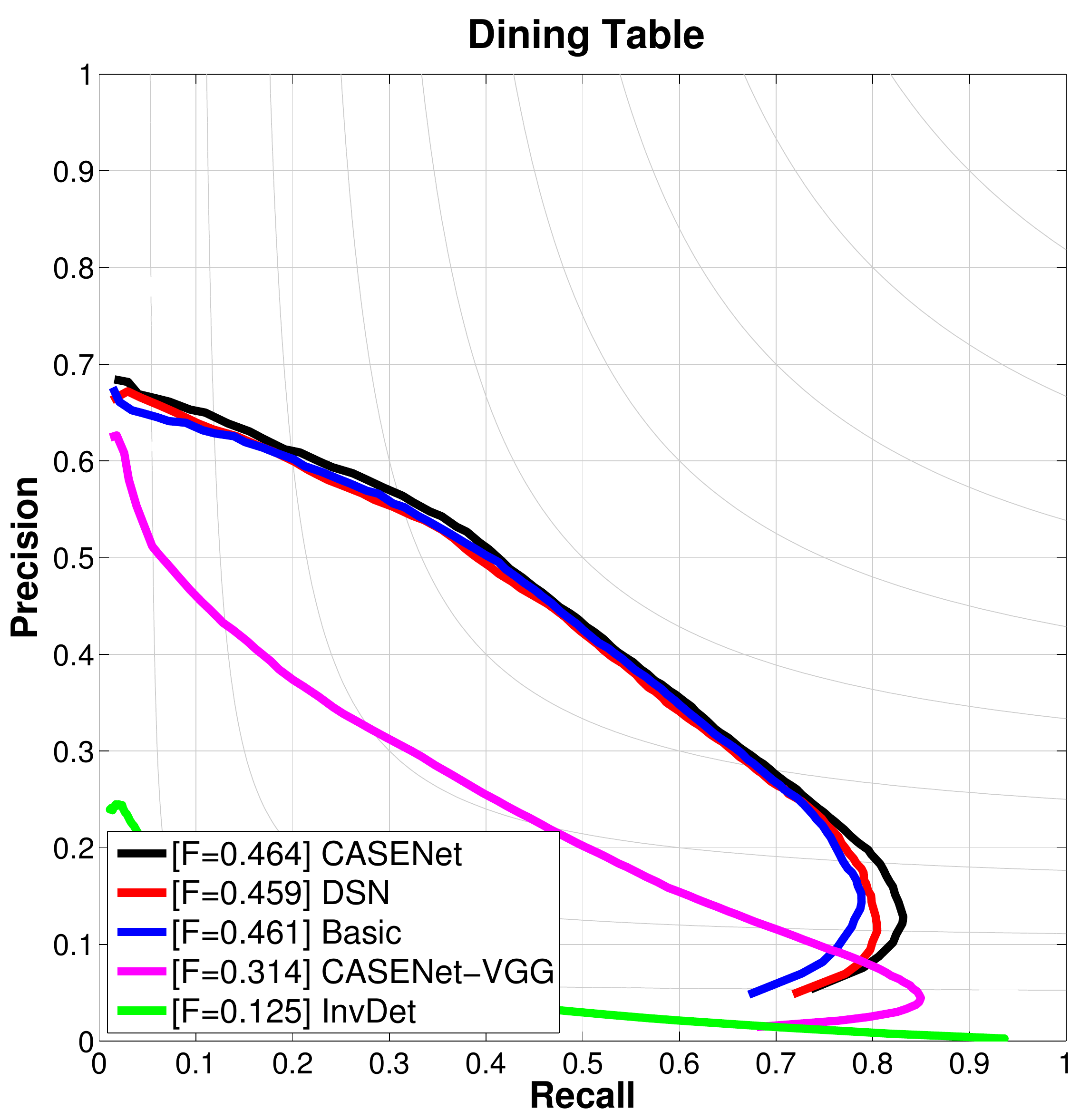}
	\includegraphics[width=0.194\textwidth]{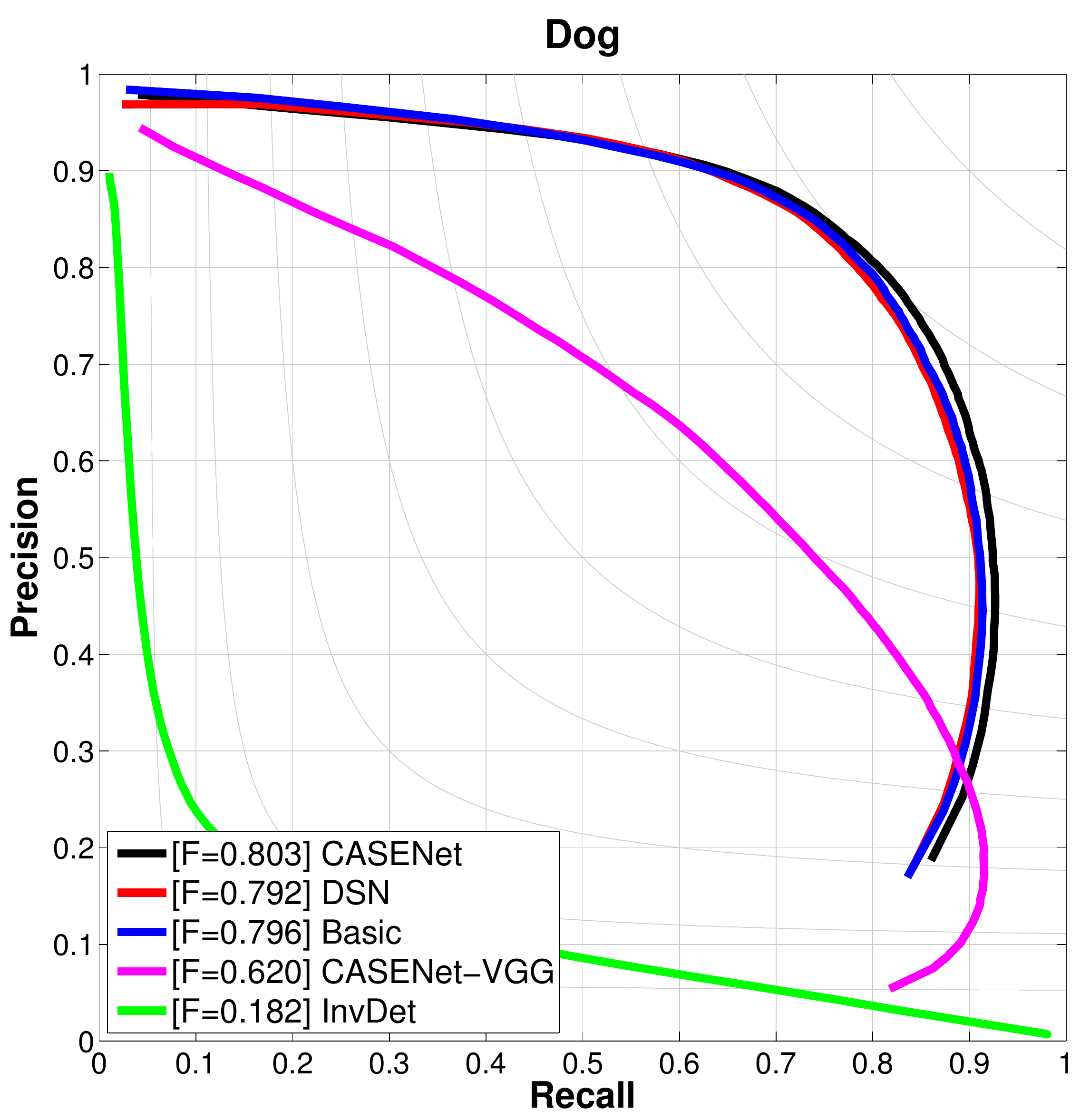}
	\includegraphics[width=0.194\textwidth]{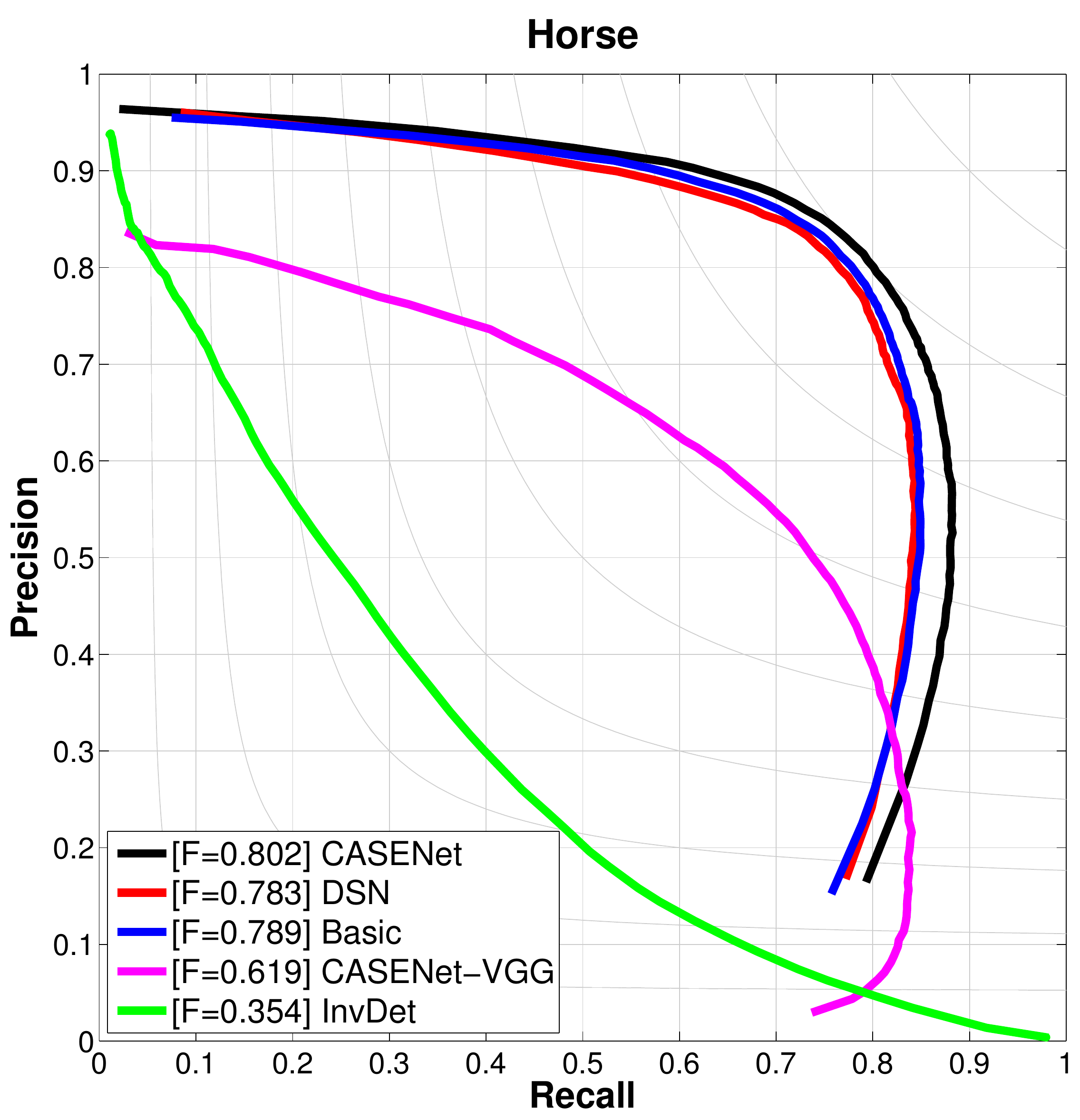}
	\includegraphics[width=0.194\textwidth]{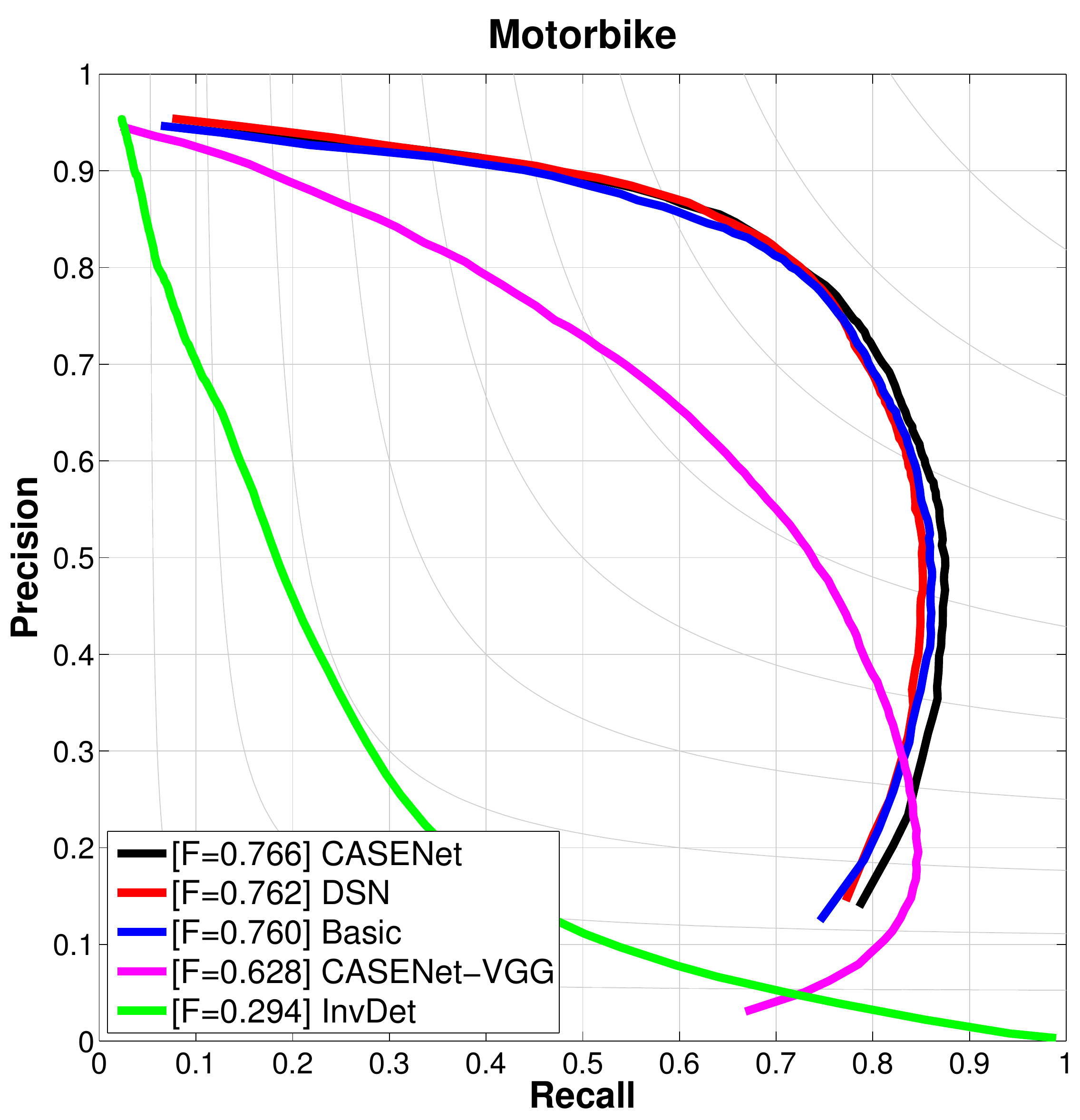}
	\includegraphics[width=0.194\textwidth]{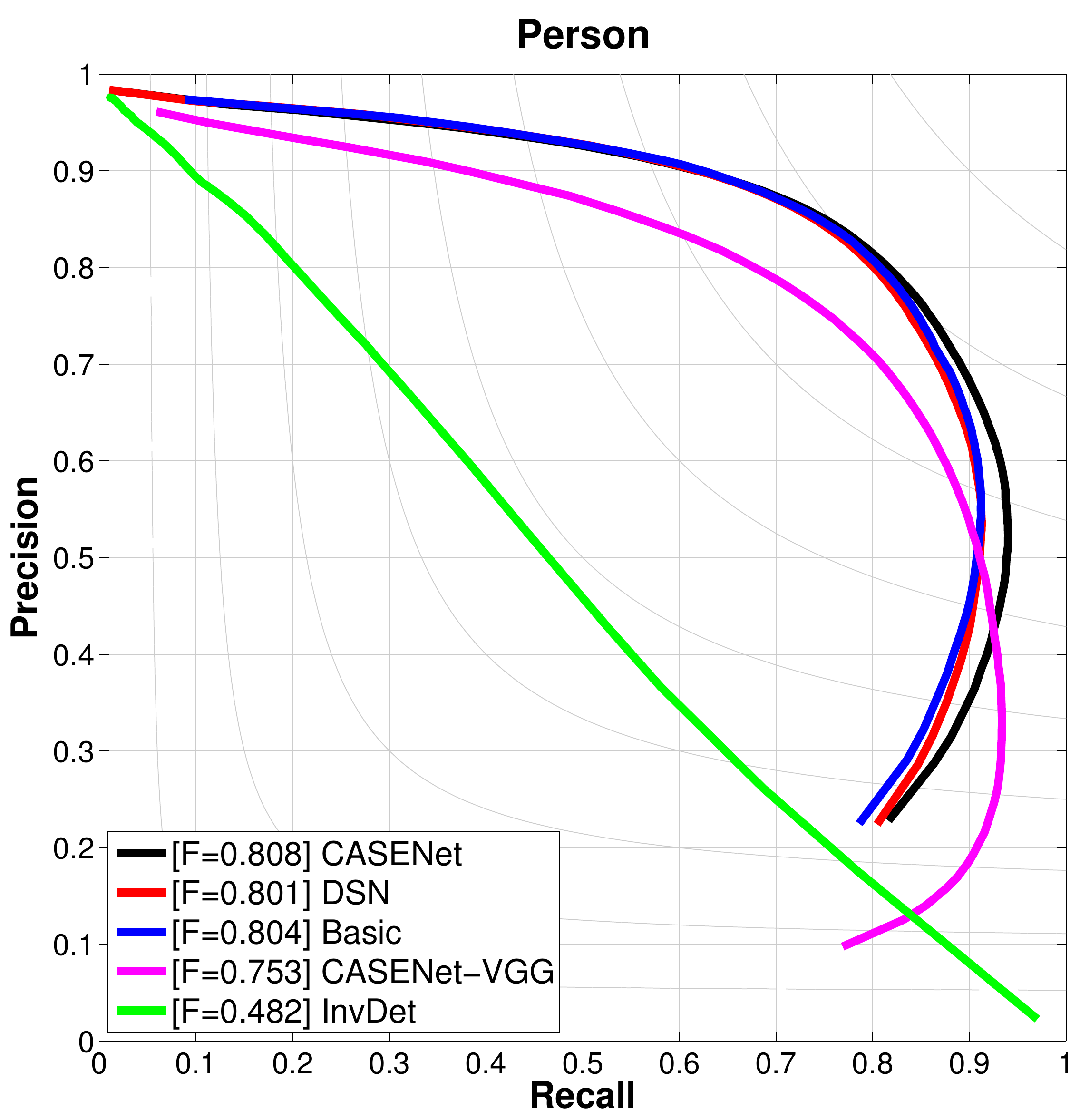}\\
	\quad\\\vspace{-0.35cm}
	\includegraphics[width=0.194\textwidth]{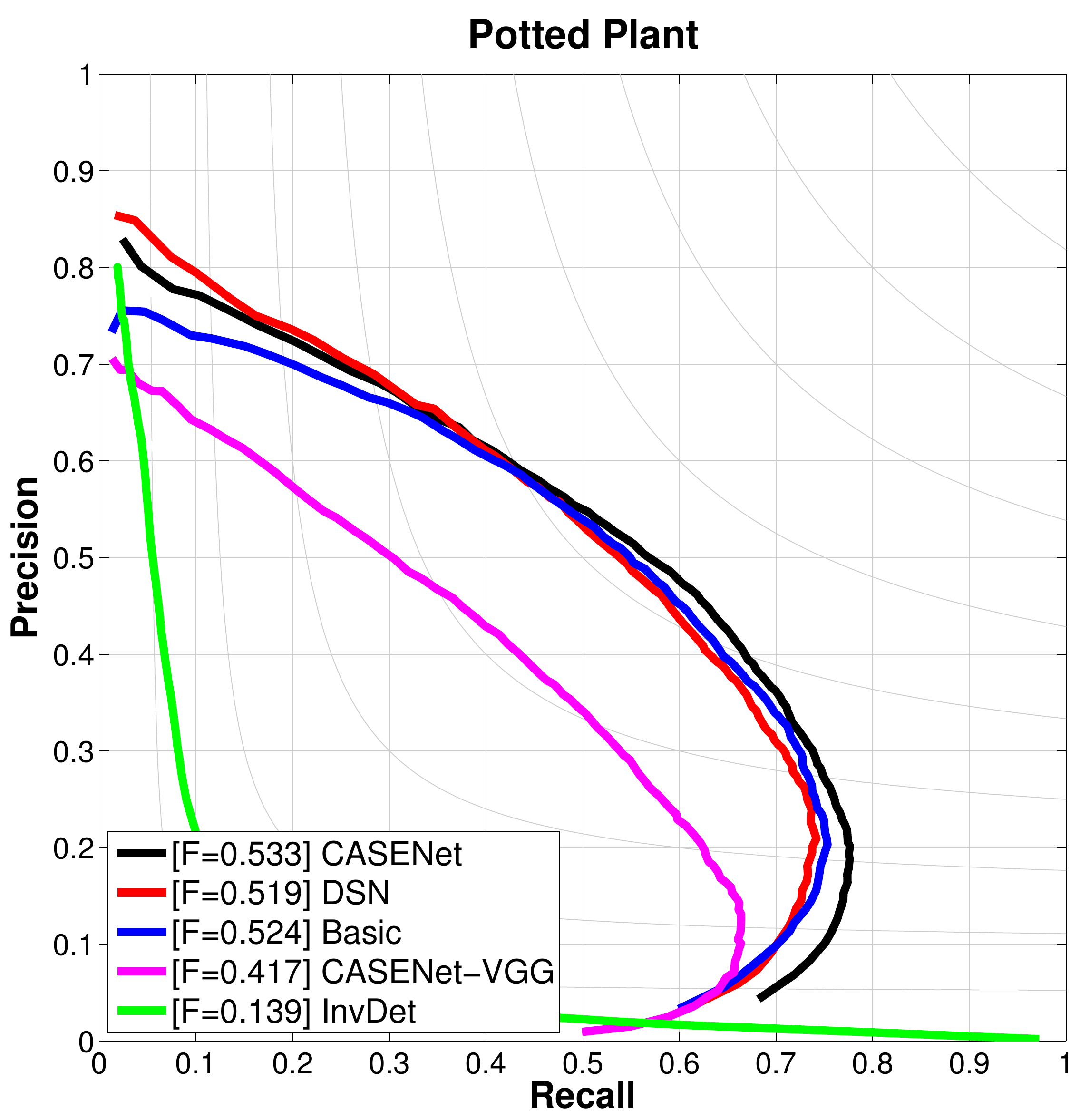}
	\includegraphics[width=0.194\textwidth]{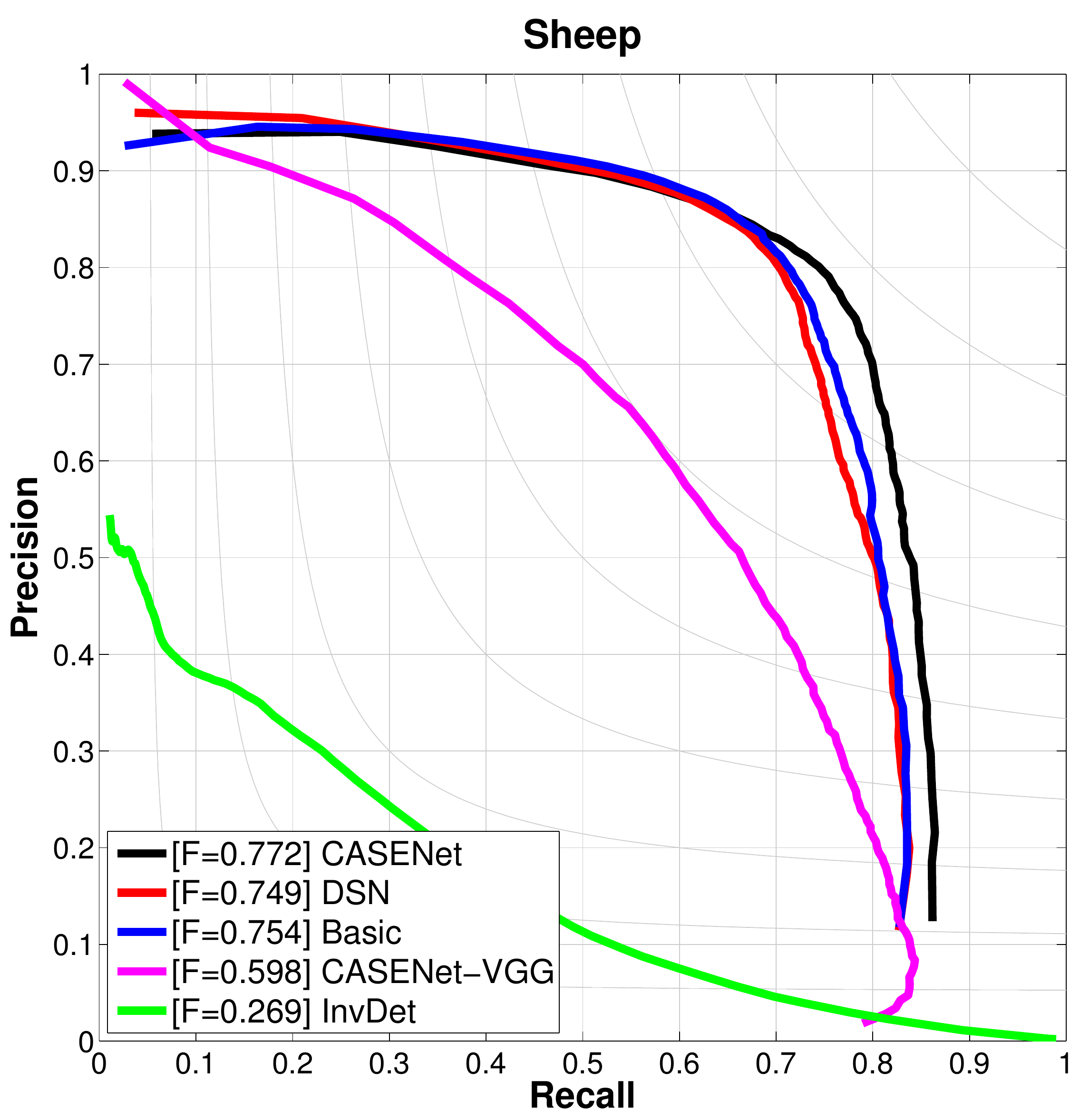}
	\includegraphics[width=0.194\textwidth]{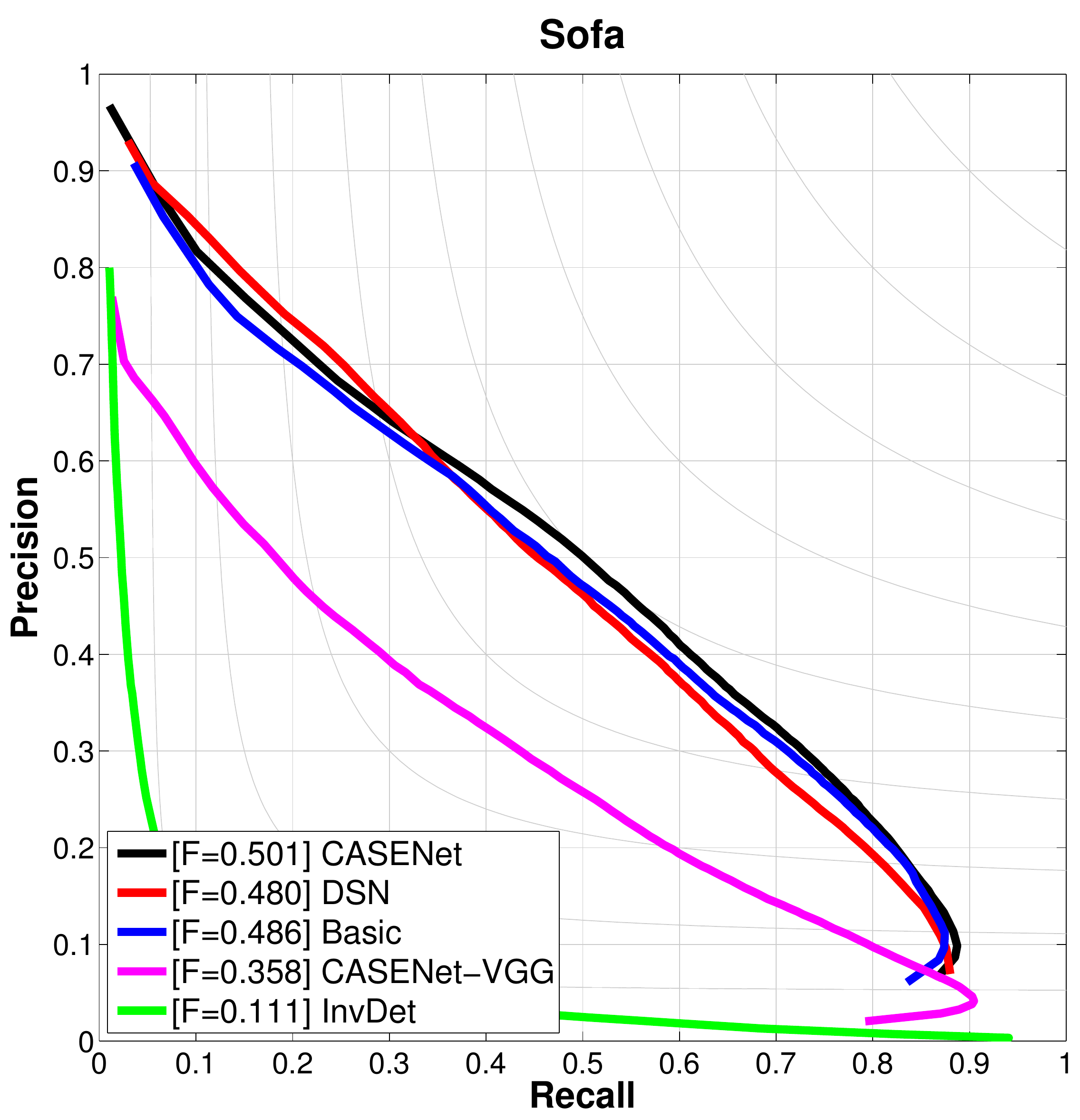}
	\includegraphics[width=0.194\textwidth]{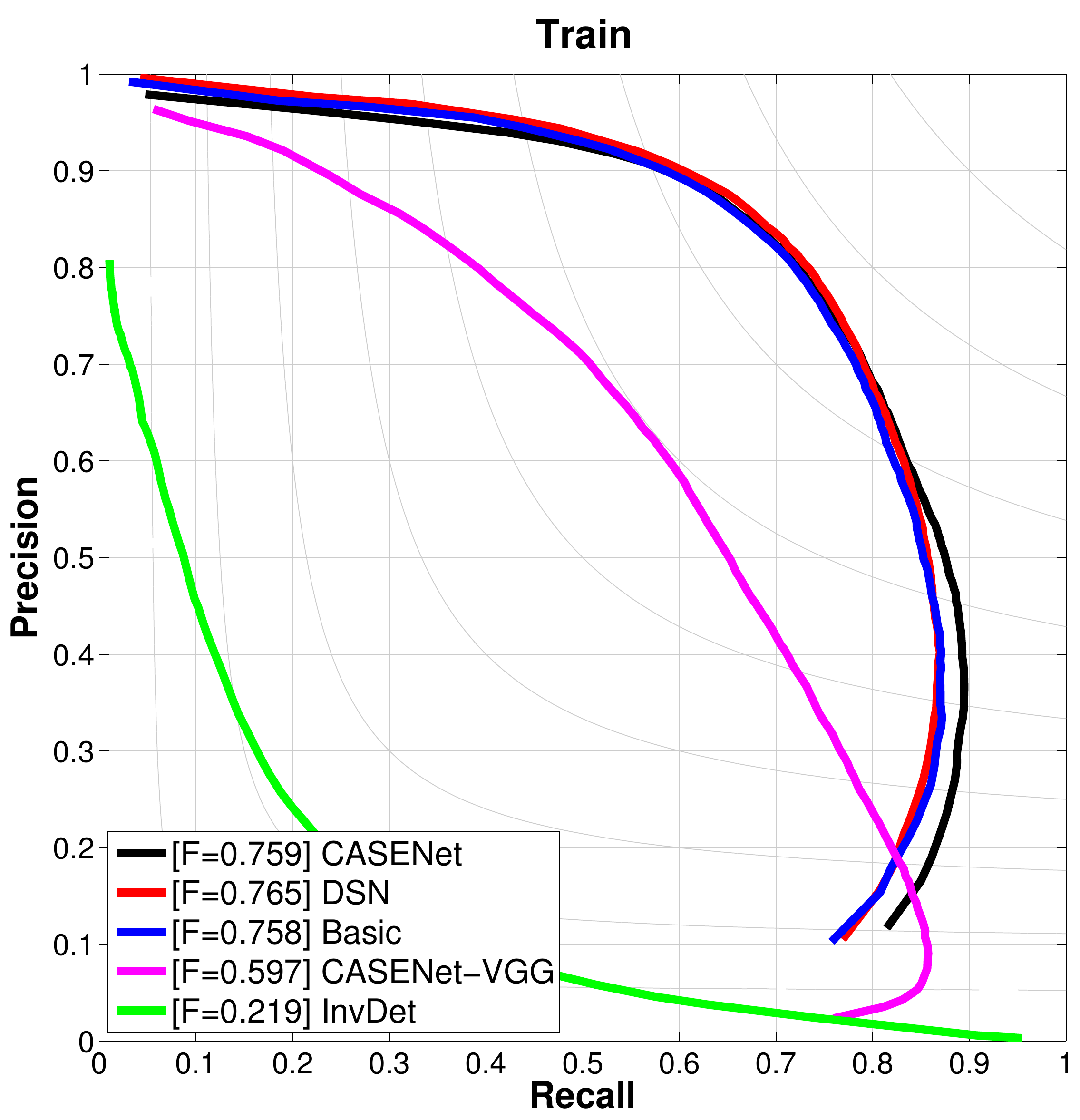}
	\includegraphics[width=0.194\textwidth]{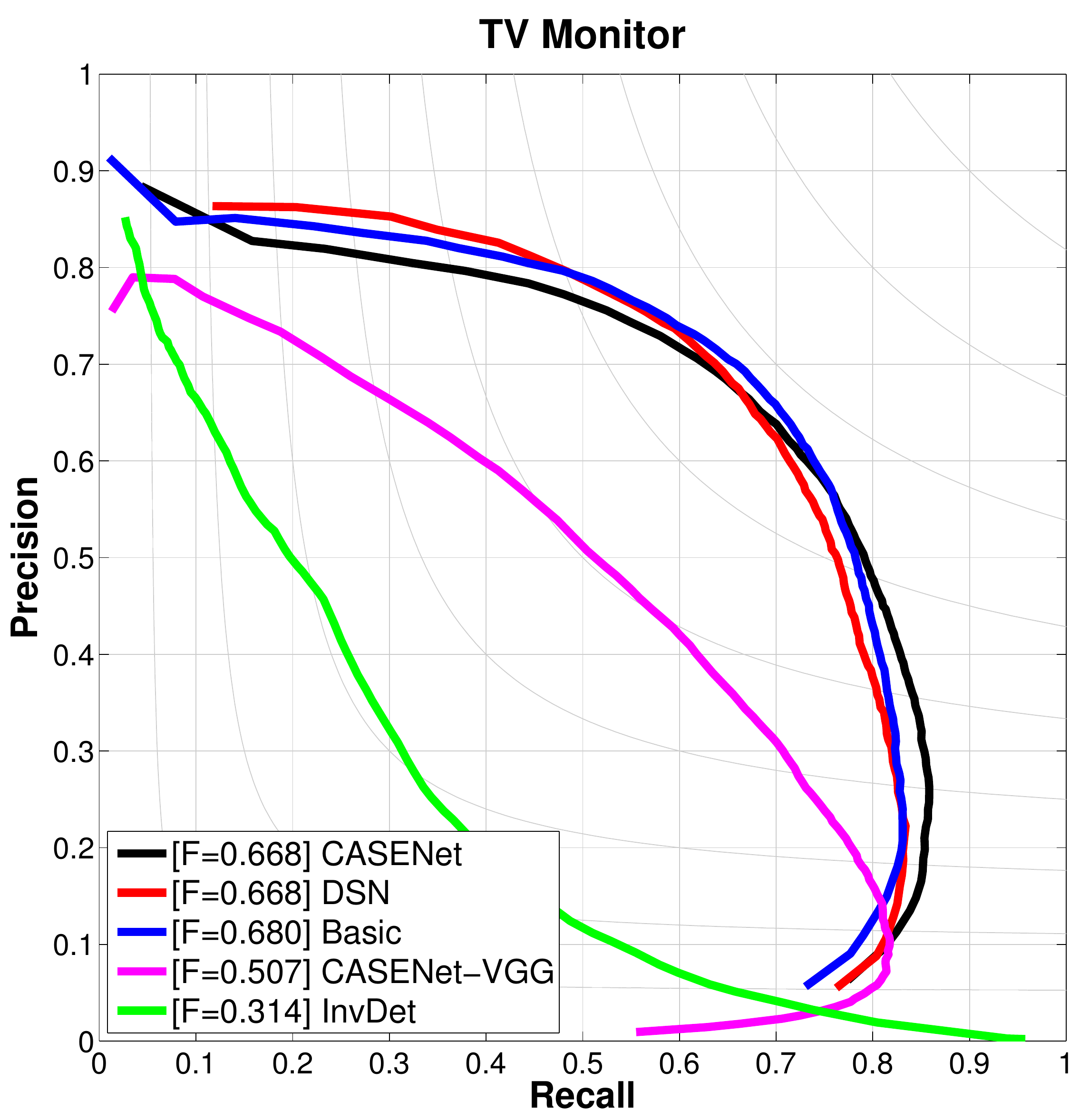}
	\caption{Class-wise precision-recall curves of the proposed methods and baselines on the SBD Dataset.}\label{fig:quan_sbd}
\end{figure*}

Fig. \ref{fig:quan_sbd} shows the precision-recall curves of each semantic class on the SBD Dataset. Note that while post-processing edge refinement may further boost the prediction performance \cite{Bertasius2015_hfl}, we evaluate only on the raw network predictions to better illustrate the network performance without introducing other factors. The evaluation is conducted fully based on the same benchmark code and ground truth files released by \cite{Hariharan2011}. Results indicate that CASENet slightly but consistently outperforms the baselines.

\subsection{Performance at different iterations}
We evaluate the Basic, DSN, CASENet on SBD for every 2000 iterations between 16000-30000, with the MF score shown in Fig.~\ref{fig:mfods_iters}. We found that the performance do not change significantly, and CASENet consistently outperforms Basic and DSN.

\begin{center}
	\includegraphics[width=1\columnwidth]{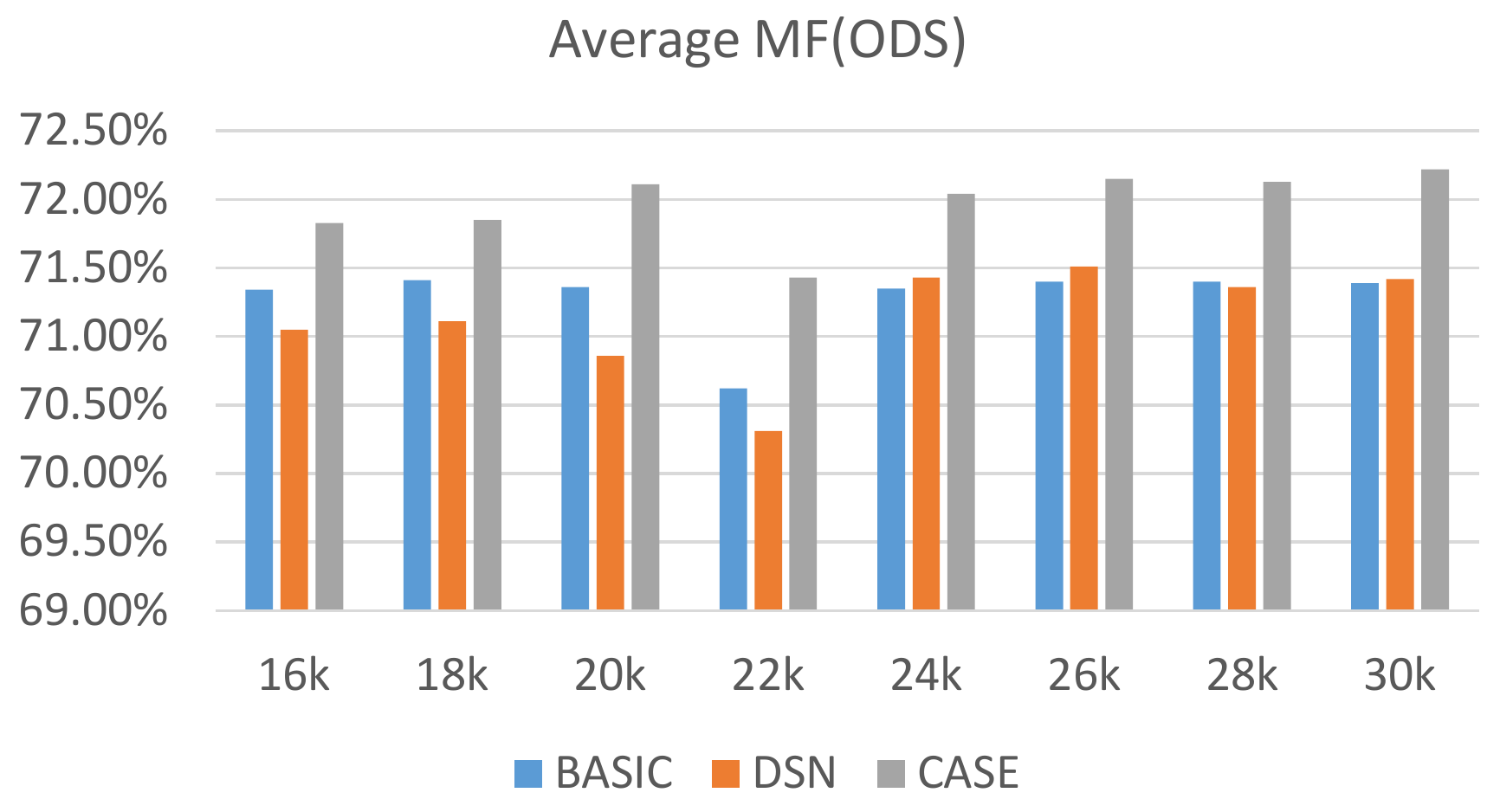}
	\captionof{figure}{Testing Performance vs. different iterations.}\label{fig:mfods_iters}
\end{center}

\subsection{Performance with a more standard split}
Considering that many datasets adopts the training + validation + test data split, we also randomly divided the SBD training set into a smaller training set and a new validation set with $1000$ images. We used the average loss on validation set to select the optimal iteration number separately for both Basic and CASENet. Their corresponding MFs on the test set are $71.22\%$ and $71.79\%$, respectively.

\section{Additional Results on Cityscapes}
\subsection{Additional qualitative results}
For more qualitative results, the readers may kindly refer to our released videos on Cityscapes validation set, as well as additional demo videos.

\subsection{Class-wise precision-recall curves}
\begin{figure*}[htbp]
	\centering
	\includegraphics[width=0.194\textwidth]{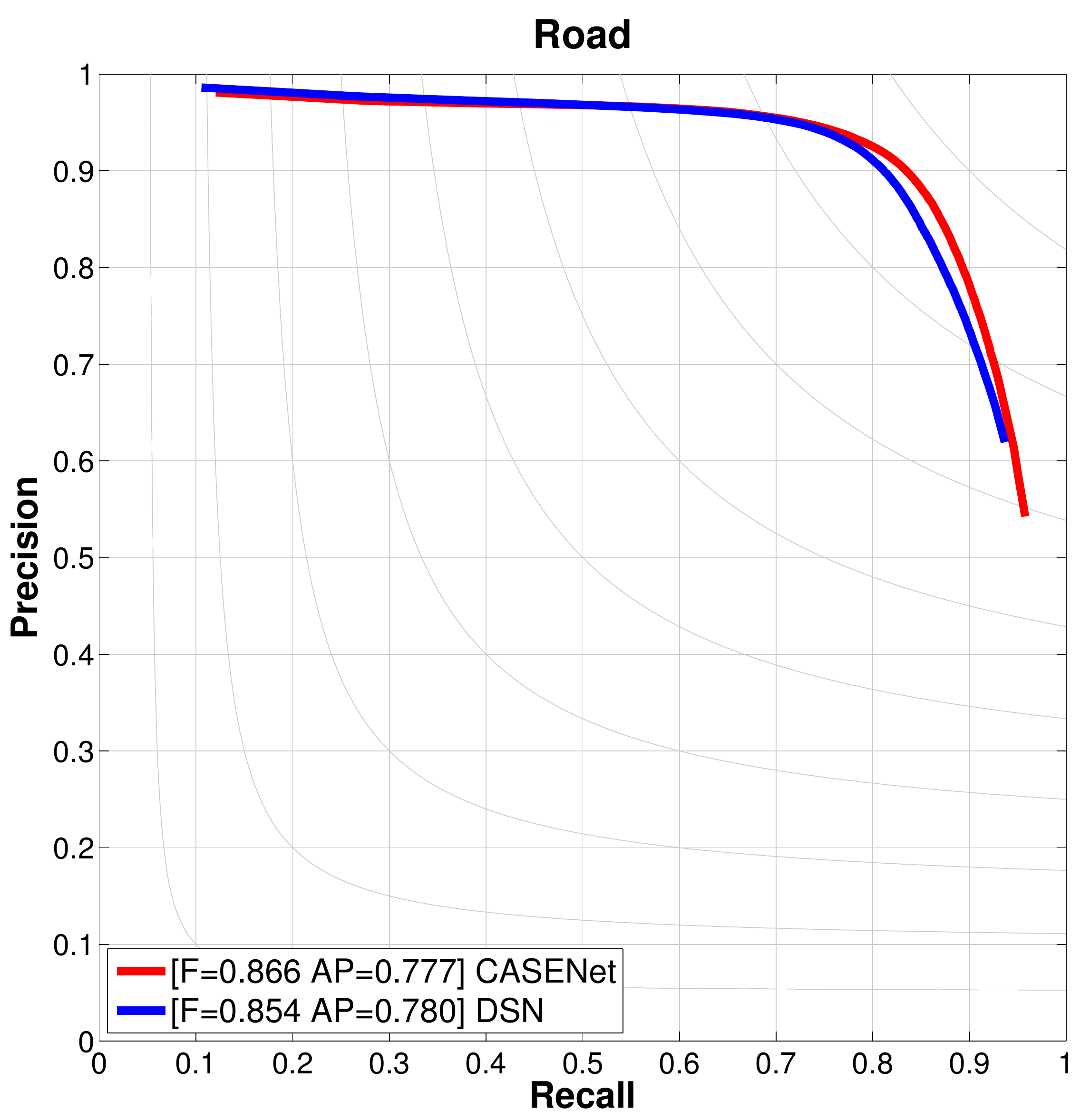}
	\includegraphics[width=0.194\textwidth]{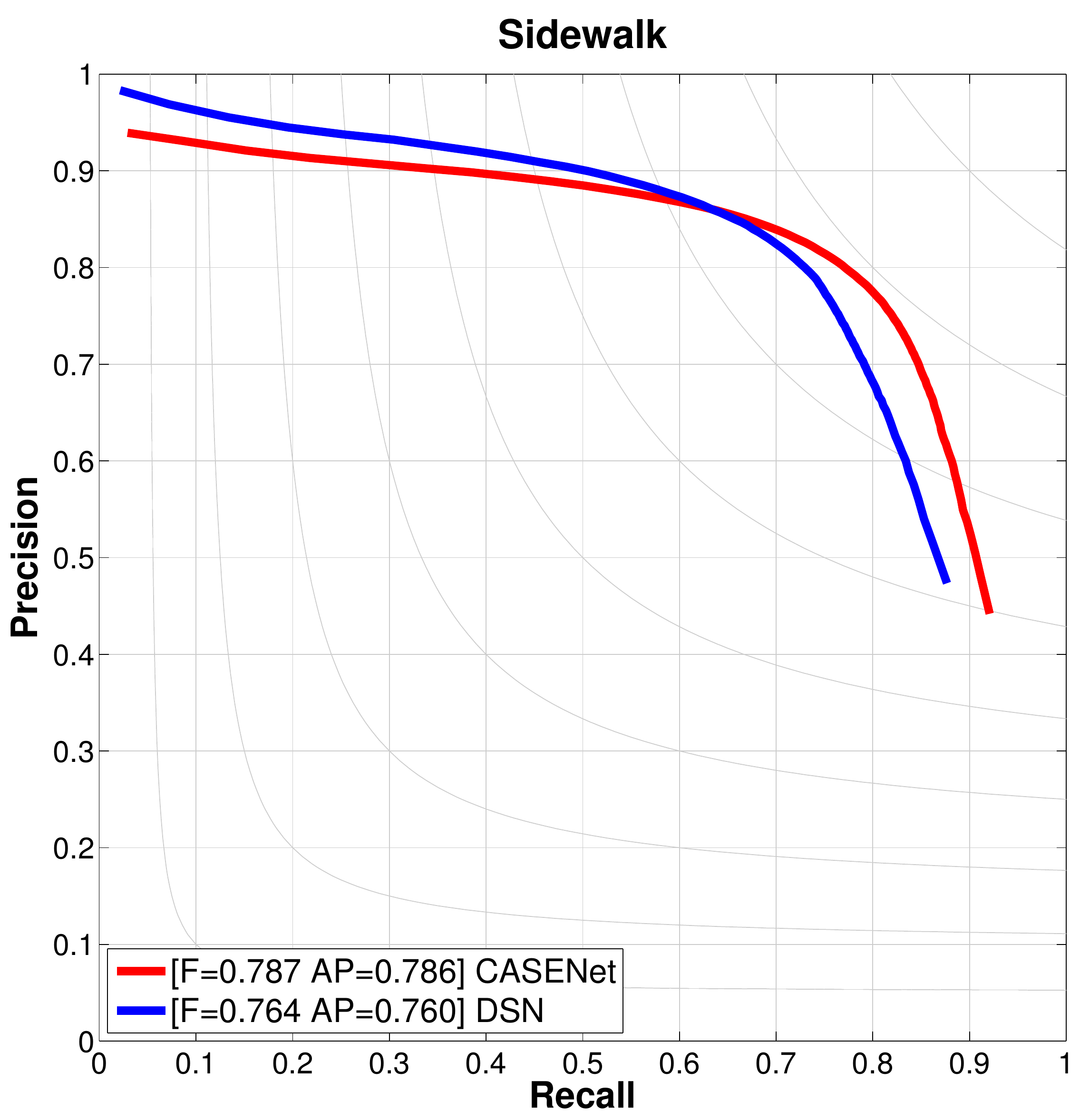}
	\includegraphics[width=0.194\textwidth]{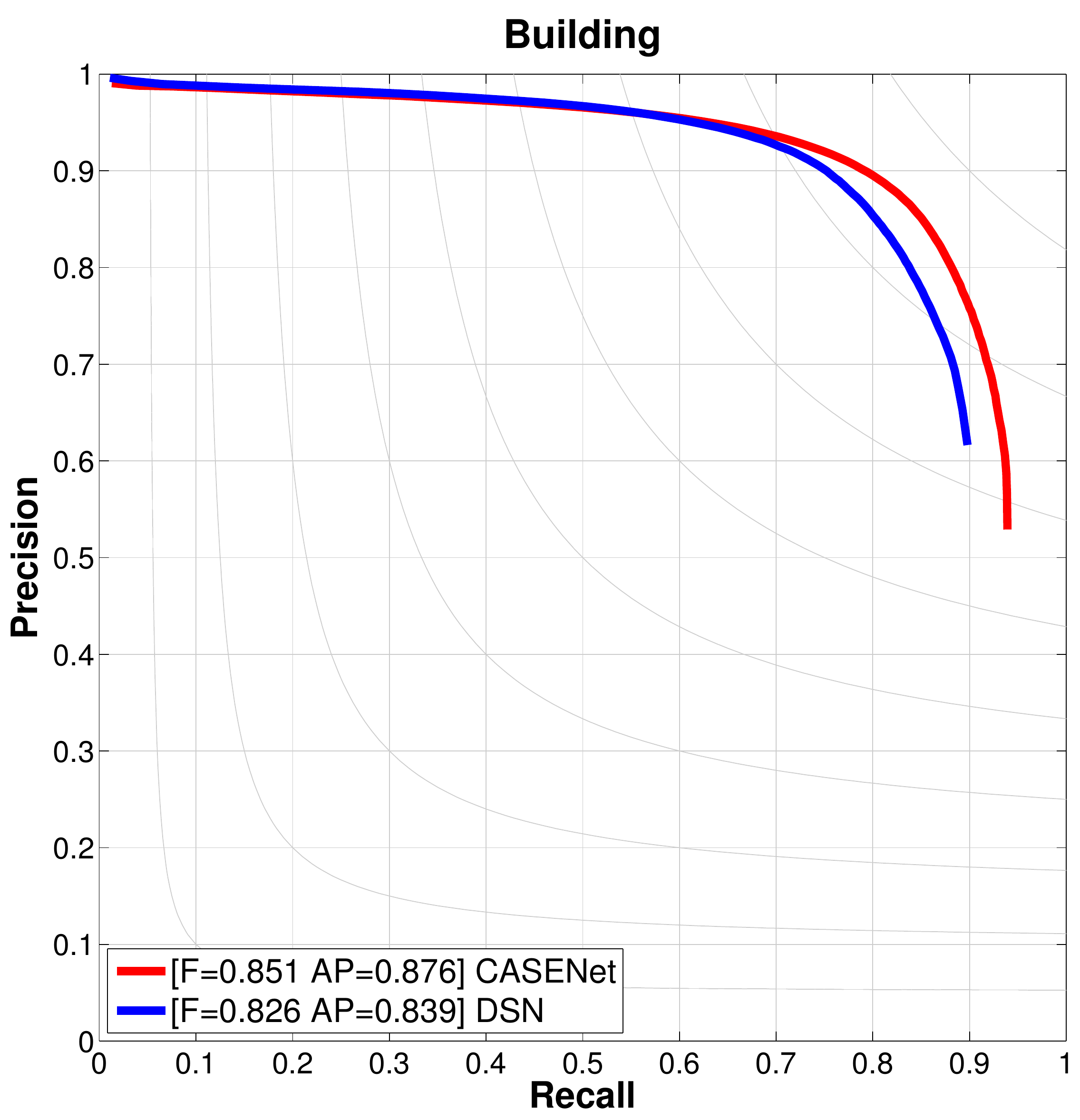}
	\includegraphics[width=0.194\textwidth]{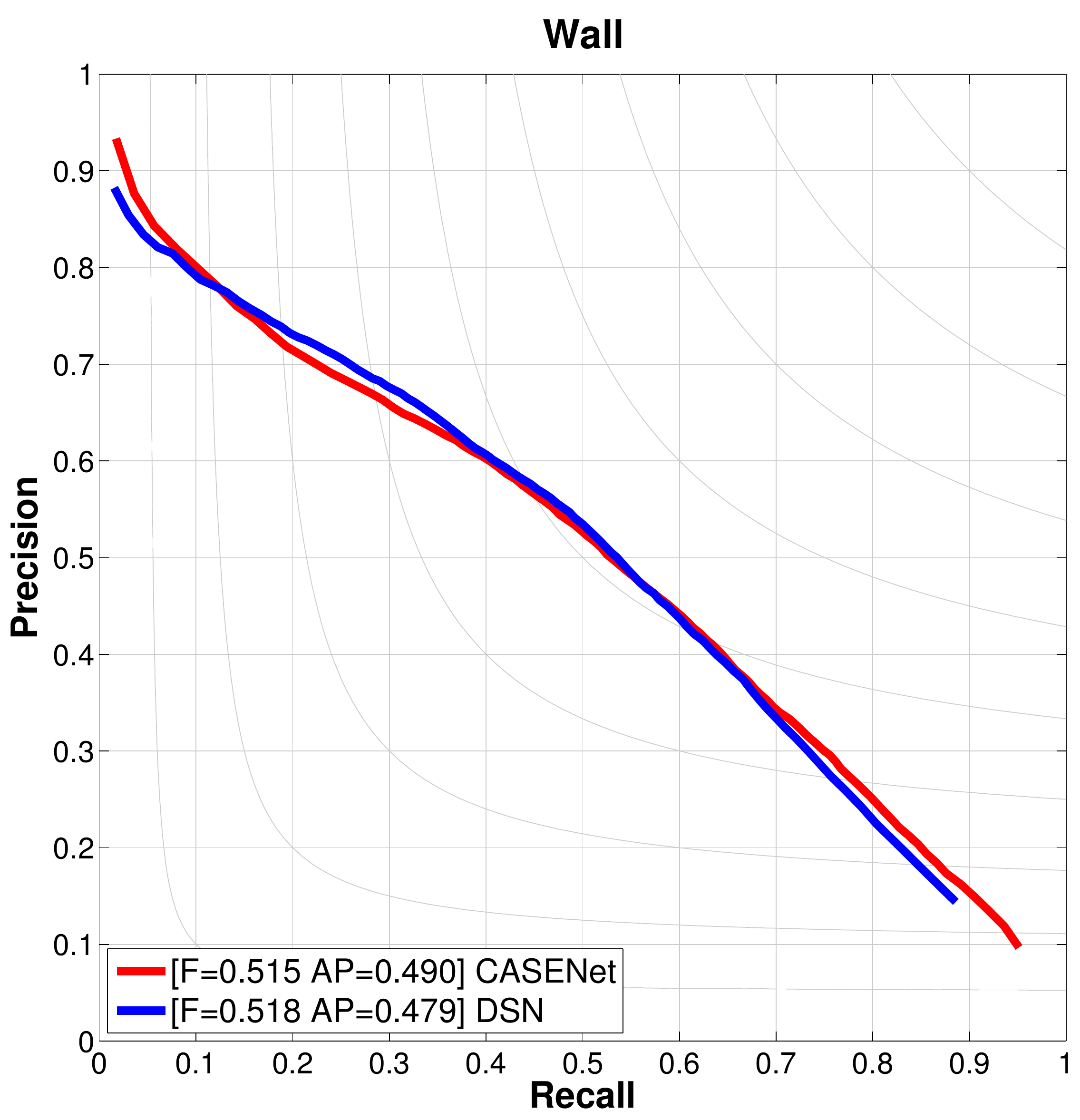}
	\includegraphics[width=0.194\textwidth]{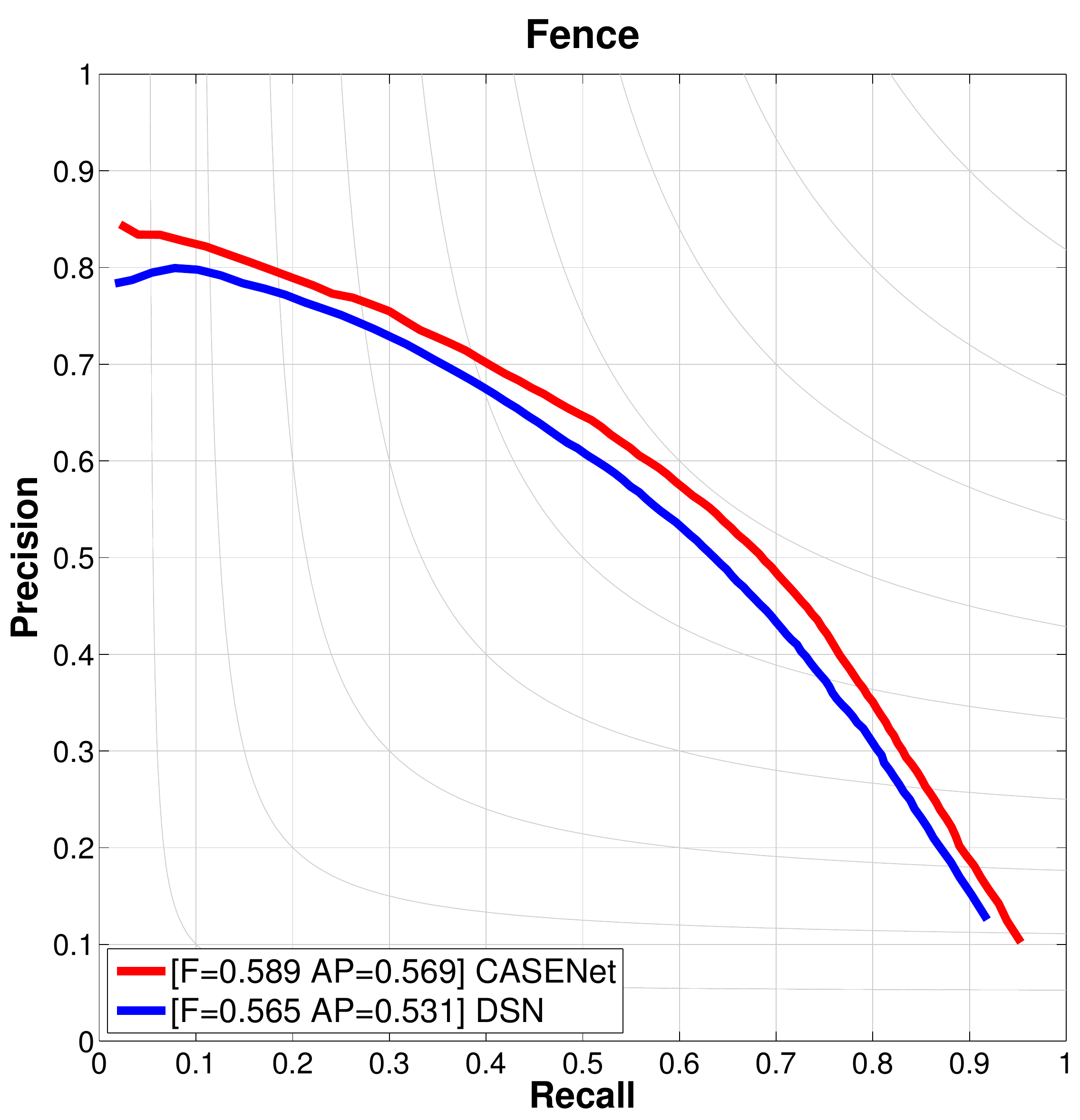}\\
	\quad\\\vspace{-0.35cm}
	\includegraphics[width=0.194\textwidth]{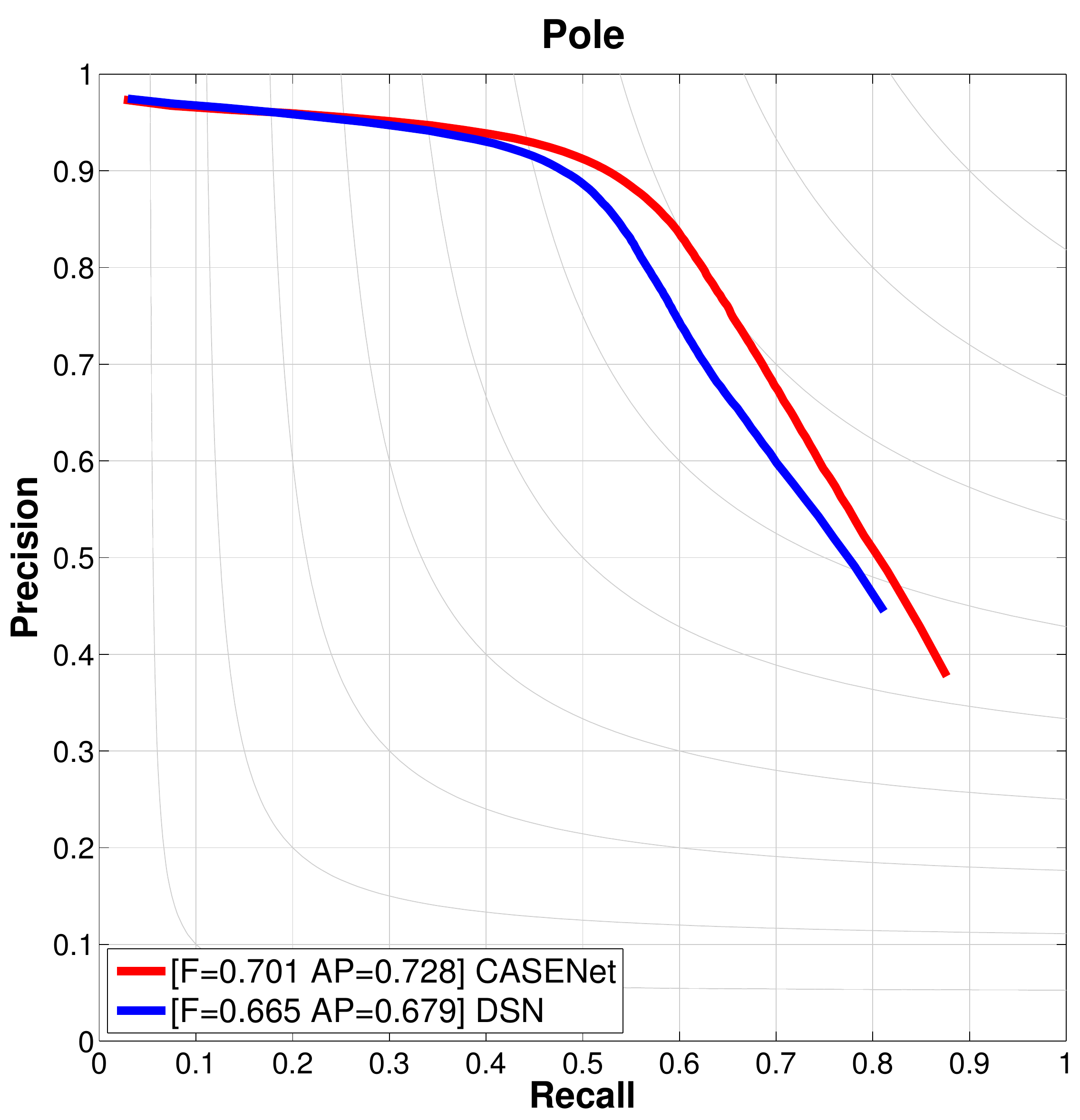}
	\includegraphics[width=0.194\textwidth]{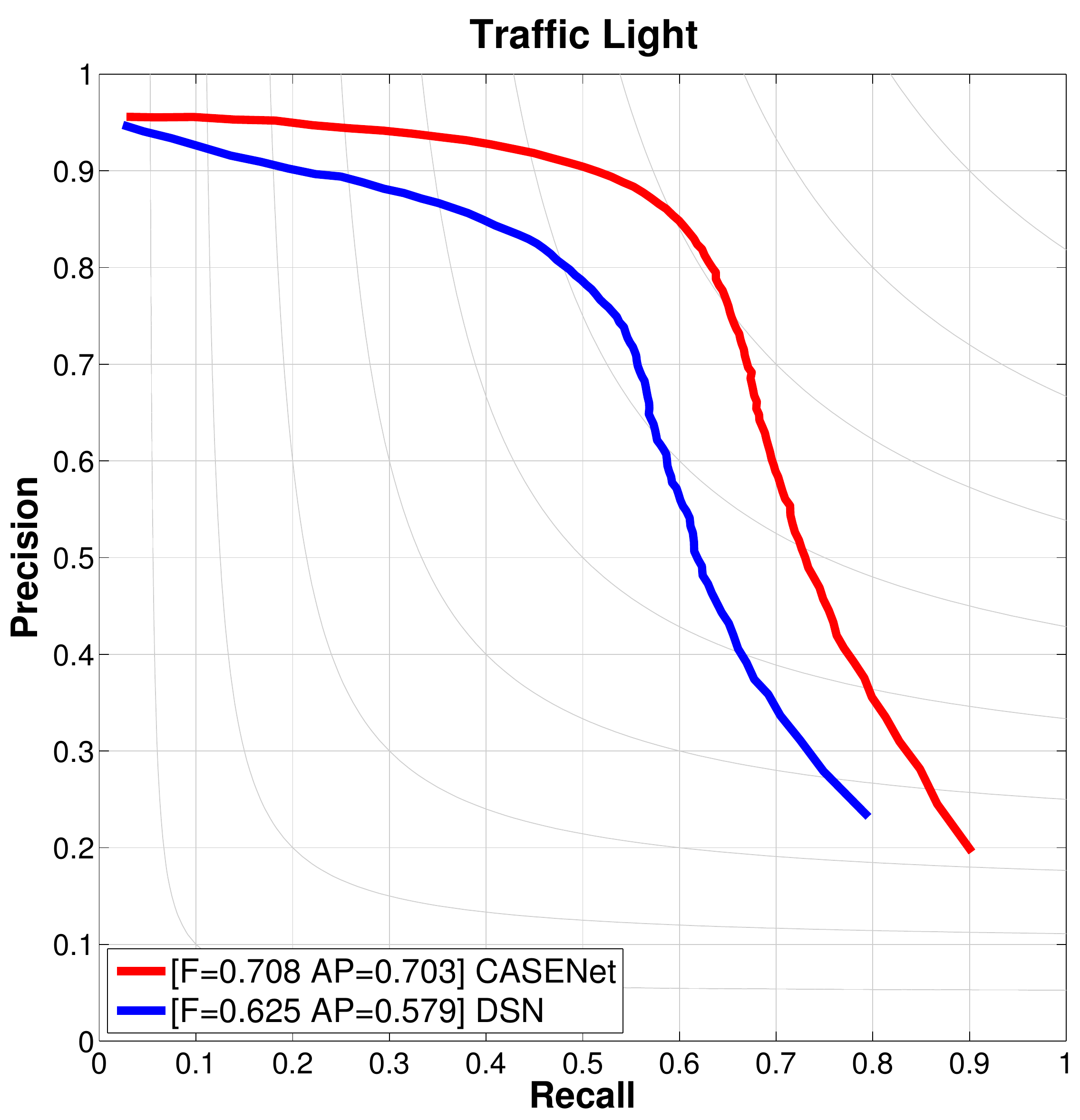}
	\includegraphics[width=0.194\textwidth]{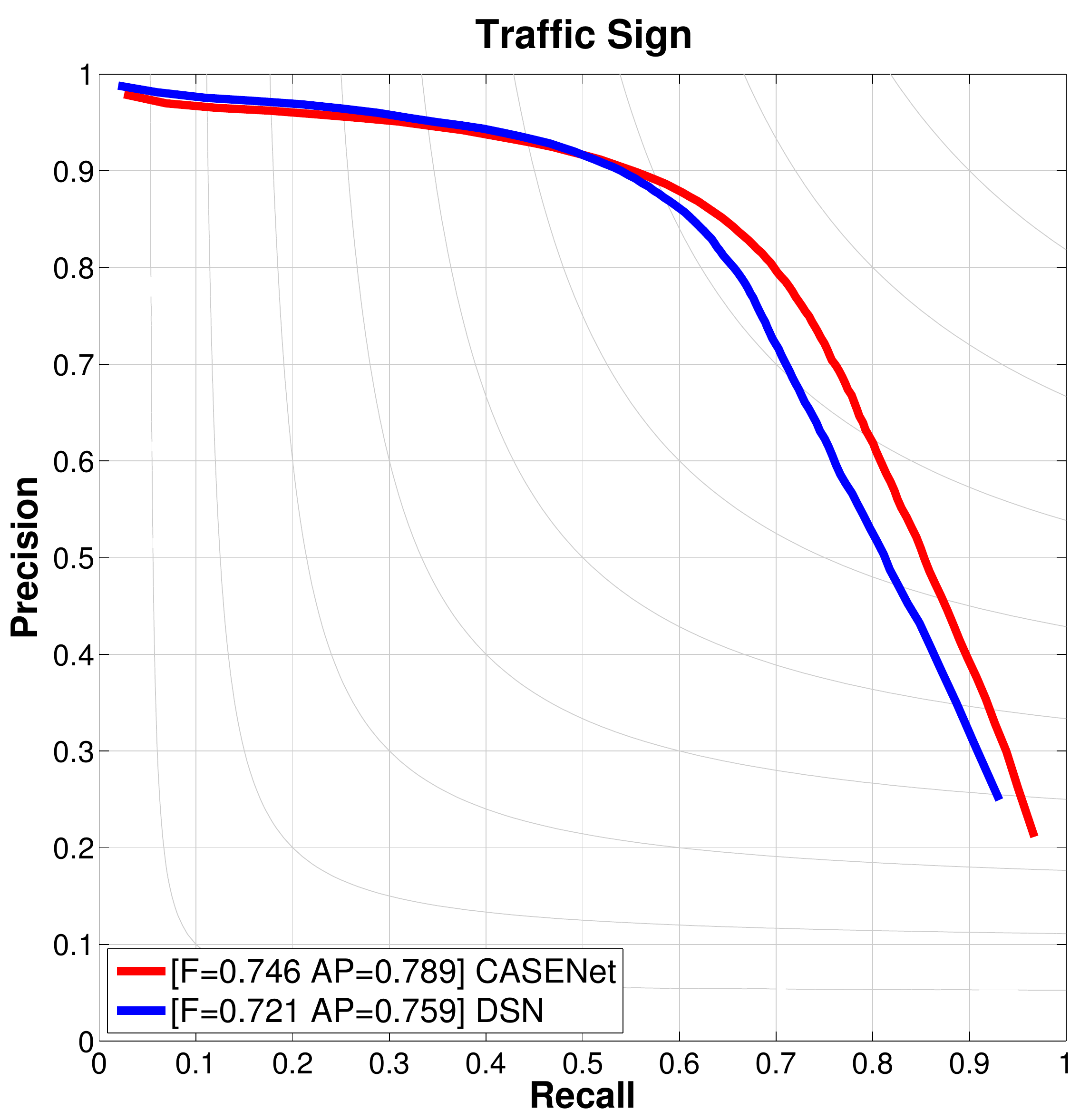}
	\includegraphics[width=0.194\textwidth]{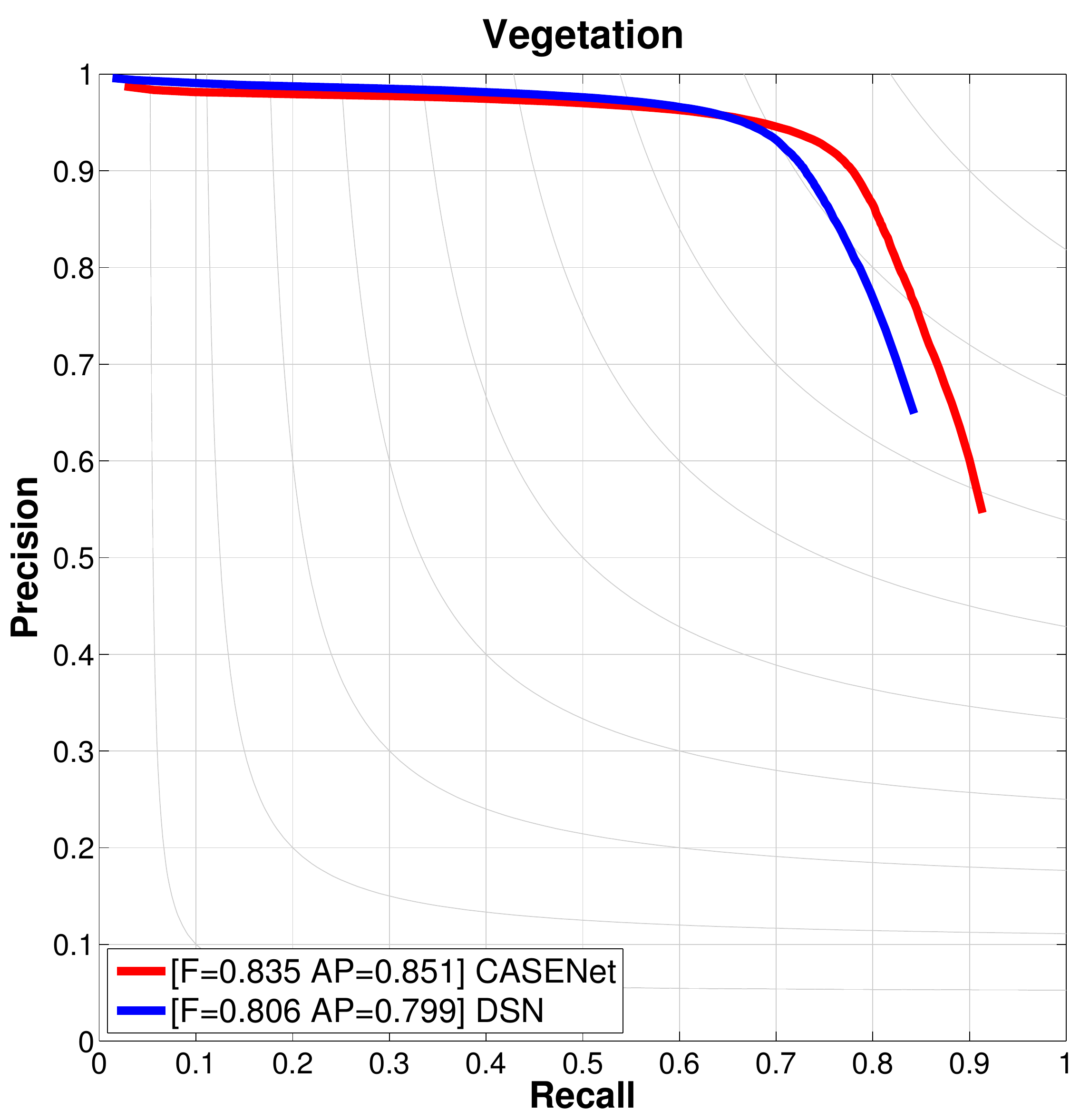}
	\includegraphics[width=0.194\textwidth]{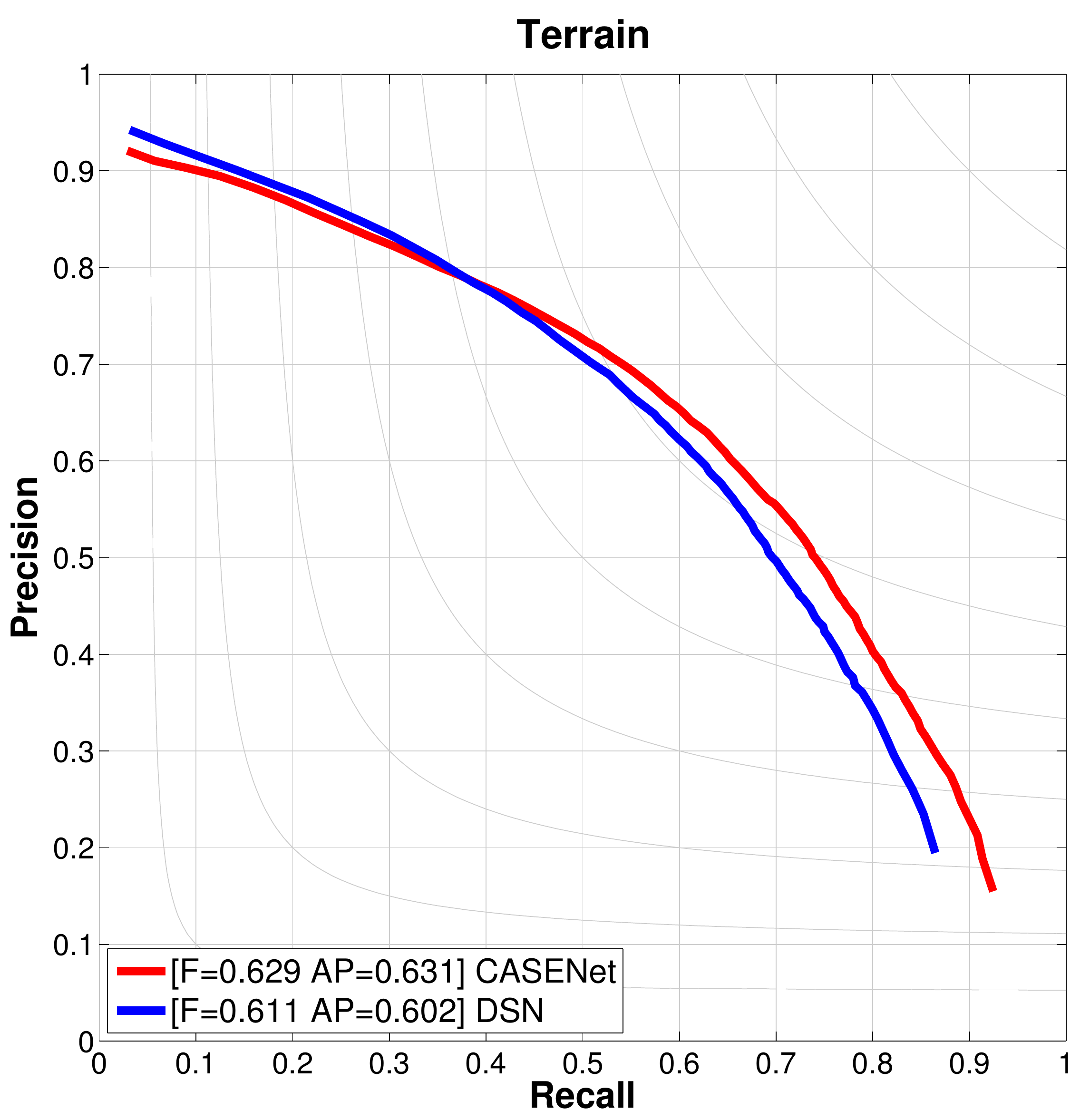}\\
	\quad\\\vspace{-0.35cm}
	\includegraphics[width=0.194\textwidth]{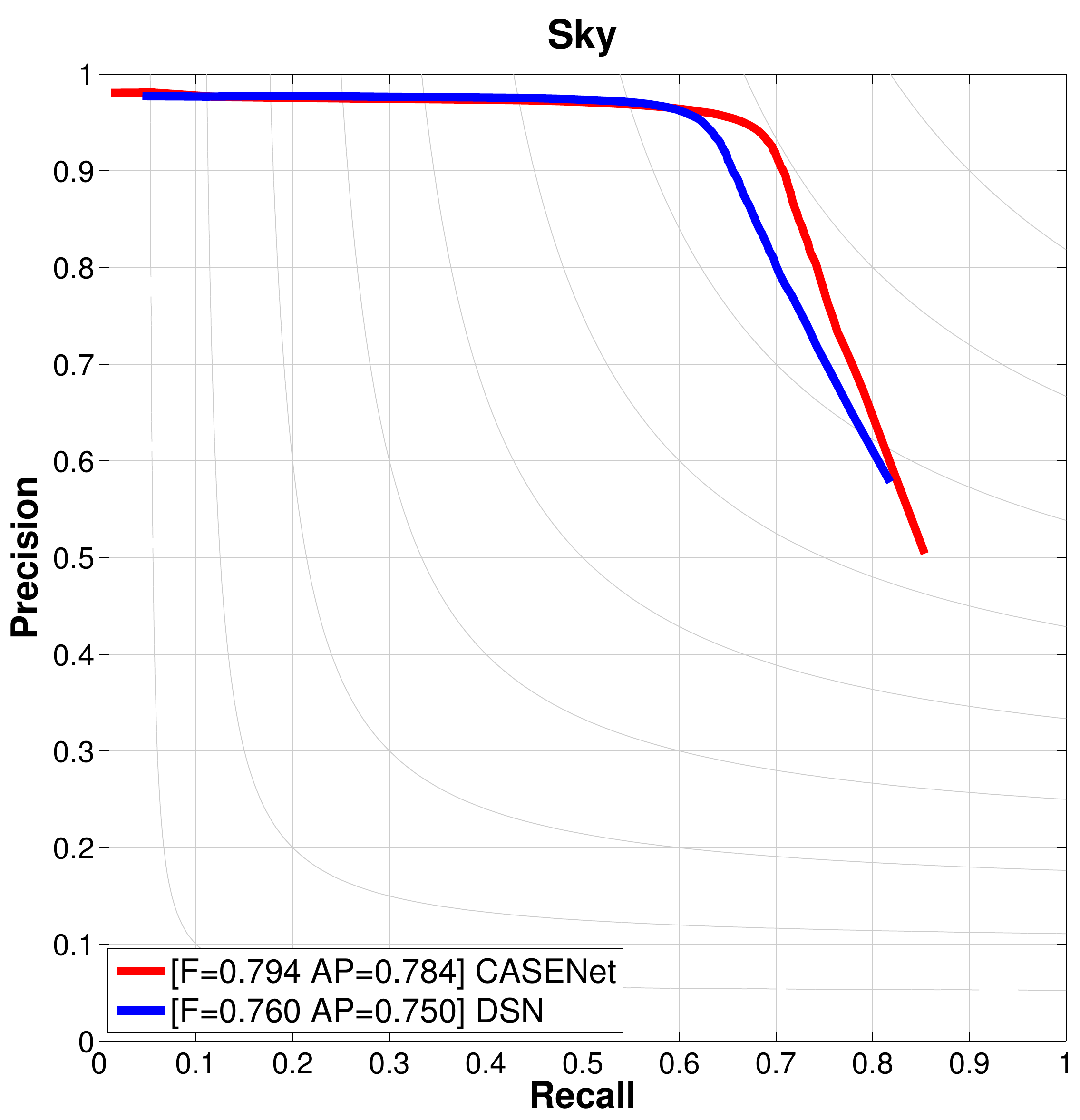}
	\includegraphics[width=0.194\textwidth]{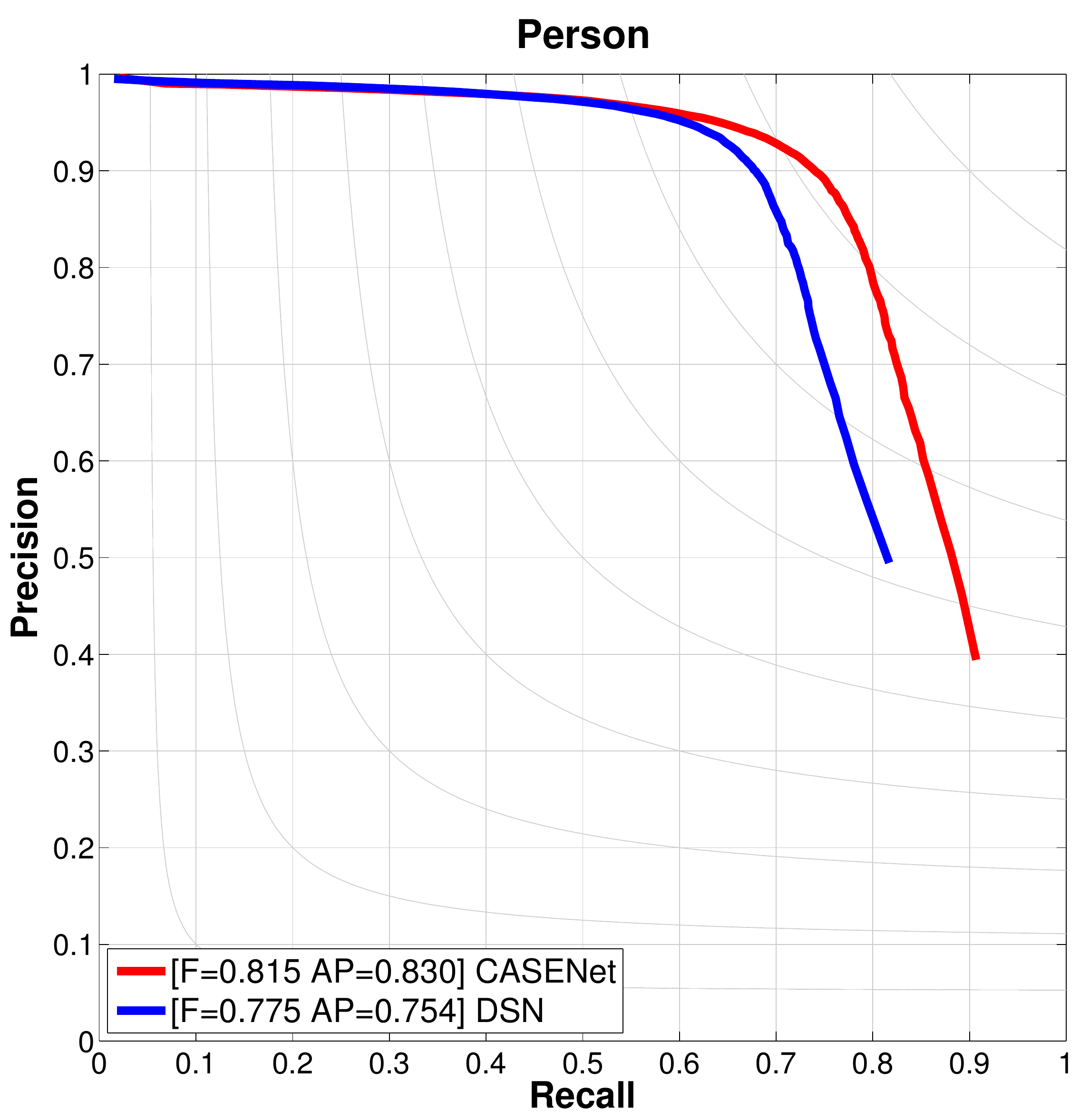}
	\includegraphics[width=0.194\textwidth]{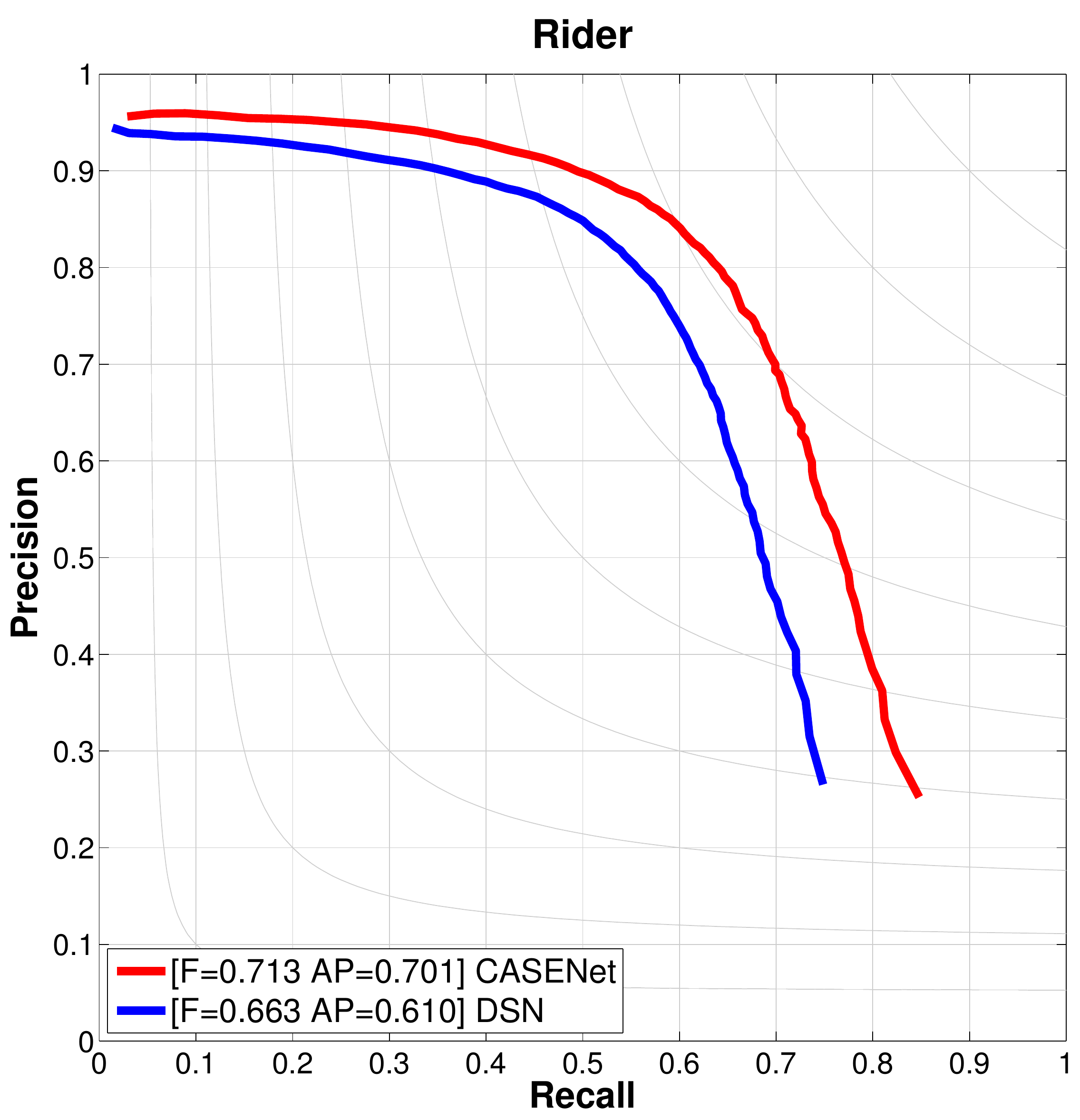}
	\includegraphics[width=0.194\textwidth]{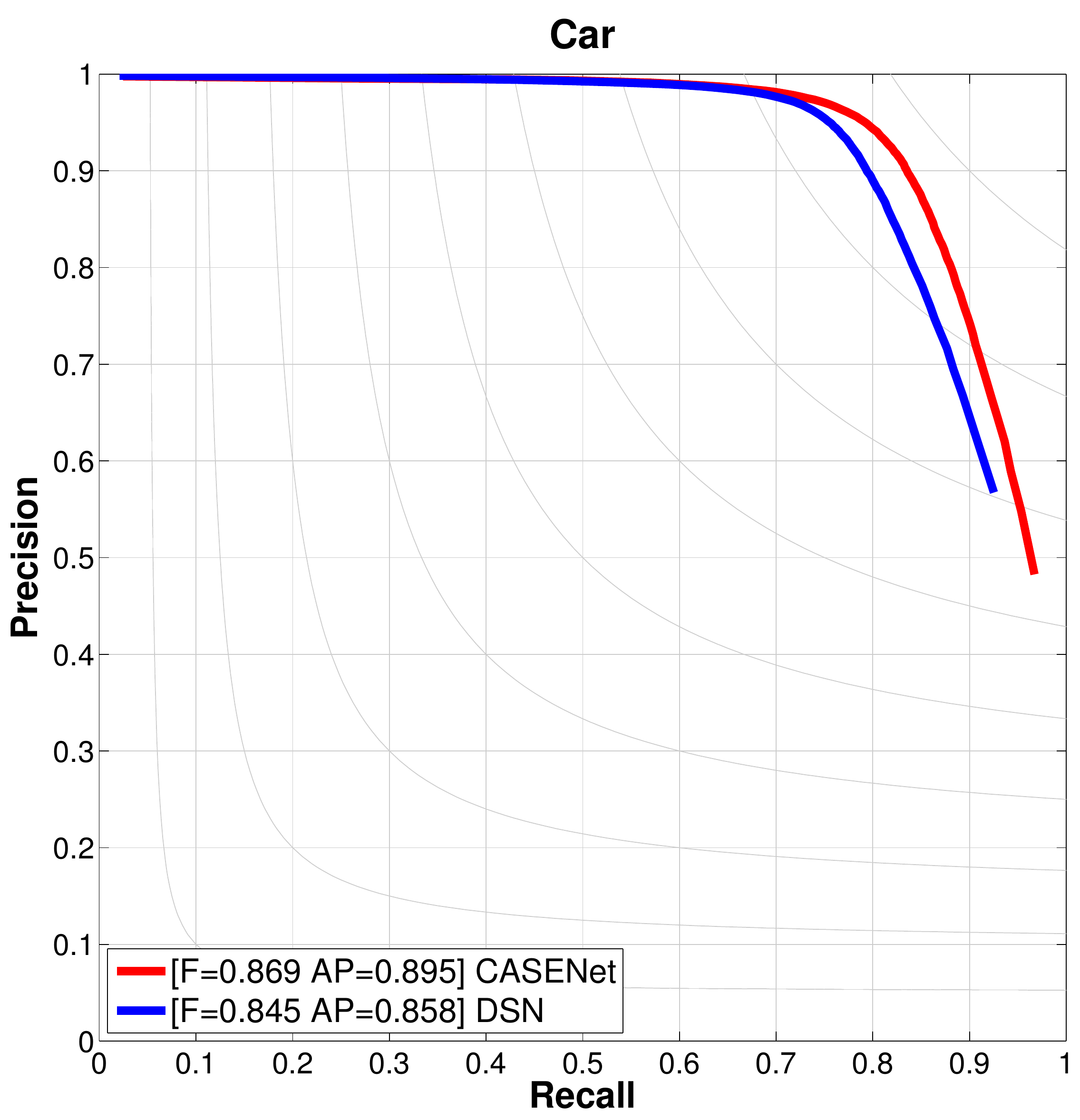}
	\includegraphics[width=0.194\textwidth]{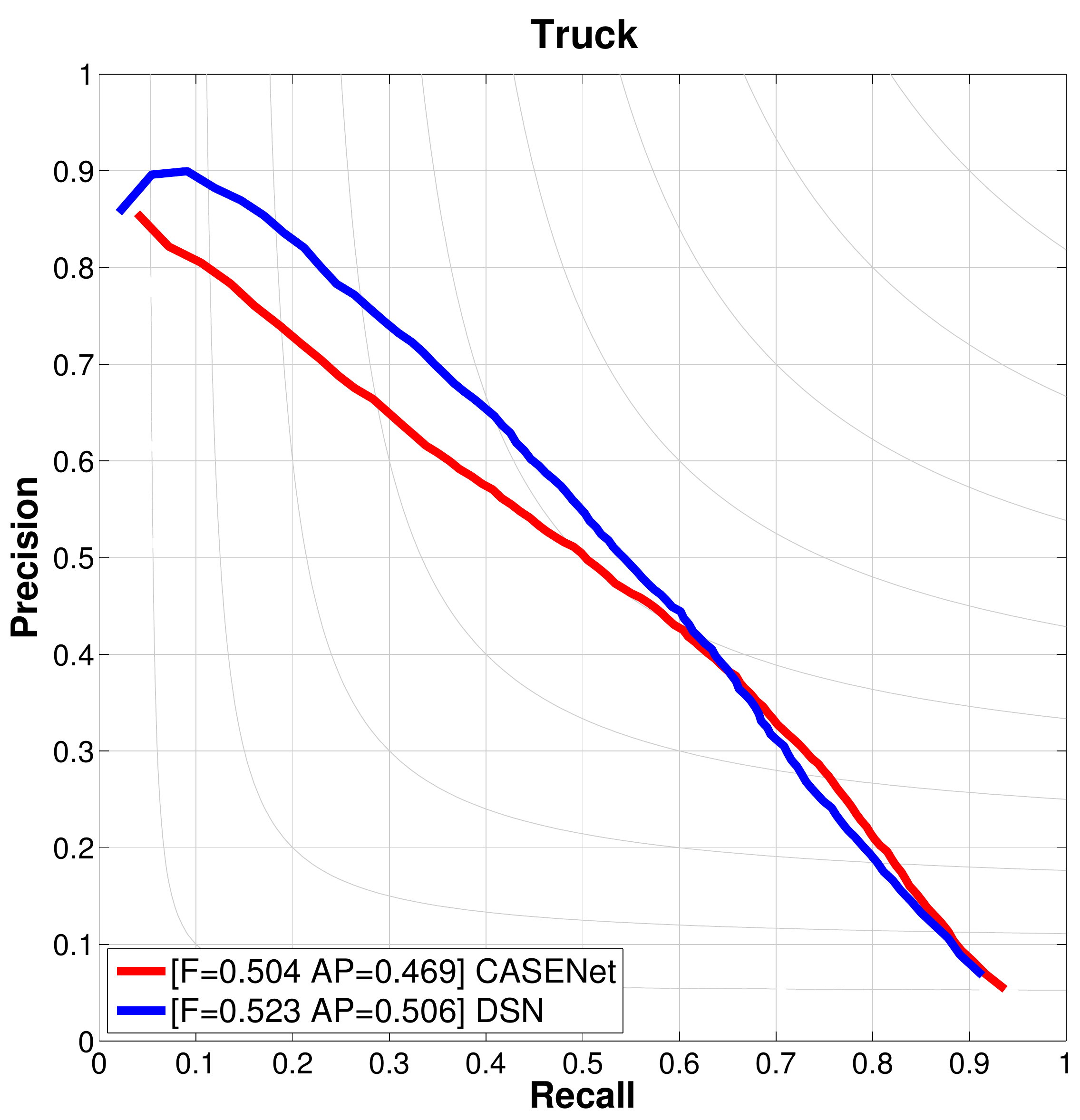}\\
	\quad\\\vspace{-0.35cm}
	\includegraphics[width=0.194\textwidth]{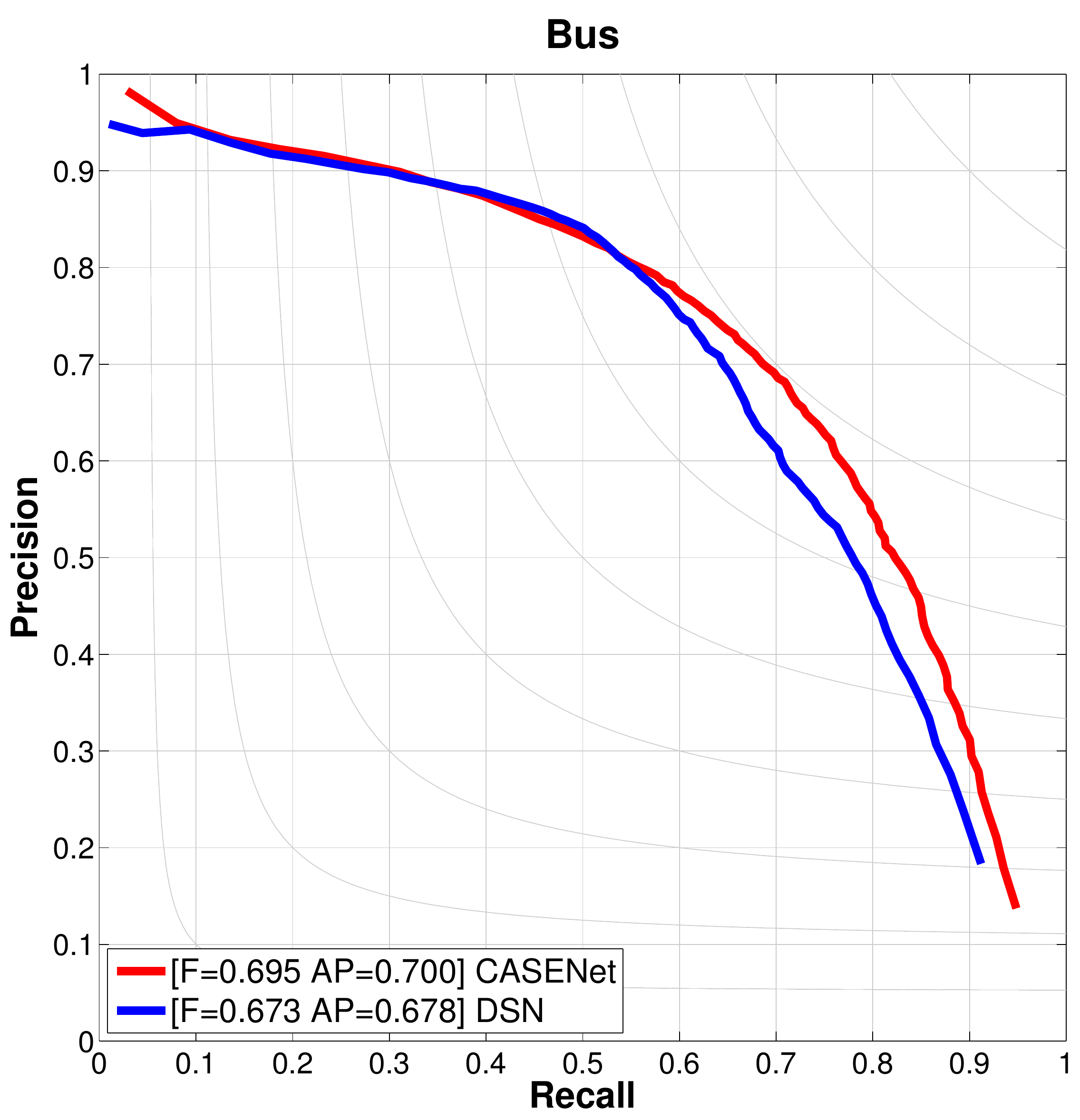}
	\includegraphics[width=0.194\textwidth]{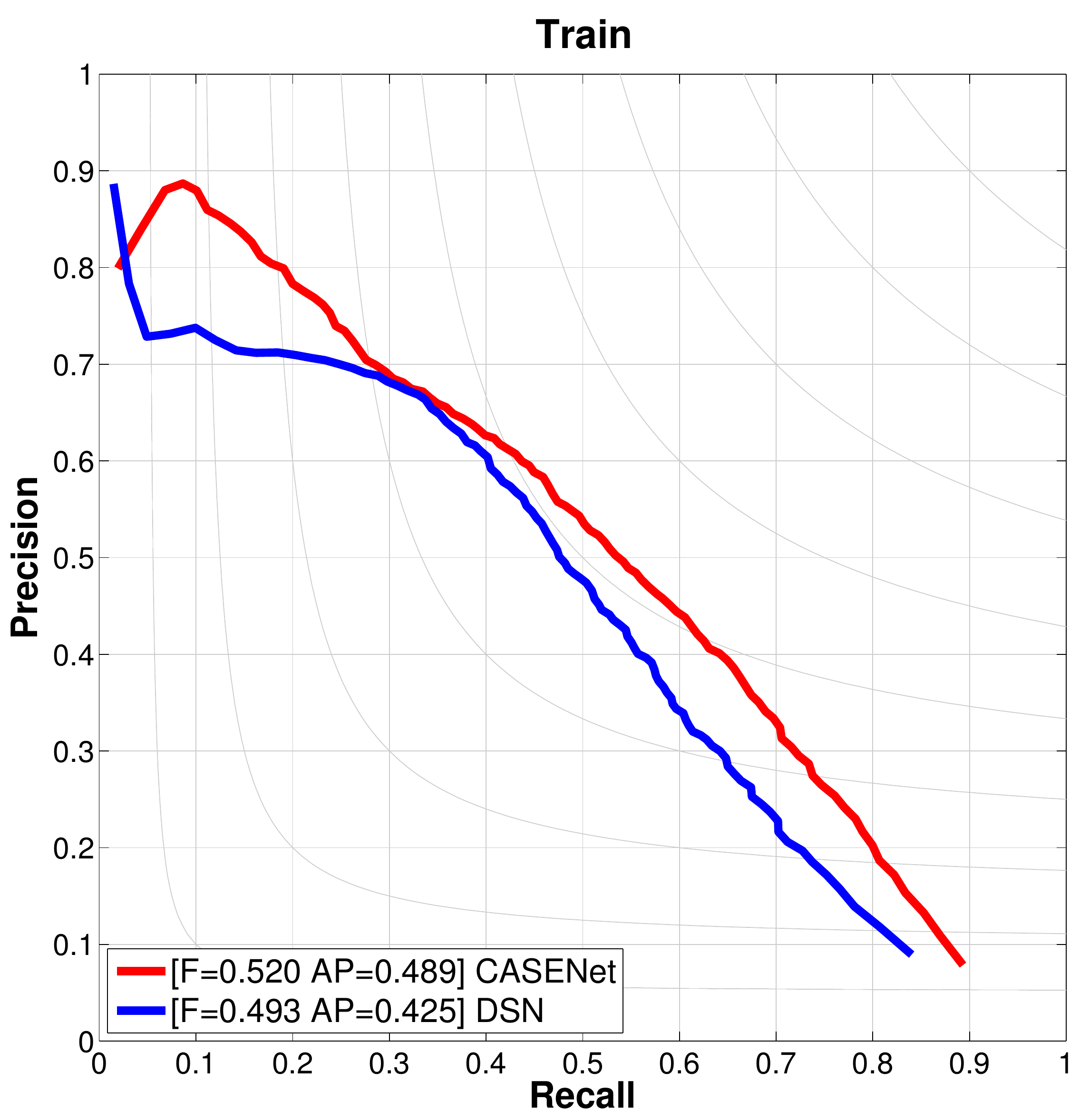}
	\includegraphics[width=0.194\textwidth]{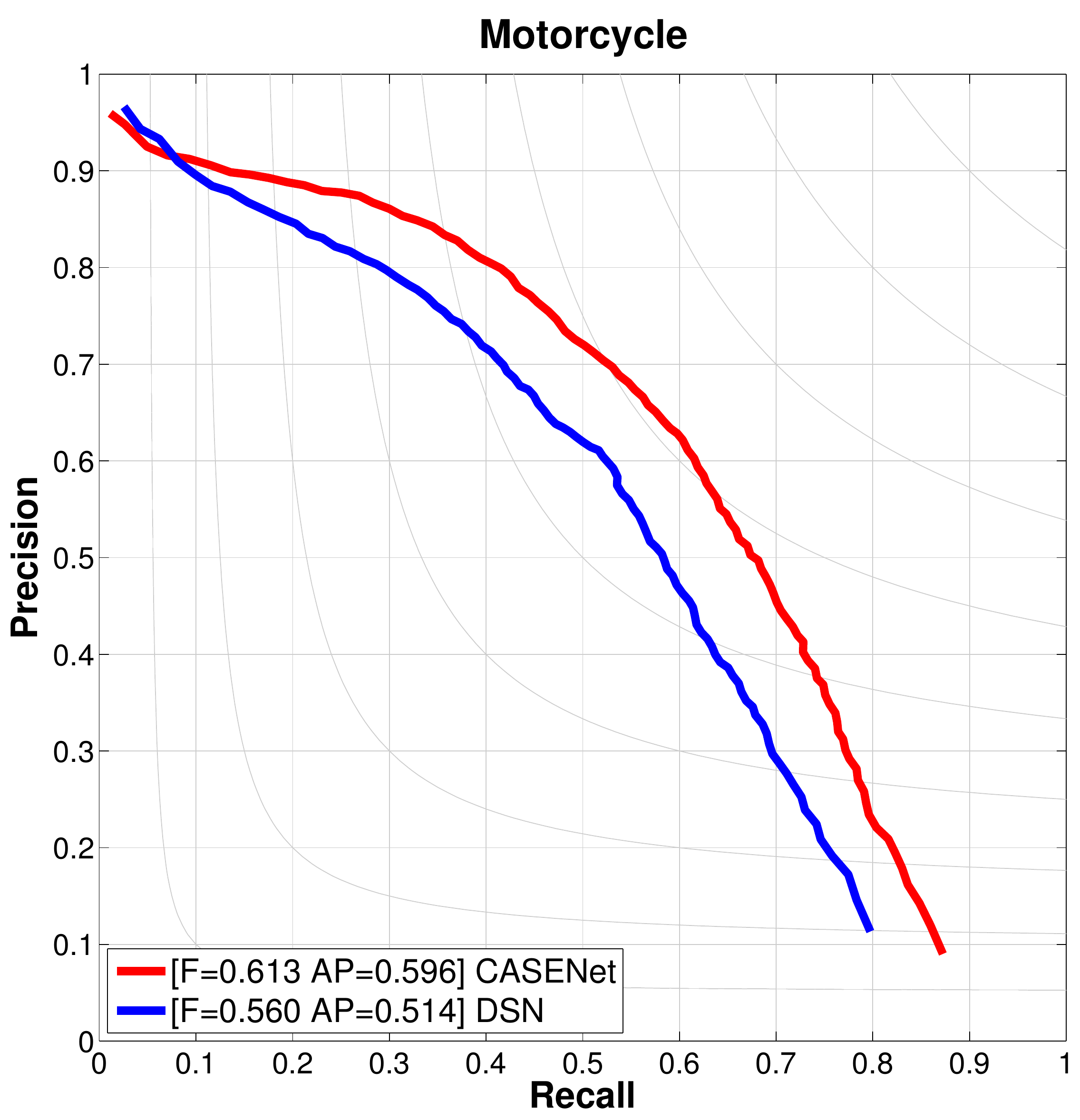}
	\includegraphics[width=0.194\textwidth]{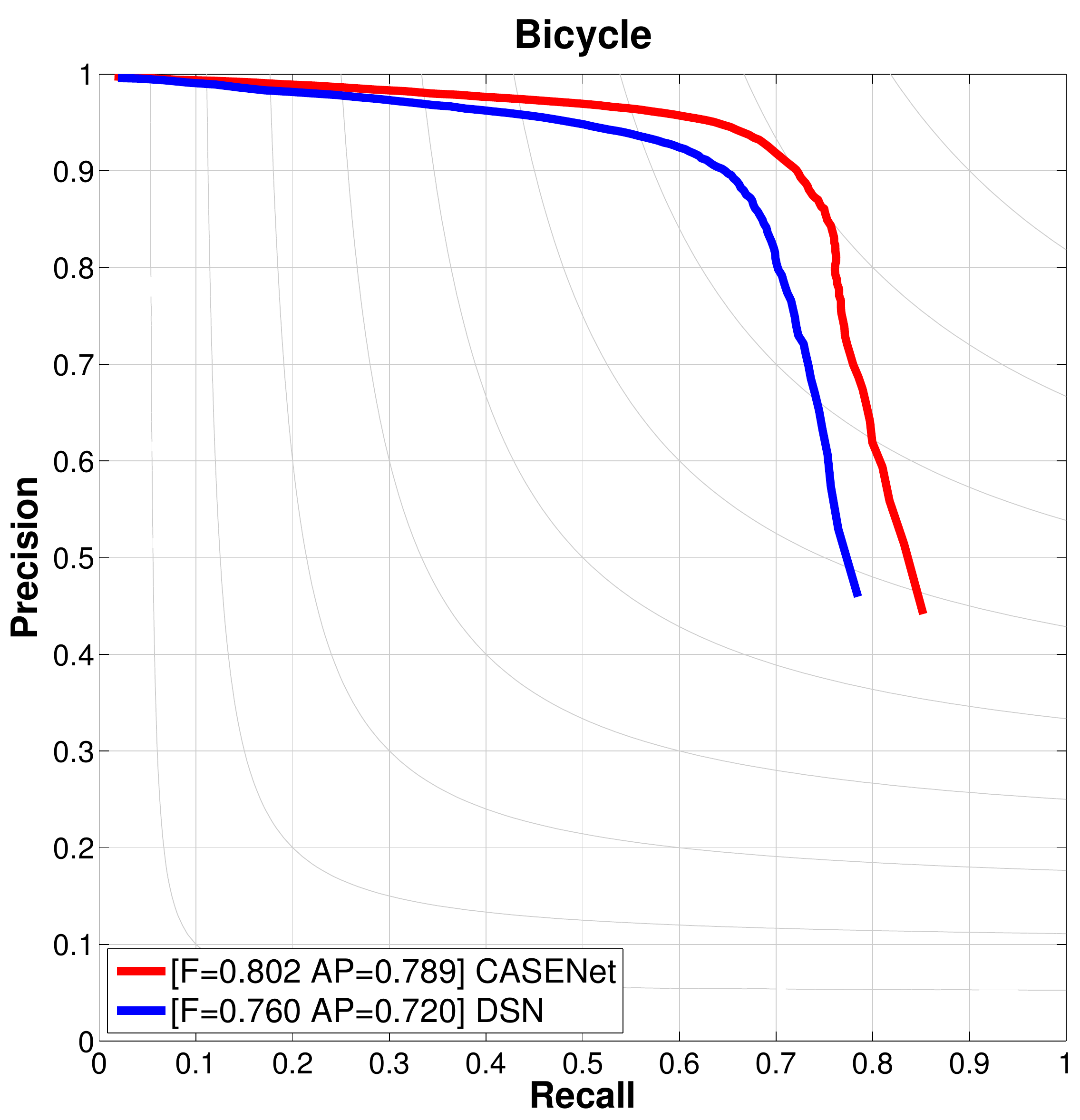}
	\caption{Class-wise precision-recall curves of CASENet and DSN on the Cityscapes Dataset.}\label{fig:quan_city}
\end{figure*}

Fig. \ref{fig:quan_city} shows the precision-recall curves of each semantic class on the Cityscapes Dataset. Again the evaluation is conducted only on the raw network predictions. Since evaluating the results at original scale ($1024 \times 2048$) is extremely slow and is not necessary, we bilinearly downsample both the edge responses and ground truths to $512 \times 1024$. Results indicate that CASENet consistently outperforms the ResNet with the DSN architecture.

\end{appendices}

\end{document}